\documentclass[12pt]{ociamthesis}  

\usepackage[numbers]{natbib}
\usepackage{commath}
\usepackage{amssymb}
\usepackage{amsmath}
\usepackage[mathscr]{euscript}
\usepackage{breqn}
\usepackage{xcolor}
\usepackage{url}
\usepackage{subcaption}
\usepackage{algorithm}
\usepackage{algpseudocode}
\usepackage{paralist}
\usepackage{algorithm}
\usepackage{algpseudocode}
\usepackage{mathrsfs}
\usepackage{booktabs}
\usepackage{ragged2e}
\usepackage{graphicx}
\usepackage{boldline}
\usepackage{bm}
\usepackage{algorithm}
\usepackage[noend]{algcompatible}
\usepackage{wrapfig}
\usepackage{dsfont}
\usepackage[final]{pdfpages}

\DeclareMathOperator*{\argminA}{arg\,min}

\newcommand{\etal}{\textit{et al}.}
\newcommand{\ie}{\textit{i}.\textit{e}.,}
\newcommand{\eg}{\textit{e}.\textit{g}.,}
\newcommand{\etc}{\textit{etc}.}
\usepackage{tikz}
\newcommand*\circled[1]{\tikz[baseline=(char.base)]{
            \node[shape=circle,draw,inner sep=0.1pt] (char) {#1};}}

\newcommand{\nicknameRecGAN}{3D-RecGAN++}
\newcommand{\nicknameAttSets}{AttSets}
\newcommand{\nicknameFASet}{FASet}
\newcommand{\nicknameBoNet}{3D-BoNet}

\title{Learning to Reconstruct and Segment 3D Objects}
\author{Bo Yang}             
\college{Exeter College}  
\degree{Doctor of Philosophy}     
\degreedate{Hilary 2020}         

\begin{document}

\baselineskip=18pt plus1pt

\setcounter{secnumdepth}{3}
\setcounter{tocdepth}{3}

\maketitle                  
\begin{dedication}
This thesis is dedicated to my parents\\
\centerline{for their indispensable, altruistic support.}
\end{dedication}
\begin{acknowledgements}
Looking back to the past three and a half years at Oxford, I feel grateful to many people who have helped me go through my DPhil journey. 

First and foremost, I would like to thank my supervisors Niki and Andrew. Their remarkable supervision has reshaped my attitude towards meaningful research, trained my skills for solid work, and created my vision for novel ideas. These are indeed essential for all the achievements in this thesis.

Second, I would like to thank all my peer collaborators, especially the seniors Dr. Hongkai Wen, Dr. Stefano Rosa, Dr. Sen Wang, and Dr. Ronald Clark. Their advice is extremely timely and beneficial to shape my research ideas and experiments. 

Third, I want to thank the labmates of Cyber-Physical System Group and my friends in the department of computer science and Exeter college. It is always enjoyable and memorable to go for tennis, formal dinners, punting, balls, drinks and so many others. 

At last, I would like to express my sincere thanks to my family. The support and love from my parents and brothers are everlasting. The sweetness from my beloved girlfriend always make my day.

\end{acknowledgements}
\begin{abstract}
To endow machines with the ability to perceive the real-world in a three dimensional representation as we do as humans is a fundamental and long-standing topic in Artificial Intelligence. Given different types of visual inputs such as images or point clouds acquired by 2D/3D sensors, one important goal is to understand the geometric structure and semantics of the 3D environment. Traditional approaches usually leverage hand-crafted features to estimate the shape and semantics of objects or scenes. However, they are difficult to generalize to novel objects and scenarios, and struggle to overcome critical issues caused by visual occlusions. By contrast, we aim to understand scenes and the objects within them by learning general and robust representations using deep neural networks, trained on large-scale real-world 3D data. To achieve these aims, this thesis makes three core contributions from object-level 3D shape estimation from single or multiple views to scene-level semantic understanding. 

In Chapter \ref{chap:rec_obj_sv}, we start by estimating the full 3D shape of an object from a single image. To recover a dense 3D shape with geometric details, a powerful encoder-decoder architecture together with adversarial learning is proposed to learn feasible geometric priors from large-scale 3D object repositories. In Chapter \ref{chap:rec_obj_mv}, we build a more general framework to estimate accurate 3D shapes of objects from an arbitrary number of images. By introducing a novel attention-based aggregation module together with a two-stage training algorithm, our framework is able to integrate a variable number of input views, predicting robust and consistent 3D shapes for objects. In Chapter \ref{chap:seg_obj_pc}, we extend our study to 3D scenes which are generally a complex collection of individual objects. Real-world 3D scenes such as point clouds are usually cluttered, unstructured, occluded and incomplete. By drawing on previous work in point-based networks, we introduce a brand-new end-to-end pipeline to recognize, detect and segment all objects simultaneously in a 3D point cloud. 

Overall, this thesis develops a series of novel data-driven algorithms to allow machines to perceive our real-world 3D environment, arguably pushing the boundaries of Artificial Intelligence and machine understanding.

\end{abstract} 

\begin{romanpages}          
\tableofcontents            
\listoffigures              
\end{romanpages}            

\chapter{Introduction}
\section{Motivation}
\begin{figure}[!b]
    \centering
    \includegraphics[width=0.95\textwidth]{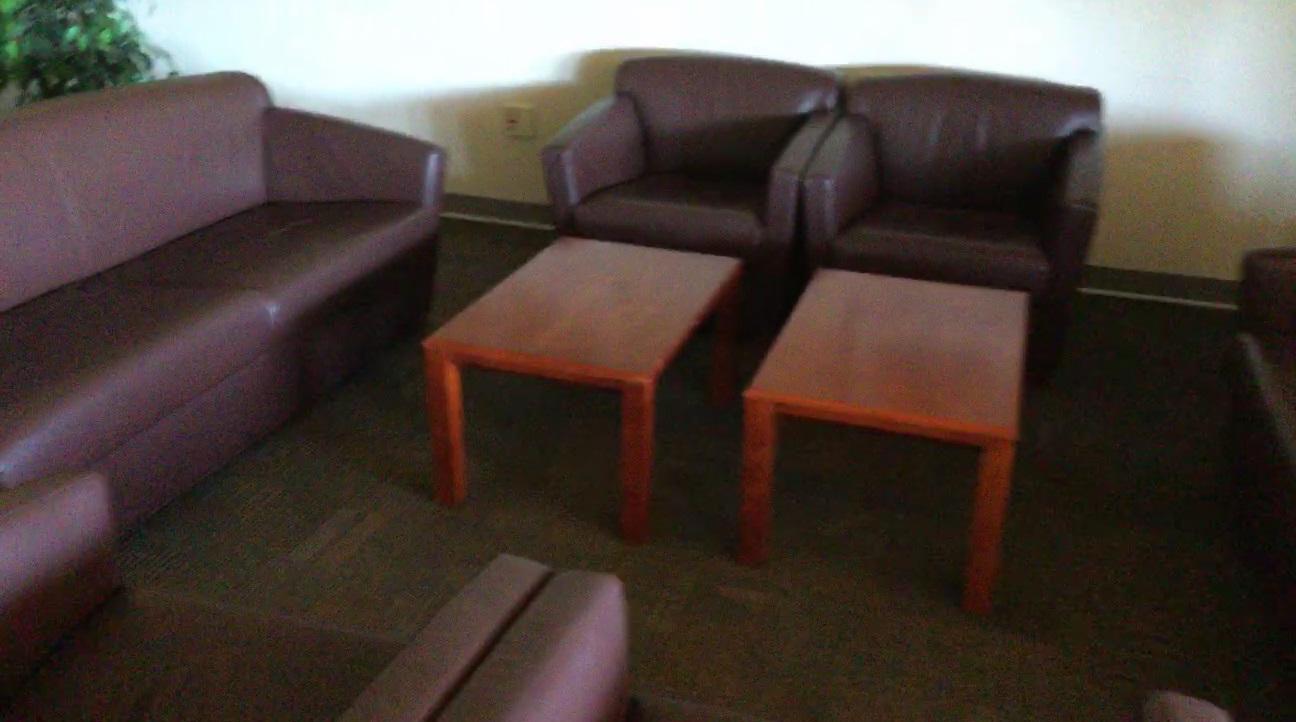}
    \caption{Motivating example. We humans can effortlessly perceive the individual objects and understand the 3D scene from visual inputs, guiding how we interact with it.}
    \label{fig:ch1_moti_human}
\end{figure}

Humans and other animals rely heavily on vision as a primary sensing modality for perceiving the world around them. Fundamentally, the brain makes sense of input from 2D retinal projections of the 3D physical world. These are sparse and incomplete, requiring the use of prior knowledge to infer scene structure and composition, and to recognize objects. As illustrated in Figure \ref{fig:ch1_moti_human}, just by taking a single glance at a sofa, we can imagine what its likely 3D shape is, guiding how we could interact with it such as sitting on it or moving it closer to the table. In fact, we not only focus on the sofa in isolation, but simultaneously perceive the complex scene. For example, we can quickly identify the total number of seats available for our guests and localize the tables where we could serve the tea or coffee.

A long-standing goal in computer vision is to build intelligent systems which have similar capabilities to infer the underlying 3D structure of individual objects as well as understand the composition of multiple objects within complex 3D scenes. These systems would inspire a wide range of applications in robotics and augmented reality (AR). For example, every family is likely to be equipped with an intelligent robot to provide daily services for people in the future. Given a single/few snapshot(s) of the kitchen from a camera or depth scanner, a robot is able to estimate the 3D shape of a mug, where the handle is, and then accurately pour hot coffee without spilling it or overfilling. Being able to understand the complex structure of the living room, the robot can naturally identify and localize all chairs, tables, couches, etc., and smoothly navigate itself to deliver the coffee into our hands. 

However, building these intelligent systems is highly challenging for two fundamental reasons. Firstly, 2D visual projections theoretically have infinite possible 3D geometries, especially when the 2D projections are sparse (\eg{} given a single or only a few views). Secondly, since real-world scenarios are usually a complex composition of objects and structures, many parts of the objects are occluded by one another, resulting in the sparsely observed scenes being incomplete and the individual objects fragmented. Overcoming these challenges requires the system to be able to effectively learn plausible geometric priors from visual inputs.

Early methods to recover the 3D shape of an object mainly leveraged hand-crafted features or explicit priors \cite{Kanade1981,Terzopoulos1988,Mukherjee1995,Thrun2005}. However, these predefined geometric regularities are only applicable to limited shapes and also unable to estimate fine-grained geometric details. Prior attempts towards the more ambitious goal of understanding complex 3D scenes primarily focused on recovering a sparse 3D point cloud to represent the structure of scenes. These systems include the classic structure from motion (SfM) \cite{Agarwal2009, Sweeney2015, Schonberger2016, Ozyesil2017} and simultaneous localization and mapping (SLAM) \cite{Davison2007, Newcombe2011b, Newcombe2011a, Cadena2016, Schops2019} pipelines, but they are unable to recognize and localize individual objects in the 3D space. 

The recent advances in the area of deep neural networks has yielded impressive results for a wide variety of tasks on 2D images, such as object recognition \cite{Krizhevsky2012b}, detection \cite{Everingham2010} and segmentation \cite{He2017a}, thanks in part to the availability of large-scale 2D datasets, \eg{} ImageNet \cite{Russakovsky2015} and COCO \cite{Lin2014}. In essence, these methods consist of multiple processing layers to automatically discover valuable representations from the raw data for classification or detection \cite{LeCun2015a}. Applying this powerful data driven approach to tackle core tasks in 3D space has emerged as a promising direction, especially since the introduction of many large-scale real-world or synthetic datasets for 3D objects and scenes. For example, Wu \etal{} introduce the ModelNet dataset in \cite{Wu2015}, Chang \etal{} present the ShapeNet dataset in \cite{Chang2015} and Koch \etal{} introduce the ABC dataset in \cite{Koch2018}. These are richly-annotated and large-scale repositories of 3D CAD models of objects. In addition to single objects, a variety of 3D indoor scenes with dense geometry and high dynamic range textures are collected in ScanNet \cite{Dai2017}, S3DIS \cite{Armeni2016}, SceneNN \cite{Hua2016}, SceneNet RGB-D \cite{McCormac2017} and Replica \cite{Straub2019}, whereas large-scale 3D outdoor scenes are scanned by LiDAR in Semantic3D \cite{Hackel2017}, SemanticKITTI \cite{Behley2019} and SemanticPOSS \cite{Pan2020}.

However, the ability to unleash the full power of deep neural nets to learn rich 3D representations is still in its infancy, in spite of the availability of large datasets. This is primarily because 3D data are usually high-dimensional, irregular and incomplete. These issues serve as the main motivation of this thesis - to design novel neural networks to address core tasks for 3D perception such as object reconstruction and segmentation.

\section{Research Challenges and Objectives}\label{sec:ch1_challenges}
This thesis aims to design a vision system that is able to understand the geometric structure and semantics of the 3D visual world, from single or multiple scans of common sensors such as a camera, Kinect device or LiDAR. Instead of solving the entire task in one go, we instead approach it from a single object level to a more complex scene level. In particular, this thesis firstly aims to reconstruct the 3D shape of a single object from a sparse number of 2D images, and then focuses on interpreting more complex 3D scenes. However, learning to infer the object-level 3D shape and the scene-level semantics is non-trivial. More specifically, the challenges are three fold as discussed below.
\begin{itemize}
\item \textbf{How to estimate the 3D full shape of an object when there is only a single view.} This is the extreme case where the sensed information is limited and serves as a fundamental proof-of-principle for the use of deep neural networks. The fundamental challenge is how  best to integrate the prior knowledge from the available datasets into the deep network, since the single view itself is unable to recover a full 3D shape.
\item \textbf{How to infer a better 3D shape when there are multiple views available.} Theoretically, given more input images, the 3D shape can be estimated more accurately because more object parts are observed from various angles and perspectives. However, to effectively aggregate the useful information from different views is not easy. 
\item \textbf{How to identify individual objects within complex 3D scenes.} To localize and recognize all 3D objects within a real-world scene is a necessity for understanding the surrounding environment. Nevertheless, 3D scenes such as point clouds are usually visually incomplete and unordered, resulting in the existing neural architectures being ineffective and inefficient.
\end{itemize}

Motivated by these challenges, the thesis aims to achieve the corresponding three primary research objectives. 
\begin{itemize}
\item The first objective is to recover the accurate 3D structure of individual objects from a single view. Particularly, we aim at estimating a dense and full 3D shape of an object from only one depth image acquired by a Kinect scanner. Initially, a single depth view together with camera parameters captures the partial shape of a 3D model. However, most object parts are occluded by the object itself. In order to recover the full shape, the main objective is to learn the prior geometric knowledge of possible object shapes, so as to complete the occluded parts. This objective is achieved in Chapter \ref{chap:rec_obj_sv}.
\item The second objective is to extend the single-view reconstruction method to multi-view scenarios. Traditionally, the SfM and visual SLAM pipelines usually fail when the multiple views are separated by large baselines, because feature registration across views is prone to failure. Ideally, the useful visual features across different input views should be aggregated automatically, steadily improving the estimated 3D shape as supplied with increasing information. This objective is studied in Chapter \ref{chap:rec_obj_mv}.
\item The third objective is to identify all object instances from complex 3D scenes. In particular, given real-world 3D point clouds, obtained from multiple images or LiDAR scans, we aim to precisely recognize and segment all objects at the point level. This objective is studied in Chapter \ref{chap:seg_obj_pc}.
\end{itemize}
\section{Contributions}
In this section, the main contributions of each chapter in this thesis are summarized as follows.

\begin{itemize}
\item In Chapter \ref{chap:rec_obj_sv}, we propose a deep neural architecture based on the generative adversarial network (GAN) to learn a dense 3D shape of an object from a \emph{single} depth view. Compared with existing approaches, our architecture is able to reconstruct a more compelling 3D shape with fine-grained geometric details. In particular, the estimated 3D shape is represented with a high  resolution $256^3$ voxel grid, thanks to our capable encoder-decoder and stable discriminator, outperforming state-of-the-art techniques. Extensive experiments on both synthetic and real-world datasets show the high performance of our approach. The work is published in:

\textit{
\noindent\fbox{\parbox{0.99\linewidth}{
Bo Yang, Hongkai Wen, Sen Wang, Ronald Clark, Andrew Markham and Niki Trigoni. 3D Object Reconstruction from a Single Depth View with Adversarial Learning.
International Conference on Computer Vision Workshops (ICCVW), 2017 \cite{Yang2017b}.
}}}

\textit{
\noindent\fbox{\parbox{0.99\linewidth}{
Bo Yang, Stefano Rosa, Andrew Markham, Niki Trigoni and Hongkai Wen. Dense 3D Object Reconstruction from a Single Depth View.
IEEE Transactions on Pattern Analysis and Machine Intelligence (TPAMI), 2018 \cite{Yang2018}. 
}}}

\item In Chapter \ref{chap:rec_obj_mv}, we propose a new neural module based on an attention mechanism to infer better 3D shapes of objects from multiple views. Compared with existing methods, our approach learns to attentively aggregate useful information from different images. We also introduce a two-stage training algorithm to guarantee the estimated 3D shapes being robust given an arbitrary number of input images. Experiments on multiple datasets demonstrate the superiority of our approach to recover accurate 3D shapes of objects. The work is published in:

\textit{
\noindent\fbox{\parbox{0.99\linewidth}{
Bo Yang, Sen Wang, Andrew Markham and Niki Trigoni. Robust Attentional Aggregation of Deep Feature Sets for Multi-view 3D Reconstruction.
International Journal of Computer Vision (IJCV), 2019 \cite{Yang2020}.
}}}

\item In Chapter 5, we introduce a new framework to identify all individual 3D objects within large-scale 3D scenes. Compared with existing works, our framework is able to directly and simultaneously detect, segment and recognize all object instances, without requiring any heavy pre/post processing steps. We demonstrate significant improvements over baselines on multiple large-scale real-world datasets. The work is published in:

\textit{
\noindent\fbox{\parbox{0.99\linewidth}{
Bo Yang, Jianan Wang, Ronald Clark, Qingyong Hu, Sen Wang, Andrew Markham and Niki Trigoni. Learning Object Bounding Boxes for 3D Instance Segmentation on Point Clouds.
Advances in Neural Information Processing Systems (NeurIPS Spotlight), 2019 \cite{Yang2019d}.
}}}
\end{itemize}

In addition to the above contributions which address the research challenges discussed in Section \ref{sec:ch1_challenges}, I have also made contributions to the following coauthored publications, which are related to my thesis, but will not be included. 

\begin{itemize}
    \item Bo Yang*, Zihang Lai*, Xiaoxuan Lu, Shuyu Lin, Hongkai Wen, Andrew Markham and Niki Trigoni. Learning 3D Scene Semantics and Structure from a Single Depth Image. Computer Vision and Pattern Recognition Workshops (CVPRW), 2018 \cite{Yang2018b}. 
    \item Zhihua Wang, Stefano Rosa, Linhai Xie, Bo Yang, Niki Trigoni and Andrew Markham. \textit{Defo-Net: Learning body deformation using generative adversarial networks}. In International Conference on Robotics and Automation (ICRA), 2018 \cite{Wang2018a}.
    \item Zhihua Wang, Stefano Rosa, Bo Yang, Sen Wang, Niki Trigoni and Andrew Markham. \textit{3D-PhysNet: Learning the intuitive physics of non-rigid object deformations}. In International Joint Conference on Artificial Intelligence (IJCAI), 2018 \cite{Wang2018l}.
    \item Shuyu Lin, Bo Yang, Robert Birke and Ronald Clark. \textit{Learning Semantically Meaningful Embeddings Using Linear Constraints}. In Computer Vision and Pattern Recognition Workshops (CVPRW), 2019 \cite{Lin2018c}.
    \item Wei Wang, Muhamad Risqi U Saputra, Peijun Zhao, Pedro Gusmao, Bo Yang, Changhao Chen, Andrew Markham and Niki Trigoni. \textit{DeepPCO: End-to-End Point Cloud Odometry through Deep Parallel Neural Network}. In International Conference on Robotics and Automation (ICRA),2019 \cite{Wang2019f}.
    \item Qingyong Hu, Bo Yang*, Linhai Xie, Stefano Rosa, Yulan Guo, Zhihua Wang, Niki Trigoni and Andrew Markham. \textit{RandLA-Net: Efficient Semantic Segmentation of Large-Scale Point Clouds}. Computer Vision and Pattern Recognition (CVPR), 2020 \cite{Hu2020}. 
\end{itemize}
\section{Thesis Structure}
The remainder of this thesis is organised as follows: 
\begin{itemize}
\item Chapter \ref{chap:all_review} presents a comprehensive review of 3D perception of objects and scenes.
\item Chapter \ref{chap:rec_obj_sv} introduces our \nicknameRecGAN{} approach, a generative adversarial network based method that predicts a dense 3D shape of an object from a single view.
\item Chapter \ref{chap:rec_obj_mv} presents our \nicknameAttSets{} module and \nicknameFASet{} algorithm, an attention based pipeline which steadily improves the estimated 3D shapes given more input views.
\item Chapter \ref{chap:seg_obj_pc} presents our \nicknameBoNet{} framework that simultaneously recognizes, detects and segments all individual 3D objects in a complex scene.
\item The last Chapter \ref{chap:conclusion} concludes the entire thesis and identifies a number of future directions.
\end{itemize}

\chapter{Literature Review}
\label{chap:all_review}
In this chapter, I discuss previous work related to 3D object reconstruction and segmentation. In particular, I will start reviewing existing efforts on 3D shape recovery from single and multiple views in Sections \ref{sec:ch2_liter_sv} and \ref{sec:ch2_liter_mv}, followed by deep neural algorithms designed for 3D point clouds in Section \ref{sec:ch2_liter_pc}. Since the framework presented in Chapter 3 utilizes generative adversarial networks and the proposed neural module in Chapter 4 involves the attention mechanism and deep learning on sets, I therefore review generative adversarial frameworks in Section \ref{sec:ch2_liter_gen}, attention mechanisms in Section \ref{sec:ch2_liter_att} and deep neural networks for sets in Section \ref{sec:ch2_liter_sets}. Lastly, Section \ref{sec:ch2_liter_relation} clarifies the relation and novelty of this thesis with regards to previous work. 
\section{Single View 3D Object Reconstruction} \label{sec:ch2_liter_sv}
In this section, different pipelines for single-view 3D reconstruction or shape completion are reviewed. Both conventional geometry based techniques and state-of-the-art deep learning approaches are covered.

\textbf{(1) 3D Model/Shape Completion.} 
Monszpart \etal{} use plane fitting to complete small missing regions in \cite{Monszpart2015a}, while shape symmetry is applied in \cite{Mitra2006,Pauly2008,Sipiran2014,Speciale2016,Thrun2005} to fill in voids. Although these methods show good results, relying on predefined geometric regularities fundamentally limits the structure space to hand-crafted shapes. Besides, these approaches are likely to fail when the missing or occluded regions are relatively large. Another similar fitting pipeline is to leverage database priors. Given a partial shape input, an identical or most likely 3D model is retrieved and aligned with the partial scan \cite{Kim2012,Li2015a,Nan,Shao2012,Shi2016,Rock2015}. However, these approaches explicitly assume the database contains identical or very similar shapes, thus being unable to generalize to novel objects or categories.

\textbf{(2) Single RGB Image Reconstruction.}
Predicting a complete 3D object model from a single view is a long-standing and extremely challenging task. When reconstructing a specific object category, model templates can be used. For example, morphable 3D models are exploited for face recovery \cite{Blanz2003,Dou2017}. This concept was extended to reconstruct simple objects in \cite{Kar2015a}. For general and complex object reconstruction from a single RGB image, recent works \cite{Gwak2017,Tulsiani2017,Yan2016} aim to infer 3D shapes using multiple RGB images for weak supervision. Shape prior knowledge is utilized in \cite{Kong2017,Kurenkov2017,Murthy2016} for shape estimation. To recover high resolution 3D shapes, Octree representation is introduced in \cite{Tatarchenko2017,Riegler2017,Christian2017} to reduce computational burden, while an inverse discrete cosine transform (IDCT) technique is proposed in \cite{Johnston2017} along similar lines. Lin \etal{} \cite{Lin2017a} designed a pseudo-renderer to predict dense 3D shapes, whilst 2.5D sketches and dense 3D shapes are sequentially estimated from a single RGB image in \cite{Wu2017d}. 

\textbf{(3) Single Depth View Reconstruction.} 
The task of reconstruction from a single depth view is to complete the 3D structure, where visible parts occlude other parts of the object. 3D ShapeNets \cite{Wu2015} is amongst some of  the earliest work using deep neural nets to estimate 3D shapes from a single depth view. Firman \etal{} \cite{Firman2016} trained a random decision forest to infer unknown voxels. Originally designed for shape denoising, VConv-DAE \cite{B2016e} can also be used for shape completion. To facilitate robotic grasping, Varley \etal{} proposed a neural network to infer the full 3D shape from a single depth view in \cite{Varley2017}. However, all these approaches are only able to generate low resolution voxel grids which are less than $40^3$ and unlikely to capture fine geometric details. Subsequent works \cite{Dai2017b,Song2017,Han2017,Wang2017b} can infer higher resolution 3D shapes. However, the pipeline in \cite{Dai2017b} relies on a shape database to synthesize a higher resolution shape after learning a small $32^3$ voxel grid from a depth view, while SSCNet \cite{Song2017} requires voxel-level annotations for supervised scene completion and semantic label prediction.
Both \cite{Han2017} and \cite{Wang2017b} were originally designed for shape inpainting instead of directly reconstructing the complete 3D structure from a partial depth view. 

\textbf{(4) Different 3D Shape Representations.}
The works discussed above usually aim to reconstruct a 3D voxel grid to represent the object shape. Being concurrent to these approaches, the proposed neural algorithm in Chapter \ref{chap:rec_obj_sv} is also based on voxel grids. Since 3D voxels are memory inefficient, more recent pipelines tend to learn point clouds, meshes, implicit surfaces and other intermediate representations for 3D shape reconstruction. In particular, PointSet \cite{Fan2017} is amongst the first works to learn a \textbf{point-based} 3D shape from a single image. A number of recent works further improve its performance using adversarial learning \cite{Jiang2018}, shape priors \cite{Li2019b} and Silhouettes \cite{Zou2020}, and some other works aim to recover denser point clouds as in \cite{Lin2017a, Lu2019, Mandikal2019, Peng2020, Liu2020}. A number of unsupervised/weakly-supervised frameworks \cite{Insafutdinov2018, Cha2019, L2019} are also proposed for point-based 3D reconstruction. To recover a \textbf{3D mesh} from a single image, early works learn to deform a template mesh in \cite{Kato2017,Wang2018f,Wen2019,Tang2019, Liu2019e, Pan2019,Groueix2018,Deprelle2019}, while a number of recent works learn to generate polygon meshes directly from images in \cite{Chen2020, Nash2020}. Instead of recovering explicit 3D shapes, a number of recent works learn \textbf{implicit surfaces} in \cite{Park2019,Mescheder2019,Chen2019g}. This pipeline is further improved by \cite{Liu2019b,Niemeyer2019,Sitzmann2019} in which the 3D supervision is no longer required. Some other \textbf{intermediate representations}, such as shape primitives \cite{Zou2017} and multi-layer depth \cite{Shin2019,Nicastro2019}, are also studied.
\section{Multi-view 3D Object Reconstruction}\label{sec:ch2_liter_mv}
In this section, both the classical SfM/SLAM pipelines and learning based approaches are reviewed, with a stronger focus towards recent work in learning based methods, for the purpose of multi-view 3D object recontruction.

\textbf{(1) Traditional SfM/SLAM.}
To estimate the underlying 3D shape from multiple images, classic SfM \cite{Ozyesil2017} and SLAM \cite{Cadena2016} algorithms firstly extract and match hand-crafted geometric features \cite{Hartley2004} and then apply bundle adjustment \cite{Triggs2000} for both shape and camera motion estimation. Existing SfM strategies include incremental \cite{Agarwal2009, Frahm2010,Wu2013,Schonberger2016}, hierarchical \cite{Gherardi2010}, and global approaches \cite{Crandall2011,Sweeney2015}. Classic SLAM systems usually consist of feature-based \cite{Davison2007,Mur-Artal2015,Mur-Artal2016,Dai2017a} and direct approaches \cite{Newcombe2011b,Newcombe2011a,Henry2012,Salas-Moreno2013,Delaunoy2014,Goldlucke2014,Slavcheva2016,Schops2019}. These systems are recently integrated with deep neural networks in \cite{Bloesch2018,McCormac2018,Czarnowski2019,Xu2019c,Zhi2019}. Although they can reconstruct visually satisfactory 3D models, the recovered shapes are usually sparse point clouds and the occluded regions are unable to be estimated.

\textbf{(2) Learning to Integrate Multi-views.}
Recent deep neural net based approaches tend to recover dense 3D shapes through learnt features from multiple color images and achieve compelling results. To fuse the deep features from multiple images, 3D-R2N2 \cite{Chan2016}, LSM \cite{Kar2017} and 3D2SeqViews \cite{Han2019b} apply the recurrent unit GRU, resulting in the networks being permutation variant and inefficient for aggregating a long sequence of images. SilNet \cite{Wiles2017,Wiles2018a}, DeepMVS \cite{Huang2018} and 3DensiNet \cite{Wang2017i} simply use max pooling to preserve the first order information of multiple images, while RayNet \cite{Paschalidou2018}  and \cite{Sridhar2019} apply average pooling to retain the first moment information of multiple deep features. MVSNet \cite{Yao2018} proposes a variance-based approach to capture the second moment information for multiple feature aggregation. These pooling techniques only capture partial information, ignoring the majority of the deep features. In addition, the geometric consistency is not explicitly considered. To overcome this,  recent works \cite{Lin2019,Wen2019,Ghosh2020,Chen2019e,Xie2019} learn multi-view stereo by applying the multi-view consistency or taking the depth prior into account.

Object shapes can also be recovered from multiple depth scans - the traditional volumetric fusion method \cite{Curless1996,Cao2018} integrates multiple viewpoint information by averaging truncated signed distance functions (TSDF). The recent learning based OctNetFusion \cite{Riegler2017} also adopts a similar strategy to integrate multiple depth information. However, this integration might result in information loss since TSDF values are averaged \cite{Riegler2017}. PSDF \cite{Dong2018} was recently proposed to learn a probabilistic distribution through Bayesian updating in order to fuse multiple depth images, but it is not straightforward to include the module into existing encoder-decoder networks.

\textbf{(3) Learning Two-view Stereos.}
SurfaceNet \cite{Ji2017b}, SuperPixel Soup \cite{Kumar2017} and Stereo2Voxel \cite{Xie2019a} learn to reconstruct 3D shapes from two images. Although demonstrating the viability of recovering the 3D models, they are unable to process an arbitrary number of input images.
\section{Segmentation for 3D Point Clouds}\label{sec:ch2_liter_pc}

To extract features from 3D point clouds, traditional approaches usually crafted features manually \cite{Chua1997,Rusu2009a,Thomas2018a}. However, these features are unable to adapt to more complex shapes and scenes. Recent learning based approaches mainly include projection-based, voxel-based \cite{Rusu2009a,Huang2016a,Tchapmi2017,Riegler2017a,Lea,Rethage2018,Liu2017d,Meng2019} and point-based schemes \cite{Qi2016,Klokov2017,Hermosilla2018,Hua2018,Su2018}, which are widely employed for the core tasks of 3D point cloud perception, such as object recognition, semantic segmentation, object detection and instance segmentation. Basically, these tasks are similar to that of 2D images. In particular, the task of object recognition aims to estimate the category of a small set of 3D points, whereas the semantic segmentation aims to predict the category of each 3D point of a large-scale point cloud. More than simply classifying individual points, both the tasks of object detection and instance segmentation seek to localize each object, but the object detection only infers a 3D bounding box for an object whereas the instance segmentation needs to precisely identify an object instance that each 3D point belongs to.

\textbf{(1) 3D Semantic Segmentation.}
Point clouds can be voxelized into 3D grids and then powerful 3D CNNs are applied as in \cite{Graham2018, Lea, Choy2019, Meng2019, Chen2019f}. Although they achieve leading results on semantic segmentation, their primary limitation is the heavy computation cost, especially when processing large-scale point clouds.

The recent point-based method PointNet \cite{Qi2016} shows leading results on classification and semantic segmentation, but it does not capture context features. To overcome this, many recent works introduced sophisticated neural modules to learn per-point local features. These modules can be generally classified as 1) neighbouring feature pooling \cite{Qi2017, Li2018,Zhao2019a, Zhang2019c, Duan2019}, 2) graph message passing \cite{Wang2018c,Shen2017a,Wang2018e, Wang2019b, Chen2019c, Jiang2019a, Liu2019f, Te2018,Xu2020,Wang2019e}, 3) kernel-based convolution 
\cite{Su2018, Xu2018a,Li2018f, Hua2018, Wu2019, Lei2019, Komarichev2019, Lan2019, Thomas2019, Mao2019,Rao2019,Liu2019c,Boulch2020,He2020,Lin2020}, 4) attention-based aggregation \cite{Liu2018a, Zhang2019, Yang2019b, Paigwar2019,Wang2019e} and 5) recurrent-based learning \cite{Liu2018g,Huang2018a,Ye2018,Wu2019a}.

To further enable the networks to consume large-scale point clouds, the multi-scale methods \cite{Engelmann2017,Guerrero2018} and graph-based
SPG \cite{Landrieu2018} are introduced to preprocess the large point clouds to learn per super-point semantics. The recent FCPN \cite{Rethage2018} and PCT \cite{Chen2019} apply both voxel-based and point-based networks to process the massive point clouds. Although achieving promising results, these approaches also require extremely high computation and memory resources. To overcome this, the recent RandLA-Net \cite{Hu2020} is able to efficiently and effectively process large-scale point clouds by introducing a novel local feature aggregation module together with an efficient random point feature down-sampling strategy.

\textbf{(2) 3D Object Detection.}
 The most common way to detect objects in 3D point clouds is to project points onto 2D images to regress bounding boxes \cite{Li2016f,Asvadi2017,Vaquero2017,Chen2017b,Ali2018,Yang2018d,Zeng2018,Wu2017e,Beltran2018,Minemura2018,Simon2018,Hu2020a}, by using existing 2D detectors. Detection performance is further improved by fusing RGB images in \cite{Chen2017b,Xu2018,Ku2018,Qi2018,Wang2018b,Meyer2019,Rashed2019, Qi2020}, which require the 2D images to be well aligned with the 3D point clouds within the field of view. 
 Point clouds can be also divided into voxels for object detection \cite{Engelcke2016,Li2017j,Zhou2018a,Yang2019c,Chen2019f,Lang2019,Chen2019h}. The detection strategies usually follow the mature frameworks for object detection in 2D images. However, most of these approaches rely on predefined anchors and the two-stage region proposal network \cite{Ren2015a}. It is inefficient to extend them to 3D point clouds. Without relying on anchors, the recent PointRCNN \cite{Shi2019} learns to detect via foreground point segmentation, and VoteNet \cite{Qi2019} detects objects via point feature grouping, sampling and voting. 
 
\textbf{(3) 3D Instance Segmentation.}
SGPN \cite{Wang2018d} is the first neural algorithm to segment instances on 3D point clouds by grouping the point-level embeddings. The subsequent ASIS \cite{Wang2019}, JSIS3D \cite{Pham2019}, MASC \cite{Liu2019}, 3D-BEVIS \cite{Elich2019}, MTML \cite{Lahoud2019}, JSNet \cite{Zhao2020}, MPNet \cite{He2020a} and \cite{Liang2019, Arase2019} use the same strategy to group point-level features for instance segmentation. Mo \etal{} introduce a segmentation algorithm in PartNet \cite{Mo2019} by classifying point features. However, the learnt segments of these proposal-free methods do not have high objectness as they do not explicitly detect object boundaries. In addition, these methods usually require a post-processing algorithm, \eg{} mean shift \cite{Comaniciu2002}, to cluster the learnt per-point features to obtain the final object instance labels, thereby resulting in an extremely heavy computation burden.

Another set of approaches to learn the object instances from 3D point clouds are proposal-based methods. By drawing on the successful 2D RPN \cite{Ren2015a} and RoI \cite{He2017a}, GSPN \cite{Yi2019} and 3D-SIS \cite{Hou2019} learn a large number of candidate object bounding boxes followed by per-point mask prediction. However, these approaches usually rely on two-stage training and a post-processing step for dense proposal pruning.
\section{Generative Adversarial Networks} \label{sec:ch2_liter_gen}
Generative Adversarial Networks (GANs) \cite{Goodfellow2014} are a novel framework to model complex, real-world data distributions. Inspired by game theory, GANs consist of a generator and a discriminator, which compete with each other. By transforming a source data distribution, the generator learns to synthesize a new data distribution to mimic the target distribution, while the discriminator learns to distinguish between the synthesized and the real target samples. GANs have achieved impressive success in  image generation \cite{Radford2016, Karras2017}, natural language \cite{Yu2017c}, and time-series synthesis \cite{Donahue2018}.

GANs can be extended to a conditional model if either the generator or the discriminator is conditioned on some extra data distribution \cite{Mirza2014}. Based on conditional GANs, images can be generated conditioned on class labels \cite{Odena2017}, text \cite{Reed2016}, and other information \cite{Reed2016a}. Conditional GANs are also used for photo-realistic image synthesis \cite{Zhang2017e}, image super-resolution \cite{Ledig2016}, and  image translation between domains \cite{Zhu2017b}.

GANs and conditional GANs are also applied in \cite{Gadelha2016, Smith2017,Soltani2017,Wu2016a} to generate low resolution 3D structures from noise or images. However, incorporating generative adversarial learning to estimate high resolution 3D shapes is not straightforward, as it is difficult to generate samples for high dimensional and complex data distributions. Fundamentally, this is because GANs are notoriously hard to train, suffering from instability issues \cite{Arjovsky2017a,Arjovsky2017c}.

\section{Attention Mechanisms}\label{sec:ch2_liter_att}
The attention mechanism was originally proposed for natural language processing in \citep{Bahdanau2015}. In a nutshell, it learns to weight deep features by importance scores for a specific task, and then uses this weighting mechanism to improve that task. Compared with the traditional encoder-decoder RNN models, the attention-based approach is able to learn a more complicated dependency that ranges across a long input sequence. This dependency is not only important for the sequential domain of language processing, but also the spatial domain of many visual tasks. Being coupled with RNNs, the attention mechanism achieves compelling results in neural machine translation \citep{Bahdanau2015}, image captioning \citep{Xu2015b}, image question answering \citep{Yang2016}, \etc. However, all these RNN-based attention approaches are permutation variant to the order of input sequences. In addition, they are computationally time-consuming when the input sequence is long due to the recurrent processing.

Dispensing with recurrence and convolutions entirely and solely relying on attention mechanism, Transformer \citep{Vaswani2017} achieves superior performance in machine translation tasks. Similarly, being decoupled from RNNs, attention mechanisms are also applied for visual recognition \citep{JieHu2018,Rodriguez2018,Liu2018e,Sarafianos2018,Zhu2018a,Nakka2018,Girdhar2017a}, semantic segmentation \citep{Li2018a}, long sequence learning \citep{Raffel2016}, multi-task learning \cite{Liu2019g}, and image generation \citep{Zhang2018b}. Although the above decoupled attention modules can be used to aggregate variable sized deep feature sets, they are literally designed to operate on fixed sized features for tasks such as image recognition and generation. The robustness of attention modules in the context of dynamic deep feature sets has not been investigated yet.
\section{Deep Learning on Sets}\label{sec:ch2_liter_sets}
In contrast to traditional approaches operating on fixed dimensional vectors or matrices, deep learning tasks defined on sets usually require learning functions to be permutation invariant and able to process an arbitrary number of elements in a set.

Zaheer \etal{} introduce general permutation invariant and equivariant models in \cite{Zaheer2017}, and they end up with a \textbf{sum pooling} for permutation invariant tasks such as population statistics estimation and point cloud classification. In the recent GQN \cite{Eslami2018}, sum pooling is also used to aggregate an arbitrary number of orderless images for 3D scene representation. Gardner \etal{} \cite{Gardner2017a} use \textbf{average pooling} to integrate an unordered deep feature set for a classification task. Su \etal{} \cite{Su2015} use \textbf{max pooling} to fuse the deep feature set of multiple views for 3D shape recognition. Similarly, PointNet \cite{Qi2016} also uses max pooling to aggregate the set of features learnt from point clouds for 3D classification and segmentation. In addition, the higher-order statistics based  pooling approaches are widely used for 3D object recognition from multiple images. Vanilla \textbf{bilinear pooling} is applied for fine-grained recognition in \cite{Lin2015} and is further improved in \cite{Lin2017b}. Concurrently, \textbf{log-covariance pooling} is proposed in \cite{Ionescu2015}, and is recently generalized by \textbf{harmonized bilinear pooling} in \cite{Yu2018a}. Bilinear pooling techniques are further improved in recent work \cite{Yu2018b,Lin2018}. However, both first-order and higher-order pooling operations ignore a majority of the information of a set. In addition, the first-order poolings do not have trainable parameters, while the higher-order poolings have only a few parameters available for the network to learn. These limitations lead to the pooling based neural networks to be optimized with regards to the specific statistics of data batches during training, and therefore unable to be robust and generalize well to variable sized deep feature sets during testing. 
\section{Novelty with Respect to State of the Art}\label{sec:ch2_liter_relation}
This section hightlights the novel aspects of this thesis compared with state of the art in the context of single/multi view 3D reconstruction and segmentation of 3D point clouds.

\textbf{(1) Single View 3D Reconstruction.} 
To recover the 3D shape of an object from a single depth view, the 3D-RecGAN++ is proposed in Chapter \ref{chap:rec_obj_sv}. The neural architecture consists of a generator which synthesizes a 3D shape conditioned on an input partial depth view, and a discriminator to distinguish whether the input 3D shape is synthesized or real. The overall pipeline extends from conditional GANs \cite{Mirza2014} as discussed in Section \ref{sec:ch2_liter_gen}. However the existing adversarial loss functions, such as the original GAN \cite{Goodfellow2014}, WGAN \cite{Arjovsky2017a} and WGAN-GP \cite{Arjovsky2017c}, are unable to converge to synthesize a high dimensional 3D shape. To overcome this, the proposed 3D-RecGAN++ introduces a mean feature layer for the discriminator to stabilize the entire framework. 

There are many other neural algorithms to estimate the 3D shape from a single view as discussed in Section \ref{sec:ch2_liter_sv}. The most similar works to 3D-RecGAN++ are 3D-EPN \cite{Dai2017b}, 3D-GAN \cite{Wu2016a}, Varley \etal{} \cite{Varley2017} and Han \etal{} \cite{Han2017}. However, all of them are only able to generate a low resolution 3D voxel grid, less than $128^3$, to represent the shape of an object. By contrast, the proposed 3D-RecGAN++ directly generates a 3D shape within a $256^3$ voxel grid, which is able to recover fine-grained geometric details. Since 3D-RecGAN++ uses a voxel grid to represent 3D shape, its  memory consumption is generally less efficient than the latest approaches (published after 3D-RecGAN++) based on point clouds, meshes and implicit surfaces as discussed in Section \ref{sec:ch2_liter_sv}. 

\textbf{(2) Multi-view 3D Reconstruction.}
In Chapter \ref{chap:rec_obj_mv}, we propose an AttSets module together with a FASet algorithm to integrate a variable number of views for more accurate shape estimation. The AttSets module extends the general idea of attention mechanism discussed in Section \ref{sec:ch2_liter_att}. Similar works include Transformer \cite{Vaswani2017}, SENet \cite{JieHu2018} and \cite{Raffel2016,Girdhar2017a}. However, the existing attention mechanisms are only able to be applied to a fixed number of input elements. To integrate an arbitrary number of input images, we formulate this multi-view 3D reconstruction as an aggregation process and propose AttSets with a series of carefully designed neural functions. Fundamentally, our AttSets involves deep learning techniques for sets. In this regard, the recent works \cite{Zaheer2017,Lee2019a,Wagstaff2019,Ilse2018} are similar to AttSets. However, these existing approaches neglect an important issue. In particular, the existing neural algorithms are not robust to a variable number of input elements, and their performance drops if the cardinality of testing sets is significantly different from that of training sets. To overcome this, we further propose the FASet algorithm to separately optimize the AttSets module and the base feature extractor, guaranteeing the robustness of the entire network with regards to a variable number of input images.

As discussed in Section \ref{sec:ch2_liter_mv}, the common way to fuse multiple views for 3D shape estimation leverages RNNs \cite{Chan2016,Kar2017} or heuristic poolings such as max/mean/sum poolings \cite{Wiles2017, Huang2018}. However, the RNN approaches formulate the multiple views as an ordered sequence and the reconstructed shape varies given a different order of the same image set. The heuristic poolings usually discard the majority of the information, thereby being unable to obtain better 3D shapes even if more images are given. By contrast, our AttSets and FASet are able to attentively aggregate useful information from an arbitrary number of views and guarantee the final recovered shape is robust to the number of input images. 

\textbf{(3) Segmentation of 3D Point Clouds.}
Beyond recovering the 3D shape of a single object, we aim to identify all individual 3D objects from large-scale real-world point clouds in Chapter \ref{chap:seg_obj_pc}. The proposed 3D-BoNet is a general framework to recognize, detect and segment all object instances simultaneously. The backbone of the framework extends the basic idea of shared MLPs invented by PointNet/PointNet++ \cite{Qi2016,Qi2017}. 

To recognize 3D objects, 3D-BoNet simply leverages any of the front-ends of existing point-based networks such as PointNet++ \cite{Qi2017} and SparseConv \cite{Graham2018}. To detect and segment individual objects, existing works either follow the idea of RPN \cite{Ren2015a} to extensively localize objects in 3D space, or learn to directly cluster per-point features as discussed in Section \ref{sec:ch2_liter_pc}. However, these methods have a number of limitations. First, they usually require post-processing steps such as non-maximum suppression or mean shift clustering, which are extremely computationally heavy. Second, the learnt instances do not have high objectness as they do not explicitly learn object boundaries and the low-level point features are very likely to be incorrectly clustered. To overcome these shortcomings, our 3D-BoNet framework offers the following unique aspects: 1) the object bonding boxes are directly learnt from the global features of a point cloud via a carefully designed neural optimal association layer, guaranteeing high objectness of all detected instances; 2) each object is further precisely segmented within a bounding box via a simple binary classifier, without requiring any post-processing steps. The aspects above make 3D-BoNet simpler and more efficient than existing competing pipelines.

\chapter{Learning to Reconstruct 3D Object from a Single View}
\label{chap:rec_obj_sv}
\section{Introduction}
To reconstruct the complete and precise 3D geometry of an object is essential for many graphics and robotics applications, from augmented reality (AR)/virtual reality (VR) \cite{B2016e} and semantic understanding, to object deformation \cite{Wang2018a}, robot grasping \cite{Varley2017} and obstacle avoidance. Classic approaches use off-the-shelf low-cost depth sensing devices such as Kinect and RealSense cameras to recover the 3D shape of an object from captured depth images. These approaches typically require multiple depth images from different viewing angles of an object to estimate the complete 3D structure \cite{Newcombe2011a}\cite{Niener2013}\cite{Steinbr2013}. However, in practice it is not always feasible to scan all surfaces of an object before reconstruction, which leads to incomplete 3D shapes with occluded regions and large holes. In addition, acquiring and processing multiple depth views require more computing power, which is not ideal in many applications that require real-time performance.

We aim to tackle the problem of estimating the complete 3D structure of an object using a single depth view. This is a very challenging task, since the partial observation of the object (\ie{} a depth image from one viewing angle) can be theoretically associated with an infinite number of possible 3D models. Traditional reconstruction approaches typically use interpolation techniques such as plane fitting, Laplacian hole filling \cite{Nealen2006}\cite{Zhao2007}, or Poisson surface estimation \cite{Kazhdan2006}\cite{Kazhdan2013} to infer the underlying 3D structure. However, they can only recover very limited occluded or missing regions, \eg{} small holes or gaps due to quantization artifacts, sensor noise and insufficient geometry information.

Interestingly, humans are surprisingly good at solving such ambiguity by implicitly leveraging prior knowledge. For example, given a view of a chair with two rear legs occluded by front legs, humans are easily able to guess the most likely shape behind the visible parts. Recent advances in deep neural networks and data driven approaches show promising results in dealing with such a task.

In this chapter, we aim to acquire the complete and high-resolution 3D shape of an object given a single depth view. By leveraging the high performance of 3D convolutional neural nets and large open datasets of 3D models, our approach learns a smooth function that maps a 2.5D view to a complete and dense 3D shape. In particular, we train an end-to-end model which estimates full volumetric occupancy from a single 2.5D depth view of an object.

While state-of-the-art deep learning approaches \cite{Wu2015}\cite{Dai2017b}\cite{Varley2017} for 3D shape reconstruction from a single depth view achieve encouraging results, they are limited to very small resolutions, typically at the scale of $32^3$ voxel grids. As a result, the learnt 3D structure tends to be coarse and inaccurate. In order to generate higher resolution 3D objects with efficient computation, Octree representation has been recently introduced in \cite{Tatarchenko2017}\cite{Riegler2017}\cite{Christian2017}. However, increasing the density of output 3D shapes would also inevitably pose a great challenge to learn the geometric details for high resolution 3D structures, which has yet to be explored.

Recently, deep generative models achieve impressive success in modeling complex high-dimensional data distributions, among which Generative Adversarial Networks (GANs) \cite{Goodfellow2014} and Variational Autoencoders (VAEs) \cite{Kingma2014} emerge as two powerful frameworks for generative learning, including image and text generation \cite{Hu2017d}\cite{Karras2017}, and latent space learning \cite{Chen2016a}\cite{Kulkarni2015}. In the past few years, a number of works \cite{B2016f}\cite{Girdhar}\cite{Huang2015a}\cite{Wu2016a} applied such generative models to learn latent space to represent 3D object shapes, in order to solve tasks such as new image generation, object classification, recognition and shape retrieval.

In this chapter, we propose \nicknameRecGAN{}, a simple yet effective model that combines a skip-connected 3D encoder-decoder with adversarial learning to generate a complete and fine-grained 3D structure conditioned on a single 2.5D view. In particular, our model firstly encodes the 2.5D view to a compressed latent representation which implicitly represents general 3D geometric structures, then decodes it back to the most likely full 3D shape. Skip-connections are applied between the encoder and decoder to preserve high frequency information. The rough 3D shape is then fed into a conditional discriminator which is adversarially trained to distinguish whether the coarse 3D structure is plausible or not. The encoder-decoder is able to approximate the corresponding shape, while the adversarial training tends to add fine details to the estimated shape. To ensure the final generated 3D shape corresponds to the input single partial 2.5D view, adversarial training of our model is based on a conditional GAN \cite{Mirza2014} instead of random guessing. The above network excels the competing approaches \cite{Varley2017}\cite{Dai2017b}\cite{Han2017}, which either use a single fully connected layer \cite{Varley2017}, a low capacity decoder without adversarial learning \cite{Dai2017b}, or the multi-stage and ineffective LSTMs \cite{Han2017} to estimate the full 3D shapes.

Our contributions are as follows:
\begin{itemize}
    \item We propose a simple yet effective model to reconstruct the complete and accurate 3D structure using a single arbitrary depth view. Particularly, our model takes a simple occupancy grid map as input without requiring object class labels or any annotations, while predicting a compelling shape within a high resolution of \textbf{ $256^3$ } voxel grid. By drawing on both 3D encoder-decoder and adversarial learning, our approach is end-to-end trainable with high level of generality.
    \item We exploit conditional adversarial training to refine the 3D shape estimated by the encoder-decoder. Our contribution here is that we use the mean value of a latent vector feature, instead of a single scalar, as the output of the discriminator to stabilize GAN training.
    \item  We conduct extensive experiments for single category and multi-category object reconstruction, outperforming the state of the art. Importantly, our approach is also able to generalize to previously unseen object categories. Lastly, our model also is shown to perform robustly on real-world data, after being trained purely on synthetic datasets.
    \item To the best of our knowledge, there are no good open datasets which have the ground truth for occluded/missing parts and holes for each 2.5D view in real-world scenarios. We therefore collect and release our real-world testing dataset to the community.
\end{itemize}

Our code and data are available at:\textit{ https://github.com/Yang7879/3D-RecGAN-extended}
\section{Method Overview}\label{sec:ch3_overview}
Our method aims to estimate a complete and dense 3D structure of an object, which only takes an arbitrary single 2.5D depth view as input. The output 3D shape is automatically aligned with the corresponding 2.5D partial view. To achieve this task, each object model is represented by a high resolution 3D voxel grid. We use the simple occupancy grid for shape encoding, where $1$ represents an occupied cell and $0$ an empty cell. Specifically, the input 2.5D partial view, denoted as $\boldsymbol{x}$, is a $64^3$ occupancy grid, while the output 3D shape, denoted as $\boldsymbol{y}$, is a high resolution $256^3$ probabilistic voxel grid. The input partial shape is directly calculated from a single depth image given camera parameters. We use the ground truth dense 3D shape with aligned orientation as same as the input partial 2.5D depth view to supervise our network.

\begin{figure}[t]
    \centering
    \includegraphics[width=0.8\textwidth]{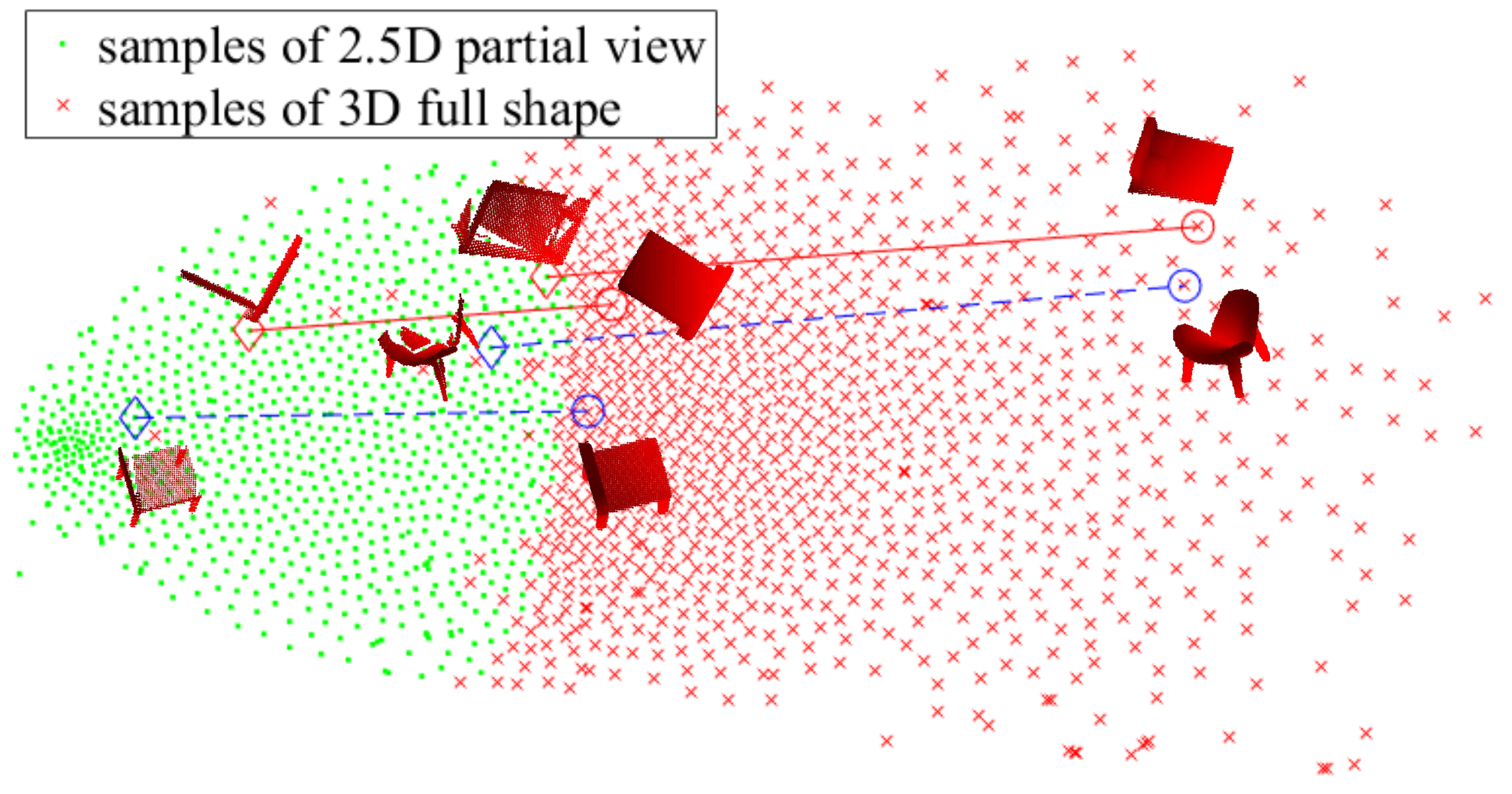}
    \caption{t-SNE embeddings of 2.5D partial views and 3D complete shapes of multiple object categories.}
    \label{fig:ch3_25d_3d}
\end{figure}

To generate ground truth training and evaluation pairs, we virtually scan 3D objects from ShapeNet \cite{Chang2015}. Figure \ref{fig:ch3_25d_3d} is the t-SNE visualization \cite{Maaten2008} of partial 2.5D views and the corresponding full 3D shapes for multiple general chair and bed models. Each green dot represents the t-SNE embedding of a 2.5D view, whilst a red dot is the embedding of the corresponding 3D shape. It can be seen that multiple categories inherently have similar 2.5D to 3D mapping relationships. Essentially, our neural network is to learn a smooth function, denoted as $f$, which maps green dots to red dots as close as possible in high dimensional space as shown in Equation \ref{eq:f25d3d}. The function $f$ is parametrized by neural layers in general.
\begin{equation}
\label{eq:f25d3d}
\boldsymbol{y} = f(\boldsymbol{x}) \quad \left( \boldsymbol{x} \in Z^{64^3}, where {\ } Z=\{0,1\} \right)
\end{equation}

After generating training pairs, we feed them into our network. The first part of our network loosely follows the idea of a 3D encoder-decoder with the U-net connections \cite{Ronneberger2015}. The skip-connected encoder-decoder serves as an initial coarse generator which is followed by an up-sampling module to further generate a higher resolution 3D shape within a $256^3$ voxel grid. This whole generator learns a correlation between partial and complete 3D structures. With the supervision of complete 3D labels, the generator is able to learn a function $f$ and infer a reasonable 3D shape given a brand new partial 2.5D view. During testing, however, the results tend to be grainy and without fine details.

\begin{figure}[t]
    \centering
    \includegraphics[width=0.98\textwidth]{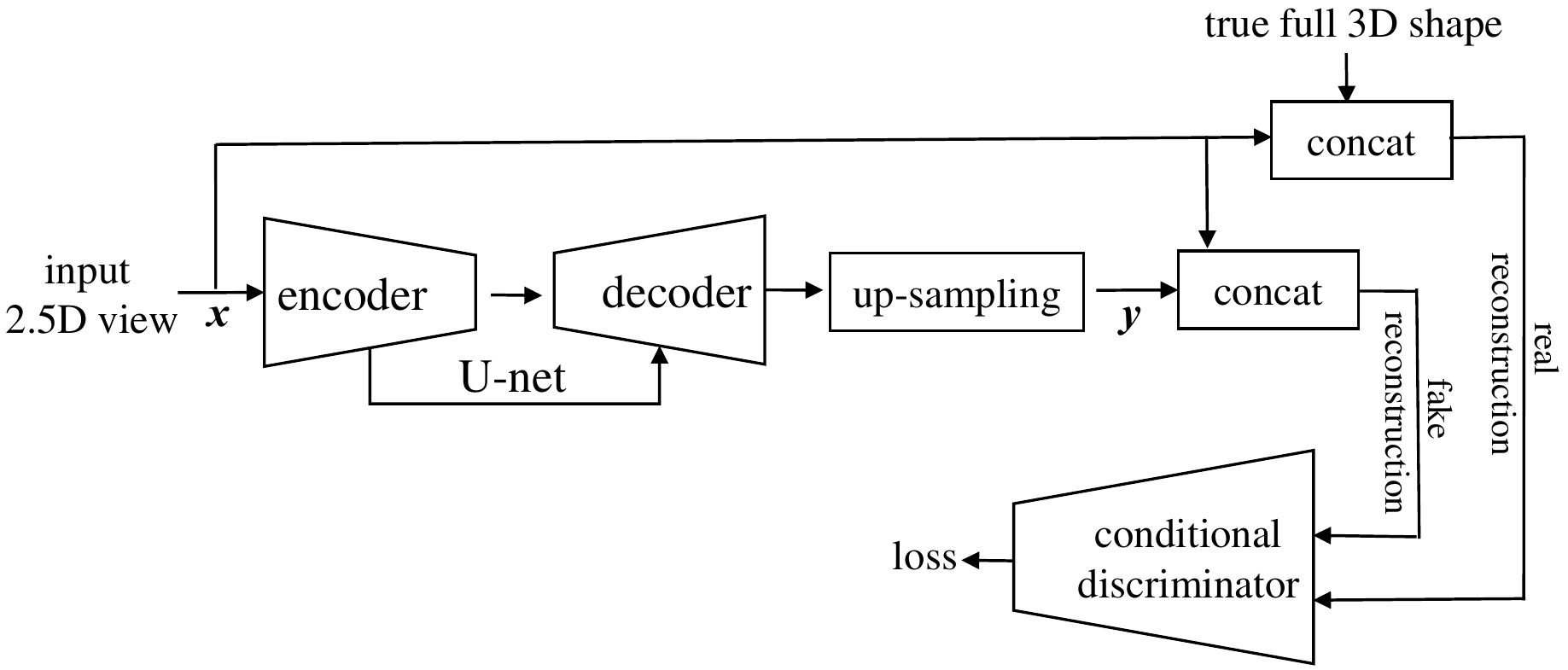}
    \caption{Overview of the network architecture for training.}
    \label{fig:ch3_arch_train}
\end{figure}
\begin{figure}[t]
    \centering
    \includegraphics[width=0.98\textwidth]{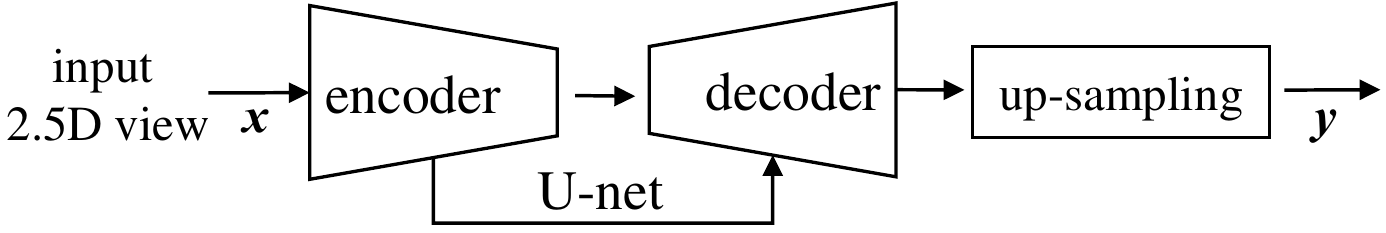}
    \caption{Overview of the network architecture for testing.}
    \label{fig:ch3_arch_test}
\end{figure}

To address this issue, in the training phase, the reconstructed 3D shape from the generator is further fed into a conditional discriminator to verify its plausibility. In particular, a partial 2.5D input view is paired with its corresponding complete 3D shape, which is called the `real reconstruction', while the partial 2.5D view is paired with its corresponding output 3D shape from generator, which is called the `fake reconstruction'. The discriminator aims to discriminate all `fake reconstruction' from `real reconstruction'. In the original GAN framework \cite{Goodfellow2014}, the task of the discriminator is to simply classify real and fake inputs, but its Jensen-Shannon divergence-based loss function is difficult to converge. The recent WGAN \cite{Arjovsky2017a} leverages Wasserstein distance with weight clipping as a loss function to stabilize the training procedure, whilst the extended work WGAN-GP \cite{Gulrajani2017} further improves the training process using a gradient penalty with respect to its input. In our \nicknameRecGAN{}, we apply WGAN-GP as the loss function on top of the mean feature of our conditional discriminator, which guarantees fast and stable convergence. The overall network architecture for training is shown in Figure \ref{fig:ch3_arch_train}, while the testing phase only needs the well trained generator as shown in Figure \ref{fig:ch3_arch_test}.
\section{Method Details}
\begin{figure*}[h]
\centering
\begin{subfigure}[t]{0.98\textwidth}
   \includegraphics[width=1\linewidth]{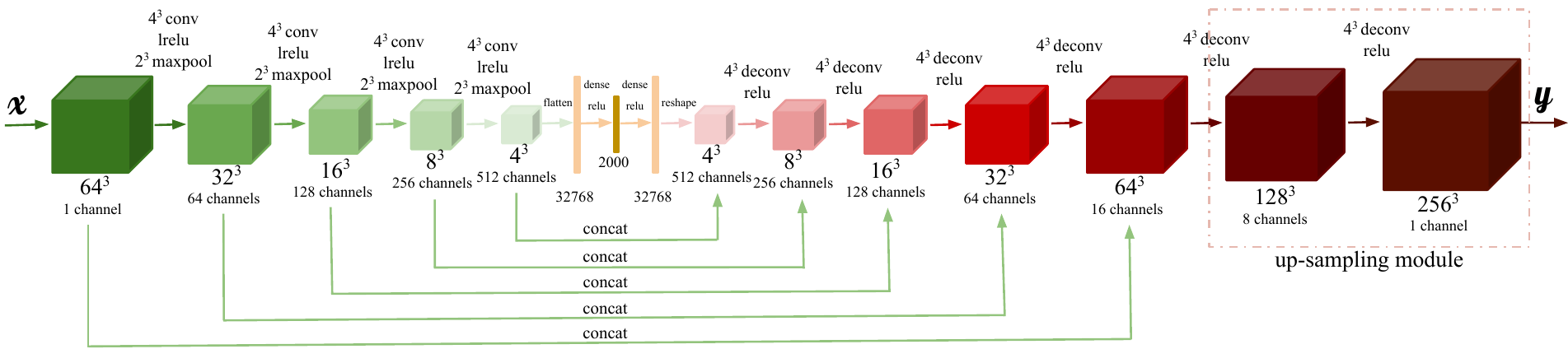}
   \caption{Generator for 3D shape estimation from a single depth view.}
   \label{fig:ch3_gen} 
\end{subfigure}
\begin{subfigure}[t]{0.98\textwidth}
   \includegraphics[width=1\linewidth]{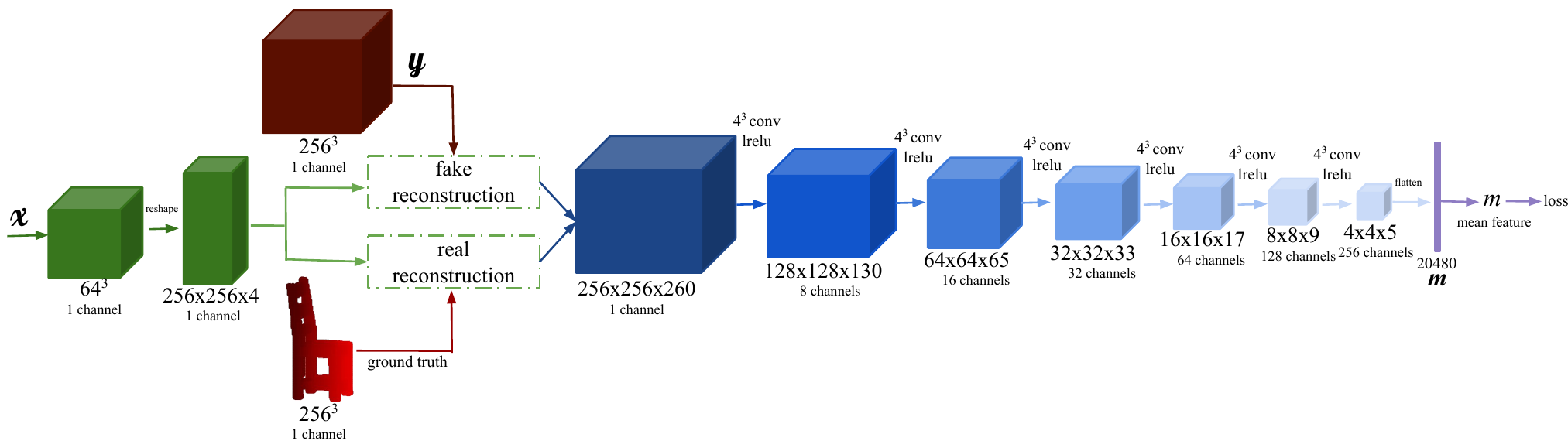}
   \caption{Discriminator for 3D shape refinement.}
   \label{fig:ch3_dis}
\end{subfigure}
\caption{Detailed architecture of \nicknameRecGAN{}, showing the two main building blocks. Note that, although these are shown as two separate modules, they are trained end-to-end.}
\label{fig:ch3_net_detail}
\end{figure*}

\subsection{Network Architecture}
Figure \ref{fig:ch3_net_detail} shows the detailed architecture of our proposed \nicknameRecGAN{}. It consists of two main networks: the generator as in Figure \ref{fig:ch3_gen} and the discriminator as in Figure \ref{fig:ch3_dis}.

\textbf{The generator} consists of a skip-connected encoder-decoder and an up-sampling module. Unlike the vanilla GAN generator which generates data from arbitrary latent distributions, our \nicknameRecGAN{} generator synthesizes data from 2.5D views. Particularly, the encoder has five 3D convolutional layers, each of which has a bank of $4\times4\times4$ filters with strides of $1\times1\times1$, followed by a leaky ReLU activation function and a max pooling layer with $2\times2\times2$ filters and strides of $2\times2\times2$. The number of output channels of max pooling layer starts with 64, doubling at each subsequent layer and ends up with 512. The encoder is lastly followed by two fully-connected layers to embed semantic information into a latent space. The decoder is composed of five symmetric up-convolutional layers which are followed by ReLU activations. Skip-connections between encoder and decoder guarantee propagation of local structures of the input 2.5D view. The skip-connected encoder-decoder is followed by the up-sampling module which simply consists of two layers of up-convolutional layers as detailed in Figure \ref{fig:ch3_gen}. This simple up-sampling module directly upgrades the output 3D shape to a higher resolution of $256^3$ without requiring complex network design and operations. It should be noted that without the two fully connected layers and skip-connections, the vanilla encoder-decoder would be unable to learn reasonable complete 3D structures as the latent space is limited and the local structure is not preserved. The loss function and optimization methods are described in Section \ref{sec:loss}.

\textbf{The discriminator} aims to distinguish whether the estimated 3D shapes are plausible or not. Based on the conditional GAN, the discriminator takes both real reconstruction pairs and fake reconstruction pairs as input. In particular, it consists of six 3D convolutional layers, the first of which concatenates the generated 3D shape (\ie{} a $256^3$ voxel grid) and the input 2.5D partial view (\ie{} a $64^3$ voxel grid), reshaped as a $256\times256\times4$ tensor. The reshaping process is done straightforwardly using Tensorflow `tf.reshape()'. Basically, this is to inject the condition information with a matched tensor dimension, and then leave the network itself to learn useful features from this condition input. Each convolutional layer has a bank of $4\times4\times4$ filters with strides of $2\times2\times2$, followed by a ReLU activation function except for the last layer which is followed by a sigmoid activation function. The number of output channels of the convolutional layers starts with 8, doubling at each subsequent layer and ends up with 256. The output of the last neural layer is reshaped as a latent vector which is the latent feature of discriminator, denoted as $\bm{m}$.

\subsection{Mean Feature for Discriminator}
At the early training stage of GAN, as the high dimensional real and fake distributions may not overlap, the discriminator can separate them perfectly using a single scalar output, which is analyzed in \cite{Arjovsky2017c}. In our experiments, due to the extremely high dimensionality (\ie{} $256^3+64^3$ dimensions) of the input data pair, the WGAN-GP always crashes in the early 3 epochs if we use a standard fully-connected layer followed by a single scalar as the final output for the discriminator. 

To stabilize training, we propose to use the mean feature $m$ (\ie{} mean of a vector feature $\bm{m}$) for discrimination. As the mean vector feature tends to capture more information from the input overall, it is more difficult for the discriminator to easily distinguish between fake or real inputs. This enables useful information to back-propagate to the generator. The final output of the discriminator $D()$ is defined as:
\begin{equation}
\label{eq:mean_feature}
	m = \mathbf{E}(\bm{m})
\end{equation}

Mean feature matching is also studied and applied in \cite{Bao2017b}\cite{Mroueh2017b} to stabilize GAN. However, Bao \etal{} \cite{Bao2017b} minimize the $L_2$ loss of the mean feature, as well as the original Jensen-Shannon divergence-based loss \cite{Goodfellow2014}, requiring hyper-parameter tuning to balance the two losses. By comparison, in our \nicknameRecGAN{} setting, the mean feature of discriminator is directly followed by the existing WGAN-GP loss, which is simple yet effective to stabilize the adversarial training.

Overall, our discriminator learns to distinguish the distributions of mean feature of fake and real reconstructions, while the generator is trained to make the two mean feature distributions as similar as possible.

\subsection{Loss Functions}
The objective function of \nicknameRecGAN{} includes two main parts: an object reconstruction loss $\ell_{en}$ for the generator; the objective function $\ell_{gan}$ for the conditional GAN.

(1) \textbf{$\ell_{en}$} For the generator, inspired by \cite{Brock2016}, we use modified binary cross-entropy loss function instead of the standard version. The standard binary cross-entropy weights both false positive and false negative results equally. However, most of the voxel grid tends to be empty, so the network easily gets a false positive estimation. In this regard, we impose a higher penalty on false positive results than on false negatives. Particularly, a weight hyper-parameter $\alpha$ is assigned to false positives, with (1-$\alpha$) for false negative results, as shown in  Equation \ref{eq:lae}.
\begin{equation}
\label{eq:lae}
\ell_{en} = \frac{1}{N} \sum_{i=1}^{N} \Big[-\alpha  \bar{y_i} \log(y_i) - (1-{\alpha} ) (1-\bar{y_i}) \log(1-y_i) \Big]
\end{equation}
$where$ $\bar{y_i}$ is the target value \{0,1\} of a specific $i^{th}$ voxel in the ground truth voxel grid $\boldsymbol{\bar{y}}$, and $y_i$ is the corresponding estimated value (0,1) in the same voxel from the generator output $\boldsymbol{y}$. We calculate the mean loss over the total $N$ voxels in the whole voxel grid.

(2) \textbf{$\ell_{gan}$} For the discriminator, we leverage the state of the art WGAN-GP loss functions. Unlike the original GAN loss function which presents an overall loss for both real and fake inputs, we separately represent the loss function $\ell_{gan}^{g}$ in Equation \ref{eq:lgang} for generating fake reconstruction pairs and $\ell_{gan}^{d}$ in Equation \ref{eq:lgand} for discriminating fake and real reconstruction pairs. Detailed definitions and derivation of the loss functions can be found in \cite{Arjovsky2017a}\cite{Gulrajani2017}, but we modify them for our conditional GAN settings.
\begin{equation}
\label{eq:lgang}
\ell_{gan}^{g} = -\mathbf{E}\left[ D(\boldsymbol{y}|\boldsymbol{x}) \right]
\end{equation}
\begin{align}
\label{eq:lgand}
\ell_{gan}^{d} = \mathbf{E}\left[ D(\boldsymbol{y}|\boldsymbol{x})\right] - \mathbf{E} \left[ D(\boldsymbol{\bar{y}}|\boldsymbol{x}) \right] 
+ \lambda \mathbf{E} \left[ \left(\norm{\nabla_{\boldsymbol{\hat{y}} }D(\boldsymbol{\hat{y}}|\boldsymbol{x})}_2 - 1 \right)^2  \right]
\end{align}
$where$ $\boldsymbol{\hat{y}} = \epsilon \boldsymbol{\bar{y}} +(1-\epsilon)\boldsymbol{y}, \epsilon \sim U[0,1]$, $\boldsymbol{x}$ is the input partial depth view, $\boldsymbol{y}$ is the corresponding output of the generator, $\boldsymbol{\bar{y}}$ is the corresponding ground truth. $\lambda$ controls the trade-off between optimizing the gradient penalty and the original objective in WGAN.

For the generator in our \nicknameRecGAN{} network, there are two loss functions, $\ell_{en}$ and $\ell_{gan}^{g}$, to optimize. As we discussed in Section \ref{sec:ch3_overview}, minimizing $\ell_{en}$ tends to learn the overall 3D shapes, whilst minimizing $\ell_{gan}^{g}$ estimates more plausible 3D structures conditioned on input 2.5D views. To minimize $\ell_{gan}^{d}$ is to improve the performance of discriminator to distinguish fake and real reconstruction pairs. To jointly optimize the generator, we assign weights $\beta$ to $\ell_{en}$ and $(1-\beta)$ to $\ell_{gan}^{g}$. Overall, the loss functions for generator and discriminator are as follows:
\begin{equation}
\ell_{g} = \beta \ell_{en} + (1-\beta) \ell_{gan}^{g}
\end{equation}
\begin{equation}
\ell_d = \ell_{gan}^{d}
\end{equation}

\subsection{Implementation}
We adopt an end-to-end training procedure for the whole network. To simultaneously optimize both the generator and discriminator, we alternate between one gradient descent step on the discriminator and then one step on the generator. For the WGAN-GP, $\lambda$ is set as 10 for gradient penalty as in \cite{Gulrajani2017}. $\alpha$ ends up as 0.85 for our modified cross entropy loss function, while $\beta$ is 0.2 for the joint loss function $\ell_{g}$.

The Adam solver\cite{Kingma2015a} is used for both discriminator and generator with a batch size of 4. The other three Adam parameters are set to default values. Learning rate is set to $5e^{-5}$ for the discriminator and $1e^{-4}$ for the generator in all epochs. As we do not use dropout or batch normalization, the testing phase is exactly the same as the training stage. The whole network is trained on a single Titan X GPU from scratch.

\subsection{Data Synthesis}
For the task of 3D dense reconstruction from a single depth view, obtaining a large amount of training data is an obstacle. Existing real RGB-D datasets for surface reconstruction suffer from occlusions and missing data and there is no ground truth of complete and high resolution $256^3$ 3D shapes for each view. The recent work \cite{Dai2017b} synthesizes data for 3D object completion, but the object resolution is only up to a resolution of $128^3$. 

To tackle this issue, we use the ShapeNet \cite{Chang2015} database to generate a large amount of training and testing data with synthetically rendered depth images and the corresponding complete 3D shape ground truth. Interior parts of individual objects are set to be filled, \ie{} `1', while the exterior to be empty, \ie{} `0'. A subset of object categories and CAD models are selected for our experiments. As some CAD models in ShapeNet may not be watertight, in our ray tracing based voxelization algorithm, if a specific point is inside more than 5 faces along the X, Y and Z axes, that point is deemed to be the interior of the object and set as `1', otherwise set to `0'.

For each category, to generate \textbf{training data}, around 220 CAD models are randomly selected. For each CAD model, we create a virtual depth camera to scan it from 125 different viewing angles, 5 uniformly sampled views for each of roll, pitch and yaw space ranging from $0\sim2\pi$ individually. Note that, the viewing angles for all 3D models are the same for simplicity. For each virtual scan, both a depth image and the corresponding complete 3D voxelized structure are generated with regard to the same camera angle. That depth image is simultaneously transformed to a point cloud using virtual camera parameters \cite{Khoshelham2012} followed by voxelization which generates a partial 2.5D voxel grid. Then a pair of partial 2.5D view and the complete 3D shape is synthesized. Overall, around 26K training pairs are generated for each 3D object category.

For each category, to synthesize \textbf{testing data}, around 40 CAD models are randomly selected. For each CAD model, two groups of testing data are generated. \textbf{Group 1}, each model is virtually scanned from 125 viewing angles which are the same as used in training dataset. Around 4.5k testing pairs are generated in total. This group of testing dataset is denoted as \textbf{same viewing} (SV) angles testing dataset. \textbf{Group 2}, each model is virtually scanned from 216 different viewing angles, 6 uniformly sampled views from each of roll, pitch and yaw space ranging from $0\sim2\pi$ individually. Note that, these viewing angles for all testing 3D models are completely different from training pairs. Around 8k testing pairs are generated in total. This group of testing dataset is denoted as \textbf{cross viewing} (CV) angles testing dataset. Similarly, we also generate around 1.5k SV and 2.5k CV \textbf{validation data} split from another 12 CAD models, which are used for hyperparameter searching.

As our network is initially designed to predict an aligned full 3D model given a depth image from an arbitrary viewing angle, these two SV and CV testing datasets are generated separately to evaluate the viewing angle robustness and generality of our model.

\begin{figure}[t]
   \centering
   \includegraphics[width=0.98\columnwidth]{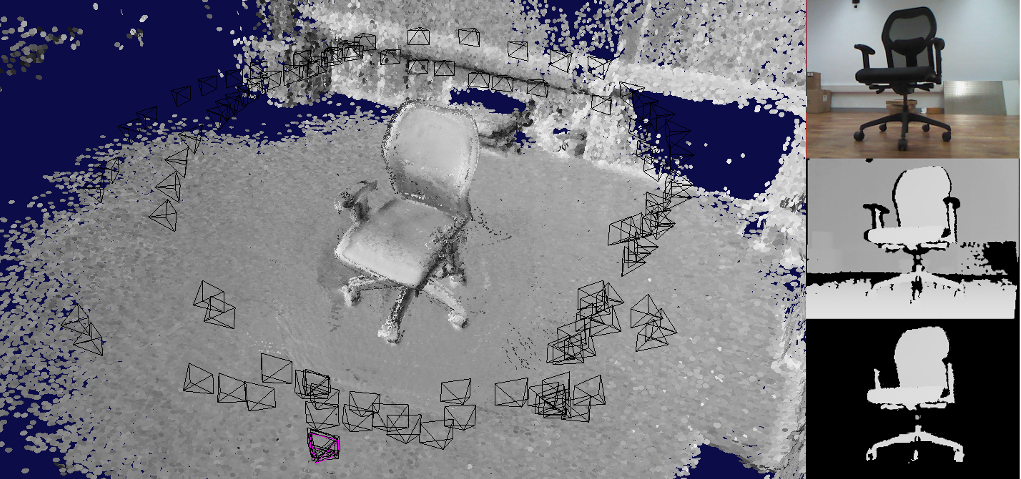}
   \caption{An example of ElasticFusion for generating real world data. Left: reconstructed object; sampled camera poses are shown in black. Right: Input RGB, depth image and segmented depth image.}
   \label{fig:ch3_elasticfusion}
\end{figure}

Besides the large quantity of synthesized data, we also collect a \textbf{real-world dataset} in order to test the proposed network in a realistic scenario. We use a Microsoft Kinect camera to manually scan a total of 20 object instances belonging to 4 classes \{bench, chair, couch, table\}, with 5 instances per class from different environments, including offices, homes, and outdoor university parks. For each object we acquire RGB-D images of the object from multiple angles by moving the camera around the object. Then, we use the dense visual SLAM algorithm ElasticFusion \cite{Whelan2015} in order to reconstruct the full 3D shape of each object, as well as the camera pose in each scan. 

We sample 50 random views from the camera trajectory, and for each one we obtain the depth image and the relative camera pose. In each depth image the 3D object is segmented from the background, using a combination of floor removal and manual segmentation. We finally generate ground truth information by aligning the full 3D objects with the partial 2.5D views.

It should be noted that, due to noise and quantization artifacts of low-cost RGB-D sensors, and the inaccuracy of the SLAM algorithm, the full 3D ground truth is not 100\% accurate, but can still be used as a reasonable approximation.
The real-world dataset highlights the challenges related to shape reconstruction from realistic data: noisy depth estimates, missing depth information, depth quantization. In addition, some of the objects are acquired outdoors (\eg{} $bench$), which is challenging for the near-infrared depth sensor of the Micorsoft Kinect. However, we argue that a real-world benchmark for shape reconstruction is necessary for a thorough validation of future approaches. Figure \ref{fig:ch3_elasticfusion} shows an example of the reconstructed object and camera poses in ElasticFusion.

\section{Experiments}
In this section, we evaluate our \nicknameRecGAN{} with comparison to the state of the art approaches and an ablation study to fully investigate the proposed network. 

\subsection{Metrics}
To evaluate the performance of 3D reconstruction, we consider two metrics. The first metric is the mean Intersection-over-Union (IoU) between predicted 3D voxel grids and their ground truth. The IoU for an individual voxel grid is formally defined as follows:
\begin{equation*}
IoU = \frac{\sum_{i=1}^{N} \left[  I (y_i>p) * I(\bar{y_i}) \right] }{ \sum_{i=1}^{N}   \left[I  \left( I(y_{i} >p) + I(\bar{y_i}) \right) \right] } 
\end{equation*}
$where$ $I(\cdot)$ is an indicator function, $y_{i}$ is the predicted value for the $i^{th}$ voxel, $\bar{y_i}$ is the corresponding ground truth, $p$ is the threshold for voxelization, $N$ is the total number of voxels in a whole voxel grid. In all our experiments, $p$ is searched using the validation data split per category for each approach. Particularly, $p$ is searched in the range $[0.1, 0.9]$ with a step size $0.05$ using the validation datasets. The higher the IoU value, the better the reconstruction of a 3D model.

The second metric is the mean value of standard Cross-Entropy loss (CE) between a reconstructed shape and the ground truth 3D model. It is formally defined as: 
\begin{equation*}
CE =-\frac{1}{N} \sum_{i=1}^{N} \left[\bar{y_i}\log(y_i) + (1 - \bar{y_i})\log(1-y_i)\right]
\end{equation*}
$where$ $y_i$, $\bar{y_i}$ and $N$ are the same as defined in above IoU. The lower CE value is, the closer the prediction to be either `1' or `0', the more robust and confident the 3D predictions are. 

We also considered the Chamfer Distance (CD) or Earth Mover's Distance (EMD) as an additional metric. However, it is computationally heavy to calculate the distance between two high resolution voxel grids due to the large number of points. In our experiments, it takes nearly 2 minutes to calculate either CD or EMD between two $256^3$ shapes on a single Titan X GPU. Although the $256^3$ dense shapes can be downsampled to sparse point clouds on object surfaces to quickly compute CD or EMD, the geometric details are inevitably lost due to the extreme downsampling process. Therefore, we did not use CD or EMD for evaluation in our experiments.

\begin{figure*}[h]
\centering
\begin{subfigure}[t]{0.98\textwidth}
   \includegraphics[width=1\linewidth]{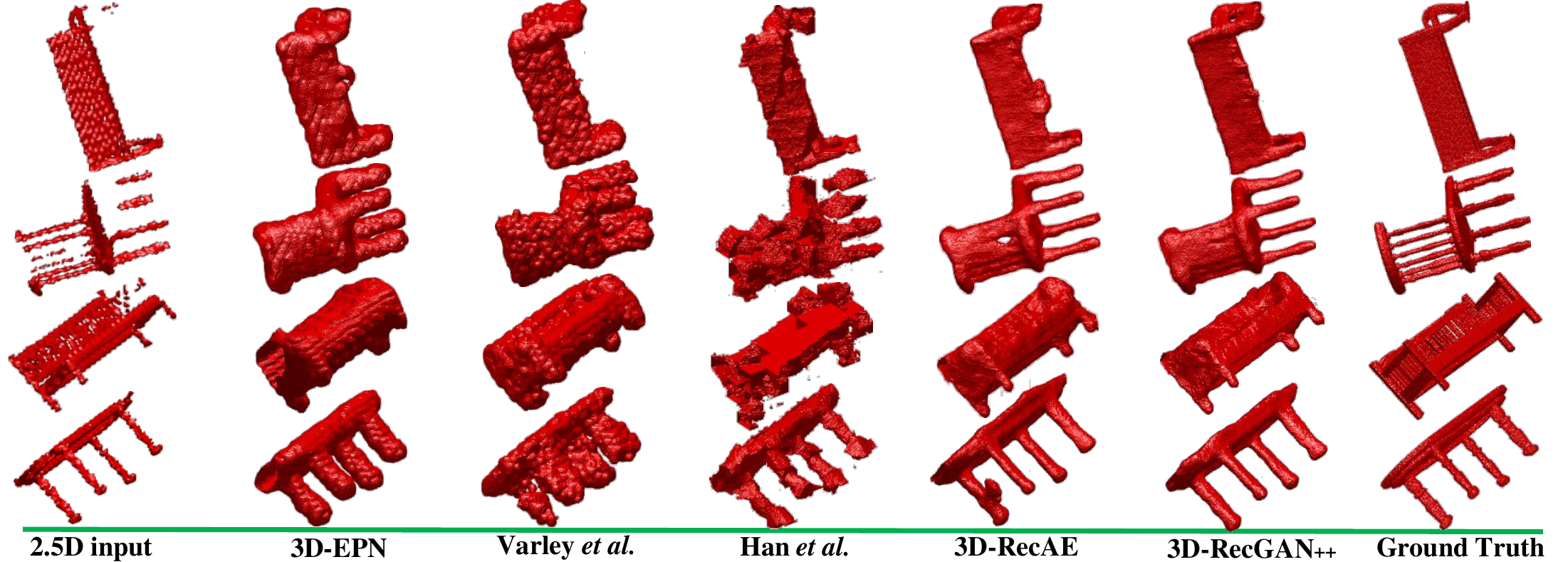}
   \caption{Qualitative results of single category reconstruction on testing datasets with same viewing angles.}
   \label{fig:ch3_per_cat_125} 
\end{subfigure}
\begin{subfigure}[t]{0.98\textwidth}
   \includegraphics[width=1\linewidth]{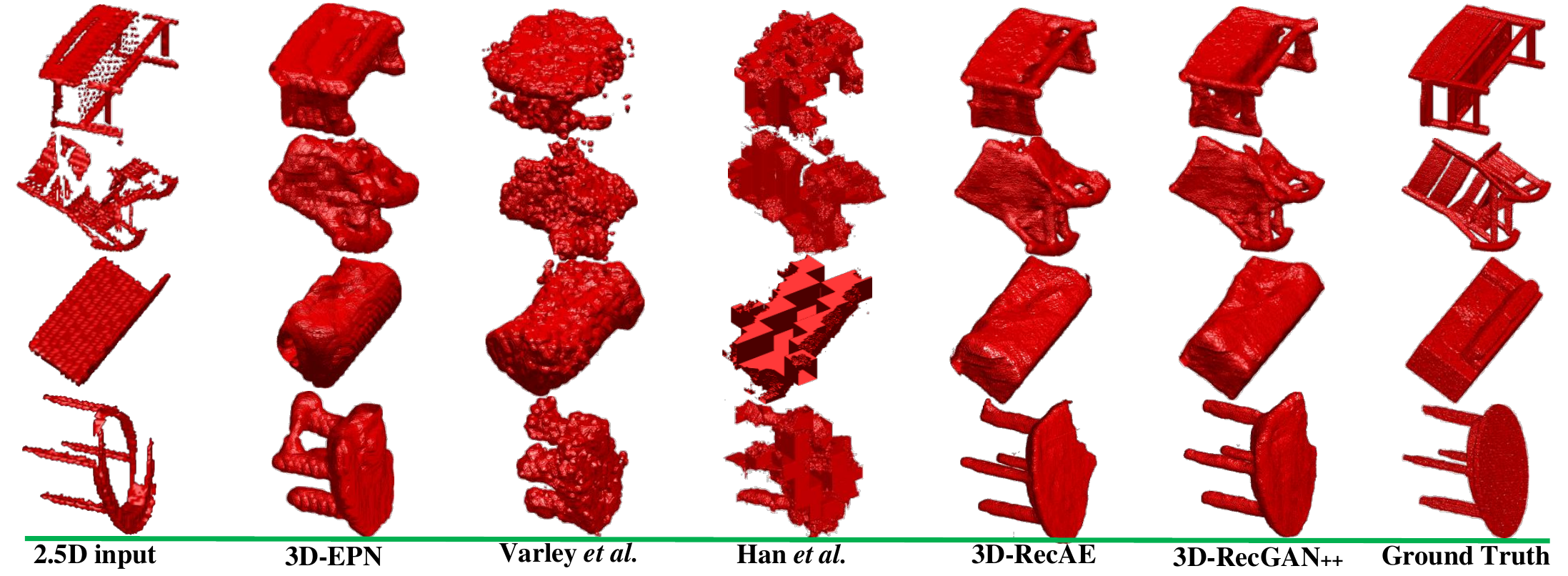}
   \caption{Qualitative results of single category reconstruction on testing datasets with cross viewing angles.}
   \label{fig:ch3_per_cat_216}
\end{subfigure}
\caption{Qualitative results of single category reconstruction on testing datasets with same and cross viewing angles.}
\label{fig:ch3_per_cat_125_216}
\end{figure*}

\subsection{Competing Approaches}
We compare against three state of the art deep learning based approaches for single depth view reconstruction. We also compare against the generator alone in our network, \ie{} without the GAN, named 3D-RecAE for short.

\begin{itemize}
    \item \textbf{3D-EPN.} In \cite{Dai2017b}, Dai \etal{} proposed a neural network, called ``3D-EPN'', to reconstruct the 3D shape up to a $32^3$ voxel grid, after which a high resolution shape is retrieved from an existing 3D shape database, called ``Shape Synthesis''. In our experiment, we only compared with their neural network (\ie{} 3D-EPN) performance because we do not have an existing shape database for similar shape retrieval during testing. Besides, occupancy grid representation is used for the network training and testing.
    
    \item \textbf{Varley \etal{}} In \cite{Varley2017}, a network was designed to complete the 3D shape from a single 2.5D depth view for robot grasping. The output of their network is a $40^3$ voxel grid.
    
    Note that, the low resolution voxel grids generated by 3D-EPN and Varley \etal{} are all upsampled to $256^3$ voxel grids using trilinear interpolation before calculating the IoU and CE metrics. The linear upsampling is a widely used post-processing technique for fair comparison in cases where the output resolution is not identical \cite{Tatarchenko2017}. However, as both 3D-EPN and Varley \etal{} are trained using lower resolution voxel grids for supervision, while the below Han \etal{} and our \nicknameRecGAN{} are trained using $256^3$ shapes for supervision, it is not strictly fair comparison in this regard. Considering both 3D-EPN and Varley \etal{} are among the early works and also solid competing approaches regarding the single depth view reconstruction task, we therefore include them as baselines.
    
    \item \textbf{Han \etal{}} In \cite{Han2017}, a global structure inference network and a local geometry refinement network are proposed to complete a high resolution shape from a noisy shape. The network is not originally designed for single depth view reconstruction, but its output shape is up to a $256^3$ voxel grid and is comparable to our network. For fair comparison, the same occupancy grid representation is used for their network. It should be noted that \cite{Han2017} involves convoluted designs, thus the training procedure is slower and less efficient due to the many integrated LSTMs.
    
    \item \textbf{3D-RecAE.} As for our \nicknameRecGAN{}, we remove the discriminator and only keep the generator to infer the complete 3D shape from a single depth view. This comparison illustrates the benefits of adversarial learning.
\end{itemize}

\begin{figure*}[h]
\centering
\begin{subfigure}[t]{0.98\textwidth}
   \includegraphics[width=1\linewidth]{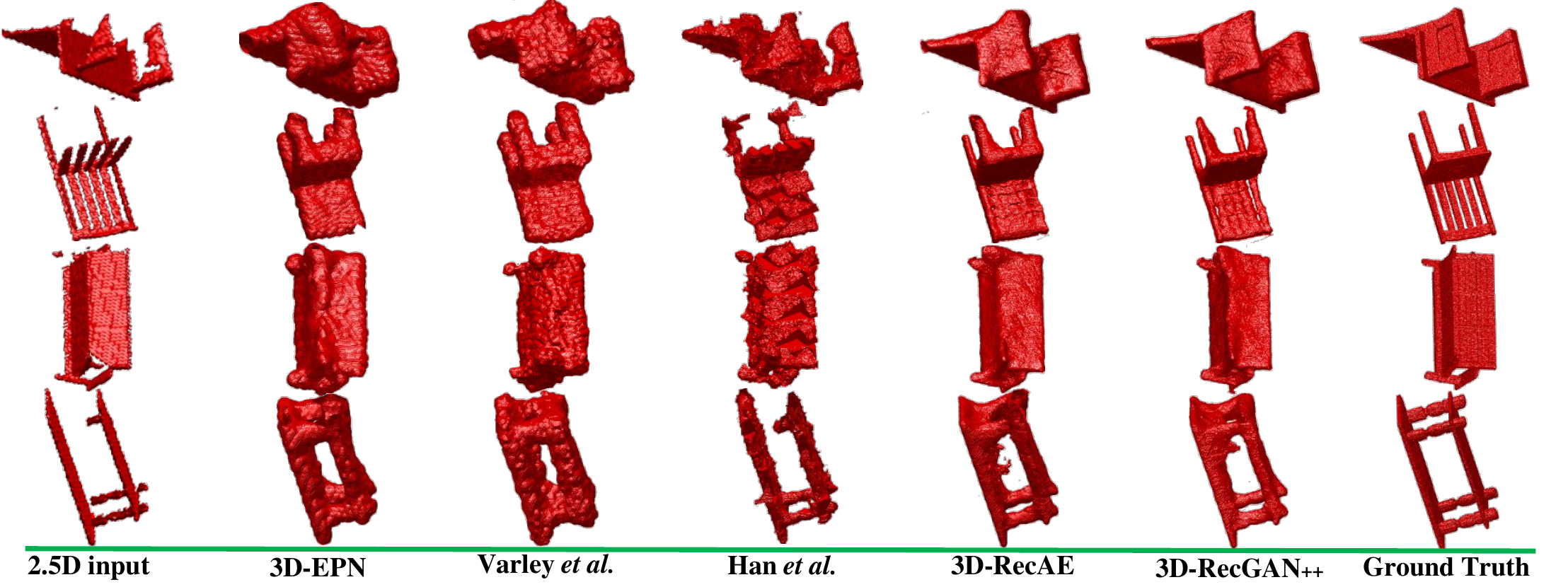}
   \caption{Qualitative results of multiple category reconstruction on testing datasets with the same viewing angles.}
   \label{fig:ch3_multi_cat_125} 
\end{subfigure}
\begin{subfigure}[t]{0.98\textwidth}
   \includegraphics[width=1\linewidth]{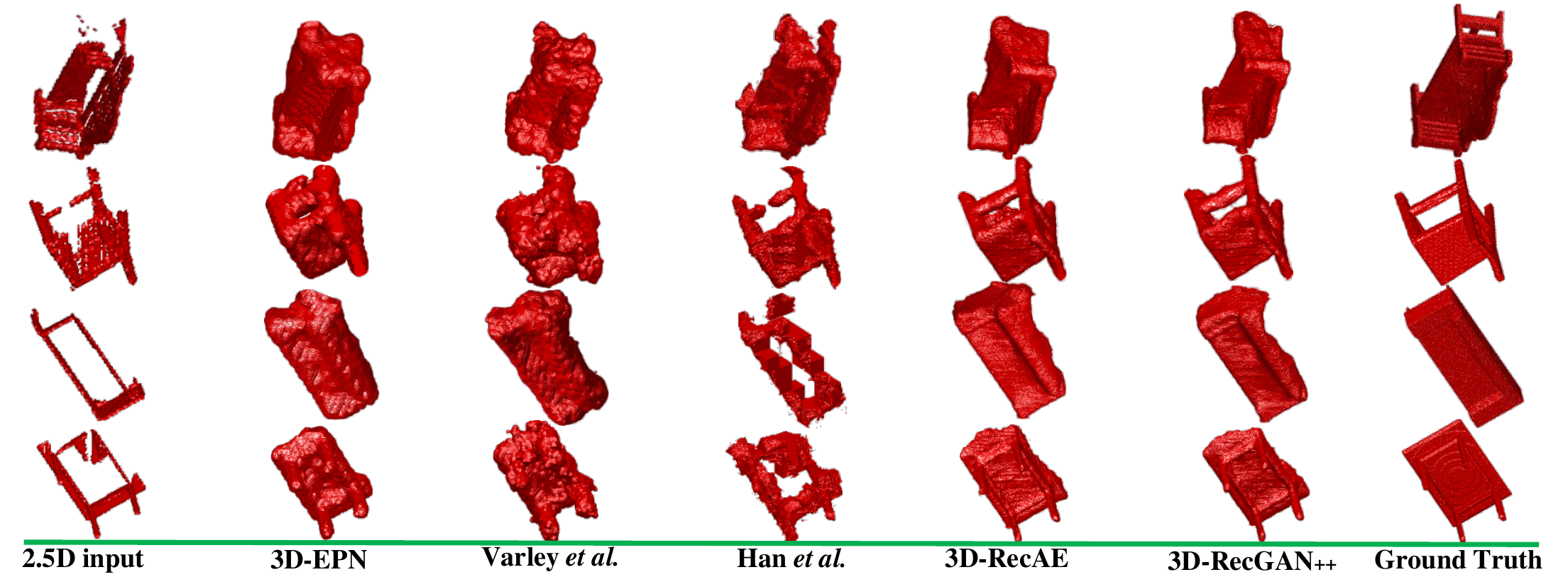}
   \caption{Qualitative results of multiple category reconstruction on testing datasets with cross viewing angles.}
   \label{fig:ch3_multi_cat_216}
\end{subfigure}
\caption{Qualitative results of multiple category reconstruction on testing datasets with same and cross viewing angles.}
\label{fig:ch3_multi_cat_125_216}
\end{figure*}

\subsection{Single-category Results}
(1) \textbf{Results.} All networks are separately trained and tested on four different categories, \{bench, chair, couch, table\}, with the same network configuration.  
Table \ref{tab:per_cat_iou_ce_sv_256} shows the IoU and CE loss of all methods on the testing dataset with same viewing angles on $256^3$ voxel grids, while Table \ref{tab:per_cat_iou_ce_cv_256} shows the IoU and CE loss comparison on testing dataset with cross viewing angles. Figure \ref{fig:ch3_per_cat_125_216} shows the qualitative results of single category reconstruction on testing datasets with same and cross viewing angles. The meshgrid function in Matlab is used to plot all 3D shapes for better visualization.

\begin{table}[ht]
\caption{Per-category IoU and CE loss on testing dataset with same viewing angles ($256^3$ voxel grids).}
\centering
\label{tab:per_cat_iou_ce_sv_256}
\tabcolsep=0.17cm
\begin{tabular}{c|cccc|cccc}
	\hline
    	& \multicolumn{4}{c|}{IoU} & \multicolumn{4}{c}{CE Loss} \\ \hline
  	             & bench  & chair & couch & table & bench  & chair & couch & table \\ 
    \hlineB{3}
     3D-EPN \cite{Dai2017b}   & 0.423 & 0.488 & 0.631 & 0.508 &0.087 &0.105 &0.144 &0.101\\
    
     Varley \etal{} \cite{Varley2017}   & 0.227 & 0.317 & 0.544 & 0.233 &0.111 &0.157 &0.195 &0.191\\
    
     Han \etal{} \cite{Han2017}  & 0.441 & 0.426 & 0.446 & 0.499 & 0.045 & 0.081 &0.165 &0.058\\
    
     \textbf{\scriptsize{3D-RecAE (ours)}}  & 0.577 & 0.641 & 0.749 & 0.675 &0.036 &0.063 &0.067 &0.043\\
    
     \textbf{\scriptsize{\nicknameRecGAN{} (ours)}}  & \textbf{0.580} & \textbf{0.647} & \textbf{0.753} & \textbf{0.679} & \textbf{0.034} &\textbf{0.060}&\textbf{0.066} &\textbf{0.040} \\
\hline
\end{tabular}
\end{table}
\begin{table}[ht]
\caption{Per-category IoU and CE loss on testing dataset with cross viewing angles ($256^3$ voxel grids).}
\centering
\label{tab:per_cat_iou_ce_cv_256}
\tabcolsep=0.17cm
\begin{tabular}{c|cccc|cccc}
	\hline
    	& \multicolumn{4}{c|}{IoU} & \multicolumn{4}{c}{CE Loss} \\ \hline
  	                          & bench  & chair & couch & table & bench  & chair & couch & table \\ 
    \hlineB{3}
     3D-EPN \cite{Dai2017b}   & 0.408 & 0.446 & 0.572 & 0.482 &0.086 &0.112 &0.163 &0.103\\
    
     Varley \etal{} \cite{Varley2017}   & 0.185 & 0.278 & 0.475 & 0.187 &0.108 &0.171 &0.210 &0.186\\
    
     Han \etal{} \cite{Han2017}  & 0.439 & 0.426 & 0.455 & 0.482 & 0.047 & 0.090 &0.163 &0.060\\
    
     \textbf{\scriptsize{3D-RecAE (ours)}}  & 0.524 & 0.588 & 0.639 & 0.610 &0.045 &0.079 &0.117 &0.058\\
     
     \textbf{\scriptsize{\nicknameRecGAN{} (ours)}}  & \textbf{0.531} & \textbf{0.594} & \textbf{0.646} & \textbf{0.618} & \textbf{0.041} &\textbf{0.074}&\textbf{0.111} &\textbf{0.053} \\
\hline
\end{tabular}
\end{table}

(2) \textbf{Analysis.} Both \nicknameRecGAN{} and 3D-RecAE significantly outperform the competing approaches in terms of IoU and CE loss on both the SV and CV testing datasets for dense 3D shape reconstruction ($256^3$ voxel grids). Although our approach is trained on depth input with a limited set of viewing angles, it still performs well to predict aligned 3D shapes from novel viewing angles. The 3D shapes generated by \nicknameRecGAN{} and 3D-RecAE are much more visually compelling than others.

Compared with 3D-RecAE, \nicknameRecGAN{} achieves better IoU scores and smaller CE loss. Basically, adversarial learning of the discriminator serves as a regularizer for fine-grained 3D shape estimation, which enables the output of \nicknameRecGAN{} to be more robust and confident. We also notice that the increase of \nicknameRecGAN{} in IoU and CE scores is not dramatic compared with 3D-RecAE. This is primarily because the main object shape can be reasonably predicted by 3D-RecAE, while the finer geometric details estimated by \nicknameRecGAN{} are usually smaller parts of the whole object shape. Therefore, \nicknameRecGAN{} only obtains a reasonable better IoU and CE scores than 3D-RecAE. The $4^{th}$ row of Figure \ref{fig:ch3_per_cat_125} shows a good example in terms of finer geometric details prediction of \nicknameRecGAN{}. In fact, in all the remaining experiments, \nicknameRecGAN{} is constantly, but not significantly, better than 3D-RecAE.

\begin{table}[ht]
\caption{Multi-category IoU and CE loss on testing dataset with same viewing angles ($256^3$ voxel grids).}
\centering
\label{tab:multi_cat_iou_ce_sv_256}
\tabcolsep=0.17cm
\begin{tabular}{c|cccc|cccc}
	\hline
    	& \multicolumn{4}{c|}{IoU} & \multicolumn{4}{c}{CE Loss} \\ \hline
  	                          & bench  & chair & couch & table & bench  & chair & couch & table \\ 
    \hlineB{3}
     3D-EPN \cite{Dai2017b}   & 0.428 & 0.484 & 0.634 & 0.506 &0.087 &0.107 &0.138 &0.102\\
    
     Varley \etal{} \cite{Varley2017}   & 0.234 & 0.317 & 0.543 & 0.236 &0.103 &0.132 &0.197 &0.170\\
    
     Han \etal{} \cite{Han2017}  & 0.425 & 0.454 & 0.440 & 0.470 & 0.045 & 0.087 &0.172 &0.065\\
    
     \textbf{\scriptsize{3D-RecAE (ours)}}  & 0.576 & 0.632 & 0.740 & 0.661 &0.037 &0.060 &0.069 &0.044\\
     
     \textbf{\scriptsize{\nicknameRecGAN{} (ours)}}  & \textbf{0.581} & \textbf{0.640} & \textbf{0.745} & \textbf{0.667} & \textbf{0.030} &\textbf{0.051}&\textbf{0.063} &\textbf{0.039} \\
\hline
\end{tabular}
\end{table}

\begin{table}[ht]
\caption{Multi-category IoU and CE loss on testing dataset with cross viewing angles ($256^3$ voxel grids).}
\centering
\label{tab:multi_cat_iou_ce_cv_256}
\tabcolsep=0.17cm
\begin{tabular}{c|cccc|cccc}
	\hline
    	& \multicolumn{4}{c|}{IoU} & \multicolumn{4}{c}{CE Loss} \\ \hline
  	                          & bench  & chair & couch & table & bench  & chair & couch & table \\ 
    \hlineB{3}
     3D-EPN \cite{Dai2017b}   & 0.415 & 0.452 & 0.531 & 0.477 &0.091 &0.115 &0.147 &0.111\\
    
     Varley \etal{} \cite{Varley2017}   & 0.201 & 0.283 & 0.480 & 0.199 &0.105 &0.143 &0.207 &0.174\\
    
     Han \etal{} \cite{Han2017}  & 0.429 & 0.444 & 0.447 & 0.474 & 0.045 & 0.089 &0.172 &0.063\\
    
     \textbf{\scriptsize{3D-RecAE (ours)}}  & 0.530 & 0.587 & 0.640 & 0.610 &0.043 &0.068 &0.096 &0.055\\
     
     \textbf{\scriptsize{\nicknameRecGAN{} (ours)}}  & \textbf{0.540} & \textbf{0.594} & \textbf{0.643} & \textbf{0.621} & \textbf{0.038} &\textbf{0.061}&\textbf{0.091} &\textbf{0.048} \\
\hline
\end{tabular}
\end{table}

\subsection{Multi-category Results} \label{sec:ch3_multi_cat}
(1) \textbf{Results.} All networks are also trained and tested on multiple categories without being given any class labels. The networks are trained on four categories: \{bench, chair, couch, table\}; and then tested separately on individual categories. Table \ref{tab:multi_cat_iou_ce_sv_256} shows the IoU and CE loss comparison of all methods on testing dataset with same viewing angles for dense shape reconstruction, while Table \ref{tab:multi_cat_iou_ce_cv_256} shows the IoU and CE loss comparison on testing dataset with cross viewing angles. Figure \ref{fig:ch3_multi_cat_125_216} shows the qualitative results of all approaches on testing datasets of multiple categories with same and cross viewing angles.

(2) \textbf{Analysis.} Both \nicknameRecGAN{} and 3D-RecAE significantly outperforms the state of the art by a large margin in all categories which are trained together on a single model. Besides, the performance of our network trained on multiple categories, does not notably degrade compared with training the network on individual categories as shown in previous Table \ref{tab:per_cat_iou_ce_sv_256} and \ref{tab:per_cat_iou_ce_cv_256}. This confirms that our network has enough capacity and capability to learn diverse features from multiple categories.

\begin{table}[h]
\caption{Cross-category IoU on testing dataset with the same viewing angles ($256^3$ voxel grids).}
\centering
\label{tab:cross_cat_iou_sv_256}
\tabcolsep=0.32cm
\begin{tabular}{c|cccccc}
	\hline
  	 	 			 & car & faucet & firearm & guitar & monitor & plane \\ 
    \hlineB{3}
     3D-EPN \cite{Dai2017b}    & 0.450 & 0.442 & 0.339 & 0.351  & 0.444 & 0.314 \\
     Varley \etal{} \cite{Varley2017}   & 0.484 & 0.260 & 0.280 & 0.255  & 0.341 & 0.295 \\
     Han \etal{} \cite{Han2017}  & 0.360 & 0.402 & 0.333 & 0.353  & 0.450 & 0.306 \\
     \textbf{\scriptsize{3D-RecAE (ours)}}  & \textbf{0.557} & 0.530 & 0.422 & 0.440  & 0.556 & 0.390 \\
     \textbf{ \scriptsize{\nicknameRecGAN{} (ours)}}  & 0.555 & \textbf{0.536} & \textbf{0.426} & \textbf{0.442}  & \textbf{0.562} & \textbf{0.394} \\
\hline
\end{tabular}
\end{table}

\begin{table}[h]
\caption{Cross-category CE loss on testing dataset with the same viewing angles ($256^3$ voxel grids).}
\centering
\label{tab:cross_cat_ce_sv_256}
\tabcolsep=0.32cm
\begin{tabular}{c|cccccc}
	\hline
  	 	 	  & car & faucet & firearm & guitar & monitor & plane \\ 
    \hlineB{3}
     3D-EPN \cite{Dai2017b}    & 0.170 & 0.088 & 0.036 & 0.036  & 0.123 & 0.066 \\
     Varley \etal{} \cite{Varley2017}   & 0.173 & 0.122 & 0.029 & 0.030  & 0.130 & 0.042 \\
     Han \etal{} \cite{Han2017}  & 0.167 & 0.077 & 0.018 & 0.015  & 0.088 & 0.031 \\
     \textbf{\scriptsize{3D-RecAE (ours)}}  & 0.110 & 0.057 & 0.018 & 0.016  & 0.072 & 0.036 \\
     \textbf{ \scriptsize{\nicknameRecGAN{} (ours)}}  & \textbf{0.102} & \textbf{0.053} & \textbf{0.016} & \textbf{0.014}  & \textbf{0.067} & \textbf{0.031} \\
\hline
\end{tabular}
\end{table}

\subsection{Cross-category Results}
\begin{figure*}[hp]
\centering
\begin{subfigure}[t]{0.98\textwidth}
   \includegraphics[width=1\linewidth]{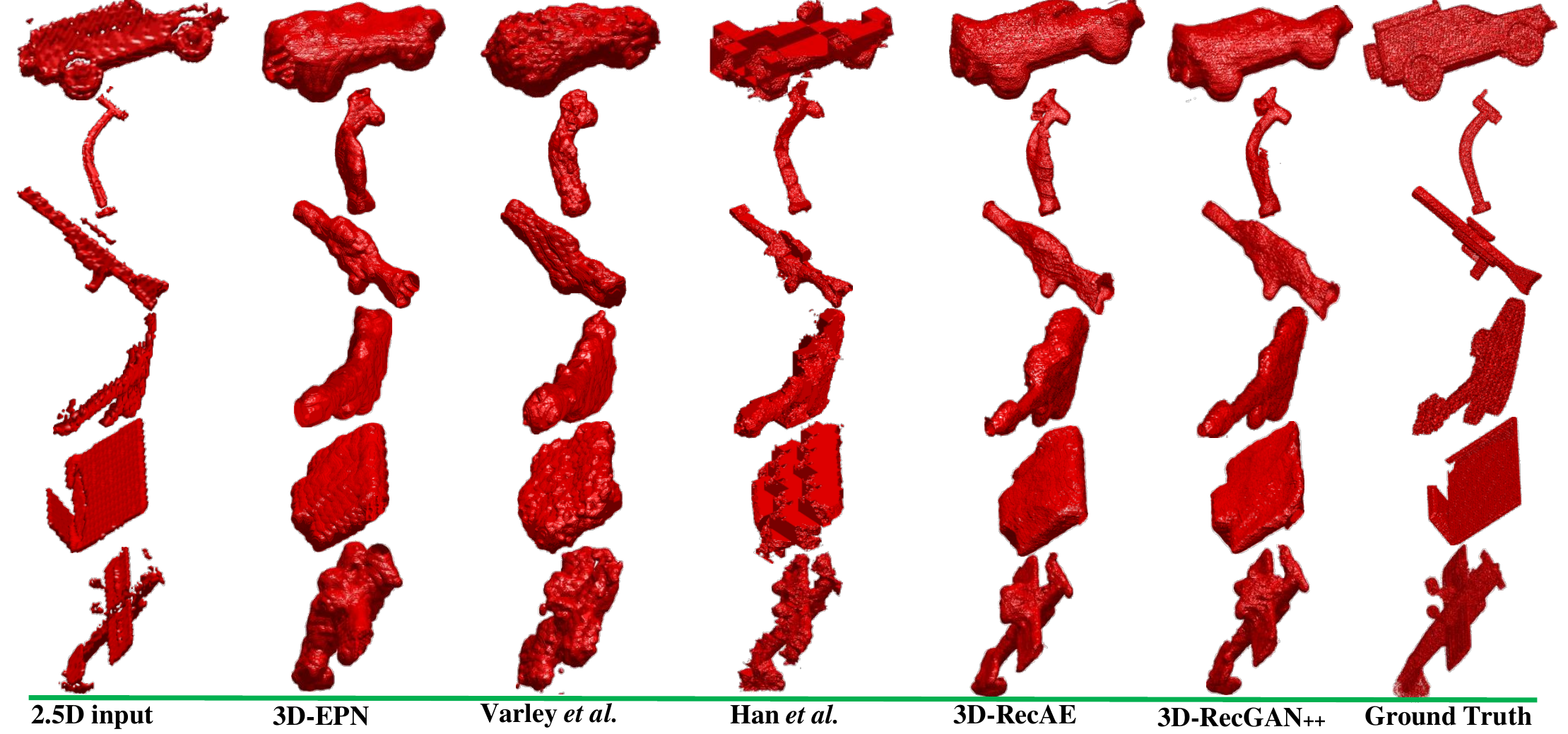}
   \caption{Qualitative results of cross category reconstruction on testing datasets with same viewing angles.}
   \label{fig:ch3_cross_cat_125} 
\end{subfigure}
\begin{subfigure}[t]{0.98\textwidth}
   \includegraphics[width=1\linewidth]{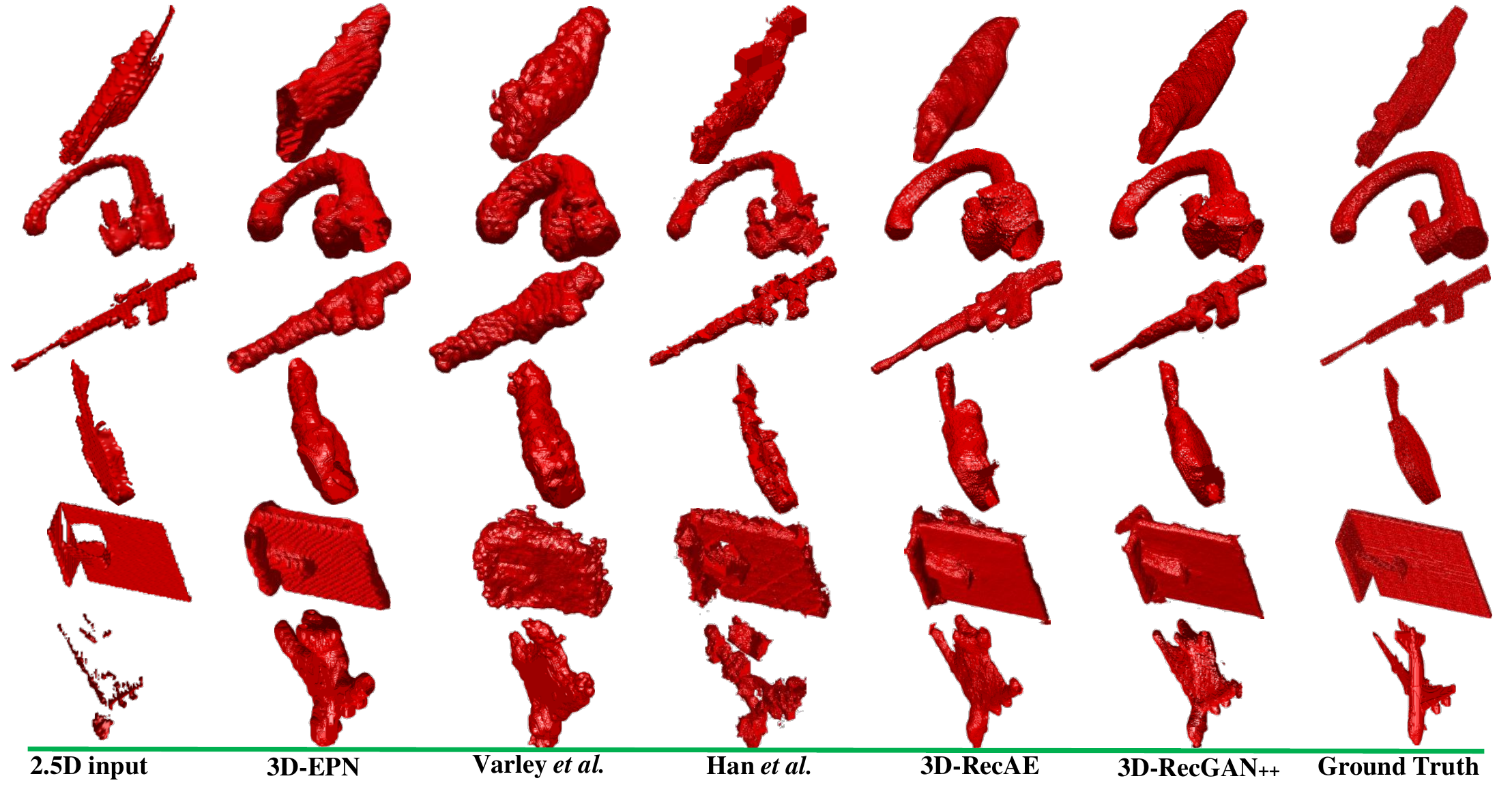}
   \caption{Qualitative results of cross category reconstruction on testing datasets with cross viewing angles.}
   \label{fig:ch3_cross_cat_216}
\end{subfigure}
\caption{Qualitative results of cross category reconstruction on testing datasets with same and cross viewing angles.}
\label{fig:ch3_cross_cat_125_216}
\end{figure*}

(1) \textbf{Results.} To further investigate the generality of networks, we train all networks on \{bench, chair, couch, table\}, and then test them on another 6 totally different categories: \{car, faucet, firearm, guitar, monitor, plane\}. For each of the 6 categories, we generate the same amount of testing datasets with same and cross viewing angles, which is similar to the previous \{bench, chair, couch, table\}. Table \ref{tab:cross_cat_iou_sv_256} and \ref{tab:cross_cat_ce_sv_256} shows the IoU and CE loss comparison of all approaches on the testing dataset with same viewing angles, while Table \ref{tab:cross_cat_iou_cv_256} and \ref{tab:cross_cat_ce_cv_256} shows the IoU and CE loss comparison on the testing dataset with cross viewing angles. Figure \ref{fig:ch3_cross_cat_125_216} shows the qualitative results of all methods on 6 unseen categories with same and cross viewing angles.

We further evaluate the generality of our \nicknameRecGAN{} on a specific category. Particularly, we conduct four groups of experiments. In the first group, we train our \nicknameRecGAN{} on bench, then separately test it on the remaining 3 categories: \{chair, couch, table\}. In the second group, the network is trained on chair and separately tested on \{bench, couch, table\}. Similarly, another two groups of experiments are conducted. Basically, this experiment is to investigate how well our approach learns features from one category and then generalizes to a different category, and vice versa. Table \ref{tab:cross_cat_single_iou_ce_sv_256} shows the cross-category IoU and CE loss of our \nicknameRecGAN{} trained on individual category and then tested on the testing dataset with same viewing angles over $256^3$ voxel grids. 

(2) \textbf{Analysis.} The proposed \nicknameRecGAN{} achieves much higher IoU and smaller CE loss across the unseen categories than competing approaches. Our network not only learns rich features from different object categories, but also is able to generalize well to completely new types of categories. Our intuition is that the network may learn geometric features such as lines, planes, curves which are common across various object categories. As our network involves skip-connections between intermediate neural layers, it is not straightforward to visualize and analyze the learnt latent features.

It can be also observed that our model trained on $bench$ tends to be more general than others. Intuitively, the bench category tends to have general features such as four legs, seats, and/or a back, which are also common among other categories \{chair, couch, table\}. However, not all chairs or couches consist of such general features that are shared across different categories. 

Overall, we may safely conclude that the more similar features two categories share, including both the low-level lines/planes/curves and the high-level shape components, the better generalization of our model achieves cross those categories.

\begin{table}[t]
\caption{Cross-category IoU on testing dataset with cross viewing angles ($256^3$ voxel grids).}
\centering
\label{tab:cross_cat_iou_cv_256}
\tabcolsep=0.32cm
\begin{tabular}{c|cccccc}
	\hline
  	 	 	  & car & faucet & firearm & guitar & monitor & plane \\ 
    \hlineB{3}
     3D-EPN \cite{Dai2017b}    & 0.446 & 0.439 & 0.324 & 0.359  & 0.448 & 0.309 \\
     Varley \etal{} \cite{Varley2017}   & 0.489 & 0.260 & 0.274 & 0.255  & 0.334 & 0.283 \\
     Han \etal{} \cite{Han2017}  & 0.349 & 0.402 & 0.321 & 0.363  & 0.455 & 0.299 \\
     \textbf{\scriptsize{3D-RecAE (ours)}}  & 0.550 & 0.521 & 0.411 & 0.441  & 0.550 & 0.382 \\
     \textbf{ \scriptsize{\nicknameRecGAN{} (ours)}}  & \textbf{0.553} & \textbf{0.529} & \textbf{0.416} & \textbf{0.449}  & \textbf{0.555} & \textbf{0.390} \\
\hline
\end{tabular}
\end{table}

\begin{table}[t]
\caption{Cross-category CE loss on testing dataset with cross viewing angles ($256^3$ voxel grids).}
\centering
\label{tab:cross_cat_ce_cv_256}
\tabcolsep=0.32cm
\begin{tabular}{c|cccccc}
	\hline
  	 	 & car & faucet & firearm & guitar & monitor & plane \\ 
    \hlineB{3}
     3D-EPN \cite{Dai2017b}    & 0.160 & 0.087 & 0.033 & 0.036  & 0.127 & 0.065 \\
     Varley \etal{} \cite{Varley2017}   & 0.171 & 0.123 & 0.028 & 0.030  & 0.136 & 0.043 \\
     Han \etal{} \cite{Han2017}  & 0.171 & 0.076 & 0.018 & 0.016  & 0.088 & 0.031 \\
     \textbf{\scriptsize{3D-RecAE (ours)}}  & 0.101 & 0.059 & 0.017 & 0.017  & 0.079 & 0.036 \\
     \textbf{ \scriptsize{\nicknameRecGAN{} (ours)}}  & \textbf{0.100} & \textbf{0.055} & \textbf{0.014} & \textbf{0.015}  & \textbf{0.074} & \textbf{0.031} \\
\hline
\end{tabular}
\end{table}

\begin{table}[h]
\caption{Cross-category IoU and CE loss of \nicknameRecGAN{} trained on individual category and then tested on the testing dataset with the same viewing angles ($256^3$ voxel grids).}
\centering
\label{tab:cross_cat_single_iou_ce_sv_256}
\tabcolsep=0.21cm
\begin{tabular}{c|cccc|cccc}
	\hline
    	& \multicolumn{4}{c|}{IoU} & \multicolumn{4}{c}{CE Loss} \\ \hline
  	   & bench  & chair & couch & table & bench  & chair & couch & table \\ 
    \hlineB{3}
    \begin{tabular}{@{}c@{}} \scriptsize{\textbf{Group 1}} \\ \scriptsize{\textbf{(trained on bench)}}\end{tabular}  & \textbf{0.580} & \underline{0.510} & \underline{0.507} & \underline{0.599} &\textbf{0.034} &\underline{0.110} &\underline{0.164} &\underline{0.062}\\
    
    \begin{tabular}{@{}c@{}}\scriptsize{Group 2} \\ \scriptsize{(trained on chair)}\end{tabular}  & 0.508 & \textbf{0.647} & 0.469 & 0.564 &0.048 &\textbf{0.060} &0.184 &0.069\\
    
    \begin{tabular}{@{}c@{}}\scriptsize{Group 3} \\ \scriptsize{(trained on couch)}\end{tabular}  & 0.429 & 0.504 & \textbf{0.753} & 0.437 & 0.070 & 0.105 &\textbf{0.066} &0.126\\
    
    \begin{tabular}{@{}c@{}}\scriptsize{Group 4} \\ \scriptsize{(trained on table)}\end{tabular}  & 0.510 & 0.509 & 0.402 & \textbf{0.679} &0.049 &0.111 &0.260 &\textbf{0.040}\\
\hline
\end{tabular}
\end{table}

\begin{table}[h]
\caption{Multi-category IoU and CE loss on real-world dataset ($256^3$ voxel grids).}
\centering
\label{tab:real_cat_iou_ce_256}
\tabcolsep=0.17cm
\begin{tabular}{c|cccc|cccc}
	\hline
    	& \multicolumn{4}{c|}{IoU} & \multicolumn{4}{c}{CE Loss} \\ \hline
  	                          & bench  & chair & couch & table & bench  & chair & couch & table \\ 
    \hlineB{3}
     3D-EPN \cite{Dai2017b}   & 0.162 & 0.190 & 0.508 & 0.140 &0.090 &0.158 &0.413 &0.187\\

     Varley \etal{} \cite{Varley2017}   & 0.118 & 0.152 & 0.433 & 0.075 &0.073 &0.155 &0.436 &0.191\\

     Han \etal{} \cite{Han2017}  & 0.166 & 0.164 & 0.235 & 0.146 & 0.083 & 0.167 &0.352 &0.194\\

     \textbf{\scriptsize{3D-RecAE (ours)}}  & 0.173 & 0.203 & 0.538 & 0.151 &0.065 &0.156 &0.318 &0.180\\

     \textbf{\scriptsize{\nicknameRecGAN{} (ours)}}  & \textbf{0.177} & \textbf{0.208} & \textbf{0.540} & \textbf{0.156} & \textbf{0.061} &\textbf{0.153}&\textbf{0.314} &\textbf{0.177} \\
\hline
\end{tabular}
\end{table}

\subsection{Real-world Experiment Results}
\begin{figure*}[h]
    \centering
    \includegraphics[width=0.98\textwidth]{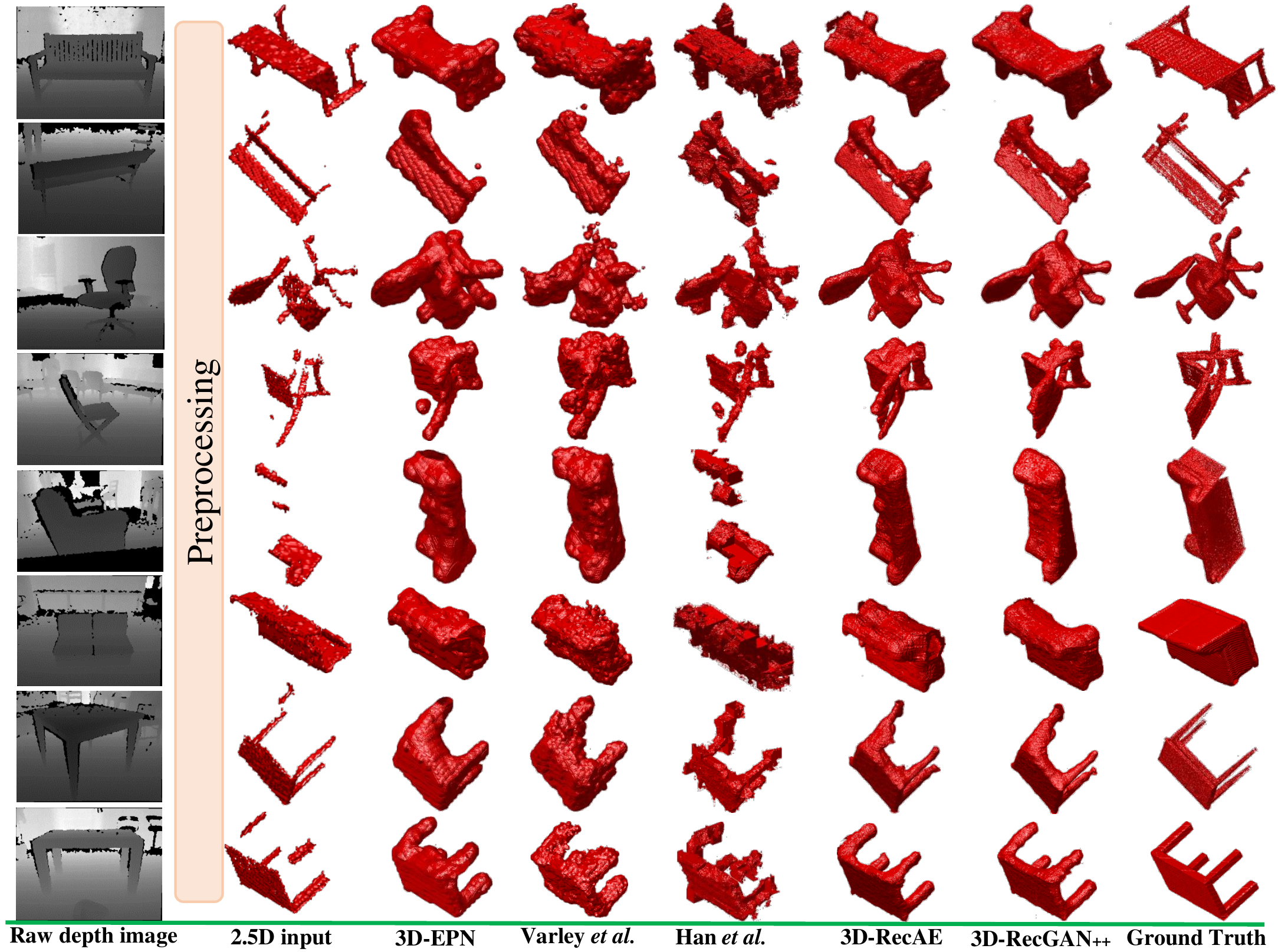}
    \caption{Qualitative results of real-world objects reconstruction from different approaches. The object instance is segmented from the raw depth image in preprocessing step.}
    \label{fig:ch3_multi_cat_real}
\end{figure*}

(1) \textbf{Results.} Lastly, in order to evaluate the domain adaptation capability of the networks, we train all networks on synthesized data of categories \{bench, chair, couch, table\}, and then test them on real-world data collected by a Microsoft Kinect camera. Table \ref{tab:real_cat_iou_ce_256} compares the IoU and CE loss of all approaches on the real-world dataset. Figure \ref{fig:ch3_multi_cat_real} shows some qualitative results for all methods.

(2) \textbf{Analysis.} There are two reasons why the IoU is significantly lower compared with testing on the synthetic dataset. First, the ground truth objects obtained from ElasticFusion are not as solid as the synthesized datasets. However, all networks predict dense and solid voxel grids, so the interior parts may not match though the overall object shapes are satisfactorily recovered as shown in Figure \ref{fig:ch3_multi_cat_real}. Secondly, the input 2.5D depth view from real-world dataset is noisy and incomplete, due to the limitation of the RGB-D sensor (\eg{} reflective surfaces, outdoor lighting). In some cases, the input 2.5D view does not capture the whole object and only contains a part of the object, which also leads to inferior reconstruction results (\eg{} the $5^{th}$ row in Figure \ref{fig:ch3_multi_cat_real}) and a lower IoU scores overall. However, our proposed network is still able to reconstruct reasonable 3D dense shapes given the noisy and incomplete 2.5D input depth views, while the competing algorithms (\eg{} Varley \etal{}) are not robust to real-world noise and unable to generate compelling results.

\subsection{Impact of Adversarial Learning}
(1) \textbf{Results.} In all above experiments, the proposed \nicknameRecGAN{} tends to outperform the ablated network 3D-RecAE which does not include the adversarial learning of GAN part. In all visualization of experiment results, the 3D shapes from \nicknameRecGAN{} are also more compelling than 3D-RecAE. To further quantitatively investigate how the adversarial learning improves the final 3D results comparing with 3D-RecAE, we calculate the mean precision and recall from the previous multi-category experiment results in Section \ref{sec:ch3_multi_cat}. Table \ref{tab:multi_cat_pre_rec_sv_256} compares the mean precision and recall of \nicknameRecGAN{}{} and 3D-RecAE on individual categories using the network trained on multiple categories.

(2) \textbf{Analysis.} It can be seen that the results of \nicknameRecGAN{} tend to consistently have higher precision scores than 3D-RecAE, which means \nicknameRecGAN{} has less false positive estimations. Therefore, the estimated 3D shapes from 3D-RecAE are likely to be `fatter' and `bigger', while \nicknameRecGAN{} tends to predict `thinner' shapes with much more shape details being exposed. Both \nicknameRecGAN{} and 3D-RecAE can achieve high recall scores (\ie{} above 0.8), which means both \nicknameRecGAN{} and 3D-RecAE are capable of estimating the major object shapes without too many false negatives. In other words, the ground truth 3D shape tends to be a subset of the estimated shape result.

Overall, with regard to experiments on per-category, multi-category, and cross-category experiments, our \nicknameRecGAN{} outperforms others by a large margin, although all other approaches can reconstruct reasonable shapes. In terms of the generality, Varley \etal{} \cite{Varley2017} and  Han \etal{} \cite{Han2017} are inferior because Varley \etal{} \cite{Varley2017} use a single fully connected layers, instead of 3D ConvNets, for shape generation which is unlikely to be general for various shapes, and Han \etal{} \cite{Han2017} apply LSTMs for shape blocks generation which is inefficient and unable to learn general 3D structures. However, our \nicknameRecGAN{} is superior thanks to the generality of the 3D encoder-decoder and the adversarial discriminator. Besides, the 3D-RecAE tends to over estimate the 3D shape, while the adversarial learning of \nicknameRecGAN{} is likely to remove the over-estimated parts, so as to leave the estimated shape to be clearer with more shape details.

\begin{table}[ht]
\caption{Multi-category  mean precision and recall on testing dataset with the same viewing angles ($256^3$ voxel grids).}
\centering
\label{tab:multi_cat_pre_rec_sv_256}
\tabcolsep=0.25cm
\begin{tabular}{c|cccc|cccc}
	\hline
    	& \multicolumn{4}{c|}{mean precision} & \multicolumn{4}{c}{mean recall} \\ \hline
  	          & bench  & chair & couch & table & bench  & chair & couch & table \\ 
    \hlineB{3}
     \scriptsize{3D-RecAE}  & 0.668 & 0.740 & 0.800 & 0.750 &\textbf{0.808} &0.818 &0.907 &0.845\\

     \scriptsize{\nicknameRecGAN{}}  & \textbf{0.680} & \textbf{0.747} &\textbf{0.804} & \textbf{0.754} & 0.804 &\textbf{0.820}&\textbf{0.910} &\textbf{0.853} \\
\hline
\end{tabular}
\end{table}

\subsection{Computation Analysis}
Table \ref{tab:para_time} compares the computation efficiency of all approaches regarding the total number of model parameters and the average time consumption to recover a single object. 

The model proposed by Han \etal{} \cite{Han2017} has the least number of parameters because most of the parameters are shared to predict different blocks of an object. Our \nicknameRecGAN{} has reasonable 167.1 millions parameters, which is on a similar scale to VGG-19 (\ie{} 144 millions) \cite{Simonyan2015}.

To evaluate the average time consumption for a single object reconstruction, we implement all networks in Tensorflow 1.2 and Python 2.7 with CUDA 8.0 and cuDNN 7.1 as the back-end driver and library. All models are tested on a single Titan X GPU in the same hardware and software environments. 3D-EPN \cite{Dai2017b} takes the shortest time to predict a $32^3$ object on GPU, while our \nicknameRecGAN{} only needs around 40 milliseconds to recover a dense $256^3$ object. Comparatively, Han \etal{} takes the longest GPU time to generate a dense object because of the time-consuming sequential processing of LSTMs. The low resolution objects predicted by 3D-EPN and Varley \etal{} are further upsampled to $256^3$ using existing SciPy library on a CPU server (Intel E5-2620 v4, 32 cores). It takes around 7 seconds to finish the upsampling for a single object.

\begin{table}[ht]
\caption{Comparison of model parameters and average time consumption to reconstruction a single object.}
\centering
\label{tab:para_time}
\tabcolsep=0.42cm
\begin{tabular}{c|c|c|c}
	\hline
    	& \multicolumn{1}{c|}{\begin{tabular}{@{}c@{}} \textbf{parameters} \\ (millions) \end{tabular}} & \multicolumn{1}{c|}{ \begin{tabular}{@{}c@{}} \textbf{GPU time} \\ \scriptsize{(milliseconds)} \end{tabular} }  &  \multicolumn{1}{c}{ \begin{tabular}{@{}c@{}} \textbf{predicted 3D shapes} \\ \scriptsize{(resolution)} \end{tabular} }\\ 
    \hlineB{3}
    3D-EPN \cite{Dai2017b}   & 52.4 & \textbf{15.8} & $32^3$ \\ 
 
    Varley \etal{} \cite{Varley2017} & 430.3 & 16.1 & $40^3$\\
 
    Han \etal{} \cite{Han2017} & \textbf{7.5} & 276.4 &\boldsymbol {$256^3$}\\

    \nicknameRecGAN{} & 167.1 & 38.9 & \boldsymbol{$256^3$} \\
\hline
\end{tabular}
\end{table}
\section{Conclusion}
In this chapter, we proposed a framework \nicknameRecGAN{}{} that reconstructs the full 3D structure of an object from an arbitrary depth view. By leveraging the generalization capabilities of 3D encoder-decoder and generative adversarial networks, our \nicknameRecGAN{} predicts dense and accurate 3D structures with fine details, outperforming the state of the art in single-view shape completion for individual object category. We further tested our network's ability to reconstruct multiple categories without providing any object class labels during training or testing, and it showed that our network is still able to predict precise 3D shapes. Furthermore, we investigated the network's reconstruction performance on unseen categories, showing that our proposed approach can also predict satisfactory 3D structures. Finally, our model is robust to real-world noisy data and can infer accurate 3D shapes although the model is purely trained on synthesized data. This confirms that our network has the capability of learning general 3D latent features of the objects, rather than simply fitting a function for the training datasets, and the adversarial learning of \nicknameRecGAN{} learns to add geometric details for estimated 3D shapes. In summary, our network only requires a single depth view to recover a dense and complete 3D shape with fine details.

Although our \nicknameRecGAN{} achieves the state of the art performance in 3D object reconstruction from a single depth view, it has limitations. Firstly, our network takes the volumetric representation of a single depth view as input, instead of taking a raw depth image. Therefore, a preprocessing of raw depth images is required for our network. However, in many application scenarios such as robot grasping, such preprocessing would be trivial and straightforward given the depth camera parameters. Secondly, the input depth view of our network only contains a clean object information without cluttered background. One possible solution is to leverage an existing segmentation algorithm such as Mask-RCNN \cite{He2017a} to clearly segment the target object instance from the raw depth view.

\chapter{Learning to Reconstruct 3D Objects from Multiple Views}
\label{chap:rec_obj_mv}
\section{Introduction}
The problem of recovering a geometric representation of the 3D world given a set of images is classically defined as multi-view 3D reconstruction in computer vision. Traditional pipelines such as Structure from Motion (SfM) \cite{Ozyesil2017} and visual Simultaneous Localization and Mapping (vSLAM) \cite{Cadena2016} typically rely on hand-crafted feature extraction and matching across multiple views to reconstruct the underlying 3D model. However, if the multiple viewpoints are separated by large baselines, it can be extremely challenging for the feature matching approach due to significant changes of appearance or self occlusions \cite{Lowe2004}. Furthermore, the reconstructed 3D shape is usually a sparse point cloud without geometric details.

Recently, a number of deep learning approaches, such as 3D-R2N2 \cite{Chan2016}, LSM \cite{Kar2017}, DeepMVS \cite{Huang2018} and RayNet \cite{Paschalidou2018} have been proposed to estimate the 3D dense shape from multiple images and have shown encouraging results. Both 3D-R2N2 \cite{Chan2016} and LSM \cite{Kar2017} formulate multi-view reconstruction as a sequence learning problem, and leverage recurrent neural networks (RNNs), particularly GRUs, to fuse the multiple deep features extracted by a shared encoder from input images. However, there are three limitations. First, the recurrent network is permutation variant, \ie{} different permutations or orderings of the input image sequence give different reconstruction results \cite{Vinyals2016a}. Therefore, inconsistent 3D shapes are estimated from the same image set with different permutations. Second, it is difficult to capture long-term dependencies in the sequence because of the gradient vanishing or exploding \cite{Bengio1994,Kolen2001}, so the estimated 3D shapes are unlikely to be refined even if more images are given during training and testing. Third, the RNN unit is inefficient as each element of the input sequence must be sequentially processed without parallelization \cite{Martin2018}, so it is time-consuming to generate the final 3D shape given a sequence of images. 

The recent DeepMVS \cite{Huang2018} applies max pooling to aggregate deep features across a set of unordered images for multi-view stereo reconstruction, while RayNet \cite{Paschalidou2018} adopts average pooling to aggregate the deep features corresponding to the same voxel from multiple images to recover a dense 3D model. The very recent GQN \cite{Eslami2018} uses sum pooling to aggregate an arbitrary number of orderless images for 3D scene representation. Although max, average and summation poolings do not suffer from the above limitations of the RNN, they tend to be `hard attentive', since they only capture the max/mean values or the summation without learning to attentively preserve the useful information. In addition, the above pooling based neural nets are usually optimized with a specific number of input images during training, therefore they are not robust or general to a dynamic number of input images during testing. This critical issue is also observed in GQN \cite{Eslami2018}.

\begin{figure*}[t]
\centering
   \includegraphics[width=1\linewidth]{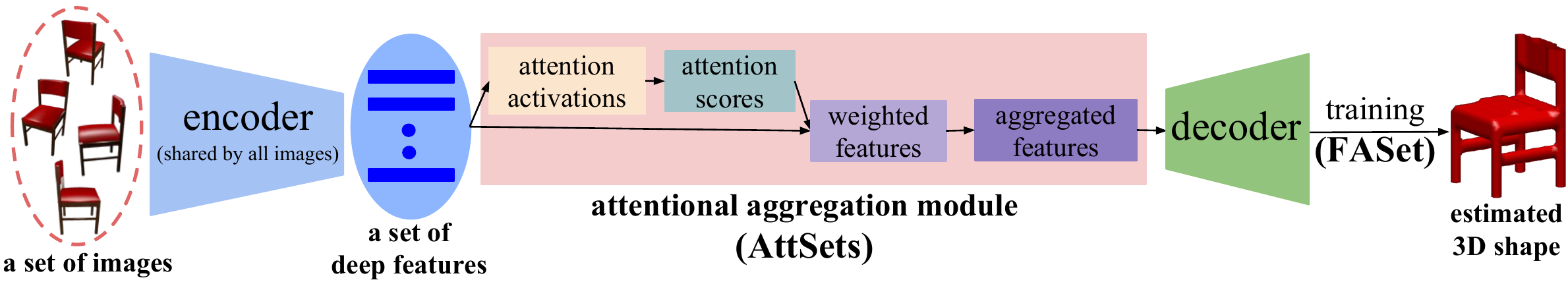}
\caption{Overview of our attentional aggregation module for multi-view 3D reconstruction. A set of $N$ images is passed through a common encoder to derive a set of deep features, one element for each image. The network is trained with our \nicknameFASet{} algorithm.}
\label{fig:ch4_attsets_view}
\end{figure*}

In this paper, we introduce a simple yet efficient attentional aggregation module, named \textbf{\nicknameAttSets{}}\footnote{\small{Code is available at \textit{https://github.com/Yang7879/AttSets}}}. It can be easily included in an existing multi-view 3D reconstruction network to aggregate an arbitrary number of elements of a deep feature set. Inspired by the attention mechanism, which shows great success in natural language processing \cite{Bahdanau2015,Raffel2016}, image captioning \cite{Xu2015b}, \textit{etc.}, we design a feed-forward neural module that can automatically learn to aggregate each element of the input deep feature set. In particular, as shown in Figure \ref{fig:ch4_attsets_view}, given a variable sized deep feature set, which encodes view-invariant visual representations from a shared encoder \cite{Paschalidou2018}, our \nicknameAttSets{} module firstly learns an \textbf{attention activation} for each latent feature through a standard neural layer (\eg{}  a fully connected layer, a 2D or 3D convolutional layer), after which an \textbf{attention score} is computed for the corresponding feature. Subsequently, the attention scores are simply multiplied by the original elements of the deep feature set, generating a set of \textbf{weighted features}. Lastly, the weighted features are summed across different elements of the deep feature set, producing a fixed size vector of \textbf{aggregated features} which are then fed into a decoder to estimate 3D shapes. Basically, this \nicknameAttSets{} module can be seen as a natural extension of sum pooling into a ``weighted'' sum pooling with learnt feature-specific weights. \nicknameAttSets{} shares similar concepts with the concurrent work \cite{Ilse2018}, but it does not require the additional gating mechanism in \cite{Ilse2018}. Notably, our simple feed-forward design allows the attention module to be separately trainable according to the property of its gradients.

In addition, we propose a new \textbf{Feature-Attention Separate training (FASet)} algorithm that elegantly decouples the base encoder-decoder (to learn deep features) from the \nicknameAttSets{} module (to learn attention scores for features). This allows the \nicknameAttSets{} module to learn desired attention scores for deep feature sets and forces the \nicknameAttSets{} based neural networks to be robust and general to dynamic sized deep feature sets. Basically, in the proposed training algorithm, the base encoder-decoder neural layers are only optimized when the number of input images is \textbf{1}, while the \nicknameAttSets{} module is only optimized where there are more than \textbf{1} input images. Eventually, the whole optimized \nicknameAttSets{} based neural network achieves superior performance with a large number of input images, while simultaneously being extremely robust and able to generalize to a small number of input images, even to a single image in the most extreme case. Comparing with the widely used feed-forward attention mechanisms for visual recognition \cite{JieHu2018,Rodriguez2018,Liu2018e,Sarafianos2018,Girdhar2017a}, our \nicknameFASet{} algorithm is the first to investigate and improve the robustness of attention modules to dynamically sized input feature sets, whilst existing works are only applicable to fixed sized input data.

Overall, our novel \nicknameAttSets{} module and \nicknameFASet{} algorithm are distinguished from all existing aggregation approaches in three ways. 1) Compared with RNN approaches, \nicknameAttSets{} is permutation invariant and computationally efficient. 2) Compared with the widely used pooling operations, \nicknameAttSets{} learns to attentively select and weigh important deep features, thereby being more effective to aggregate useful information for better 3D reconstruction. 3) Compared with existing visual attention mechanisms, our \nicknameFASet{} algorithm enables the whole network to be general to variable sized sets, being more robust and suitable for realistic multi-view 3D reconstruction scenarios where the number of input images usually varies dramatically.

Our key contributions are:
\begin{itemize}
\item We propose an efficient feed-forward attention module, \nicknameAttSets{}, to effectively aggregate deep feature sets. Our design allows the attention module to be separately optimizable according to the property of the gradients of \nicknameAttSets{}. 
\item We propose a new two-stage training algorithm, \nicknameFASet{}, to decouple the base encoder/decoder and the attention module, driving the whole network to be robust and general to an arbitrary number of input images.
\item We conduct extensive experiments on multiple public datasets, demonstrating consistent improvement over existing aggregation approaches for 3D object reconstruction from either single or multiple views.
\end{itemize}
\section{Attentional Aggregation Module}
\subsection{Problem Definition}
We consider the problem of aggregating an arbitrary number of elements of a set $\mathcal{A}$ into a fixed single output $\boldsymbol{y}$. Each element of set $\mathcal{A}$ is a feature vector extracted from a shared encoder, and the fixed dimension output $\boldsymbol{y}$ is fed into a subsequent decoder, such that the whole network can process an arbitrary number of input elements. 

Given $N$ elements in the input deep feature set $\mathcal{A} = \{\boldsymbol{x}_1, \boldsymbol{x}_2, \cdots, \boldsymbol{x}_N\}$, $\boldsymbol{x}_n \in \mathbb{R}^{1\times D}$, where $N$ is an arbitrary value, while $D$ is fixed for a specific encoder, and the output $\boldsymbol{y} \in \mathbb{R}^{1\times D}$, which is then fed into the subsequent decoder, our task is to design an aggregation function $f$ with learnable weights $\boldsymbol{\mathit{W}}$: $\boldsymbol{y} = f(\mathcal{A}, \boldsymbol{\mathit{W}})$, which should be permutation invariant, \ie{} for any permutation $\pi$:

\begin{equation}
f(\{ \boldsymbol{x}_1, \cdots, \boldsymbol{x}_N \}, \boldsymbol{\mathit{W}}) = f(\{\boldsymbol{x}_{\pi(1)}, \cdots, \boldsymbol{x}_{\pi(N)} \}, \boldsymbol{\mathit{W}})
\end{equation}

The common pooling operations, \eg{} max/mean/sum, are the simplest instantiations of function $f$ where $\boldsymbol{\mathit{W}} \in \emptyset$. However, these pooling operations are predefined and only capture partial information. 

\begin{figure*}[t]
\centering
   \includegraphics[width=1\linewidth]{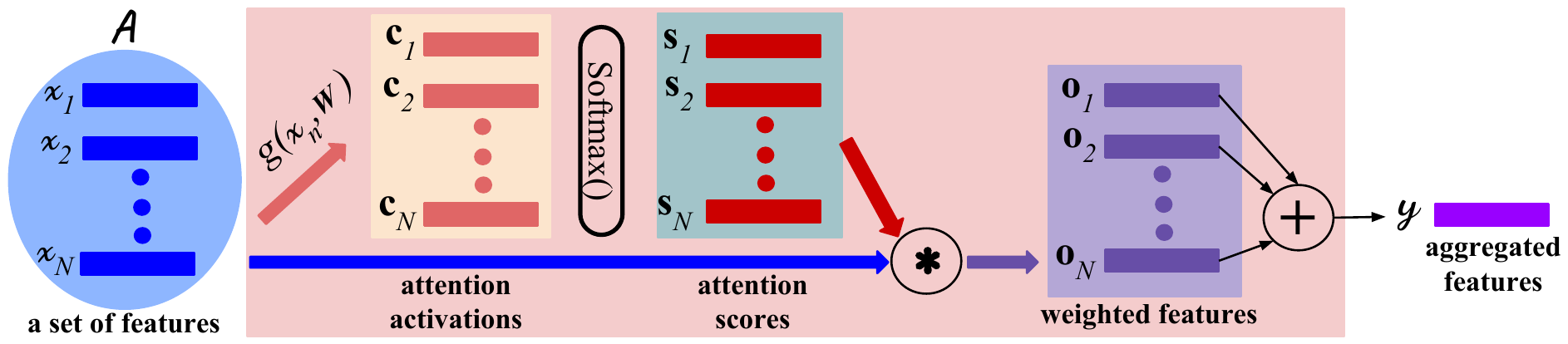}
\caption{Attentional aggregation module on sets. This module learns an attention score for each individual deep feature.}
\label{fig:ch4_attsets_f}
\end{figure*}

\subsection{Attentional Aggregation}\label{sec:ch4_attagg}
The key idea of our \nicknameAttSets{} module is to learn an attention score for each latent feature of the whole deep feature set. In this chapter, `each latent feature' refers to each entry of an individual element of the feature set, with an individual element usually represented by a latent vector, \ie{} $\boldsymbol{x}_n$. The learnt scores can be regarded as a mask that automatically selects useful latent features across the set. The selected features are then summed across multiple elements of the set.

As shown in Figure \ref{fig:ch4_attsets_f}, given a set of features $\mathcal{A} = \{\boldsymbol{x}_1, \boldsymbol{x}_2, \cdots, \boldsymbol{x}_N\}$, $\boldsymbol{x}_n \in \mathbb{R}^{1\times D}$, \nicknameAttSets{} aims to fuse it into a fixed dimensional output $\boldsymbol{y}$, where $\boldsymbol{y} \in \mathbb{R}^{1\times D}$.

To build the \nicknameAttSets{} module, we first feed each element of the feature set $\mathcal{A}$ into a shared function $g$ which can be a standard neural layer, \ie{} a linear transformation layer without any non-linear activation functions. Here we use a fully connected layer as an example, and the bias term is dropped for simplicity. The output of function $g$ when applied to all inputs is a set of learnt attention activations $\mathcal{C}=\{\boldsymbol{c}_1, \boldsymbol{c}_2, \cdots, \boldsymbol{c}_N\}$, where
\begin{equation}
\begin{gathered}
\boldsymbol{c}_n = g(\boldsymbol{x}_n, \boldsymbol{W}) = \boldsymbol{x}_n\boldsymbol{W},\\ 
(\boldsymbol{x}_n \in \mathbb{R}^{1\times D}, \quad \boldsymbol{W} \in \mathbb{R}^{D\times D}, \quad \boldsymbol{c}_n \in \mathbb{R}^{1\times D} )
\end{gathered}
\end{equation}

Secondly, the learnt attention activations are normalized across the $N$ elements of the set, computing a set of attention scores $\mathcal{S}=\{\boldsymbol{s}_1, \boldsymbol{s}_2, \cdots, \boldsymbol{s}_N\}$. We choose $softmax$ as the normalization operation, so the attention scores for the $n^{th}$ feature element are
\begin{equation}
\begin{gathered}
\boldsymbol{s}_n = [s^1_n, s^2_n, \cdots, s^d_n, \cdots, s^D_n],\\
s^d_n = \frac{e^{c^d_n}}{\sum^N_{j=1}{ e^{c^d_j}}}, \qquad \textit{where $c^d_n$, $c^d_j$ are the $d^{th}$ entry of $\boldsymbol{c}_n$, $\boldsymbol{c}_j$.}
\end{gathered}
\end{equation}

Thirdly, the computed attention scores $\mathcal{S}$ are multiplied by their corresponding original feature set $\mathcal{A}$, generating a new set of deep features, denoted as weighted features $\mathcal{O}=\{\boldsymbol{o}_1, \boldsymbol{o}_2, \cdots, \boldsymbol{o}_N\}$, where
\begin{align}
\boldsymbol{o}_n = \boldsymbol{x}_n * \boldsymbol{s}_n
\end{align}

Lastly, the set of weighted features $\mathcal{O}$ are summed up across the total $N$ elements to get a fixed size feature vector, denoted as $\boldsymbol{y}$, where
\begin{equation}
\begin{gathered}
\boldsymbol{y} =[y^1, y^2, \cdots, y^d, \cdots, y^D],\\
y^d = \sum^N_{n=1}o^d_n, \qquad \textit{where $o^d_n$ is the $d^{th}$ entry of $\boldsymbol{o}_n$.}
\end{gathered}
\end{equation}

In the above formulation, we show how \nicknameAttSets{} gradually aggregates a set of $N$ feature vectors $\mathcal{A}$ into a single vector $\boldsymbol{y}$, where $\boldsymbol{y} \in \mathbb{R}^{1\times D}$. 

\subsection{Permutation Invariance}
The output of \nicknameAttSets{} module $\boldsymbol{y}$ is permutation invariant with regard to the input deep feature set $\mathcal{A}$. Here is the simple proof. 
\begin{equation}
\label{eq:yd}
[y^1, \cdots y^d \cdots, y^D] = f(\{\boldsymbol{x}_1, \cdots \boldsymbol{x}_n \cdots, \boldsymbol{x}_N\}, \boldsymbol{W})
\end{equation}

In Equation \ref{eq:yd}, the $d^{th}$ entry of the output $\boldsymbol{y}$ is computed as follows:
\begin{align}
\label{eq:yd2}
y^d &= \sum^N_{n=1}o^d_n  
= \sum^N_{n=1}(x^d_n*s^d_n) \nonumber \\
&= \sum^N_{n=1}\left( x^d_n * \frac{e^{c^d_n}}{\sum^N_{j=1} e^{c^d_j}  }  \right) \nonumber \\
&= \sum^N_{n=1}\left( x^d_n * \frac{e^{(\boldsymbol{x}_n\boldsymbol{w}^d)} } 
{ \sum^N_{j=1}e^{(\boldsymbol{x}_j\boldsymbol{w}^d})} \right)  \nonumber \\
&=\frac{\sum^N_{n=1} \left( x^d_n * e^ {(\boldsymbol{x}_n\boldsymbol{w}^d)} \right) }
{\sum^N_{j=1}e^{(\boldsymbol{x}_j\boldsymbol{w}^d)}}, \qquad 
\end{align}
\textit{where} $\boldsymbol{w}^d$ is the $d^{th}$ column of the weights $\boldsymbol{W}$. In Equation \ref{eq:yd2}, both the denominator and numerator are a summation of a permutation equivariant term. Therefore the value $y^d$, and also the full vector $\boldsymbol{y}$, is also invariant to different permutations of the deep feature set $\mathcal{A}=\{\boldsymbol{x}_1, \boldsymbol{x}_2, \cdots, \boldsymbol{x}_n, \cdots, \boldsymbol{x}_N\}$ \cite{Zaheer2017}. 

\begin{figure*}[t]
\centering
   \includegraphics[width=1.0\linewidth]{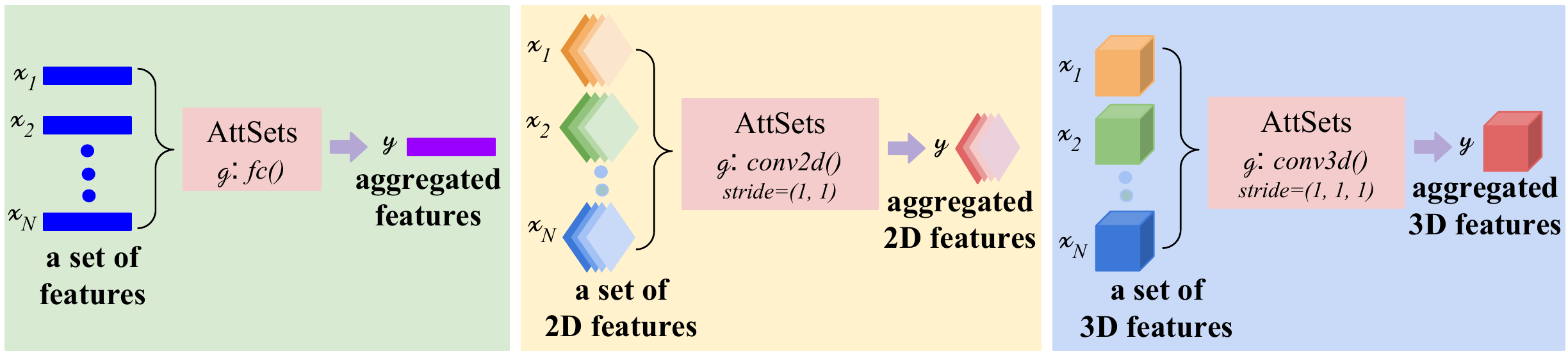}
\caption{Different implementations of \nicknameAttSets{} with a fully connected layer, 2D ConvNet, and 3D ConvNet. These three variants of \nicknameAttSets{} can be flexibly plugged into different locations of an existing encoder-decoder network.}
\label{fig:ch4_attsets_imp}
\end{figure*}

\subsection{Implementation}\label{sec:ch4_impl}
In Section \ref{sec:ch4_attagg}, we described how our \nicknameAttSets{} aggregates an arbitrary number of vector features into a single vector, where the attention activation learning function $g$ embeds a fully connected ($fc$) layer. \nicknameAttSets{} can also be easily implemented with both 2D and 3D convolutional neural layers to aggregate both 2D and 3D deep feature sets, thus being flexible to be easily included into a 2D encoder/decoder or 3D encoder/decoder. Particularly, as shown in Figure \ref{fig:ch4_attsets_imp}, to aggregate a set of 2D features, \ie{} a tensor of $(width\times height\times channels)$, the attention activation learning function $g$ embeds a standard $conv2d$ layer with a stride of $(1\times 1)$. Similarly, to fuse a set of 3D features, \ie{} a tensor of $(width\times height\times depth\times channels)$, the function $g$ embeds a standard $conv3d$ layer with a stride of $(1\times 1\times 1)$. For the above $conv2d$/$conv3d$ layer, the filter size can be 1, 3 or many. The larger the filter size, the larger the local spatial area the learnt attention score is considered to be correlated with.

Instead of embedding a single neural layer, the function $g$ is also flexible to include multiple layers, with the restriction that the tensor shape of the output of function $g$ is required to be consistent with the input element $\boldsymbol{x}_n$. This guarantees each individual feature of the input set $\mathcal{A}$ will be associated with a learnt and unique weight. For example, a standard 2-layer or 3-layer ResNet module \cite{He2016b} could be a candidate of the function $g$. The more the layers that $g$ embeds, the greater the capability of the \nicknameAttSets{} module is expected to be.

Compared with $fc$ enabled \nicknameAttSets{}, the $conv2d$ or $conv3d$ based \nicknameAttSets{} variants tend to have fewer learnable parameters. Note that both the $conv2d$ and $conv3d$ based \nicknameAttSets{} are still permutation invariant, as the function $g$ is shared across all elements of the deep feature set and it does not depend on the order of the elements \cite{Zaheer2017}.
\section{Feature Attention Separate Training}
\subsection{Motivation}\label{sec:ch4_optim_motiv}
Our \nicknameAttSets{} module can be included in an existing encoder-decoder multi-view 3D reconstruction network, replacing the RNN units or pooling operations. Essentially, in an \nicknameAttSets{} enabled encoder-decoder net, the encoder-decoder serves as the base architecture to learn visual features for shape estimation, while the \nicknameAttSets{} module learns to assign different attention scores to combine those features. As such, the base network tends to have robustness and generality with regard to different input image content, while the \nicknameAttSets{} module tends to be generally applicable to an arbitrary number of input images. 

However, to achieve this robustness is not straightforward. The standard end-to-end joint optimization approach is unable to force that the base encoder-decoder and \nicknameAttSets{} are able to learn visual features and the corresponding scores separately, because there are no explicit feature score labels available to directly supervise the \nicknameAttSets{} module.

Let us revisit Equation \ref{eq:yd2} below and draw insights from it.
\begin{align}
\label{eq:yd2_2}
y^d =\frac{\sum^N_{n=1} \left( x^d_n * e^ {(\boldsymbol{x}_n\boldsymbol{w}^d)} \right) }
{\sum^N_{j=1}e^{(\boldsymbol{x}_j\boldsymbol{w}^d)}}
\end{align}
\textit{where} $N$ is the size of an arbitrary input set and $\boldsymbol{w}^d$ are the \nicknameAttSets{} parameters to be optimized. If $N$ is 1, then the equation can be simplified as
\begin{align}
\label{eq:yd2_3}
y^d &= x^d_n \\
\frac{\partial y^d}{\partial x^d_n} =1, \qquad   \label{eq:yd2_33}
\frac{\partial y^d}{\partial \boldsymbol{w}^d} &=\boldsymbol{0}, \qquad N=1 
\end{align}

This shows that none of the parameters, \ie{} $\boldsymbol{w}^d$, of the \nicknameAttSets{} module will be optimized when the size of the input feature set is 1. 

However, if $N>1$, Equation \ref{eq:yd2_2} is unable to be simplified to Equation \ref{eq:yd2_3}. Therefore, 
\begin{align}
\label{eq:yd2_44}
\frac{\partial y^d}{\partial x^d_n} \neq 1, \qquad
\frac{\partial y^d}{\partial \boldsymbol{w}^d } \neq \boldsymbol{0}, \qquad N>1
\end{align}
This shows that both the parameters of \nicknameAttSets{} and the base encoder-decoder layers will be optimized simultaneously, if the whole network is trained in the standard end-to-end fashion.

Here arises the critical issue. When $N>1$, all derivatives of the parameters in the \textbf{encoder} are different from the derivatives when $N=1$ due to the chain rule of differentiation applied backwards from $\frac{\partial y^d}{\partial x^d_n}$. Put simply, the derivatives of encoder are \textit{$N$-dependent}. As a consequence, the encoded visual features and the learnt attention scores would be \textit{$N$-biased} if the whole network is jointly trained. This biased network is unable to generalize to an arbitrary value of $N$ during testing.

To illustrate the above issue, assuming the base encoder-decoder and the \nicknameAttSets{} module are jointly trained given $5$ images to reconstruct every object, the base encoder will be eventually optimized towards $5$-view object reconstruction during training. The trained network can indeed perform well given 5 views during testing, but it is unable to predict a satisfactory object shape given only 1 image.

To alleviate the above problem, a naive approach is to enumerate various values of $N$ during the joint training, such that the final optimized network can be somewhat robust and general to arbitrary $N$ during testing. However, this approach would inevitably optimize the encoder to learn the \textit{mean} features of input data for varying $N$. The overall performance will hence not be optimal. In addition, it is impractical and also time-consuming to enumerate all values of $N$ during training.

\begin{algorithm*}[h]
\caption{ Feature-Attention Separate training of an \nicknameAttSets{} enabled network. $M$ is batch size, $N$ is image number.
}
\label{alg:faset}
\begin{algorithmic} 
\vspace{2mm}
\STATE{\textbf{Stage 1:}}
\FOR{number of training iterations}{}
    \STATE{$\bullet$ Sample $M$ sets of images $\{ \mathcal{I}_1, \cdots, \mathcal{I}_m, \cdots, \mathcal{I}_M \}$ and sample $N$ images for each set, \ie{} $\mathcal{I}_m = \{\boldsymbol{i}_m^1, \cdots, \boldsymbol{i}_m^n, \cdots, \boldsymbol{i}_m^N \}$. 
 Sample $M$ 3D shape labels $\{ \boldsymbol{v}_1, \cdots, \boldsymbol{v}_m, \cdots, \boldsymbol{v}_M \}$.}
    
    \vspace{2mm}
    \STATE{$\bullet$ Update the base network by descending its stochastic gradient:\\ 
        \[
        	\nabla_{\Theta_{base}} \frac{1}{MN} \sum_{m=1}^M \sum_{n=1}^N \left[ \ell(\boldsymbol{\hat{v}}_m^n, \boldsymbol{v}_m) \right], \text{ $where$ $\boldsymbol{\hat{v}}_m^n$ is the estimated 3D shape of image $\boldsymbol{i}_m^n$.}
        \]
        }
    \ENDFOR
\STATE{\textbf{Stage 2:}}
\FOR{number of training iterations}{}
    \STATE{$\bullet$ Sample $M$ sets of images $\{ \mathcal{I}_1, \cdots, \mathcal{I}_m, \cdots, \mathcal{I}_M \}$ and sample $N$ images for each set, \ie{} $\mathcal{I}_m = \{\boldsymbol{i}_m^1, \cdots, \boldsymbol{i}_m^n, \cdots, \boldsymbol{i}_m^N \}$. 
 Sample $M$ 3D shape labels $\{ \boldsymbol{v}_1, \cdots, \boldsymbol{v}_m, \cdots, \boldsymbol{v}_M \}$.}
 
   \vspace{2mm}
    \STATE{$\bullet$ Update the \nicknameAttSets{} module by descending its stochastic gradient:\\
        \[
        	\nabla_{\Theta_{att}} \frac{1}{M} \sum_{m=1}^M \left[ \ell(\boldsymbol{\hat{v}}_m, \boldsymbol{v}_m) \right],         \text{$where$ $\boldsymbol{\hat{v}}_m$ is the estimated 3D shape of the image set $\mathcal{I}_m$.}
        \]
        }
\ENDFOR
  \vspace{-1mm}
  \\The gradient-based updates can use any gradient optimization algorithm.
\end{algorithmic}
\end{algorithm*}

\subsection{Algorithm}
To resolve the critical issue discussed in Section \ref{sec:ch4_optim_motiv}, we propose a \textbf{Feature-Attention Separate training (FASet)} algorithm that decouples the base encoder-decoder and the \nicknameAttSets{} module, enabling the base encoder-decoder to learn robust deep features and the \nicknameAttSets{} module to learn the desired attention scores for the feature sets.

In particular, the base encoder-decoder neural layers are only optimized when a single input image is supplied, while the \nicknameAttSets{} module is only optimized where there are more than one input images. In this regard, the parameters of the base encoding layers have consistent derivatives over the whole training stage, thus being steadily optimized. In the meantime, the \nicknameAttSets{} module would be optimized solely based on multiple elements of learnt visual features from the shared encoder. 

The trainable parameters of the base encoder-decoder are denoted as $\boldsymbol{\Theta}_{base}$, and the trainable parameters of \nicknameAttSets{} module are denoted as $\boldsymbol{\Theta}_{att}$, and the loss function of the whole network is represented by $\ell$ which is determined by the specific supervision signal of the base network. Our \nicknameFASet{} is shown in Algorithm \ref{alg:faset}. It can be seen that $\boldsymbol{\Theta}_{base}$ and $\boldsymbol{\Theta}_{att}$ are completely decoupled from one another, thus being separately optimized in two stages. In stage 1, the $\boldsymbol{\Theta}_{base}$ is firstly well optimized until convergence, which guarantees the base encoder-decoder is able to learn robust and general visual features. In stage 2, the $\boldsymbol{\Theta}_{att}$ is optimized to learn attention scores for individual visual features. Basically, this module learns to select and weigh important deep features automatically. 

In \nicknameFASet{} algorithm, once the $\boldsymbol{\Theta}_{base}$ is well optimized in stage 1, it is not necessary to train it again, since the two-stage algorithm guarantees that optimizing $\boldsymbol{\Theta}_{base}$ is agnostic to the attention module. The \nicknameFASet{} algorithm is a crucial component to maintain the superior robustness of the \nicknameAttSets{} module, as is shown in Section \ref{sec:ch4_sig_faset}. Without it, the feed-forward attention mechanism is ineffective with respect to dynamically sized input sets.
\section{Experiments}
\subsection{Base Networks}
\begin{figure}[t]
\centering
   \includegraphics[width=1\linewidth]{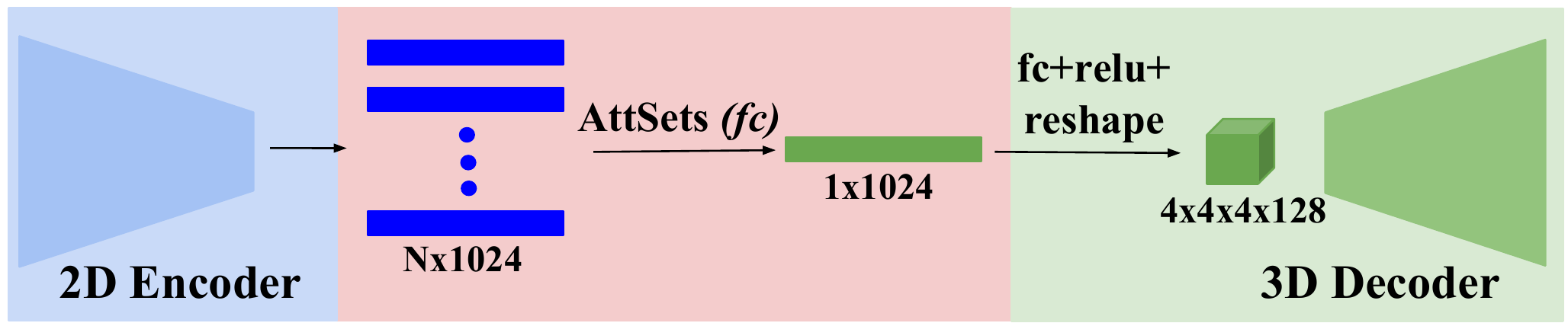}
\caption{The architecture of Base$_{\textrm{r2n2}}$-AttSets for multi-view 3D reconstruction network. The base encoder-decoder is the same as 3D-R2N2.}
\label{fig:ch4_attsets_base_r2n2_fc}
\end{figure}

\begin{figure}[t]
\centering
   \includegraphics[width=1\linewidth]{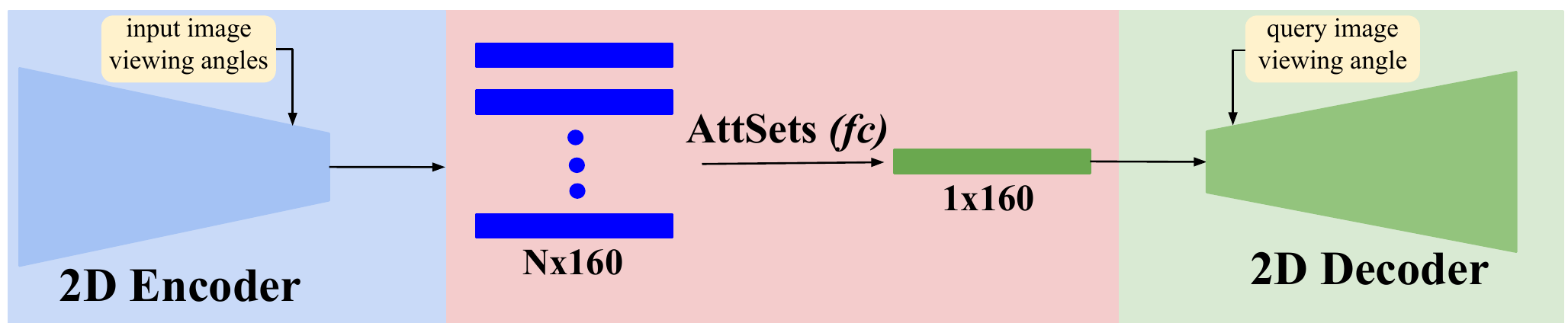}
\caption{The architecture of Base$_{\textrm{silnet}}$-AttSets for multi-view 3D shape learning. The base encoder-decoder is the same as SilNet.}
\label{fig:ch4_attsets_base_silnet_fc}
\end{figure}

To evaluate the performance and various properties of \nicknameAttSets{}, we choose the encoder-decoders of 3D-R2N2 \cite{Chan2016} and SilNet \cite{Wiles2017} as two base networks. 
\begin{itemize}
\item Encoder-decoder of 3D-R2N2. The original 3D-R2N2 consists of (1) a shared ResNet-based 2D encoder which encodes a size of $127\times 127 \times 3$ images into $1024$ dimensional latent vectors, (2) a GRU module which fuses $N$ $1024$ dimensional latent vectors into a single $4\times 4\times 4\times 128$ tensor, and (3) a ResNet-based 3D decoder which decodes the single tensor into a $32\times 32\times 32$ voxel grid representing the 3D shape. Figure \ref{fig:ch4_attsets_base_r2n2_fc} shows the architecture of \nicknameAttSets{} based multi-view 3D reconstruction network where the only difference is that the original GRU module is replaced by \nicknameAttSets{} in the middle. This network is called Base$_{\textrm{r2n2}}$-AttSets.
\item Encoder-decoder of SilNet. The original SilNet consists of (1) a shared 2D encoder which encodes a size of $127\times 127\times 3$ images together with image viewing angles into $160$ dimensional latent vectors, (2) a max pooling module which aggregates $N$ latent vectors into a single vector, and (3) a 2D decoder which estimates an object silhouette ($57\times 57$) from the single latent vector and a new viewing angle. Instead of being explicitly supervised by 3D shape labels, SilNet aims to implicitly learn a 3D shape representation from multiple images via the supervision of 2D silhouettes. Figure \ref{fig:ch4_attsets_base_silnet_fc} shows the architecture of \nicknameAttSets{} based SilNet where the only difference is that the original max pooling is replaced by \nicknameAttSets{} in the middle. This network is called Base$_{\textrm{silnet}}$-AttSets.
\end{itemize}

\subsection{Competing Approaches}
We compare our \nicknameAttSets{} and \nicknameFASet{} with three groups of competing approaches. Note that all the following competing approaches are connected at the same location of the base encoder-decoder shown in the pink block of Figures \ref{fig:ch4_attsets_base_r2n2_fc} and \ref{fig:ch4_attsets_base_silnet_fc}, with the same network configurations and training/testing settings.

\begin{itemize}
\item RNNs. The original 3D-R2N2 makes use of the \textbf{GRU} \cite{Chan2016,Kar2017} unit for feature aggregation and serves as a solid baseline.
\item First-order poolings. The widely used \textbf{max}/\textbf{mean}/ \textbf{sum} pooling operations \cite{Huang2018,Paschalidou2018,Eslami2018} are all implemented for comparison.
\item Higher-order poolings. We also compare with the state-of-the-art higher-order pooling approaches, including bilinear pooling (\textbf{BP}) \cite{Lin2015}, and the very recent \textbf{MHBN} \cite{Yu2018a} and \textbf{SMSO} poolings \cite{Yu2018b}. 
\end{itemize}

\subsection{Datasets}
All approaches are evaluated on four large open datasets.
\begin{itemize}
\item ShapeNet$_{\textrm{r2n2}}$ Dataset \cite{Chan2016}. The released 3D-R2N2 dataset consists of $13$ categories of $43,783$ common objects with synthesized RGB images from the large scale ShapeNet 3D repository \cite{Chang2015}. For each 3D object, $24$ images are rendered from different viewing angles circling around each object. The train/test dataset split is $0.8:0.2$.
\item ShapeNet$_{\textrm{lsm}}$ Dataset \cite{Kar2017}. Both LSM and 3D-R2N2 datasets are generated from the same 3D ShapeNet repository \cite{Chang2015}, \ie{} they have the same ground truth labels regarding the same object. However, the ShapeNet$_{\textrm{lsm}}$ dataset has totally different camera viewing angles and lighting sources for the rendered RGB images. Therefore, we use the ShapeNet$_{\textrm{lsm}}$ dataset to evaluate the robustness and generality of all approaches. All images of ShapeNet$_{\textrm{lsm}}$ dataset are resized from $224\times 224$ to $127\times 127$ through linear interpolation.
\item ModelNet40 Dataset. ModelNet40 \cite{Wu2015} consists of $12,311$ objects belonging to 40 categories. The 3D models are split into $9,843$ training samples and $2,468$ testing samples. For each 3D model, it is voxelized into a $30\times30\times30$ shape in \cite{Qi2016a}, and 12 RGB images are rendered from different viewing angles.
All 3D shapes are zero-padded to be $32\times32\times32$, and the images are linearly resized from $224\times224$ to $127\times127$ for training and testing.
\item Blobby Dataset \cite{Wiles2017}. It contains $11,706$ blobby objects. Each object has $5$ RGB images paired with viewing angles and the corresponding silhouettes, which are generated from Cycles in Blender under different lighting sources and texture models. 
\end{itemize}

\subsection{Metrics}
The explicit 3D voxel reconstruction performance of Base$_{\textrm{r2n2}}$-AttSets and the competing approaches is evaluated on three datasets: ShapeNet$_{\textrm{r2n2}}$, ShapeNet$_{\textrm{lsm}}$ and ModelNet40. We use the mean Intersection-over-Union (IoU) \cite{Chan2016} between predicted 3D voxel grids and their ground truth as the metric. The IoU for an individual voxel grid is formally defined as follows:
\begin{align*}
IoU = \frac{\sum_{i=1}^{L} \left[  I (h_i>p) * I(\bar{h_i}) \right] }{ \sum_{i=1}^{L}   \left[I  \left( I(h_{i} >p) + I(\bar{h_i}) \right) \right] } 
\end{align*}
$where$ $I(\cdot)$ is an indicator function, $h_{i}$ is the predicted value for the $i^{th}$ voxel, $\bar{h_i}$ is the corresponding ground truth, $p$ is the threshold for voxelization, $L$ is the total number of voxels in a whole voxel grid. As there is no validation split in the above three datasets, to calculate the IoU scores, we independently search the optimal binarization threshold value from $0.2 \sim 0.8$ with a step $0.05$ for all approaches for fair comparison. In our experiments, we found that all optimal thresholds of different approaches end up with $0.3$ or $0.35$.

The implicit 3D shape learning performance of Base$_{\textrm{silnet}}$-AttSets and the competing approaches is evaluated on the Blobby dataset. The mean IoU between predicted 2D silhouettes and the ground truth is used as the metric \cite{Wiles2017}.

\begin{table*}[t]
\caption{ Group 1: mean IoU for multi-view reconstruction of all 13 categories in ShapeNet$_{\textrm{r2n2}}$ testing split. All networks are firstly trained given only 1 image for each object in Stage 1. The \nicknameAttSets{} module is further trained given \textbf{2 images} per object in Stage 2, while other competing approaches are fine-tuned given \textbf{2 images} per object in Stage 2.}
\centering
\small
\label{tab:iou_r2n2_02v}
\tabcolsep=0.03cm
\begin{tabular}{ l|cccccccccc}
\hline
number of views &1&2&3&4&5&8&12&16&20&24 \\
\hline
Base$_{\textrm{r2n2}}$-GRU &0.574&0.608&0.622&0.629&0.633&0.639&0.642&0.642&0.641&0.640 \\
Base$_{\textrm{r2n2}}$-max pooling &0.620&0.651&0.660&0.665&0.666&0.671&0.672&0.674&0.673&0.673 \\
Base$_{\textrm{r2n2}}$-mean pooling &0.632&0.654&0.661&0.666&0.667&0.674&0.676&0.680&0.680&0.681 \\
Base$_{\textrm{r2n2}}$-sum pooling &0.633&0.657&0.665&0.669&0.669&0.670&0.666&0.667&0.666&0.665 \\
Base$_{\textrm{r2n2}}$-BP pooling &0.588&0.608&0.615&0.620&0.621&0.627&0.628&0.632&0.633&0.633 \\
Base$_{\textrm{r2n2}}$-MHBN pooling &0.578&0.599&0.606&0.611&0.612&0.618&0.620&0.623&0.624&0.624 \\
Base$_{\textrm{r2n2}}$-SMSO pooling &0.623&0.654&0.664&0.670&0.672&0.679&0.679&0.682&0.680&0.678 \\
\textbf{Base$_{\textrm{r2n2}}$-\nicknameAttSets{}(Ours)} &\textbf{0.642}&\textbf{0.665}&\textbf{0.672}&\textbf{0.677}&\textbf{0.678}&\textbf{0.684}
&\textbf{0.686}&\textbf{0.690}&\textbf{0.690}&\textbf{0.690} \\
\hline
\end{tabular}
\end{table*}

\begin{table*}[t]
\caption{Group 2: mean IoU for multi-view reconstruction of all 13 categories in ShapeNet$_{\textrm{r2n2}}$ testing split. All networks are firstly trained given only 1 image for each object in Stage 1. The \nicknameAttSets{} module is further trained given \textbf{8 images} per object in Stage 2, while other competing approaches are fine-tuned given \textbf{8 images} per object in Stage 2.}
\centering
\small
\label{tab:iou_r2n2_08v}
\tabcolsep=0.03cm
\begin{tabular}{ l|cccccccccc}
\hline
number of views&1&2&3&4&5&8&12&16&20&24 \\
\hline
Base$_{\textrm{r2n2}}$-GRU &0.580&0.616&0.629&0.637&0.641&0.649&0.652&0.652&0.652&0.652 \\
Base$_{\textrm{r2n2}}$-max pooling &0.524&0.615&0.641&0.655&0.661&0.674&0.678&0.683&0.684&0.684 \\
Base$_{\textrm{r2n2}}$-mean pooling &0.632&0.657&0.665&0.670&0.672&0.679&0.681&0.685&0.686&0.686 \\
Base$_{\textrm{r2n2}}$-sum pooling &0.580&0.628&0.644&0.656&0.660&0.672&0.677&0.682&0.684&0.685 \\
Base$_{\textrm{r2n2}}$-BP pooling &0.544&0.599&0.618&0.628&0.632&0.644&0.648&0.654&0.655&0.656 \\
Base$_{\textrm{r2n2}}$-MHBN pooling &0.570&0.596&0.606&0.612&0.614&0.621&0.624&0.628&0.629&0.629 \\
Base$_{\textrm{r2n2}}$-SMSO pooling &0.570&0.621&0.641&0.652&0.656&0.668&0.673&0.679&0.680&0.681 \\
\textbf{Base$_{\textrm{r2n2}}$-\nicknameAttSets{}(Ours)} &\textbf{0.642}&\textbf{0.662}&\textbf{0.671}&\textbf{0.676}&\textbf{0.678}&\textbf{0.686}
&\textbf{0.688}&\textbf{0.693}&\textbf{0.694}&\textbf{0.694} \\
\hline
\end{tabular}
\end{table*}

\begin{table*}[t]
\caption{Group 3: mean IoU for multi-view reconstruction of all 13 categories in ShapeNet$_{\textrm{r2n2}}$ testing split. All networks are firstly trained given only 1 image for each object in Stage 1. The \nicknameAttSets{} module is further trained given \textbf{16 images} per object in Stage 2, while other competing approaches are fine-tuned given \textbf{16 images} per object in Stage 2.}
\centering
\small
\label{tab:iou_r2n2_16v}
\tabcolsep=0.03cm
\begin{tabular}{ l|cccccccccc}
\hline
number of views&1&2&3&4&5&8&12&16&20&24 \\
\hline
Base$_{\textrm{r2n2}}$-GRU &0.579&0.614&0.628&0.636&0.640&0.647&0.651&0.652&0.653&0.653 \\
Base$_{\textrm{r2n2}}$-max pooling &0.511&0.604&0.633&0.649&0.656&0.671&0.678&0.684&0.686&0.686 \\
Base$_{\textrm{r2n2}}$-mean pooling &0.594&0.637&0.652&0.662&0.667&0.677&0.682&0.687&0.688&0.689 \\
Base$_{\textrm{r2n2}}$-sum pooling &0.570&0.629&0.647&0.657&0.664&0.678&0.684&0.690&0.692&0.692 \\
Base$_{\textrm{r2n2}}$-BP pooling &0.545&0.593&0.611&0.621&0.627&0.637&0.642&0.647&0.648&0.649 \\
Base$_{\textrm{r2n2}}$-MHBN pooling &0.570&0.596&0.606&0.612&0.614&0.621&0.624&0.627&0.628&0.629 \\
Base$_{\textrm{r2n2}}$-SMSO pooling &0.580&0.627&0.643&0.652&0.656&0.668&0.673&0.679&0.680&0.681 \\
\textbf{Base$_{\textrm{r2n2}}$-\nicknameAttSets{}(Ours)} &\textbf{0.642}&\textbf{0.660}&\textbf{0.668}&\textbf{0.673}&\textbf{0.676}&\textbf{0.683}
&\textbf{0.687}&\textbf{0.691}&\textbf{0.692}&\textbf{0.693} \\
\hline
\end{tabular}
\end{table*}

\begin{table*}[t]
\caption{Group 4: mean IoU for multi-view reconstruction of all 13 categories in ShapeNet$_{\textrm{r2n2}}$ testing split. All networks are firstly trained given only 1 image for each object in Stage 1. The \nicknameAttSets{} module is further trained given \textbf{24 images} per object in Stage 2, while other competing approaches are fine-tuned given \textbf{24 images} per object in Stage 2.}
\centering
\small
\label{tab:iou_r2n2_24v}
\tabcolsep=0.03cm
\begin{tabular}{ l|cccccccccc}
\hline
number of views&1&2&3&4&5&8&12&16&20&24 \\
\hline
Base$_{\textrm{r2n2}}$-GRU &0.578 &0.613&0.627&0.635&0.639&0.647&0.651&0.653&0.653&0.654 \\
Base$_{\textrm{r2n2}}$-max pooling &0.504&0.600&0.631&0.648&0.655&0.671&0.679&0.685&0.688&0.689 \\
Base$_{\textrm{r2n2}}$-mean pooling &0.593&0.634&0.649&0.659&0.663&0.673&0.677&0.683&0.684&0.685 \\
Base$_{\textrm{r2n2}}$-sum pooling &0.580&0.634&0.650&0.658&0.660&0.678&0.682&0.689&0.690&0.691 \\
Base$_{\textrm{r2n2}}$-BP pooling &0.524&0.585&0.609&0.623&0.630&0.644&0.650&0.656&0.659&0.660 \\
Base$_{\textrm{r2n2}}$-MHBN pooling &0.566&0.595&0.606&0.613&0.616&0.624&0.627&0.631&0.632&0.632 \\
Base$_{\textrm{r2n2}}$-SMSO pooling &0.556&0.613&0.635&0.647&0.653&0.668&0.674&0.681&0.682&0.684 \\
\textbf{Base$_{\textrm{r2n2}}$-\nicknameAttSets{}(Ours)} &\textbf{0.642}&\textbf{0.660}&\textbf{0.668}&\textbf{0.674}&\textbf{0.676}&\textbf{0.684}
&\textbf{0.688}&\textbf{0.693}&\textbf{0.694}&\textbf{0.695} \\
\hline
\end{tabular}
\end{table*}

\begin{table*}[t]
\caption{Group 5: mean IoU for multi-view reconstruction of all 13 categories in ShapeNet$_{\textrm{r2n2}}$ testing split. All networks are firstly trained given only 1 image for each object in Stage 1. The \nicknameAttSets{} module is further trained given random number of images per object in Stage 2, \ie{} $N$ is uniformly sampled from \textbf{[1, 24]}, while other competing approaches are fine-tuned given random number of views per object in Stage 2.}
\centering
\small
\label{tab:iou_r2n2_allv}
\tabcolsep=0.03cm
\begin{tabular}{ l|cccccccccc}
\hline
number of views&1&2&3&4&5&8&12&16&20&24 \\
\hline
Base$_{\textrm{r2n2}}$-GRU &0.580&0.615&0.629&0.637&0.641&0.648&0.651&0.651&0.651&0.651 \\
Base$_{\textrm{r2n2}}$-max pooling &0.601&0.638&0.652&0.660&0.663&0.673&0.677&0.682&0.683&0.684 \\
Base$_{\textrm{r2n2}}$-mean pooling &0.598&0.637&0.652&0.660&0.664&0.675&0.679&0.684&0.685&0.686 \\
Base$_{\textrm{r2n2}}$-sum pooling &0.587&0.631&0.646&0.656&0.660&0.672&0.678&0.683&0.684&0.685 \\
Base$_{\textrm{r2n2}}$-BP pooling &0.582&0.610&0.620&0.626&0.628&0.635&0.638&0.641&0.642&0.643 \\
Base$_{\textrm{r2n2}}$-MHBN pooling &0.575&0.599&0.608&0.613&0.615&0.622&0.624&0.628&0.629&0.629 \\
Base$_{\textrm{r2n2}}$-SMSO pooling &0.580&0.624&0.641&0.652&0.657&0.669&0.674&0.679&0.681&0.682 \\
\textbf{Base$_{\textrm{r2n2}}$-\nicknameAttSets{}(Ours)} &\textbf{0.642}&\textbf{0.662}&\textbf{0.670}&\textbf{0.675}&\textbf{0.677}&\textbf{0.685}
&\textbf{0.688}&\textbf{0.692}&\textbf{0.693}&\textbf{0.694} \\
\hline
\end{tabular}
\end{table*}

\begin{figure*}[!h]
	\begin{minipage}[t]{0.32\textwidth}
		\centerline{
			\includegraphics[width=1\textwidth]{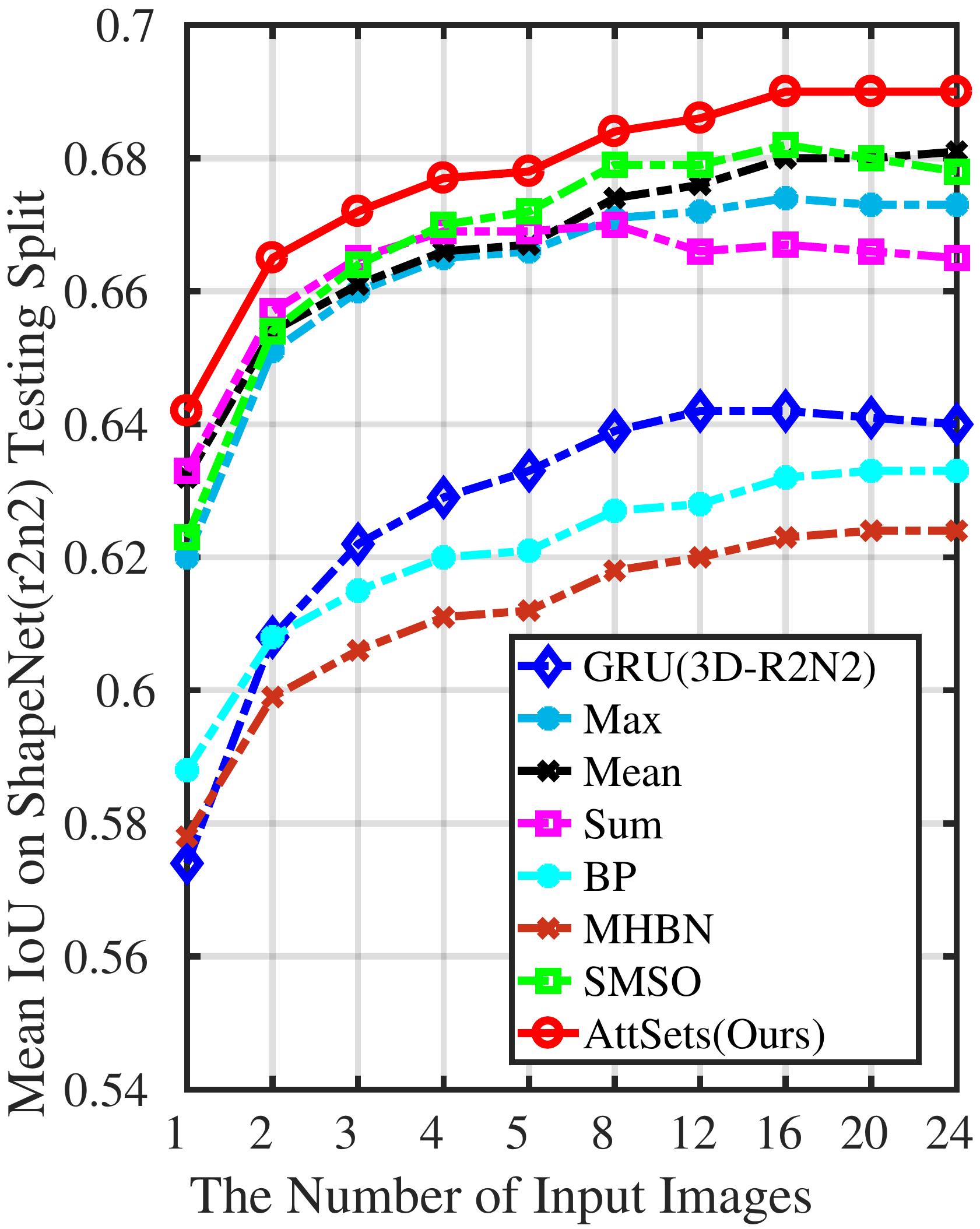}}
		\caption{IoUs: Group 1.}
		\label{fig:02v_mIoU}
	\end{minipage}
	\makebox[0.02in][]{}
	\begin{minipage}[t]{0.32\textwidth}
		\centerline{
			\includegraphics[width=1.0\textwidth]{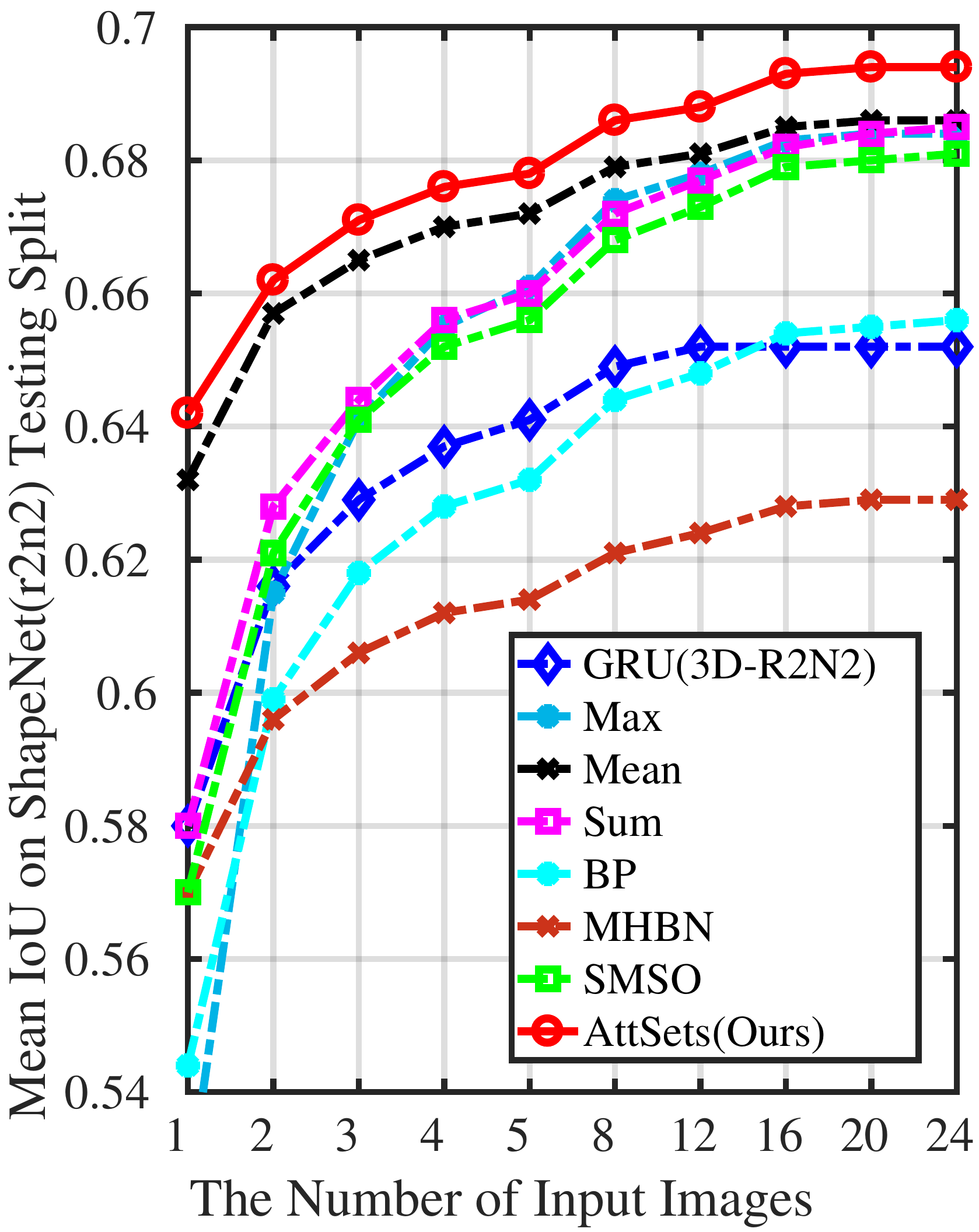}}
		\caption{IoUs: Group 2.}
        \label{fig:08v_mIoU}
	\end{minipage}
	\makebox[0.02in][]{}
	\begin{minipage}[t]{0.32\textwidth}
		\centerline{
			\includegraphics[width=1.0\textwidth]{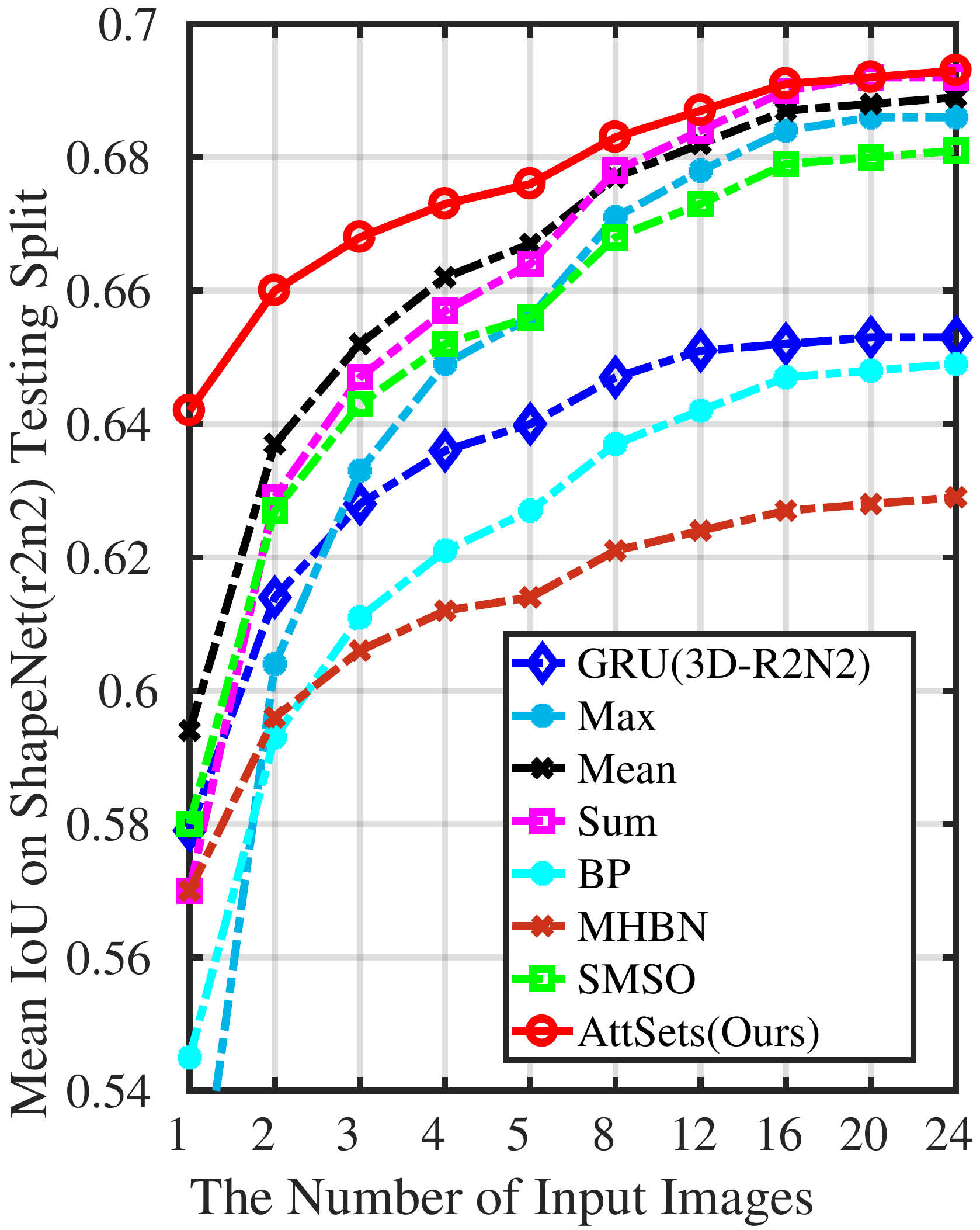}}
		\caption{IoUs: Group 3.}
        \label{fig:16v_mIoU}
	\end{minipage}
    \makebox[0.02in][]{}
	\begin{minipage}[t]{0.32\textwidth}
		\centerline{
			\includegraphics[width=1\textwidth]{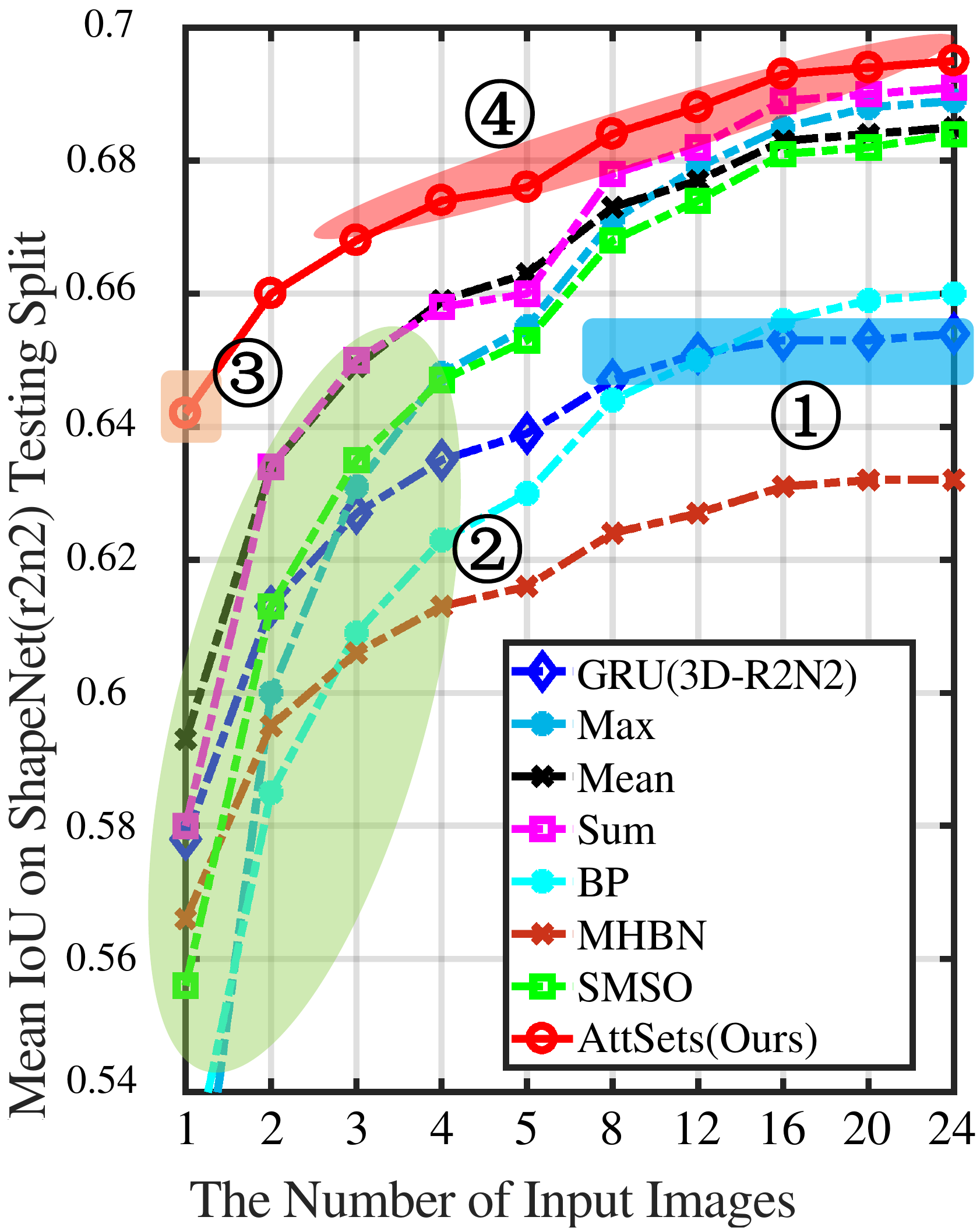}}
		\caption{IoUs: Group 4.}
        \label{fig:24v_mIoU}
	\end{minipage}
	\makebox[0.02in][]{}
	\begin{minipage}[t]{0.33\textwidth}
		\centerline{
			\includegraphics[width=1.0\textwidth]{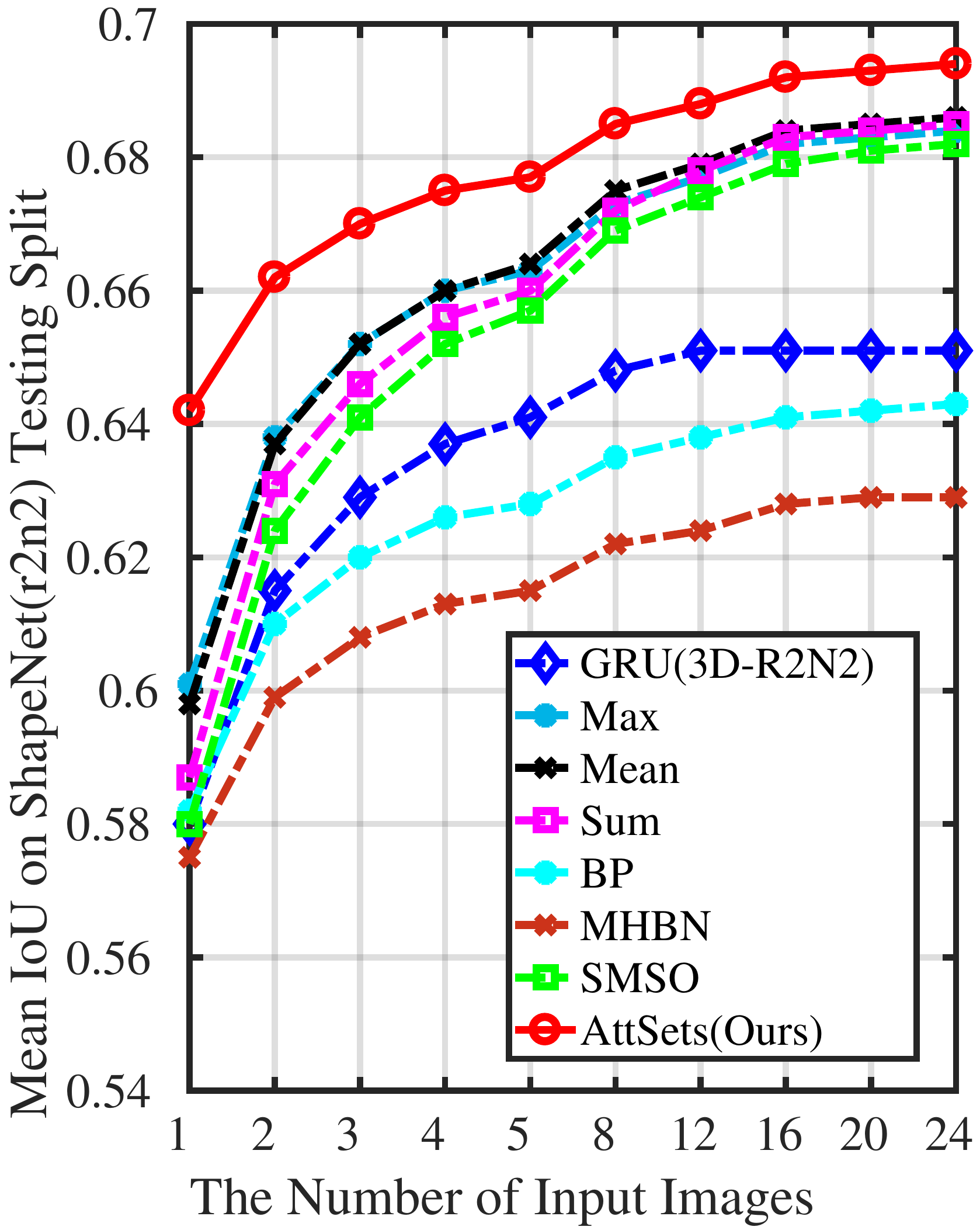}}
		\caption{IoUs: Group 5.}
        \label{fig:allv_mIoU}
	\end{minipage}
\end{figure*}

\begin{figure*}[t]
\centering
   \includegraphics[width=1\linewidth]{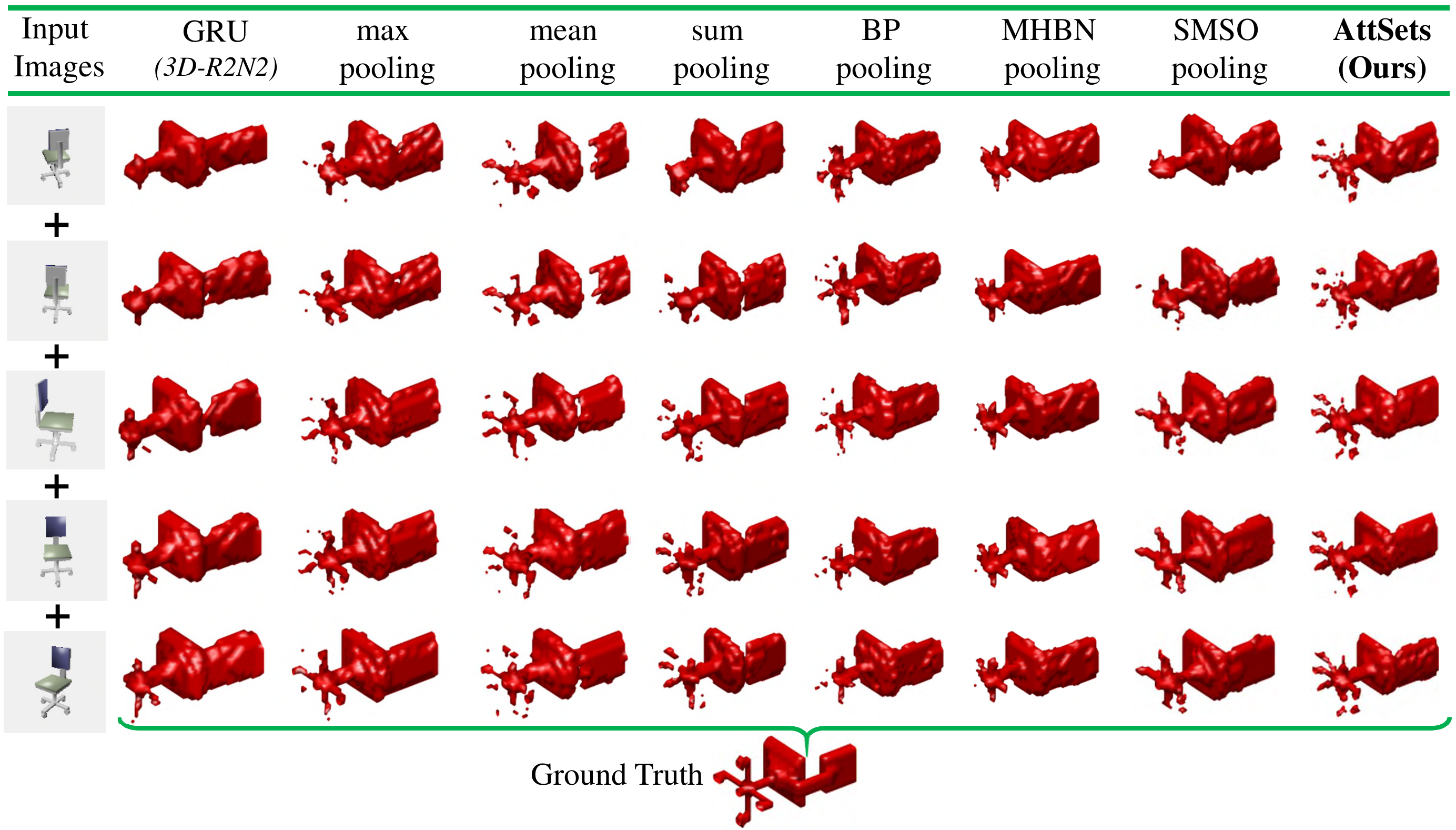}
\caption{Qualitative results of multi-view reconstruction achieved by different approaches in experiment Group 5.}
\label{fig:mv_demo_r2n2}
\end{figure*}

\begin{table*}[t]
\caption{ Per-category mean IoU for single view reconstruction on ShapeNet$_{\textrm{r2n2}}$ testing split.}
\centering
\scriptsize
\label{tab:iou_r2n2_1v}
\tabcolsep=0.0cm
\begin{tabular}{ l|cccccccccccccc}
\hline
&plane&bench&cabinet&car&chair&monitor&lamp&spk.&firearm&couch&table&phone&w.craft&mean \\
\hline
Base$_{\textrm{r2n2}}$-GRU &0.530&0.449&0.730&0.807&0.487&0.497&0.391&0.671&0.553&0.631&0.515&0.683&0.535&0.580\\
Base$_{\textrm{r2n2}}$-max pooling &0.573&0.521&0.755&0.835&0.533&0.544&0.423&0.695&0.587&0.678&0.562&0.710&0.582&0.620\\
Base$_{\textrm{r2n2}}$-mean pooling &0.582&0.540&0.773&0.837&0.547&0.550&0.440&0.713&0.595&0.695&0.576&0.718&0.593&0.632\\
Base$_{\textrm{r2n2}}$-sum pooling &0.588&0.536&0.771&0.838&0.554&0.547&0.442&0.710&0.598&0.690&0.575&0.728&0.598&0.633\\
Base$_{\textrm{r2n2}}$-BP pooling &0.536&0.469&0.747&0.816&0.484&0.499&0.398&0.678&0.556&0.646&0.528&0.681&0.550&0.588\\
Base$_{\textrm{r2n2}}$-MHBN pooling &0.528&0.451&0.742&0.812&0.471&0.487&0.386&0.677&0.548&0.637&0.515&0.674&0.546&0.578\\
Base$_{\textrm{r2n2}}$-SMSO pooling &0.572&0.521&0.763&0.833&0.541&0.548&0.433&0.704&0.581&0.682&0.566&0.721&0.581&0.623\\
OGN &0.587&0.481&0.729&0.816&0.483&0.502&0.398&0.637&0.593&0.646&0.536&0.702&\textbf{0.632}&0.596\\
AORM &\textbf{0.605}&0.498&0.715&0.757&0.532&0.524&0.415&0.623&\textbf{0.618}&0.679&0.547&0.738&0.552&0.600\\
PointSet &0.601&0.550&0.771&0.831&0.544&0.552&\textbf{0.462}&\textbf{0.737}&0.604&\textbf{0.708}&\textbf{0.606}&\textbf{0.749}&0.611&0.640\\
\textbf{Base$_{\textrm{r2n2}}$-\nicknameAttSets{}\scriptsize{(Ours)}} &0.594&\textbf{0.552}&\textbf{0.783}&\textbf{0.844}&\textbf{0.559}&\textbf{0.565}
&0.445&0.721&0.601&0.703&0.590&0.743&0.601&\textbf{0.642}\\
\hline
\end{tabular}
\end{table*}

\subsection{Evaluation on ShapeNet$_{\textrm{r2n2}}$ Dataset}\label{sec:ch4_eval_r2n2}

To fully evaluate the aggregation performance and robustness, we train the Base$_{\textrm{r2n2}}$-AttSets and its competing approaches on ShapeNet$_{\textrm{r2n2}}$ dataset. For fair comparison, all networks (the pooling/GRU/\nicknameAttSets{} based approaches) are trained according to the proposed two-stage training algorithm.

\textbf{Training Stage 1.} All networks are trained given only 1 image for each object, \ie{} $N=1$ in all training iterations, until convergence. Basically, this is to guarantee all networks are well optimized for the extreme case where there is only one input image.

\textbf{Training Stage 2.} To enable these networks to be more robust for multiple input images, all networks are further trained given more images per object. Particularly, we conduct the following five parallel groups of training experiments.

\begin{itemize}
\item Group 1. All networks are further trained given only 2 images for each object, \ie{} $N=2$ in all iterations. As with our Base$_{\textrm{r2n2}}$-AttSets, the well-trained encoder-decoder in previous stage 1 is frozen, and we only optimize the \nicknameAttSets{} module according to our \nicknameFASet{} algorithm \ref{alg:faset}. For the competing approaches, \eg{} GRU and all poolings, we fine-tune the whole networks because they do not have separate parameters suitable for special training. To be specific, we use smaller learning rate (1e-5) to carefully train these networks to achieve better performance where $N=2$ until convergence.

\item Group 2/3/4. Similarly, in these three groups of second-stage training experiments, $N$ is set to 8, 16 and 24 separately. 

\item Group 5. All networks are further trained until convergence, but $N$ is uniformly and randomly sampled from $[1, 24]$ for each object during training. In the above Group 1/2/3/4, $N$ is fixed for each object, while $N$ is dynamic for each object in Group 5. 
\end{itemize}

In the above experiment, Groups 1/2/3/4 are designed to investigate how all competing approaches would be further optimized towards the statistics of a fixed $N$ during training, thus resulting in different levels of robustness given an arbitrary number of $N$ during testing. By contrast, the paradigm in Group 5 aims at enumerating all possible $N$ values during training. Therefore the overall performance might be more robust with an arbitrary number of input images during testing, compared with the above Group 1/2/3/4 experiments.

\textbf{Testing Stage.}
All networks trained in the above five groups of experiments are separately tested given $N = \{1, 2, 3, 4, 5, 8, 12, 16, 20,24\}$. The permutations of input images are the same for all different approaches for fair comparison. Note that, we do not test the networks which are only trained in Stage 1, because the \nicknameAttSets{} module is not optimized and the corresponding Base$_{\textrm{r2n2}}$-AttSets is unable to generalize to multiple input images during testing. Therefore, it is meaningless to compare the performance when the network is solely trained on a single image.

\textbf{Results.} Tables \ref{tab:iou_r2n2_02v} $\sim$ \ref{tab:iou_r2n2_allv} show the mean IoU scores of all 13 categories for experiments of Group 1 $\sim$ 5, while Figures \ref{fig:02v_mIoU} $\sim$ \ref{fig:allv_mIoU} show the trends of mean IoU changes in different Groups. Figure \ref{fig:mv_demo_r2n2} shows the estimated 3D shapes in experiment Group 5, with an increasing number of images from 1 to 5 for different approaches. 

We notice that the reported IoU scores of ShapeNet data repository in the original LSM \cite{Kar2017} are higher than our scores. However, the experimental settings in LSM \cite{Kar2017} are quite different from ours in the following two ways. 1) The original LSM requires both RGB images and the corresponding viewing angles as input, while all our experiments do not. 2) The original LSM dataset has different styles of rendered color images and different train/test splits compared with our experimental settings. Therefore the reported IoU scores in LSM are not directly comparable with ours and we do not include the results in this paper to avoid confusion. Note that, the aggregation module of LSM \cite{Kar2017}, \ie{} GRU, is the same as used in 3D-R2N2 \cite{Chan2016}, and is indeed fully evaluated throughout our experiments.

To highlight the performance of single view 3D reconstruction, Table \ref{tab:iou_r2n2_1v} shows the optimal per-category IoU scores for different competing approaches from experiments Group 1 $\sim$ 5. In addition, we also compare with the state-of-the-art dedicated single view reconstruction approaches including OGN \cite{Tatarchenko2017}, AORM \cite{Yang2018c} and PointSet \cite{Fan2017} in Table \ref{tab:iou_r2n2_1v}. Overall, our \nicknameAttSets{} based approach outperforms all others by a large margin for either single view or multi view reconstruction, and generates much more compelling 3D shapes.

\textbf{Analysis.} We investigate the results as follows: 
\begin{itemize}
\item The GRU based approach can generate reasonable 3D shapes in all experiments Group 1 $\sim$ 5 given either few or multiple images during testing, but the performance saturates quickly after being given more images, \eg{} 8 views, because the recurrent unit is not particularly capable of capturing features from longer image sequences, as illustrated in Figure \ref{fig:24v_mIoU} \circled{1}.

\item In Group 1 $\sim$ 4, all pooling based approaches are able to estimate satisfactory 3D shapes when given a similar number of images in testing as in training, but they are unlikely to predict reasonable shapes given an arbitrary number of images. For example, in experiment Group 4, all pooling based approaches have inferior IoU scores given only few images as shown in Table \ref{tab:iou_r2n2_24v} and Figure \ref{fig:24v_mIoU} \circled{2}, because the pooled features from fewer images during testing are unlikely to be as general and representative as pooled features from more images during training. Therefore, those models trained on $24$ images fail to generalize well to only one image during testing.

\item In Group 5, as shown in Table \ref{tab:iou_r2n2_allv} and Figure \ref{fig:allv_mIoU}, all pooling based approaches are much more robust compared with Group 1$\sim$4, because the networks are generally optimized according to an arbitrary number of images during training. However, these networks tend to have the performance \textit{in the middle}. Compared with Group 4, all approaches in Group 5 tend to have better performance when $N=1$, while being worse when $N=24$. Compared with Group 1, all approaches in Group 5 are likely to be better when $N=24$, while being worse when $N=1$. Basically, these networks tend to be optimized to learn the \textit{mean} features overall.

\item  In all experiments Group 1 $\sim$ 5, all approaches tend to have better performance when given sufficiently many input images, \ie{} $N=24$, because more images are able to provide enough information for reconstruction.

\item In all experiments Group 1 $\sim$ 5, our \nicknameAttSets{} based approach clearly outperforms all others in either single or multiple view 3D reconstruction and it is more robust to a variable number of input images. Our \nicknameFASet{} algorithm completely decouples the base network to learn visual features for accurate single view reconstruction as illustrated in Figure \ref{fig:24v_mIoU} \circled{3}, while the trainable parameters of \nicknameAttSets{} module are separately responsible for learning attention scores for better multi-view reconstruction as shown in Figure \ref{fig:24v_mIoU} \circled{4}. Therefore, the whole network does not suffer from limitations of GRU or pooling approaches, and can achieve better performance for either fewer or more image reconstruction.
\end{itemize}

\begin{figure*}[h]
\centering
   \includegraphics[width=1\linewidth]{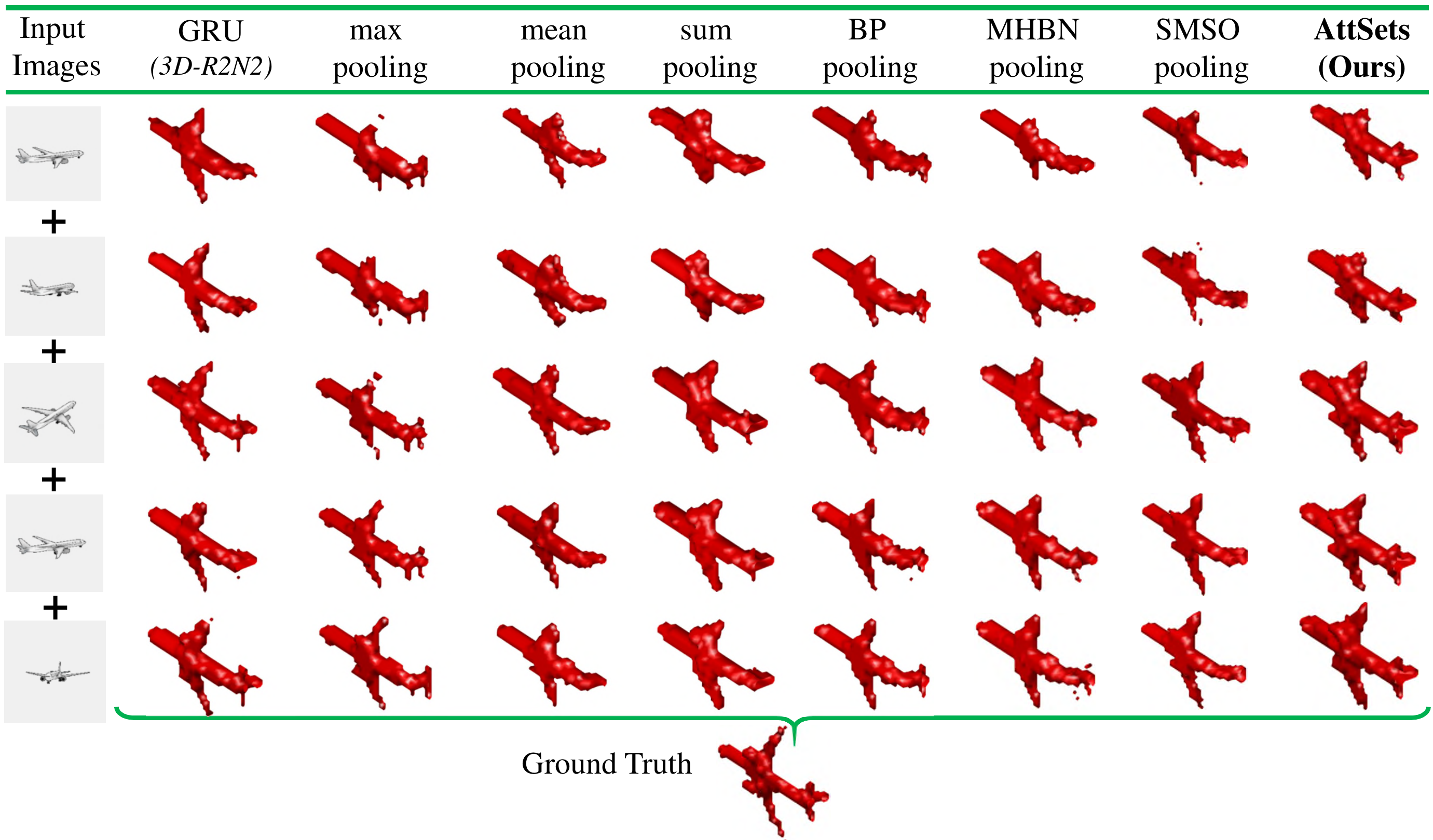}
\caption{Qualitative results of multi-view reconstruction from different approaches in ShapeNet$_{\textrm{lsm}}$ testing split.}
\label{fig:mv_demo_lsm}
\end{figure*}

\begin{table*}[ht]
\caption{Mean IoU for multi-view reconstruction of all 13 categories from ShapeNet$_{\textrm{lsm}}$ dataset. All networks are well trained in previous experiment Group 5 of Section \ref{sec:ch4_eval_r2n2}.}
\centering
\small
\label{tab:iou_lsm_allv}
\tabcolsep=0.08cm
\begin{tabular}{ l|ccccccccc}
\hline
number of views&1&2&3&4&5&8&12&16&20 \\
\hline
Base$_{\textrm{r2n2}}$-GRU &0.390&0.428&0.444&0.454&0.459&0.467&0.470&0.471&0.472 \\
Base$_{\textrm{r2n2}}$-max pooling &0.276&0.388&0.433&0.459&0.473&0.497&0.510&0.515&0.518 \\
Base$_{\textrm{r2n2}}$-mean pooling &0.365&0.426&0.452&0.468&0.477&0.493&0.503&0.508&0.511 \\
Base$_{\textrm{r2n2}}$-sum pooling &0.363&0.421&0.445&0.459&0.466&0.481&0.492&0.499&0.503 \\
Base$_{\textrm{r2n2}}$-BP pooling &0.359&0.407&0.426&0.436&0.442&0.453&0.459&0.462&0.463 \\
Base$_{\textrm{r2n2}}$-MHBN pooling &0.365&0.403&0.418&0.427&0.431&0.441&0.446&0.449&0.450 \\
Base$_{\textrm{r2n2}}$-SMSO pooling &0.364&0.419&0.445&0.460&0.469&0.488&0.500&0.506&0.510 \\
\textbf{Base$_{\textrm{r2n2}}$-\nicknameAttSets{}(Ours)} &\textbf{0.404}&\textbf{0.452}&\textbf{0.475}&\textbf{0.490}&\textbf{0.498}&\textbf{0.514}
&\textbf{0.522}&\textbf{0.528}&\textbf{0.531} \\
\hline
\end{tabular}
\end{table*}

\subsection{Evaluation on ShapeNet$_{\textrm{lsm}}$ Dataset}\label{sec:ch4_eval_lsm}
To further investigate how well the learnt visual features and attention scores generalize across different style of images, we use the well trained networks of previous Group 5 of Section \ref{sec:ch4_eval_r2n2} to test on the large ShapeNet$_{\textrm{lsm}}$ dataset. Note that, we only borrow the synthesized images from ShapeNet$_{\textrm{lsm}}$ dataset corresponding to the objects in ShapeNet$_{\textrm{r2n2}}$ testing split. This guarantees that all the trained models have never seen either the style of LSM rendered images or the 3D object labels before. The image viewing angles from the original ShapeNet$_{\textrm{lsm}}$ dataset are not used in our experiments, since the Base$_{\textrm{r2n2}}$ network does not require image viewing angles as input. Table \ref{tab:iou_lsm_allv} shows the mean IoU scores of all approaches, while Figure \ref{fig:mv_demo_lsm} shows the qualitative results. 

Our \nicknameAttSets{} based approach outperforms all others given either few or multiple input images. This demonstrates that our Base$_{\textrm{r2n2}}$-AttSets approach does not overfit the training data, but has better generality and robustness over new styles of rendered color images compared with other approaches.

\begin{table*}[t]
\caption{Group 1: mean IoU for multi-view reconstruction of all 40 categories in ModelNet40 testing split. All networks are firstly trained given only 1 image for each object in Stage 1. The \nicknameAttSets{} module is further trained given \textbf{12 images} per object in Stage 2, while other competing approaches are fine-tuned given \textbf{12 images} per object in Stage 2.}
\centering
\small
\label{tab:iou_modelnet_12v}
\tabcolsep=0.23cm
\begin{tabular}{ l|ccccccc}
\hline
number of views&1&2&3&4&5&8&12 \\
\hline
Base$_{\textrm{r2n2}}$-GRU &0.344&0.390&0.414&0.430&0.440&0.454&0.464\\
Base$_{\textrm{r2n2}}$-max pooling &0.393&0.468&0.490&0.504&0.511&0.523&0.525\\
Base$_{\textrm{r2n2}}$-mean pooling &0.415&0.464&0.481&0.495&0.502&0.515&0.520\\
Base$_{\textrm{r2n2}}$-sum pooling &0.332&0.441&0.473&0.492&0.500&0.514&0.520\\
Base$_{\textrm{r2n2}}$-BP pooling &0.431&0.466&0.479&0.492&0.497&0.509&0.515\\
Base$_{\textrm{r2n2}}$-MHBN pooling &0.423&0.462&0.478&0.491&0.497&0.509&0.515\\
Base$_{\textrm{r2n2}}$-SMSO pooling &0.441&0.476&0.490&0.500&0.506&0.517&0.520\\
\textbf{Base$_{\textrm{r2n2}}$-\nicknameAttSets{}{}(Ours)} &\textbf{0.487}&\textbf{0.505}&\textbf{0.511}&\textbf{0.517}&\textbf{0.521}&\textbf{0.527}
&\textbf{0.529} \\
\hline
\end{tabular}
\end{table*}

\begin{table*}[t]
\caption{Group 2: mean IoU for multi-view reconstruction of all 40 categories in ModelNet40 testing split. All networks are firstly trained given only 1 image for each object in Stage 1. The \nicknameAttSets{} module is further trained given random number of images per object in Stage 2, \ie{} $N$ is uniformly sampled from \textbf{[1, 12]}, while other competing approaches are fine-tuned given random number of views per object in Stage 2.}
\centering
\small
\label{tab:iou_modelnet_allv}
\tabcolsep=0.23cm
\begin{tabular}{ l|ccccccc}
\hline
number of views&1&2&3&4&5&8&12 \\
\hline
Base$_{\textrm{r2n2}}$-GRU &0.388&0.421&0.434&0.440&0.444&0.449&0.452\\
Base$_{\textrm{r2n2}}$-max pooling &0.461&0.489&0.498&0.506&0.509&0.515&0.517\\
Base$_{\textrm{r2n2}}$-mean pooling &0.455&0.487&0.498&0.507&0.512&0.520&0.523\\
Base$_{\textrm{r2n2}}$-sum pooling &0.453&0.484&0.494&0.503&0.506&0.514&0.517\\
Base$_{\textrm{r2n2}}$-BP pooling &0.454&0.479&0.487&0.496&0.499&0.507&0.510\\
Base$_{\textrm{r2n2}}$-MHBN pooling &0.453&0.480&0.488&0.497&0.500&0.507&0.509\\
Base$_{\textrm{r2n2}}$-SMSO pooling &0.462&0.488&0.497&0.505&0.509&0.516&0.519\\
\textbf{Base$_{\textrm{r2n2}}$-\nicknameAttSets{}(Ours)} &\textbf{0.487}&\textbf{0.505}&\textbf{0.511}&\textbf{0.518}&\textbf{0.520}&\textbf{0.525}
&\textbf{0.527} \\
\hline
\end{tabular}
\end{table*}

\begin{figure*}[t]
\centering
   \includegraphics[width=1\linewidth]{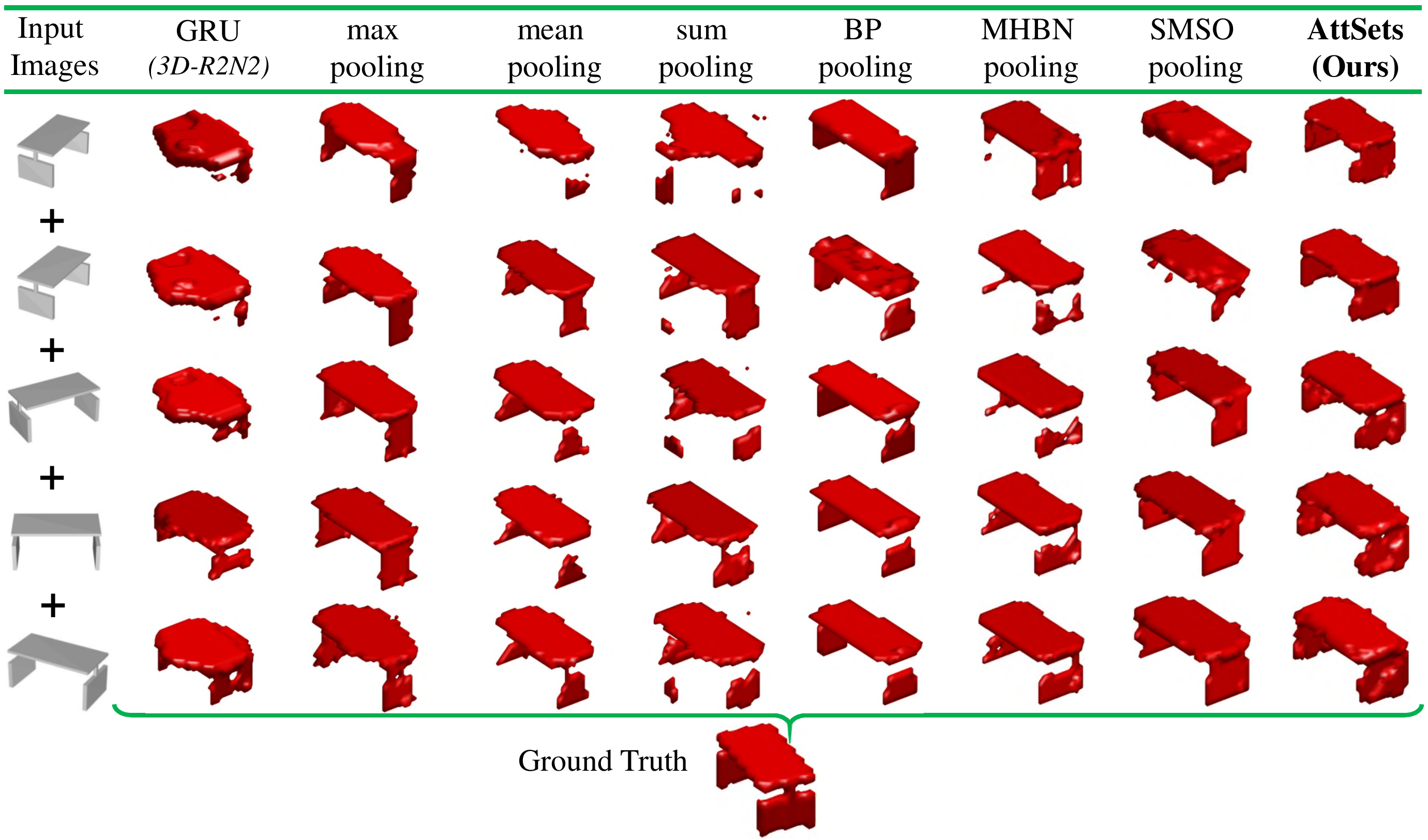}
\caption{Qualitative results of multi-view reconstruction from different approaches in ModelNet40 testing split.}
\label{fig:mv_demo_modelnet40}
\end{figure*}

\subsection{Evaluation on ModelNet40 Dataset}\label{sec:ch4_eval_modelnet40}
We train the Base$_{\textrm{r2n2}}$-AttSets and its competing approaches on ModelNet40 dataset from scratch. For fair comparison, all networks (the pooling/GRU/\nicknameAttSets{} based approaches) are trained according to the proposed \nicknameFASet{} algorithm, which is similar to the two-stage training strategy of Section \ref{sec:ch4_eval_r2n2}. 

\textbf{Training Stage 1.} All networks are trained given only 1 image for each object, \ie{} $N=1$ in all training iterations, until convergence. This guarantees all networks are well optimized for single view 3D reconstruction.

\textbf{Training Stage 2.} We further conduct the following two parallel groups of training experiments to optimize the networks for multi-view reconstruction.
\begin{itemize}
\item Group 1. All networks are further trained given all 12 images for each object, \ie{} $N=12$ in all iterations, until convergence. As with our Base$_{\textrm{r2n2}}$-AttSets, the well-trained encoder-decoder in the previous Stage 1 is frozen, and only the \nicknameAttSets{} module is trained. All other competing approaches are fine-tuned using a smaller learning rate (1e-5) in this stage.

\item Group 2. All networks are further trained until convergence, but $N$ is uniformly and randomly sampled from $[1, 12]$ for each object during training. Only the \nicknameAttSets{} module is trained, while all other competing approaches are fine-tuned in Stage 2.
\end{itemize}

\textbf{Testing Stage.} All networks trained in the above two groups are separately tested given $N=[1,2,3,4,5,8,12]$. The permutations of input images are the same for all different approaches for fair comparison.

\textbf{Results.} Tables \ref{tab:iou_modelnet_12v} and \ref{tab:iou_modelnet_allv} show the mean IoU scores of Groups 1 and 2 respectively, and Figure \ref{fig:mv_demo_modelnet40} shows qualitative results of Group 2. The Base$_{\textrm{r2n2}}$-AttSets surpasses all competing approaches by a large margin for both single and multiple view 3D reconstruction, and all the results are consistent with previous experimental results on both ShapeNet$_{\textrm{r2n2}}$ and ShapeNet$_{\textrm{lsm}}$ datasets.

\begin{table}[ht]
\caption{Group 1: mean IoU for silhouettes prediction on the Blobby dataset. All networks are firstly trained given only 1 image for each object in Stage 1. The \nicknameAttSets{} module is further trained given \textbf{2 images} per object, \ie{} $N$ =2, while other competing approaches are fine-tuned given 2 views per object in Stage 2.}
\centering
\label{tab:iou_blobby_02v}
\tabcolsep=0.52cm
\begin{tabular}{ l|cccc}
\hline
&1 view&2 views&3 views& 4 views \\
\hline
Base$_{\textrm{silnet}}$-GRU &0.857&0.860&0.860&0.860\\
Base$_{\textrm{silnet}}$-max pooling &0.922&0.923&0.924&0.924\\
Base$_{\textrm{silnet}}$-mean pooling &0.920&0.922&0.923&0.924\\
Base$_{\textrm{silnet}}$-sum pooling &0.913&0.918&0.917&0.916\\
Base$_{\textrm{silnet}}$-BP pooling &0.908&0.912&0.914&0.914\\
Base$_{\textrm{silnet}}$-MHBN pooling &0.901&0.904&0.906&0.906\\
Base$_{\textrm{silnet}}$-SMSO pooling &0.860&0.865&0.865&0.865\\
\textbf{Base$_{\textrm{silnet}}$-\nicknameAttSets{}(Ours)} &\textbf{0.924}&\textbf{0.931}&\textbf{0.933}&\textbf{0.935} \\
\hline
\end{tabular}
\end{table}

\begin{table}[h]
\caption{Group 2: mean IoU for silhouettes prediction on the Blobby dataset. All networks are firstly trained given only 1 image for each object in Stage 1. The \nicknameAttSets{} module is further trained given \textbf{4 images} per object, \ie{} $N$=4, while other competing approaches are fine-tuned given 4 views per object in Stage 2.}
\centering
\label{tab:iou_blobby_04v}
\tabcolsep=0.52cm
\begin{tabular}{ l|cccc}
\hline
&1 view&2 views&3 views& 4 views \\
\hline
Base$_{\textrm{silnet}}$-GRU &0.863&0.865&0.865&0.865\\
Base$_{\textrm{silnet}}$-max pooling &0.923&0.927&0.929&0.929\\
Base$_{\textrm{silnet}}$-mean pooling &0.923&0.925&0.927&0.927\\
Base$_{\textrm{silnet}}$-sum pooling &0.902&0.917&0.921&0.924\\
Base$_{\textrm{silnet}}$-BP pooling &0.911&0.916&0.919&0.920\\
Base$_{\textrm{silnet}}$-MHBN pooling &0.904&0.908&0.911&0.911\\
Base$_{\textrm{silnet}}$-SMSO pooling &0.863&0.865&0.865&0.865\\
\textbf{Base$_{\textrm{silnet}}$-\nicknameAttSets{}(Ours)} &\textbf{0.924}&\textbf{0.932}&\textbf{0.936}&\textbf{0.937} \\
\hline
\end{tabular}
\end{table}

\begin{figure*}[t]
\centering
   \includegraphics[width=1\linewidth]{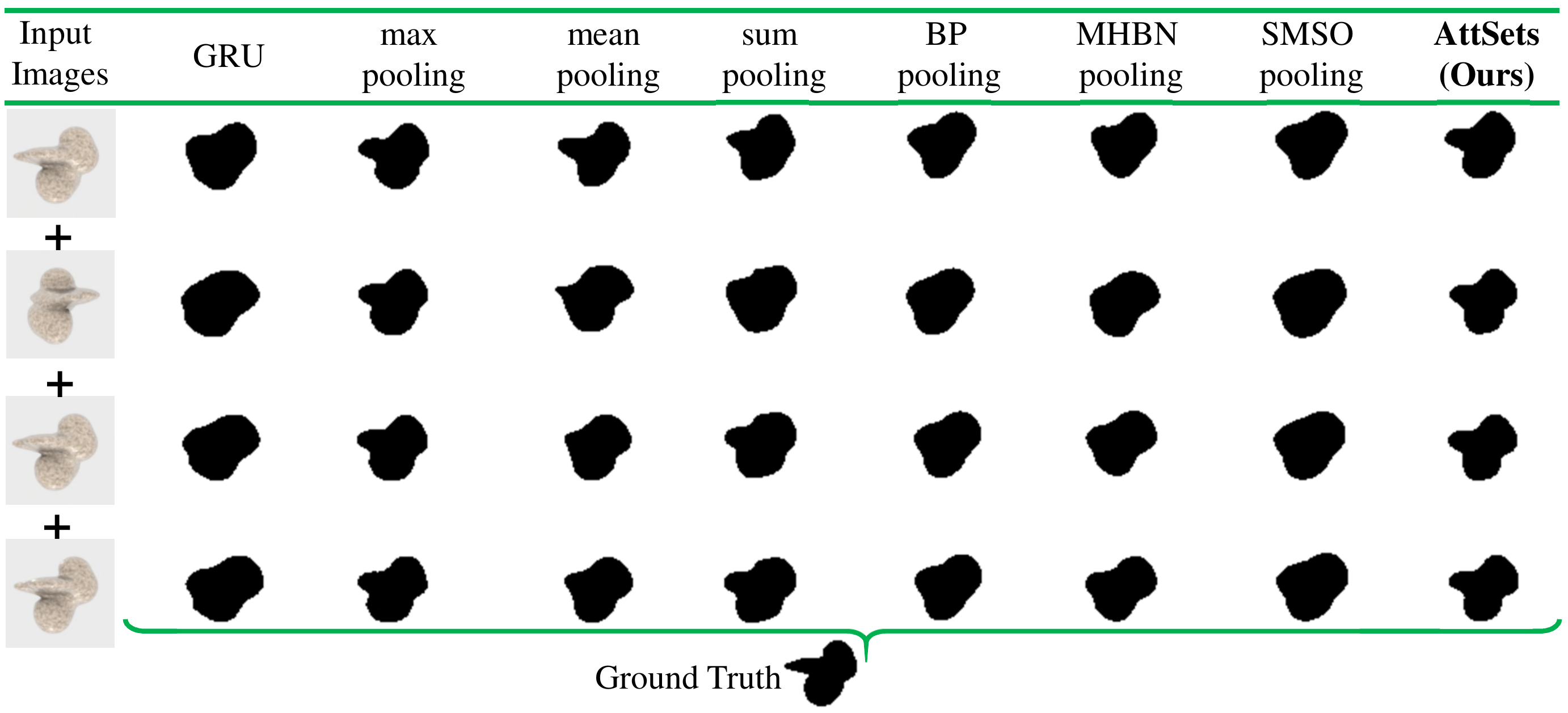}
\caption{Qualitative results of silhouettes prediction from different approaches on the Blobby dataset.}
\label{fig:mv_demo_blobby}
\end{figure*}

\subsection{Evaluation on Blobby Dataset}\label{sec:ch4_eval_blobby}
In this section, we evaluate the Base$_{\textrm{silnet}}$-AttSets and the competing approaches on the Blobby dataset. For fair comparison, the GRU module is implemented with a single fully connected layer of 160 hidden units, which has similar network capacity with our \nicknameAttSets{} based network. All networks (the pooling/GRU/\nicknameAttSets{} based approaches) are trained with the proposed two-stage \nicknameFASet{} algorithm as follows:

\textbf{Training Stage 1.} All networks are trained given only 1 image together with the viewing angle for each object, \ie{} $N$=1 in all training iterations, until convergence. This guarantees the performance of single view shape learning. 

\textbf{Training Stage 2.} Another two parallel groups of training experiments are conducted to further optimize the networks for multi-view shape learning.

\begin{itemize}
\item Group 1. All networks are further trained given only 2 images for each object, \ie{} $N$=2 in all iterations. As to Base$_{\textrm{silnet}}$-AttSets, only the \nicknameAttSets{} module is optimized with the well-trained base encoder-decoder being frozen. For fair comparison, all competing approaches are fine-tuned given 2 images per object for better performance where $N$ =2 until convergence.
\item Group 2. Similar to the above Group 1, all networks are further trained given all 4 images for each object, \ie{} $N$=4, until convergence.
\end{itemize}

\textbf{Testing Stage.} All networks trained in the above two groups are separately tested given $N$ = [1,2,3,4]. The permutations of input images are the same for all different networks for fair comparison.

\textbf{Results.} Tables \ref{tab:iou_blobby_02v} and \ref{tab:iou_blobby_04v} show the mean IoUs of the above two groups of experiments and Figure \ref{fig:mv_demo_blobby} shows the qualitative results of Group 2. Note that, the IoUs are calculated on predicted 2D silhouettes instead of 3D voxels, so they are not numerically comparable with previous experiments on ShapeNet$_{\textrm{r2n2}}$, ShapeNet$_{\textrm{lsm}}$, and ModelNet40 datasets. We do not include the IoU scores of the original SilNet \cite{Wiles2017}, because the original IoU scores are obtained from an end-to-end training strategy. In this paper, we uniformly apply the proposed two-stage \nicknameFASet{} training paradigm on all approaches for fair comparison. Our Base$_{\textrm{silnet}}$-AttSets consistently outperforms all competing approaches for shape learning from either single or multiple views.

\begin{figure*}[t]
\centering
   \includegraphics[width=1\linewidth]{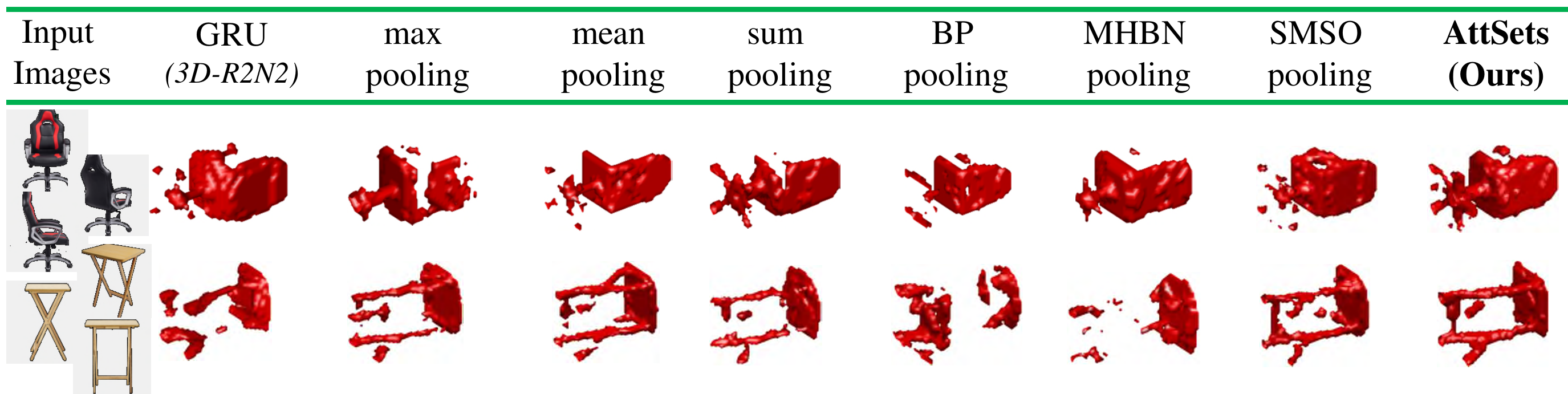}
\caption{Qualitative results of multi-view 3D reconstruction from real-world images.}
\label{fig:mv_demo_real}
\end{figure*}

\begin{figure}[t]
\centering
   \includegraphics[width=1\linewidth]{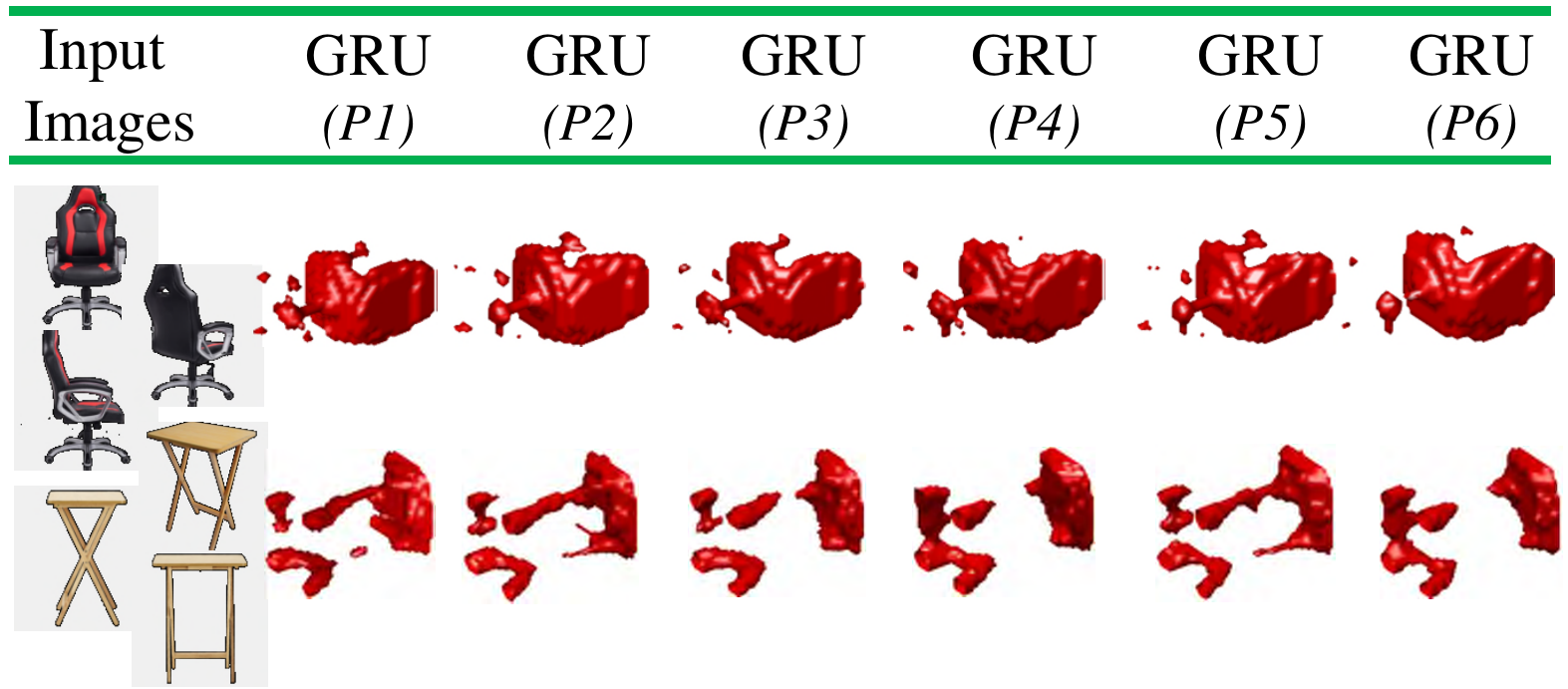}
\caption{Qualitative results of inconsistent 3D reconstruction from the GRU based approach.}
\label{fig:mv_demo_real_permu}
\end{figure}

\subsection{Qualitative Results on Real-world Images}
To the best of our knowledge, there is no public real-world dataset for multi-view 3D object reconstruction. Therefore, we manually collect real world images from Amazon online shops to qualitatively demonstrate the generality of all networks which are trained on the synthetic ShapeNet$_{\textrm{r2n2}}$ dataset in experiment Group 4 of Section \ref{sec:ch4_eval_r2n2}, as shown in Figure \ref{fig:mv_demo_real}.

In the meantime, we use these real-world images to qualitatively show the permutation invariance of different approaches. In particular, for each object, we use 6 different permutations in total for testing. As shown in Figure \ref{fig:mv_demo_real_permu}, the GRU based approach generates inconsistent 3D shapes given different image permutations. For example, the arm of a chair and the leg of a table can be reconstructed in permutation 1, but fail to be recovered in another permutation. By comparison, all other approaches are permutation invariant, as shown in Figure \ref{fig:mv_demo_real}.

\begin{table}[t]
\caption{Mean time consumption for a single object ($32^3$ voxel grid) estimation from different number of images (milliseconds).}
\centering
\label{tab:time_con}
\tabcolsep=0.3cm
\begin{tabular}{ l|ccccccc}
\hline
number of input images&1 &4 &8 &12 &16 &20 &24  \\
\hline
Base$_{\textrm{r2n2}}$-GRU &6.9&11.2&17.0&22.8&28.8&34.7&40.7\\
Base$_{\textrm{r2n2}}$-max pooling &6.4&\textbf{10.0}&\textbf{15.1}&20.2&\textbf{25.3}&\textbf{30.2}&\textbf{35.4}\\
Base$_{\textrm{r2n2}}$-mean pooling &\textbf{6.3}&10.1&\textbf{15.1}&\textbf{20.1}&\textbf{25.3}&30.3&35.5 \\
Base$_{\textrm{r2n2}}$-sum pooling &6.4&10.1&\textbf{15.1}&\textbf{20.1}&\textbf{25.3}&30.3&35.5 \\
Base$_{\textrm{r2n2}}$-BP pooling &6.5 &10.5 & 15.6 & 20.5&25.7& 30.6&35.8 \\
Base$_{\textrm{r2n2}}$-MHBN pooling &6.5 &10.3 & 15.3 & 20.3&25.5& 30.7&35.7 \\
Base$_{\textrm{r2n2}}$-SMSO pooling &6.5 &10.2 & 15.3 & 20.3&25.4&30.5&35.6 \\
\textbf{Base$_{\textrm{r2n2}}$-\nicknameAttSets{}(Ours)} &7.7&11.0&16.3&21.2&26.3&31.4&36.4\\
\hline
\end{tabular}
\end{table}

\subsection{Computational Efficiency}
To evaluate the computation and memory cost of \nicknameAttSets{}, we implement Base$_{\textrm{r2n2}}$-AttSets and the competing approaches in Python 2.7 and Tensorflow 1.2 with CUDA 9.0 and cuDNN 7.1 as the back-end driver and library. All approaches share the same Base$_{\textrm{r2n2}}$ network and run in the same Titan X and software environments. Table \ref{tab:time_con} shows the average time consumption to reconstruct a single 3D object given different number of images. Our \nicknameAttSets{} based approach is as efficient as the pooling methods, while Base$_{\textrm{r2n2}}$-GRU (\ie{} 3D-R2N2) takes more time when processing an increasing number of images due to the sequential computation mechanism of its GRU module. In terms of the total trainable weights, the max/mean/sum pooling based approaches have $16.66$ million, while \nicknameAttSets{} based net has $17.71$ million. By contrast, the original 3D-R2N2 has $34.78$ million, the BP/MHBN/SMSO have $141.57, 60.78$ and $17.71$ million respectively. Overall, our \nicknameAttSets{} outperforms the recurrent unit and pooling operations without incurring notable computation and memory cost.

\begin{table*}[t]
\caption{ Mean IoU of \nicknameAttSets{} variants on all 13 categories in ShapeNet$_{\textrm{r2n2}}$ testing split.}
\centering
\small
\label{tab:iou_variants}
\tabcolsep=0.05cm
\begin{tabular}{ l|cccccccccc}
\hline
number of views&1&2&3&4&5&8&12&16&20&24 \\
\hline
\textbf{\scriptsize{Base$_{\textrm{r2n2}}$-\nicknameAttSets{} ($conv2d$)}}&\textbf{0.642}&0.648&0.651&0.655&0.657&0.664&0.668&0.674&0.675&0.676\\
\textbf{\scriptsize{Base$_{\textrm{r2n2}}$-\nicknameAttSets{} ($conv3d$)}}&\textbf{0.642}&\textbf{0.663}&\textbf{0.671}&\textbf{0.676}&\textbf{0.677}&0.683&0.685&0.689&0.690&0.690 \\
\textbf{\scriptsize{Base$_{\textrm{r2n2}}$-\nicknameAttSets{} ($fc$)}} &\textbf{0.642}&0.660&0.668&0.674&0.676&\textbf{0.684}
&\textbf{0.688}&\textbf{0.693}&\textbf{0.694}&\textbf{0.695} \\
\hline
\end{tabular}
\end{table*}

\begin{figure}
\centering
   \includegraphics[width=1\linewidth]{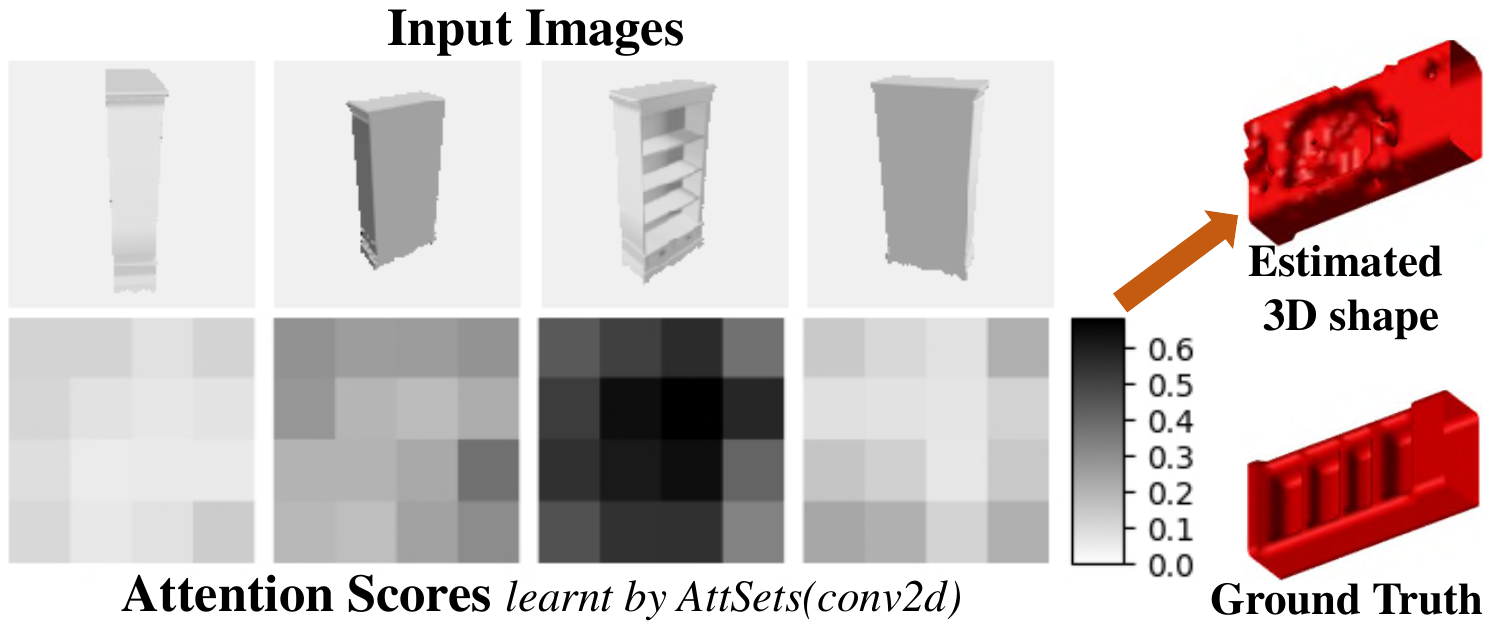}
\caption{Learnt attention scores for deep feature sets via $conv2d$ based \nicknameAttSets{}.}
\label{fig:atts_w}
\end{figure}

\subsection{Comparison between Variants of \nicknameAttSets{}}
We further compare the aggregation performance of $fc$, $conv2d$ and $conv3d$ based \nicknameAttSets{} variants in Figure \ref{fig:ch4_attsets_imp} in Section \ref{sec:ch4_impl}. The $fc$ based \nicknameAttSets{} net is the same as in Section \ref{sec:ch4_eval_r2n2}. The $conv2d$ based \nicknameAttSets{} is inserted into the middle of the 2D encoder, fusing a $(N, 4, 4, 256)$ tensor into $(1, 4, 4, 256)$, where $N$ is an arbitrary image number. The $conv3d$ based \nicknameAttSets{} is inserted into the middle of the 3D decoder, integrating a $(N, 8, 8, 8, 128)$ tensor into $(1, 8, 8, 8, 128)$. All other layers of these variants are the same. Both the $conv2d$ and $conv3d$ based \nicknameAttSets{} networks are trained using the paradigm of experiment Group 4 in Section \ref{sec:ch4_eval_r2n2}.
Table \ref{tab:iou_variants} shows the mean IoU scores of three variants on ShapeNet$_{\textrm{r2n2}}$ testing split. The $fc$ and $conv3d$ based variants achieve similar IoU scores for either single or multi view 3D reconstruction, demonstrating the superior aggregation capability of \nicknameAttSets{}. In the meantime, we observe that the overall performance of $conv2d$ based \nicknameAttSets{} net is slightly decreased compared with the other two. One possible reason is that the 2D feature set has been aggregated at an early layer of the network, resulting in features being lost early. Figure \ref{fig:atts_w} visualizes the learnt attention scores for a 2D feature set, \ie{} $(N,4,4,256)$ features, via the $conv2d$ based \nicknameAttSets{} net. To visualize 2D feature scores, we average the scores along the channel axis and then roughly trace back the spatial locations of those scores corresponding to the original input. The more visual information the input image has, the higher the attention scores that are learnt by \nicknameAttSets{} for the corresponding latent features. For example, the third image has richer visual information than the first image, so its attention scores are higher. Note that, for a specific base network, there are many potential locations to drop in \nicknameAttSets{} and it is also possible to include multiple \nicknameAttSets{} modules into the same net. How to fully evaluate these factors is suggested as an interesting direction for future work.

\begin{table*}[t]
\caption{ Mean IoU of all 13 categories in ShapeNet$_{\textrm{r2n2}}$ testing split for feature-wise and element-wise attentional aggregation.}
\centering
\label{tab:iou_fw_ew}
\tabcolsep=0.02cm
\begin{tabular}{ l|cccccccccc}
\hline
number of views&1&2&3&4&5&8&12&16&20&24 \\
\hline
\textbf{\scriptsize{Base$_{\textrm{r2n2}}$-\nicknameAttSets{}}} \scriptsize{(element)} &\textbf{0.642}&0.653&0.657&0.660&0.661&0.665&0.667&0.670&0.671&0.672\\
\textbf{\scriptsize{Base$_{\textrm{r2n2}}$-\nicknameAttSets{}} \scriptsize{(feature)}} &\textbf{0.642}&\textbf{0.660}&\textbf{0.668}&\textbf{0.674}&\textbf{0.676}&\textbf{0.684}
&\textbf{0.688}&\textbf{0.693}&\textbf{0.694}&\textbf{0.695} \\
\hline
\end{tabular}
\end{table*}

\begin{figure}
\centering
   \includegraphics[width=1\linewidth]{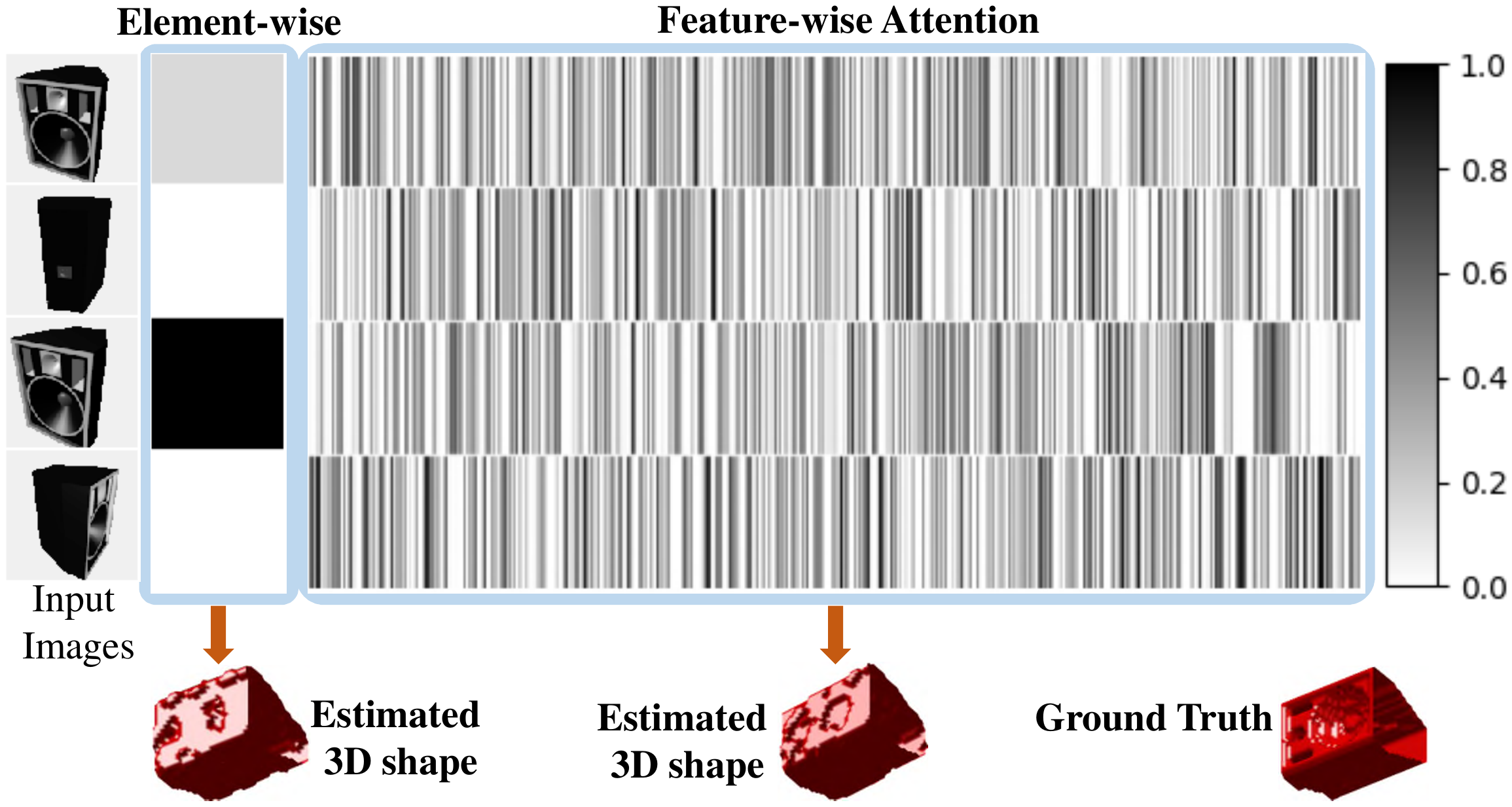}
\caption{Learnt attention scores for deep feature sets via element-wise attention and feature-wise attention \nicknameAttSets{}.}
\label{fig:atts_ele}
\end{figure}

\subsection{Feature-wise Attention $vs.$ Element-wise Attention}
Our \nicknameAttSets{} module is initially designed to learn unique feature-wise attention scores for the whole input deep feature set, and we demonstrate that it significantly improves the aggregation performance over dynamic feature sets in Sections \ref{sec:ch4_eval_r2n2}, \ref{sec:ch4_eval_lsm}, \ref{sec:ch4_eval_modelnet40} and \ref{sec:ch4_eval_blobby}. In this section, we further investigate the advantage of this feature-wise attentive pooling over element-wise attentional aggregation.

For element-wise attentional aggregation, the \nicknameAttSets{} module tends to learn a single attention score for each element of the feature set $\mathcal{A} = \{\boldsymbol{x}_1, \boldsymbol{x}_2, \cdots, \boldsymbol{x}_N\}$, followed by the $softmax$ normalization and weighted summation pooling. In particular, as shown in Figure \ref{fig:ch4_attsets_f}, the shared function $g(\boldsymbol{x}_n, \boldsymbol{W})$ now learns a scalar, instead of a vector, as the attention activation for each input element. Eventually, all features within the same element are weighted by a learnt common attention score. Intuitively, the original feature-wise \nicknameAttSets{} tends to be fine-grained aggregation, while the element-wise \nicknameAttSets{} learns coarse aggregate features.

Following the same training settings of experiment Group 4 in Section \ref{sec:ch4_eval_r2n2}, we conduct another group of experiments on the ShapeNet$_{\textrm{r2n2}}$ dataset for element-wise attentional aggregation. Table \ref{tab:iou_fw_ew} compares the mean IoU for 3D object reconstruction through feature-wise and element-wise attentional aggregation. Figure \ref{fig:atts_ele} shows an example of the learnt attention scores and the predicted 3D shapes. As expected, the feature-wise attention mechanism clearly achieves better aggregation performance compared with the coarse element-wise approach. As shown in Figure \ref{fig:atts_ele}, the element-wise attention mechanism tends to focus on a few images, while completely ignoring others. By comparison, the feature-wise \nicknameAttSets{} learns to fuse information across all images, thus achieving better aggregation performance.

\begin{table*}[t]
\caption{ Mean IoU of different training algorithms on all 13 categories in ShapeNet$_{\textrm{r2n2}}$ testing split.}
\centering
\small
\label{tab:iou_train_alg}
\tabcolsep=0.05cm
\begin{tabular}{ l|cccccccccc}
\hline
number of views&1&2&3&4&5&8&12&16&20&24 \\
\hline
\textbf{\scriptsize{Base$_{\textrm{r2n2}}$-\nicknameAttSets{}}} (JoinT)&0.307&0.437&0.516&0.563&0.595&0.639&0.659&0.673&0.677&0.680\\
\textbf{\scriptsize{Base$_{\textrm{r2n2}}$-\nicknameAttSets{} (\nicknameFASet{})}} &\textbf{0.642}&\textbf{0.660}&\textbf{0.668}&\textbf{0.674}&\textbf{0.676}&\textbf{0.684}
&\textbf{0.688}&\textbf{0.693}&\textbf{0.694}&\textbf{0.695} \\
\hline
\end{tabular}
\end{table*}

\subsection{Significance of \nicknameFASet{} Algorithm}\label{sec:ch4_sig_faset}
In this section, we investigate the impact of \nicknameFASet{} algorithm by comparing it with the standard end-to-end joint training (JoinT). Particularly, in JoinT, all parameters $\Theta_{base}$ and $\Theta_{att}$ are jointly optimized with a single loss. Following the same training settings of experiment Group 4 in Section \ref{sec:ch4_eval_r2n2}, we conduct another group of experiments on ShapeNet$_{\textrm{r2n2}}$ dataset under the JoinT training strategy. As IoU scores shown in Table \ref{tab:iou_train_alg}, the JoinT training approach tends to optimize the whole net given the training multi-view batches, thus being unable to generalize well for fewer images during testing. Effectively, the network itself is unable to dedicate its base layers to learning visual features, decoupling the \nicknameAttSets{} module to learning attention scores. The theoretical reason behind this has been discussed previously in Section \ref{sec:ch4_optim_motiv}. The \nicknameFASet{} algorithm may also be applicable to other learning based aggregation approaches, as long as the aggregation module can be decoupled from the base encoder/decoder.
\section{Conclusion}
In this chapter, we present \nicknameAttSets{} module and the \nicknameFASet{} training algorithm to aggregate elements of deep feature sets. \nicknameAttSets{} together with \nicknameFASet{} has powerful permutation invariance, computation efficiency, robustness and flexible implementation properties, along with theoretical underpinnings and extensive experiments to support its prowess for multi-view 3D reconstruction. Both quantitative and qualitative results explicitly show that \nicknameAttSets{} significantly outperforms other widely used aggregation approaches. Nonetheless, all of our experiments are dedicated to multi-view 3D reconstruction. It would be interesting to explore the generality of \nicknameAttSets{} and \nicknameFASet{} over other set-based tasks such as point cloud classification and semantic segmentation \cite{Qi2016}, multi-view object recognition \cite{Su2015} and image tagging \cite{Zaheer2017}. These tasks usually take an arbitrary number of elements as input. 

\chapter{Learning to Segment 3D Objects from Point Clouds}
\label{chap:seg_obj_pc}
\section{Introduction}
Imbuing machines with the ability to understand 3D scenes is a fundamental necessity for a number of key applications such as autonomous driving, augmented reality and robotics. Core problems over 3D geometric data, such as point clouds, include semantic segmentation, object detection and instance segmentation. Of these problems, instance segmentation has only started to be tackled in the literature. The primary obstacle is that point clouds are inherently unordered, unstructured and unevenly sampled. Widely used convolutional neural networks require the 3D point clouds to be voxelized into a regular grid, incurring high computational and memory costs.

\begin{figure}[h]
\centering
   \includegraphics[width=.8\linewidth]{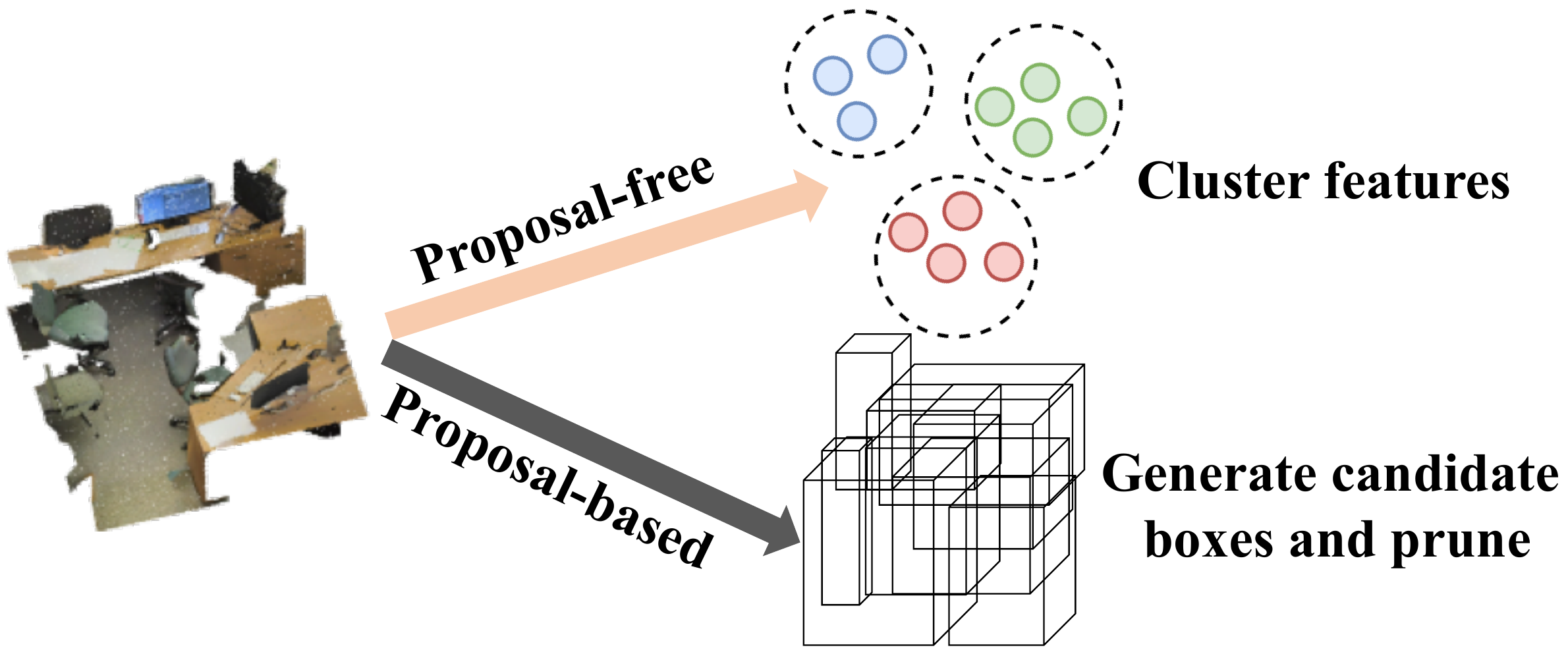}
\caption{The existing pipelines for instance segmentation on 3D point clouds.}
\label{fig:bonet_existing}
\end{figure}

The first neural algorithm to directly tackle 3D instance segmentation is SGPN \cite{Wang2018d}, which learns to group per-point features through a similarity matrix. Similarly, ASIS \cite{Wang2019}, JSIS3D \cite{Pham2019}, MASC \cite{Liu2019}, 3D-BEVIS \cite{Elich2019} and \cite{Liang2019} apply the same per-point feature grouping pipeline to segment 3D instances. Mo \etal{} formulate the instance segmentation as a per-point feature classification problem in PartNet \cite{Mo2019}. However, the learnt segments of these proposal-free methods do not have high objectness as they do not explicitly detect the object boundaries. In addition, they inevitably require a post-processing step such as mean-shift clustering \cite{Comaniciu2002} to obtain the final instance labels, which is computationally heavy. Another pipeline is the proposal-based 3D-SIS \cite{Hou2019} and GSPN \cite{Yi2019}, which usually rely on two-stage training and the expensive non-maximum suppression to prune dense object proposals. 

Figure \ref{fig:bonet_existing} illustrates the existing two pipelines for instance segmentation for 3D point clouds. Overall, the proposal-free methods learn to cluster point features, while the proposal-based approaches firstly generate candidate bounding boxes based on spatial anchors and then classify the points within each bounding box.

In this chapter, we present an elegant, efficient, and novel framework for 3D instance segmentation, where objects are loosely but uniquely detected through a single-forward stage using efficient MLPs, and then each instance is precisely segmented through a simple point-level binary classifier. To this end, we introduce a new bounding box prediction module together with a series of carefully designed loss functions to directly learn object boundaries. Our framework is significantly different from existing proposal-based and proposal-free approaches, since we are able to efficiently segment all instances with high objectness, but without relying on expensive and dense object proposals. Our code and data are available at \textit{https://github.com/Yang7879/3D-BoNet}.

\begin{figure}[t]
\centering
   \includegraphics[width=1.0\linewidth]{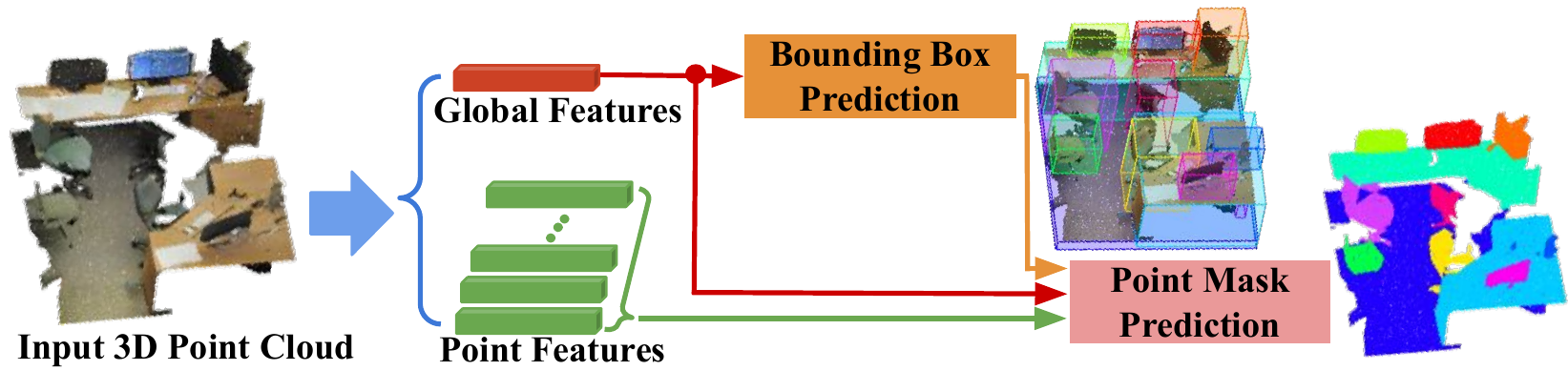}
\caption{The \nicknameBoNet{} framework for instance segmentation on 3D point clouds.}
\label{fig:bonet_view}
\end{figure}

As shown in Figure \ref{fig:bonet_view}, our framework, called \textbf{\nicknameBoNet{}}, is a single-stage, anchor-free, and end-to-end trainable neural architecture. It first uses an existing backbone network to extract a local feature vector for each point and a global feature vector for the whole input point cloud. The backbone is followed by two branches: 1) instance-level bounding box prediction, and 2) point-level mask prediction for instance segmentation. 

\setlength{\columnsep}{15pt}
\begin{wrapfigure}{R}{0.4\textwidth}
\raisebox{0pt}[\dimexpr\height-1.2\baselineskip\relax]{
\centering
   \includegraphics[width=1\linewidth]{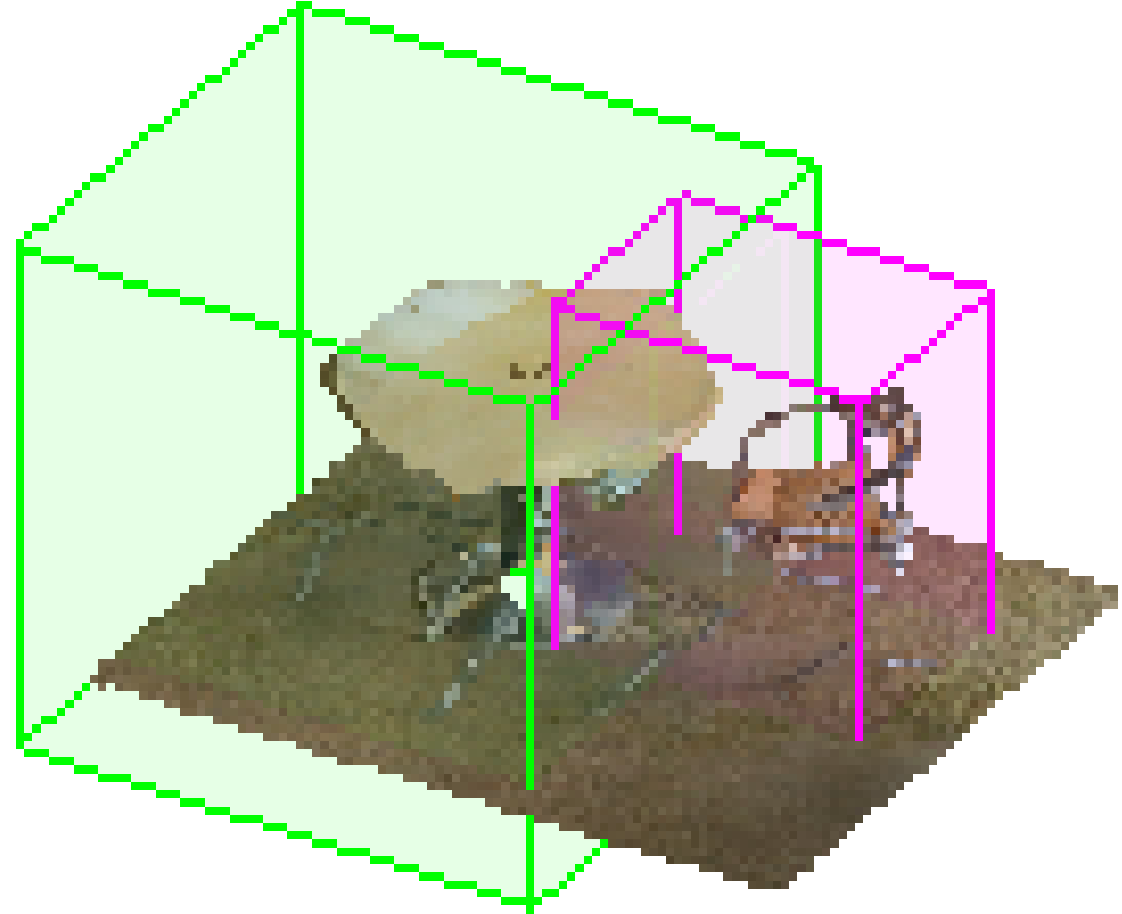}
}
\caption{Rough instance boxes.}
\label{fig:bonet_bbox_intro}
\end{wrapfigure}

The \textbf{bounding box prediction branch} is the core of our framework. This branch aims to predict a unique, unoriented and rectangular bounding box for each instance in a single forward stage, without relying on predefined spatial anchors or a region proposal network \cite{Ren2015a}. As shown in Figure \ref{fig:bonet_bbox_intro}, we believe that roughly drawing a 3D bounding box for an instance is relatively achievable, because the input point clouds explicitly include 3D geometry information.  Once the object bounding boxes are specified,  it becomes easier to tackle point-level instance segmentation since reasonable bounding boxes can guarantee high objectness for learnt segments. However, learning good instance boxes involves two critical issues: 1) the number of total instances is variable, \ie{} from 1 to many, 2) there is no fixed order for all instances. These issues pose significant challenges for correctly optimizing the network, because there is no information to directly link predicted boxes with ground truth labels to supervise the network. However, we show how to elegantly solve these issues. The proposed box prediction branch simply takes the global feature vector as input and directly outputs a large and fixed number of bounding boxes together with confidence scores. These scores are used to indicate whether it is likely that the box contains a valid instance or not. To supervise the network, we design a novel \textit{bounding box association layer} followed by a \textit{multi-criteria loss function}. Given a set of ground-truth instances, we need to determine which of the predicted boxes best fit them. We formulate this association process as an optimal assignment problem with an existing solver. After the boxes have been optimally associated, our multi-criteria loss function not only minimizes the Euclidean distance of paired boxes, but also maximizes the coverage of valid points inside the predicted boxes. 

The predicted boxes together with point and global features are then fed into the subsequent \textbf{point mask prediction branch}, in order to predict a point-level binary mask for each instance. The purpose of this branch is to classify whether each point inside a bounding box belongs to the valid instance or the background. Assuming the estimated instance box is reasonably good, it is very likely that an accurate point mask can be obtained, as the purpose of this branch is simply to reject points that are not part of the detected instance. As a degenerate example, randomly guessing could, on average, bring about $50\%$ corrections.

Overall, our framework is distinguished from all existing 3D instance segmentation approaches in three distinct ways: 1) Compared with the proposal-free pipeline, our method segments instances with high objectness by explicitly learning 3D object boundaries. 2) Compared with the widely-used proposal-based approaches, our framework does not require expensive and dense proposals. 3) Our framework is remarkably efficient, since the instance-level masks are learnt in a single-forward pass without requiring any post-processing steps. Our key contributions are:

\begin{itemize}
\item We propose a new framework for instance segmentation on 3D point clouds. The framework is single-stage, anchor-free and end-to-end trainable, without requiring any post-processing steps.
\item We design a novel bounding box association layer followed by a multi-criteria loss function to supervise the box prediction branch.
\item We demonstrate significant improvement over baselines and provide insight and intuition behind our design choices through extensive ablation studies.
\end{itemize}
\section{Method Overview}
\begin{figure*}[t]
\centering
   \includegraphics[width=1\linewidth]{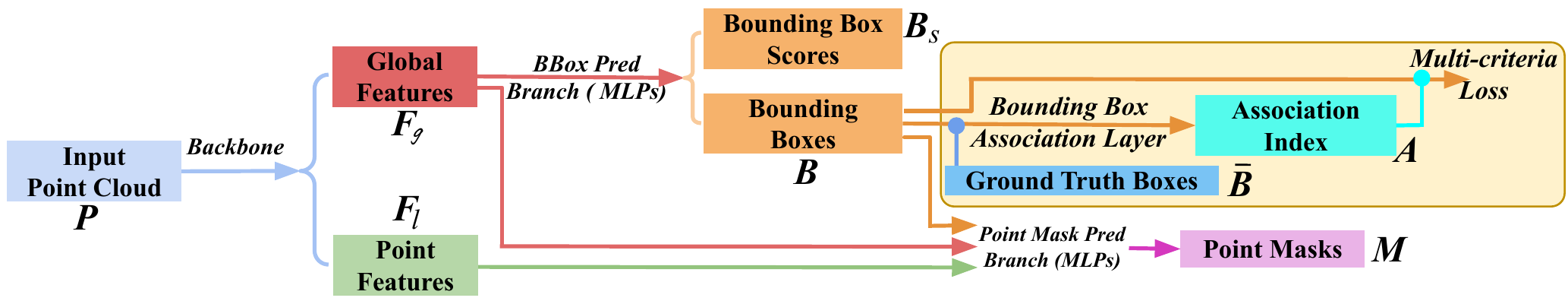}
\caption{The general workflow of \nicknameBoNet{} framework.}
\label{fig:bonet_workflow}
\end{figure*}

As shown in Figure \ref{fig:bonet_workflow}, our framework consists of two branches on top of the backbone network. Given an input point cloud $\boldsymbol{P}$ with $N$ points in total, \ie{} $\boldsymbol{P} \in \mathbb{R}^{N\times k_0} $, where $k_0$ is the number of channels such as the location $\{x,y,z\}$ and color $\{r,g,b\}$ of each point, the \textbf{backbone network} extracts per-point local features, denoted as $\boldsymbol{F}_{l} \in \mathbb{R}^{N \times k}$, and aggregates a global point cloud feature vector, denoted as $\boldsymbol{F}_{g} \in \mathbb{R}^{1\times k}$, where $k$ is the length of feature vectors.

The \textbf{bounding box prediction branch} simply takes the global feature vector $\boldsymbol{F}_{g}$ as input, and directly regresses a predefined and fixed set of bounding boxes, denoted as $\boldsymbol{B}$, and the corresponding box scores, denoted as $\boldsymbol{B}_{s}$. We use ground truth bounding box information to supervise this branch. During training, the predicted bounding boxes $\boldsymbol{B}$ and the ground truth boxes are fed into a \textit{box association layer}. This layer aims to automatically associate a unique and the most similar predicted bounding box to each ground truth box. The output of the association layer is a list of association indices $\boldsymbol{A}$. The indices reorganize the predicted boxes, such that each ground truth box is paired with a unique predicted box for subsequent loss calculation. The predicted bounding box scores are also reordered accordingly before calculating loss. The reordered predicted bounding boxes are then fed into the \textit{multi-criteria loss function}. At a high-level, this loss function aims to not only minimize the Euclidean distance between each ground truth box and the associated predicted box, but also maximize the coverage of valid points inside of each predicted box. Note that, both the bounding box association layer and multi-criteria loss function are only designed for network training. They are not used during testing. Eventually, this branch is able to predict a correct bounding box together with a box score for each instance directly.

In order to predict a point-level binary mask for each instance, every predicted box together with the previous derived local and global features, \ie{} $\boldsymbol{F}_{l}$ and $\boldsymbol{F}_{g}$, are further fed into the \textbf{point mask prediction branch}. This network branch is shared by all instances of different categories, and is therefore extremely light and compact. Such a class-agnostic approach inherently allows general segmentation across unseen categories. 
\section{Bounding Box Prediction}
\subsection{Bounding Box Encoding}
In existing object detection networks, a bounding box is usually represented by the center location and the length of three dimensions \cite{Chen2017b}, or the corresponding residuals \cite{Zhou2018a} together with orientations. Instead, we completely parameterize the rectangular bounding box by specifying the  two extreme vertices i.e. the minimum and maximum for simplicity: 
\begin{equation*}
\{[x_{min} \; y_{min} \; z_{min}],[x_{max}\; y_{max}\; z_{max}]\}
\end{equation*}

\subsection{Neural Layers} 
\begin{figure*}[t]
\centering
   \includegraphics[width=1\linewidth]{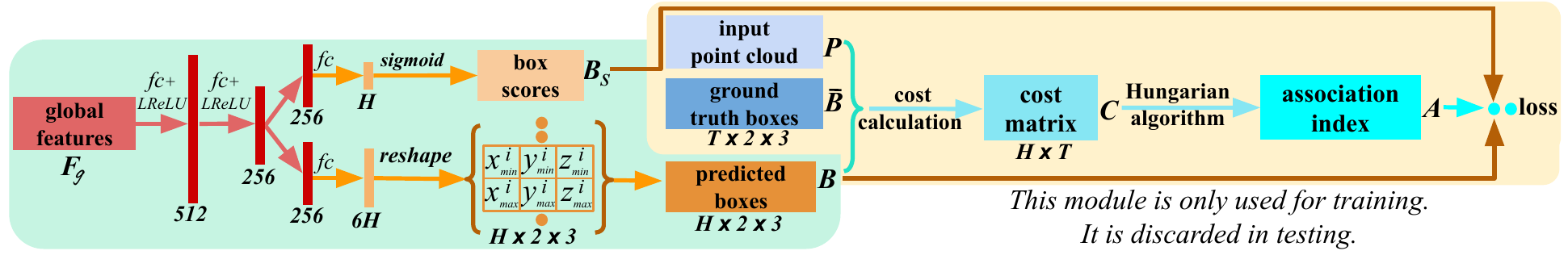}
\caption{The architecture of the bounding box regression branch. The predicted $H$ boxes are optimally associated with $T$ ground truth boxes before calculating the multi-criteria loss.}
\label{fig:bonet_bbox}
\end{figure*}

As shown in Figure \ref{fig:bonet_bbox}, the global feature vector $\boldsymbol{F}_{g}$ is fed through two fully connected layers with Leaky ReLU as the non-linear activation function. This is followed by another two parallel fully connected layers. One layer outputs a $6H$ dimensional vector, which is then reshaped as an $H\times 2\times 3$ tensor. $H$ is a predefined and fixed number of bounding boxes that the whole network is expected to predict as an upper limit. The other layer outputs an $H$ dimensional vector followed by $sigmoid$ function to represent the bounding box scores. The higher the score, the more likely it is that the predicted box contains an instance, indicating that the box is likely to be valid.

\setlength{\columnsep}{12pt}
\begin{wrapfigure}{R}{0.4\textwidth}
\raisebox{0pt}[\dimexpr\height-1.\baselineskip\relax]{
\centering
   \includegraphics[width=0.98\linewidth]{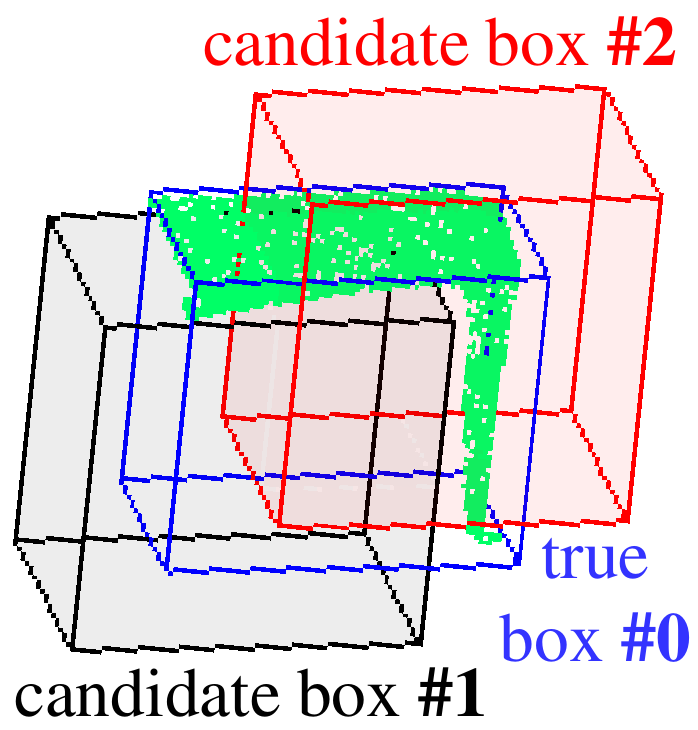}
}
\caption{\small{A sparse input point cloud.}}
\label{fig:bonet_bbox_l2iouce}
\end{wrapfigure}

\subsection{Bounding Box Association Layer}
Given the previously predicted $H$ bounding boxes, \ie{} $\boldsymbol{B} \in \mathbb{R}^{H\times 2\times 3}$, it is not straightforward to take use of the ground truth boxes, denoted as $\boldsymbol{\bar{B}} \in \mathbb{R}^{T\times 2\times 3}$, to supervise the network, because there are no predefined anchors to trace each predicted box back to a corresponding ground truth box in our framework. Besides, for each input point cloud $\boldsymbol{P}$, the number of ground truth boxes $T$ varies and it is usually different from the predefined number $H$, although we can safely assume the predefined number $H \geq T$ for all input point clouds. In addition, there is no box order for either predicted or ground truth boxes.

\textit{(1) Optimal Association Formulation:} To associate a unique predicted bounding box from $\boldsymbol{B}$ for each ground truth box of $\boldsymbol{\bar{B}}$, we formulate this association process as an optimal assignment problem. Formally, let $\boldsymbol{A}$ be a boolean association matrix where $\boldsymbol{A}_{i, j}$ =1 \textit{iff} the $i^{th}$ predicted box is assigned to the $j^{th}$ ground truth box. $\boldsymbol{A}$ is also called association index in this paper. Let $\boldsymbol{C}$ be the association cost matrix where $\boldsymbol{C}_{i, j}$ represents the cost that the $i^{th}$ predicted box is assigned to the $j^{th}$ ground truth box. Basically, the cost $\boldsymbol{C}_{i, j}$ represents the similarity between two boxes; the smaller the cost, the more similar the two boxes. Therefore, the bounding box association problem is to find the optimal assignment matrix $\boldsymbol{A}$ with the minimum overall cost:

\begin{align}
  \boldsymbol{A} &= \argminA_{\boldsymbol{A}} \sum_{i=1}^{H}\sum_{j=1}^{T} \boldsymbol{C}_{i,j} \boldsymbol{A}_{i,j} \nonumber \\ \text{subject to } \sum_{i=1}^{H} \boldsymbol{A}_{i,j} &= 1, \sum_{j=1}^{T} \boldsymbol{A}_{i,j} \leq 1, \small{j \in \{1.. T\}, i \in \{1.. H\}} 
\end{align}

To solve the above optimal association problem, the existing Hungarian algorithm \cite{Kuhn1955,Kuhn1956} is applied. 

(2) \textit{Association Matrix Calculation:} To evaluate the similarity between the $i^{th}$ predicted box and the $j^{th}$ ground truth box, a simple and intuitive criterion is the Euclidean distance between two pairs of min-max vertices. However, it is not optimal. In essence, we want the predicted box to include as many valid points as possible. As illustrated in Figure \ref{fig:bonet_bbox_l2iouce}, the input point cloud is usually sparse and distributed non-uniformly in 3D space. With regards to the ground truth box \#0 (blue), the candidate box \#2 (red) is believed to be much better than the candidate \#1 (black), because the box \#2 has more valid points overlapped with \#0. Therefore, the coverage of valid points should also be included to calculate the cost matrix $\boldsymbol{C}$. In this chapter, we consider the following three criteria:

\begin{itemize}
    \item Euclidean Distance between Vertices. Formally, the cost between the $i^{th}$ predicted box $\boldsymbol{B}_{i}$ and the $j^{th}$ ground truth box $\boldsymbol{\bar{B}}_{j}$ is calculated as follows:
    \begin{equation}
    \boldsymbol{C}_{i,j}^{ed} = \frac{1}{6} \sum (\boldsymbol{B}_i - \boldsymbol{\bar{B}}_j)^2
    \end{equation}
    
    \item Soft Intersection-over-Union on Points. Given the input point cloud $\boldsymbol{P}$ and the $j^{th}$ ground truth instance box $\boldsymbol{\bar{B}}_j$, it is able to directly obtain a hard-binary vector $\boldsymbol{\bar{q}}_j \in \mathbb{R}^N$ to represent whether each point of the whole input point cloud $\boldsymbol{P}$ is inside the box or not, where `1' indicates the point being inside and `0' outside. However, for a specific $i^{th}$ predicted box of the same input point cloud $\boldsymbol{P}$, to directly obtain a similar hard-binary vector would result in the framework being non-differentiable, due to the discretization operation. Therefore, we introduce a differentiable yet simple Algorithm \ref{alg:ppp} to obtain a similar but soft-binary vector $\boldsymbol{q}_i$, called \textbf{point-in-pred-box-probability}, where all values are in the range $(0, 1)$. The deeper the corresponding point is inside of the box, the higher the value. The farther away the point is outside, the smaller the value. Formally, the Soft Intersection-over-Union (sIoU) cost between the $i^{th}$ predicted box and the $j^{th}$ ground truth box is defined as follows:
    \begin{equation}
    \boldsymbol{C}_{i,j}^{sIoU} = \frac{ -\sum_{n=1}^N(q^n_i * \bar{q}^n_j)}{ \sum_{n=1}^N q^n_i+\sum_{n=1}^{N}\bar{q}^n_j- \sum_{n=1}^{N}(q^n_i *\bar{q}^n_j)}
    \end{equation}
    \textit{where} $q^n_i$ and $\bar{q}^n_j$ are the $n^{th}$ values of $\boldsymbol{q}_i$ and $\boldsymbol{\bar{q}}_j$, $N$ is the total number of points in $\boldsymbol{P}$. 
    
    \item Cross-Entropy Score. In addition, we also consider the cross-entropy score between $\boldsymbol{q}_i$ and $\boldsymbol{\bar{q}}_j$. Unlike sIoU cost which prefers tighter boxes, this score represents the confidence with which a predicted bounding box includes as many valid points as possible. It prefers larger and more inclusive boxes, and is formally defined as:
    \begin{equation}
    \boldsymbol{C}_{i,j}^{ces} =-\frac{1}{N} \sum_{n=1}^{N} \left[\bar{q}^n_j\log q^n_i + (1 - \bar{q}^n_j)\log(1-q^n_i)\right]
    \end{equation}
    \textit{where} $q^n_i$ and $\bar{q}^n_j$ are the $n^{th}$ values of $\boldsymbol{q}_i$ and $\boldsymbol{\bar{q}}_j$, $N$ is the total number of points in $\boldsymbol{P}$. 
\end{itemize}

Overall, the criterion (1) guarantees the geometric boundaries for learnt boxes and criteria (2)(3) maximize the coverage of valid points and overcome the non-uniformity as illustrated in Figure \ref{fig:bonet_bbox_l2iouce}. The final association cost between the $i^{th}$ predicted box and the $j^{th}$ ground truth box is defined as:
\begin{equation}\label{eq:cij}
\boldsymbol{C}_{i,j} = \boldsymbol{C}_{i,j}^{ed} + \boldsymbol{C}_{i,j}^{sIoU} + \boldsymbol{C}_{i,j}^{ces}
\end{equation}

Note that we assign an equal weight `1' to all the three criteria in Equation \ref{eq:cij} for simplicity. Since different criteria tend to favor different densities of input point clouds, it would be ideal to learn a combination of the three criteria automatically. This is suggested as an interesting direction for future work.

\begin{algorithm}[H]
\caption{ An algorithm to calculate point-in-pred-box-probability. $H$ is the number of predicted bounding boxes $\boldsymbol{B}$, $N$ is the number of points in point cloud $\boldsymbol{P}$, $\theta_1$ and $\theta_2$ are hyperparameters for numerical stability. We use $\theta_1=100$, $\theta_2=20$ in all our implementation.
}
\label{alg:ppp}
\begin{algorithmic} 
    \FOR {$i \leftarrow 1$ to $H$}{}
    \STATE{$\bullet$ the $i^{th}$ box min-vertex $\boldsymbol{B}^i_{min} = [x^i_{min} \; y^i_{min} \; z^i_{min}]$.}
    \STATE{$\bullet$ the $i^{th}$ box max-vertex $\boldsymbol{B}^i_{max} = [x^i_{max} \; y^i_{max} \; z^i_{max}]$.}
   \FOR{$n \leftarrow 1$ to $N$} 
    \STATE{$\bullet$ the $n^{th}$ point location $\boldsymbol{P}^n = [x^n \; y^n \; z^n]$.}
    \STATE {$\bullet$ step 1: $\boldsymbol{\Delta}_{xyz} \leftarrow (\boldsymbol{B}^i_{min} - \boldsymbol{P}^n)(\boldsymbol{P}^n - \boldsymbol{B}^i_{max})$.}
    \STATE {$\bullet$ step 2: $\boldsymbol{\Delta}_{xyz} \leftarrow max\left[ min(\theta_1\boldsymbol{\Delta}_{xyz}, \theta_2), -\theta_2 \right]$.}
    \STATE {$\bullet$ step 3: probability $\boldsymbol{p}_{xyz}= \frac{\boldsymbol{ 1}}{ \boldsymbol{1}+\exp(- \boldsymbol{\Delta}_{xyz} ) }$.}
    \STATE {$\bullet$ step 4: point probability $q^n_i = min(\boldsymbol{p}_{xyz})$.}
   \ENDFOR
   \STATE {$\bullet$ obtain the soft-binary vector $\boldsymbol{q}_i = [q^1_i \cdots q^N_i]$.}
  \ENDFOR
  \\The above two loops are only for illustration. They are easily replaced by standard and efficient matrix operations.
\end{algorithmic}
\end{algorithm}

\subsection{Loss Functions} 
After the bounding box association layer, both the predicted boxes $\boldsymbol{B}$ and scores $\boldsymbol{B}_s$ are reordered using the association index $\boldsymbol{A}$, such that the first predicted $T$ boxes and scores are well paired with the $T$ ground truth boxes.

\textit{(1) Multi-criteria Loss for Box Prediction}: The previous association layer finds the most similar predicted box for each ground truth box according to the minimal cost including: 1) vertex Euclidean distance, 2) sIoU cost on points, and 3) cross-entropy score. Therefore, the loss function for bounding box prediction is naturally designed to consistently minimize those cost. It is formally defined as follows:
\begin{equation}
\ell_{bbox} = \frac{1}{T}\sum_{t=1}^{T} (\boldsymbol{C}^{ed}_{t,t}+ \boldsymbol{C}^{sIoU}_{t,t} + \boldsymbol{C}^{ces}_{t,t}) 
\end{equation}
\textit{where} $\boldsymbol{C}^{ed}_{t,t}$, $\boldsymbol{C}^{sIoU}_{t,t}$ and $\boldsymbol{C}^{ces}_{t,t}$ are the cost of $t^{th}$ paired boxes. Note that, we only minimize the cost of $T$ paired boxes; the remaining $H-T$ predicted boxes are ignored because there is no corresponding ground truth for them. Therefore, this box prediction sub-branch is agnostic to the predefined value of $H$. Here however arises an issue. Since the $H-T$ negative predictions are not penalized, it might be possible that the network predicts multiple similar boxes for a single instance. Fortunately, the loss function for the parallel box score prediction is able to alleviate this problem.

\textit{(2) Loss for Box Score Prediction}: The predicted box scores aim to indicate the validity of the corresponding predicted boxes. After being reordered by the association index $\boldsymbol{A}$, the ground truth scores for the first $T$ scores are all `1', and `0' for the remaining invalid $H-T$ scores. We use cross-entropy loss for this binary classification task:
\begin{equation}
\ell_{bbs} = -\frac{1}{H} \left[ \sum^T_{t=1} \log\boldsymbol{B}^{t}_s + \sum^H_{t=T+1} \log(1-\boldsymbol{B}^t_s) \right] 
\end{equation}
\textit{where} $\boldsymbol{B}^{t}_s$ is the $t^{th}$ predicted score after being associated. Basically, this loss function rewards the correctly predicted bounding boxes, while implicitly penalizing the cases where multiple similar boxes are regressed for a single instance. 

\begin{figure*}[t]
\centering
   \includegraphics[width=0.7\linewidth]{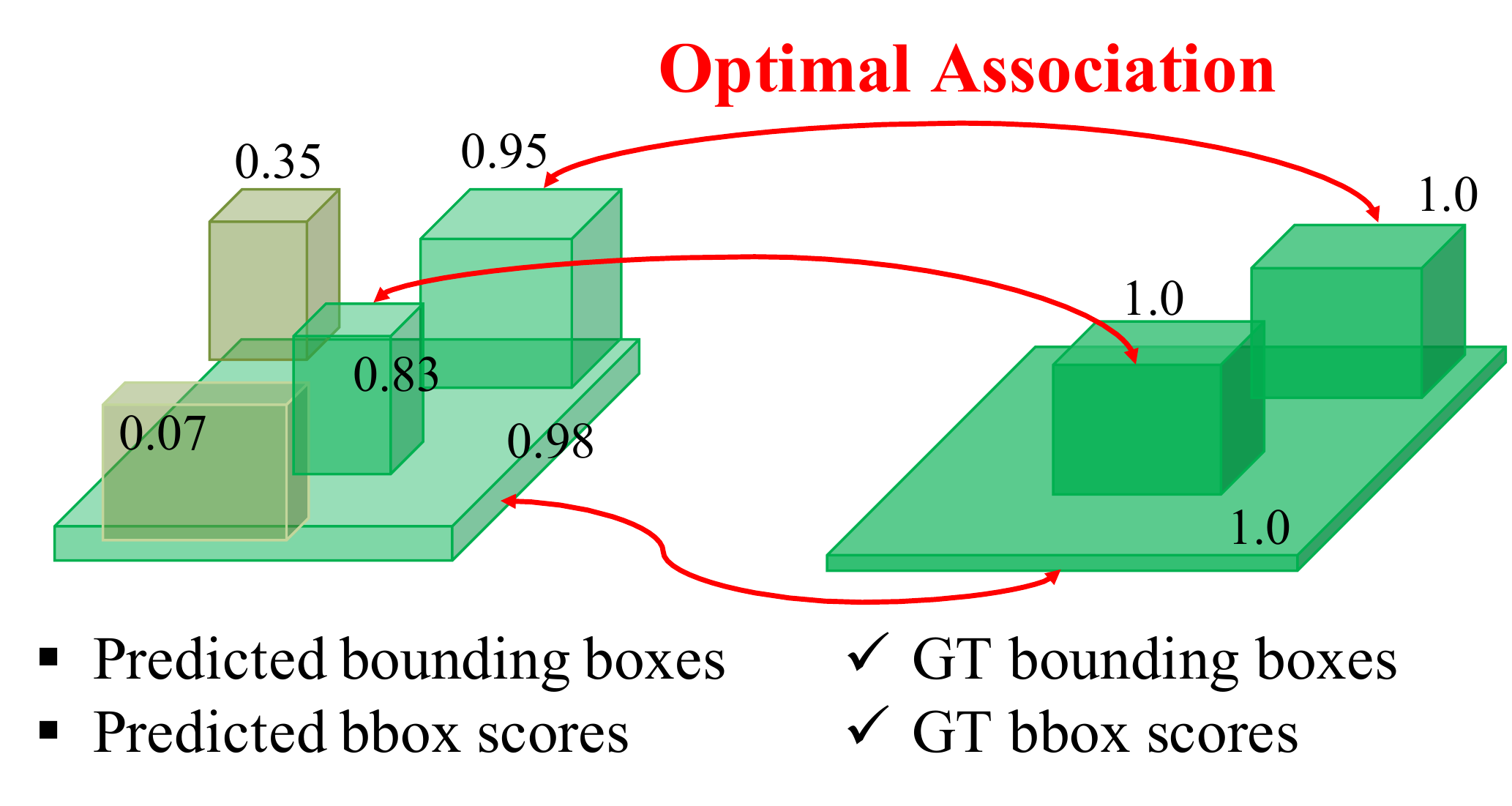}
\caption{Illustration of the proposed bounding box association layer.}
\label{fig:bonet_bbox_ass}
\end{figure*}

Overall, Figure \ref{fig:bonet_bbox_ass} illustrates the proposed bounding box association layer with the optimal assignment algorithm.  

\subsection{Gradient Estimation for Hungarian Algorithm}
Given the predicted bounding boxes, $\mathbf{B}$, and ground-truth boxes, $\mathbf{\Bar{B}}$, we compute the assignment cost matrix, $\mathbf{C}$. This matrix is converted to a permutation matrix, $\mathbf{A}$, using the Hungarian algorithm. Here we focus on the euclidean distance component of the loss, $\mathbf{C}^{ed}$. The derivative of our loss component w.r.t the network parameters, $\theta$, in matrix form is: 
\begin{equation}\label{eq:gradient}
\frac{\partial \mathbf{C}^{ed}}{\partial \theta} = -2(\mathbf{A}\mathbf{B} - \mathbf{\Bar{B}}) \left[\mathbf{A} + \frac{\partial \mathbf{A} }{\partial \mathbf{C}} \frac{\partial \mathbf{C} }{\partial \mathbf{B}} \mathbf{B} \right]^T \frac{\partial \mathbf{B}}{\partial \theta} 
\end{equation}

The components are easily computable except for $\frac{\partial \mathbf{A} }{\partial \mathbf{C}}$ which is the gradient of the permutation w.r.t the assignment cost matrix which is zero nearly everywhere. In our implementation, we found that the network is able to converge when setting this term to zero.

However, convergence could be sped up using the straight-through-estimator \cite{Bengio2013}, which assumes that the gradient of the rounding is identity (or a small constant), $\frac{\partial \mathbf{A} }{\partial \mathbf{C}} = \mathds{1}$. 
This would speed up convergence as it allows both the error in the bounding box alignment (1st term of Eq. \ref{eq:gradient}) to be backpropagated and the assignment to be reinforced (2nd term of Eq. \ref{eq:gradient}). This approach has been shown to work well in practice for many problems including for differentiating through permutations for solving combinatorial optimization problems \cite{Emami2018} and for training binary neural networks \cite{Yin2019}. More complex approaches could also be used in our framework for computing the gradient of the assignment such as \cite{Grover2019} 
which uses a Plackett-Luce distribution over permutations and a reparameterized gradient estimator.
\section{Point Mask Prediction}
Given the predicted bounding boxes $\boldsymbol{B}$, the learnt point features $\boldsymbol{F}_l$ and global features $\boldsymbol{F}_g$, the point mask prediction branch processes each bounding box individually with shared neural layers.

\begin{figure*}[t]
\centering
   \includegraphics[width=1\linewidth]{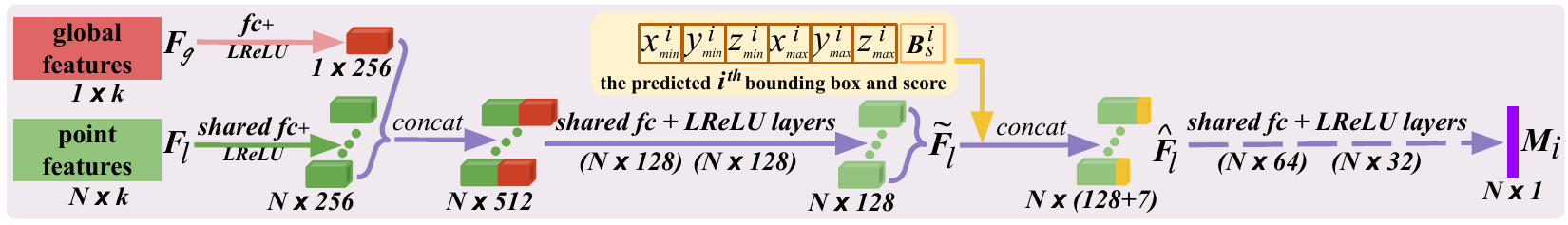}
\caption{The architecture of point mask prediction branch. The point features are fused with each bounding box and score, after which a point-level binary mask is predicted for each instance.}
\label{fig:bonet_pmask}
\end{figure*}

\subsection{Neural Layers} 
As shown in Figure \ref{fig:bonet_pmask}, both the point and global features are compressed to be $256$ dimensional vectors through fully connected layers, before being concatenated and further compressed to be $128$ dimensional mixed point features $\widetilde{\boldsymbol{F}}_{l}$. For the $i^{th}$ predicted bounding box $\boldsymbol{B}_i$, the estimated vertices and score are fused with features $\widetilde{\boldsymbol{F}}_l$ through concatenation, producing box-aware features $\widehat{\boldsymbol{F}}_l$. These features are then fed through shared layers, predicting a point-level binary mask, denoted as $\boldsymbol{M}_i$. We use $sigmoid$ as the final activation function. This simple box fusing approach is extremely computationally efficient, compared with the commonly used RoIAlign in prior art \cite{Yi2019,Hou2019,He2017a} which involves expensive point feature sampling and alignment. 

\subsection{Loss Function}
The predicted instance masks $\boldsymbol{M}$ are similarly associated with the ground truth masks according to the previous association index $\boldsymbol{A}$. Due to the imbalance of instance and background point numbers, we use focal loss \cite{Lin2017c} with default hyper-parameters instead of the standard cross-entropy loss to optimize this branch. Only the valid $T$ paired masks are used for the loss $\ell_{pmask}$. It is defined as follows:
\begin{align}
\ell_{pmask} = -\frac{1}{TN}\sum^T_{t=1}\sum^N_{n=1} \left[ \alpha (1-\boldsymbol{M}^n_t)^{\gamma} \boldsymbol{\bar{M}}^n_t \log \boldsymbol{M}^n_t + (1-\alpha) {(\boldsymbol{M}^n_t)}^{\gamma}
(1-\boldsymbol{\bar{M}}^n_t) \log(1 - \boldsymbol{M}^n_t) \right]
\end{align}
\textit{where} $\boldsymbol{M}^n_t$, $\boldsymbol{\bar{M}}^n_t$ are the predicted and ground truth point-level mask values for the $t^{th}$ paired instances. 
We use the default 0.75 and 2 for $\alpha$ and $\gamma$ in all our experiments.
\section{Implementation}
\begin{figure*}[t]
\centering
   \includegraphics[width=1\linewidth]{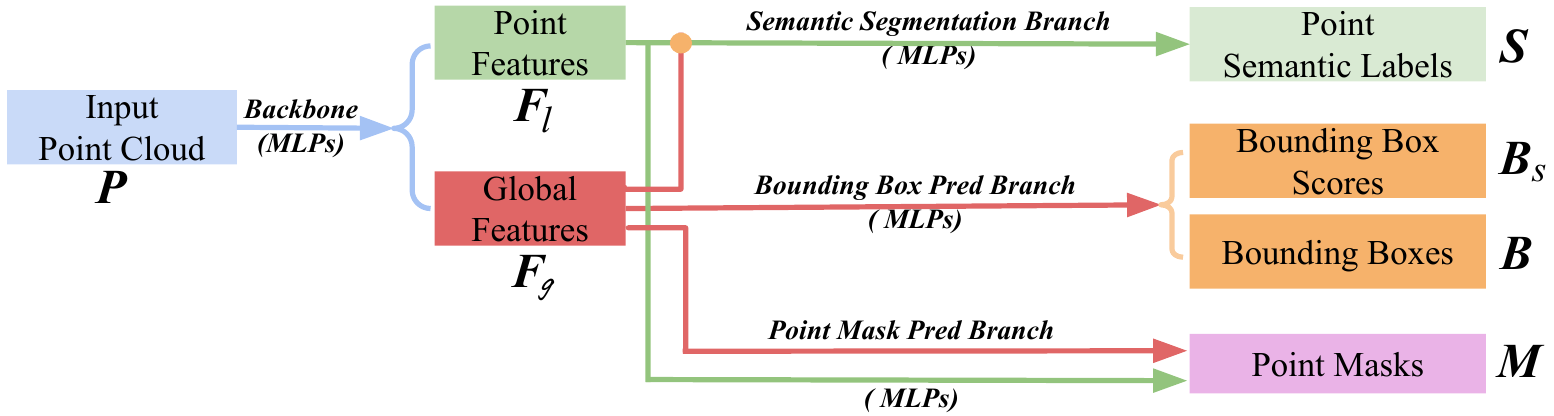}
\caption{The end-to-end implementation for semantic segmentation, bounding box prediction and point mask prediction of 3D point clouds.}
\label{fig:bonet_implementation}
\end{figure*}

While our framework is not restricted to a particular point cloud network, we adopt PointNet++ \cite{Qi2017} as the backbone to learn the local and global features. In parallel, another separate branch is implemented to learn per-point semantics with the standard $softmax$ cross-entropy loss function $\ell_{sem}$. Figure \ref{fig:bonet_implementation} shows the overall architecture for semantic segmentation, bounding box prediction and point mask prediction of 3D point clouds.  The architecture of the backbone and semantic branch is the same as used in \cite{Wang2018d}. Note that our 3D-BoNet focuses on precisely segmenting individual objects and we do not aim to improve the recognition accuracy of 3D points and objects in this chapter. Therefore we leverage the existing semantic segmentation networks as a sub-branch in our end-to-end framework to estimate the categories of segmented objects.
Given an input point cloud $\boldsymbol{P}$, the above three branches are linked and end-to-end trained using a single combined multi-task loss:
\begin{equation}
\ell_{all} = \ell_{sem} + \ell_{bbox} +\ell_{bbs} +\ell_{pmask}
\end{equation}
We use Adam solver~\cite{Kingma2015a} with its default hyper-parameters for optimization. Initial learning rate is set to $5e^{-4}$ and then divided by 2 every $20$ epochs. The whole network is trained on a Titan X GPU from scratch. We use the same settings for all experiments, which guarantees the reproducibility of our framework.
\section{Experiments}

\begin{table*}[t]
\setlength{\belowdisplayskip}{4pt}
\caption{ \small{Instance segmentation results on ScanNet(v2) benchmark (hidden test set). The metric is AP(\%) with IoU threshold $0.5$. Accessed on 2 June 2019.}}
\vspace*{0.1cm}
\tiny{
\centering
\label{tab:semins_scannet}
\tabcolsep=0.02cm
\begin{tabular}{ l|ccccccccccccccccccc}
\hline
&mean&btub&bed&bkshelf&cab&chair&counter&curt&desk&door&other&pic&refrig&showerCur&sink&sofa&table&toilet&window \\
\hline
MaskRCNN \cite{He2017a}&5.8&33.3&0.2&0.0&5.3&0.2&0.2&2.1&0.0&4.5&2.4&23.8&6.5&0.0&1.4&10.7&2.0&11.0&0.6 \\
SGPN \cite{Wang2018d}&14.3&20.8&39.0&16.9&6.5&27.5&2.9&6.9&0.0&8.7&4.3&1.4&2.7&0.0&11.2&35.1&16.8&43.8&13.8 \\
3D-BEVIS \cite{Elich2019}&24.8&66.7&56.6&7.6&3.5&39.4&2.7&3.5&9.8&9.9&3.0&2.5&9.8&37.5&12.6&60.4&18.1&85.4&17.1 \\
R-PointNet \cite{Yi2019}&30.6&50.0&40.5&31.1&34.8&58.9&5.4&6.8&12.6&28.3&29.0&2.8&21.9&21.4&33.1&39.6&27.5&82.1&24.5 \\
UNet-Backbone \cite{Liang2019}&31.9&66.7&71.5&23.3&18.9&47.9&0.8&21.8&6.7&20.1&17.3&10.7&12.3&43.8&15.0&61.5&35.5&91.6&9.3 \\
3D-SIS (5 views) \cite{Hou2019}&38.2&\textbf{100.0}&43.2&24.5&19.0&57.7&1.3&26.3&3.3&32.0&24.0&7.5&42.2&85.7&11.7&\textbf{69.9}&27.1&88.3&23.5 \\
MASC \cite{Liu2019}&44.7&52.8&55.5&38.1&\textbf{38.2}&63.3&0.2&50.9&26.0&36.1&43.2&32.7&\textbf{45.1}&57.1&36.7&63.9&38.6&\textbf{98.0}&27.6 \\
ResNet-Backbone \cite{Liang2019} &45.9&\textbf{100.0}&\textbf{73.7}&15.9&25.9&58.7&\textbf{13.8}&47.5&21.7&\textbf{41.6}&40.8&12.8&31.5&71.4&41.1&53.6&\textbf{59.0}&87.3&30.4 \\
PanopticFusion \cite{Narita2019} &47.8&66.7&71.2&\textbf{59.5}&25.9&55.0&0.0&61.3&17.5&25.0&\textbf{43.4}&\textbf{43.7}&41.1&85.7&\textbf{48.5}&59.1&26.7&94.4&35.9 \\
MTML &48.1&\textbf{100.0}&66.6&37.7&27.2&\textbf{70.9}&0.1&57.9&25.4&36.1&31.8&9.5&43.2&\textbf{100.0}&18.4&60.1&48.7&93.8&38.4 \\
\textbf{\nicknameBoNet{}(Ours)} &\textbf{48.8}&\textbf{100.0}&67.2&59.0&30.1&48.4&9.8&\textbf{62.0}&\textbf{30.6}&34.1&25.9&12.5&43.4&79.6&40.2&49.9&51.3&90.9&\textbf{43.9} \\
\hline
\end{tabular}
}
\end{table*}

\subsection{Evaluation on ScanNet}
We first evaluate our approach on ScanNet(v2) 3D semantic instance segmentation benchmark \cite{Dai2017}. ScanNet(v2) consists of 1613 complete 3D scenes acquired from real-world indoor spaces. The official split has 1201 training scenes, 312 validation scenes and 100 hidden testing scenes. The original large point clouds are divided into $1m\times 1m$ blocks with $0.5m$ overlapped between neighbouring blocks. This data preprocessing step is the same as the one used by PointNet \cite{Qi2016} and SGPN \cite{Wang2018d} for the S3DIS dataset. We sample $4096$ points from each block for training, but use all points of a block for testing followed by the BlockMerging algorithm \cite{Wang2018d} to assemble blocks into complete 3D scenes. Each point is represented by a 9D vector \textit{(normalized xyz in the block, RGB, normalized xyz in the room)}. $H$ is set as $20$ in our experiments. In our experiment, we observe that the performance of the vanilla PointNet++ based semantic prediction sub-branch is limited and unable to provide satisfactory semantics. Thanks to the flexibility of our framework, we therefore easily train a parallel SCN network \cite{Graham2018} to estimate more accurate per-point semantic labels for the predicted instances of our \nicknameBoNet{}. The average precision (AP) with an IoU threshold 0.5 is used as the evaluation metric.

We compare with the leading approaches on 18 object categories in Table \ref{tab:semins_scannet}. Particularly, SGPN \cite{Wang2018d}, 3D-BEVIS \cite{Elich2019}, MASC \cite{Liu2019} and \cite{Liang2019} are point feature clustering based approaches; R-PointNet \cite{Yi2019} learns to generate dense object proposals followed by point-level segmentation; 3D-SIS \cite{Hou2019} is a proposal-based approach using both point clouds and color images as input. PanopticFusion \cite{Narita2019} learns to segment instances on multiple 2D images by Mask-RCNN \cite{He2017a} and then uses the SLAM system to reproject back to 3D space. Our approach surpasses them all using point clouds only. Remarkably, our framework offers satisfactory performance on all categories without preferring specific classes. 

\begin{figure*}[hp]
\centering
   \includegraphics[width=1\linewidth]{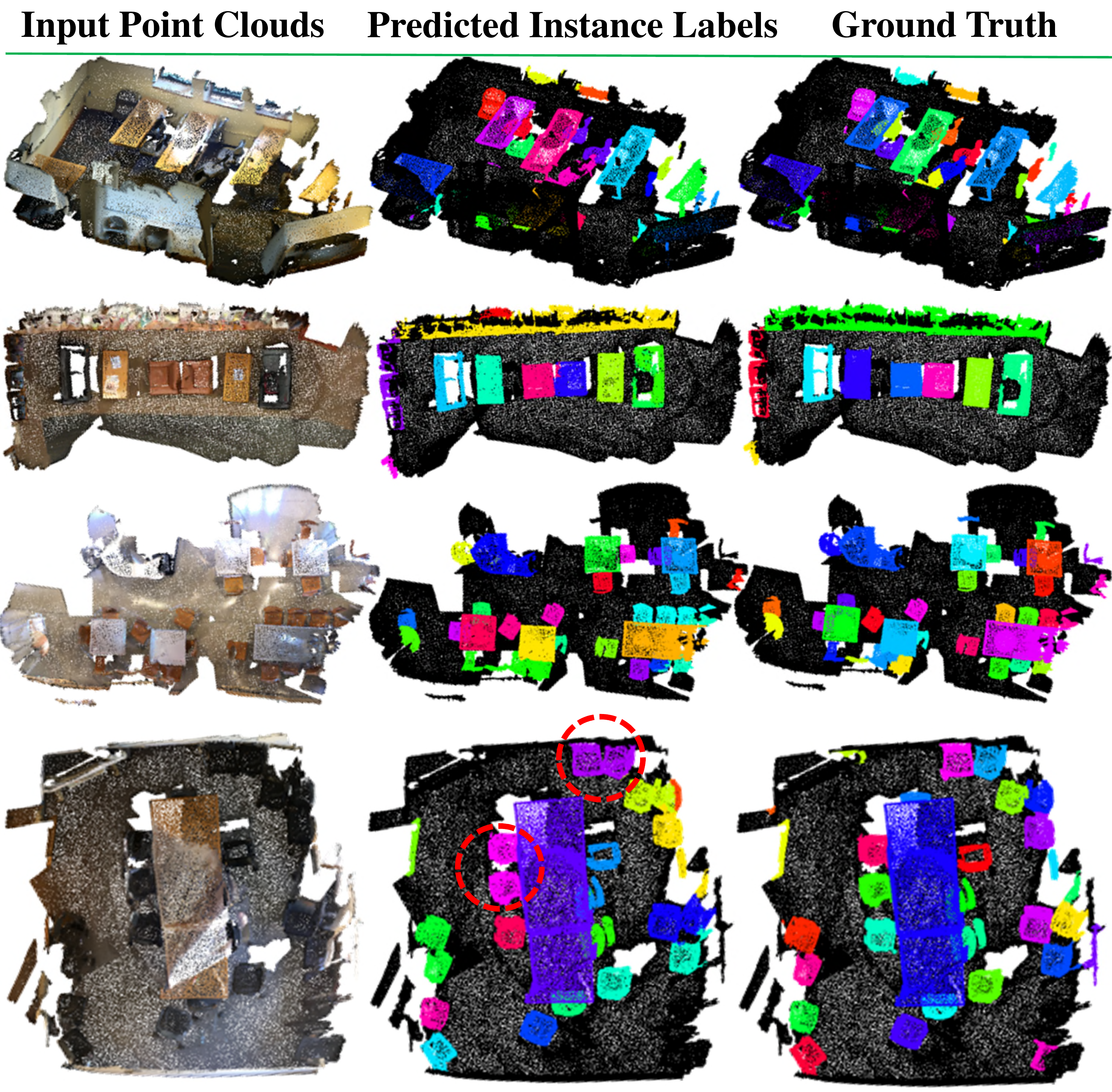}
\caption{Qualitative results of our approach for instance segmentation on ScanNet(v2) validation split. Black points are not of interest as they do not belong to any of the 18 object categories.}
\label{fig:scannet_demo}
\end{figure*}

Figure \ref{fig:scannet_demo} shows qualitative results of our \nicknameBoNet{} for instance segmentation on ScanNet validation split. It can be seen that our approach tends to predict complete object instances, instead of inferring tiny and but invalid fragments. This demonstrates that our framework indeed guarantees high objectness for segmented instances. The red circles showcase the failure cases, where the very similar instances are unable to be well segmented by our approach. 

\subsection{Evaluation on S3DIS}
We further evaluate the semantic instance segmentation of our framework on S3DIS \cite{Armeni2016}, which consists of 3D complete scans from 271 rooms belonging to 6 large areas. Our data preprocessing and experimental settings strictly follow PointNet \cite{Qi2016}, SGPN \cite{Wang2018d}, ASIS \cite{Wang2019}, and JSIS3D \cite{Pham2019}. In particular, the original large point clouds are divided into $1m\times 1m$ blocks with $0.5m$ overlapped between neighbouring blocks. We sample $4096$ points from each block for training, but use all points of a block for testing. Each point is represented by a 9D vector \textit{(normalized xyz in the block, rgb, normalized xyz in the room)}. In our experiments, $H$ is set as $24$ and we follow the 6-fold evaluation \cite{Armeni2016,Wang2019}. We train our \nicknameBoNet{} to predict object bounding boxes and point-level masks, and in parallel train a vanilla PointNet++ based sub-branch to predict point-level semantic labels. Particularly, all the semantic, bounding box and point mask sub-branches share the same PointNet++ backbone to extract point features, and are end-to-end trained from scratch. 

\begin{figure*}[t]
\centering
   \includegraphics[width=1\linewidth]{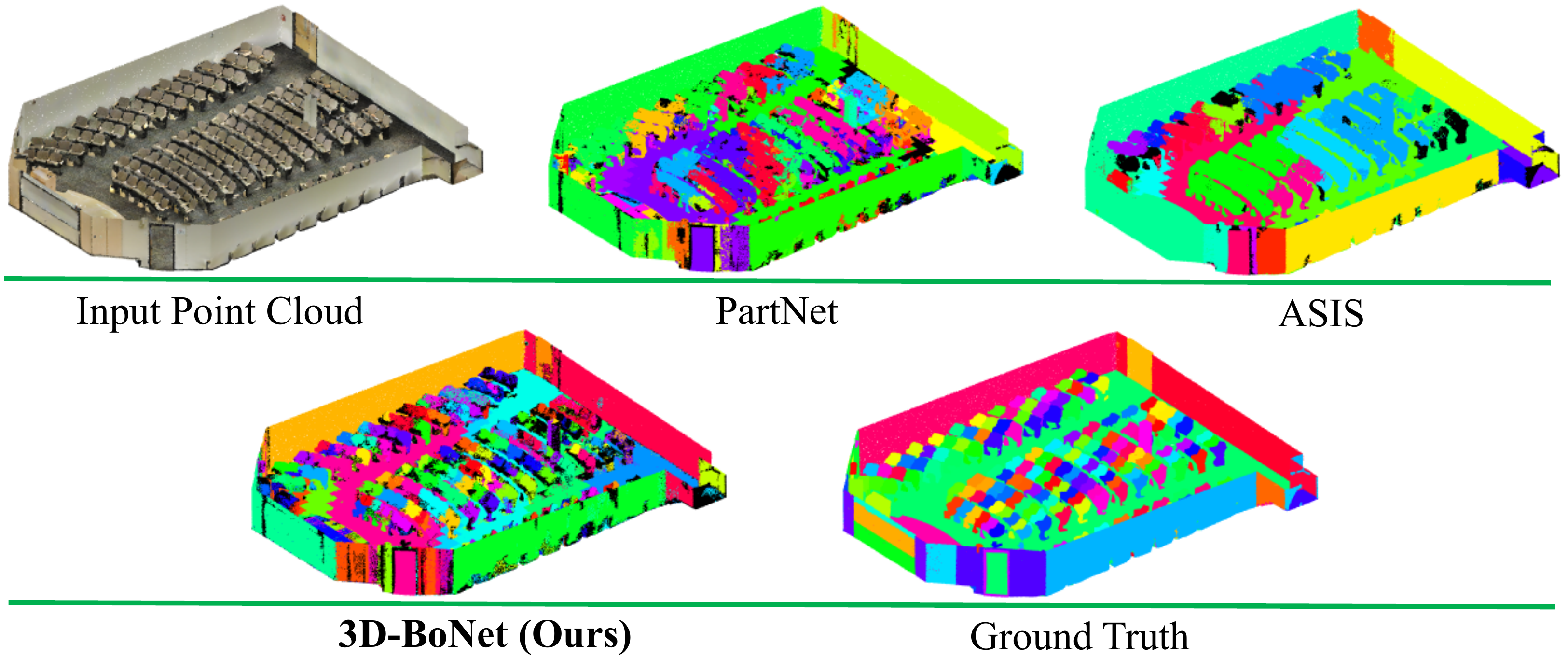}
\caption{This shows a lecture room with hundreds of objects (\eg{} chairs, tables), highlighting the challenge of instance segmentation. Different colors indicates different instances. Our framework predicts more precise instance labels than other techniques.}
\label{fig:bonet_demo_s3dis}
\end{figure*}

We compare the proposed approach with ASIS \cite{Wang2019}, the state of art on S3DIS, and the PartNet baseline \cite{Mo2019}. For fair comparison, we carefully train the PartNet baseline with the same PointNet++ backbone and other settings as used in our framework. For evaluation, the classical metrics mean precision (mPrec) and mean recall (mRec) with IoU threshold 0.5 are reported. Note that, we use the same BlockMerging algorithm \cite{Wang2018d} to merge the instances from different blocks for both our approach and the PartNet baseline. The final scores are averaged across the total 13 categories. Table \ref{tab:prerec_s3dis} presents the mPrec/mRec scores and Figure \ref{fig:bonet_demo_s3dis} shows qualitative results. Our method surpasses PartNet baseline \cite{Mo2019} by large margins, and also outperforms ASIS \cite{Wang2019}, but not significantly, mainly because our semantic prediction branch (vanilla PointNet++ based) is inferior to ASIS which tightly fuses semantic and instance features for mutual optimization. We leave the feature fusion as our future exploration.

\begin{table}
\centering
\caption{Instance segmentation results on S3DIS dataset.}
\tabcolsep= 0.4cm 
\begin{tabular}{l|cc} \hline
 & mPrec & mRec \\ \hline
PartNet \cite{Mo2019} &56.4 & 43.4\\  
ASIS \cite{Wang2019} &63.6 & 47.5\\  
\textbf{\nicknameBoNet{} (Ours)} &\textbf{65.6} & \textbf{47.6}\\  \hline
\end{tabular}
\label{tab:prerec_s3dis}
\end{table}

\begin{figure*}[h]
\centering
   \includegraphics[width=1\linewidth]{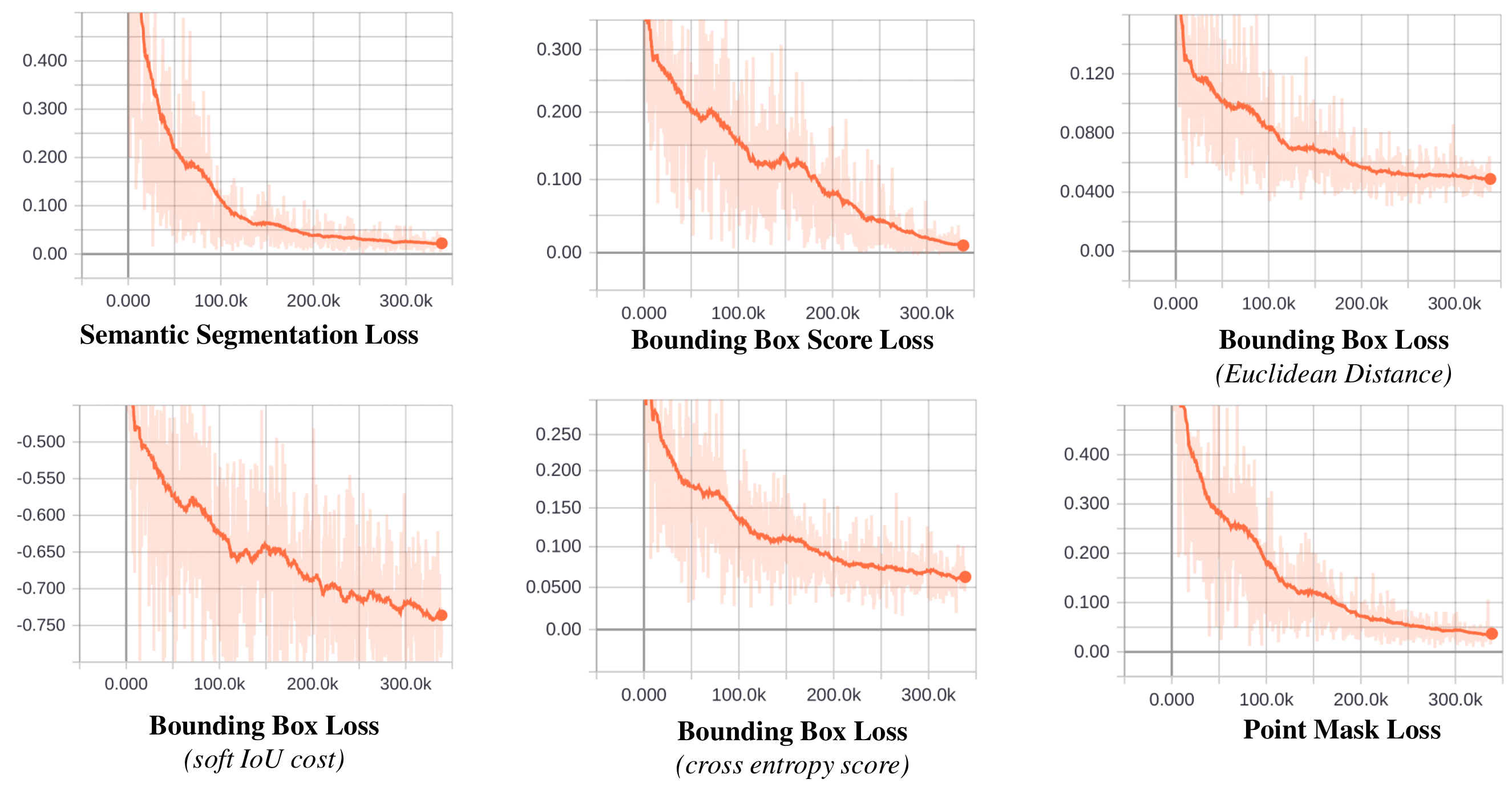}
\caption{Training losses on S3DIS Areas (1,2,3,4,6).}
\label{fig:s3dis_traincurv}
\end{figure*}

\begin{figure*}[h]
\centering
   \includegraphics[width=1\linewidth]{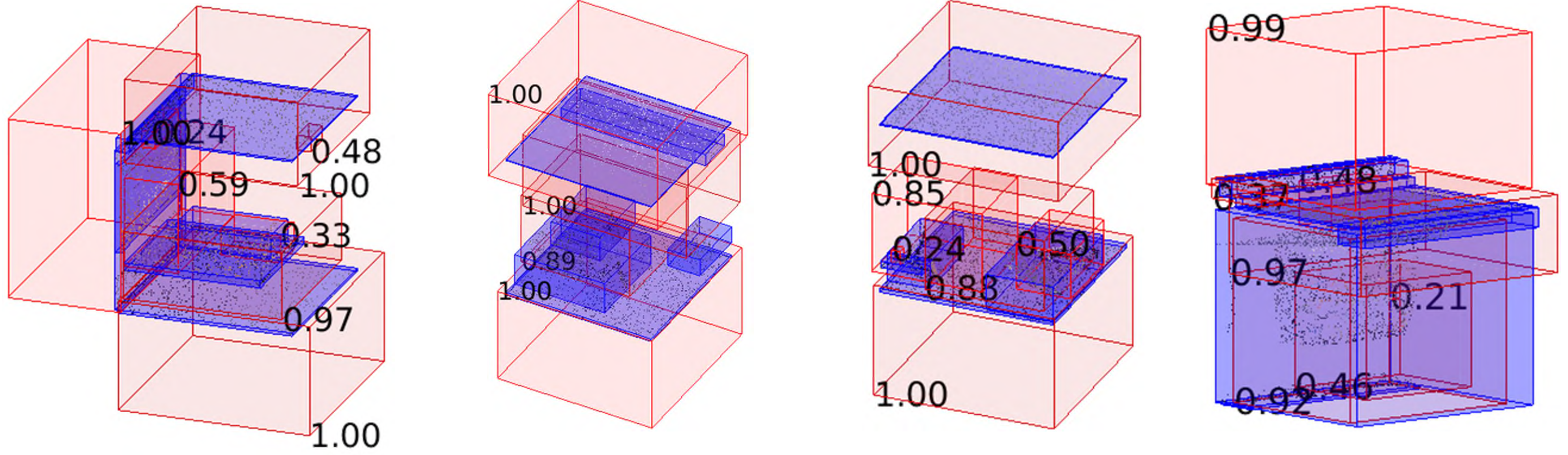}
\caption{Qualitative results of predicted bounding boxes and scores on S3DIS Area 2. The point clouds inside of the blue boxes are fed into our framework which then estimates the red boxes to roughly detect instances. The tight blue boxes are the ground truth. }
\label{fig:s3dis_bbox_demo}
\end{figure*}

\begin{figure*}[h]
\centering
   \includegraphics[width=1\linewidth]{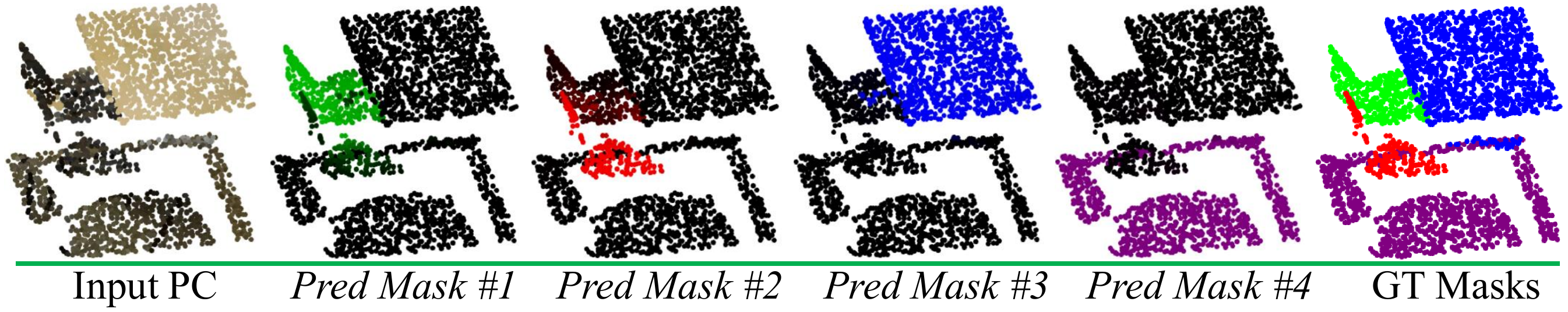}
\caption{Qualitative results of predicted instance masks.}
\label{fig:s3dis_pmask_demo}
\end{figure*}

\begin{figure*}[tp]
\centering
   \includegraphics[width=1\linewidth]{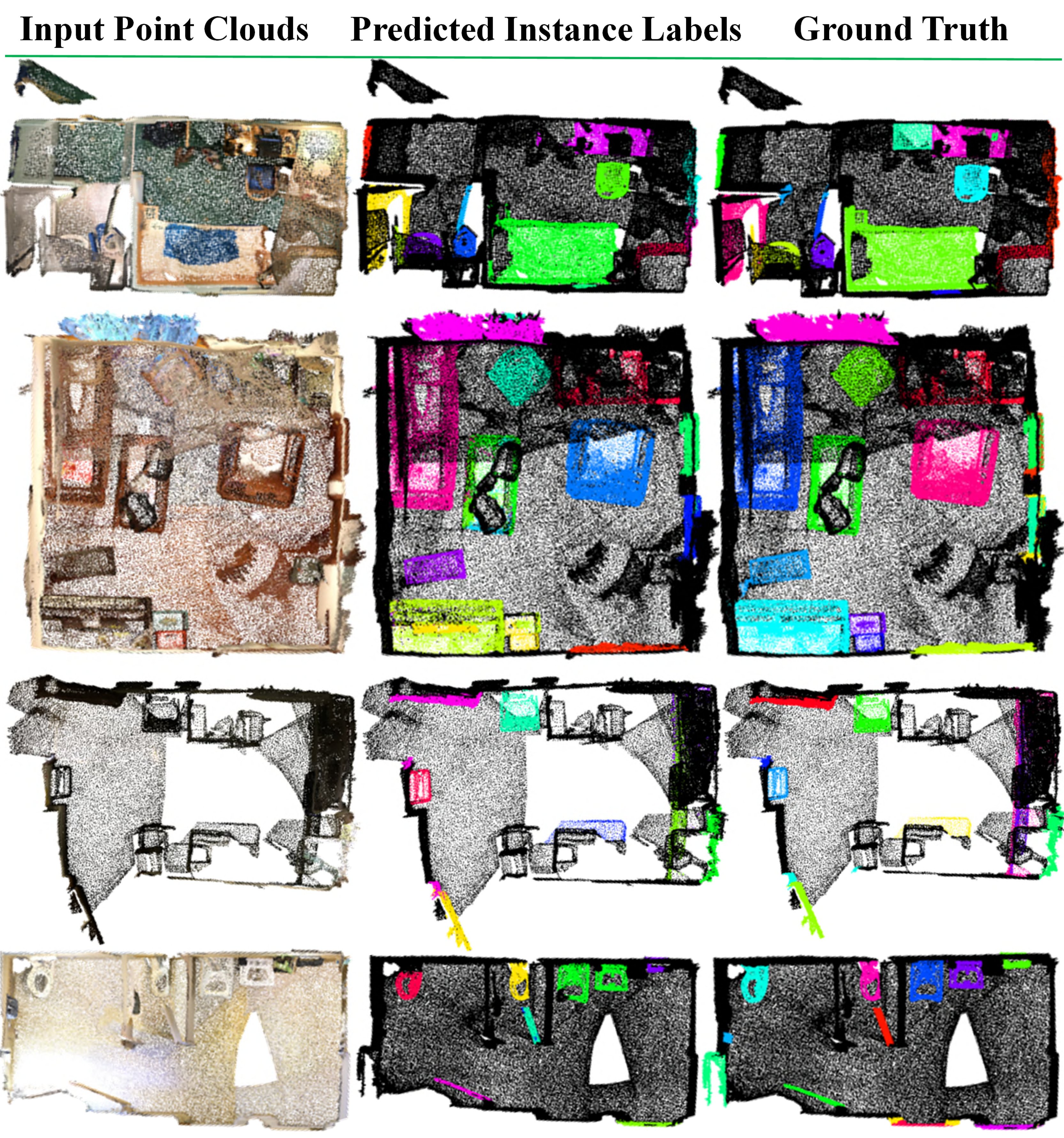}
\caption{Qualitative results of instance segmentation on ScanNet dataset. Although the model is trained on S3DIS dataset and then directly tested on ScanNet validation split, it is still able to predict high-quality instance labels. }
\label{fig:scannet_crossCat}
\end{figure*}

Figure \ref{fig:s3dis_traincurv} shows the training curves of our proposed loss functions on Areas (1,2,3,4,6) of S3DIS dataset. It demonstrates that all the proposed loss functions are able to converge consistently, thus jointly optimizing the semantic segmentation, bounding box prediction, and point mask prediction branches in an end-to-end fashion. 

Figure \ref{fig:s3dis_bbox_demo} presents the qualitative results of predicted bounding boxes and scores. It can be seen that the predicted boxes are not necessarily tight and precise. Instead, they tend to be inclusive but with high objectness. Fundamentally, this highlights the simple but effective concept of our bounding box prediction network. Given these bounded points, it is extremely easy to segment the instance inside. 

Figure \ref{fig:s3dis_pmask_demo} visualizes the predicted instance masks, where the black points have $\sim$ 0 probability and the brighter points have $\sim$ 1 probability to be an instance within each predicted mask. In particular, the predicted masks \#1 $\sim$ \#4 show the results of point mask branch from 4 predicted bounding boxes. It can be seen that both the table (blue) and the floor (purple) are clearly segmented, while there is ambiguity between the predicted two chairs (green and red masks).

\subsection{Generalization to Unseen Scenes and Categories}
Our framework learns the object bounding boxes and point masks from raw point clouds without direct coupling to the semantic information, which inherently allows for generalization to new categories and scenes. We conduct further experiments to qualitatively demonstrate the generality of our framework. In particular, we use the well-trained model from S3DIS dataset (Areas 1/2/3/4/6) to directly test on the validation split of ScanNet(v2) dataset. Since the ScanNet dataset consists of many more object categories than S3DIS dataset, there are a number of categories (\eg{} toilet, desk, sink, bathtub) that the trained model has never seen before. 

As shown in Figure \ref{fig:scannet_crossCat}, our model is still able to predict high-quality instance labels even though the scenes and some object categories have not been seen before. This shows that our model does not simply fit the training dataset. Instead, it tends to learn the underlying geometric features which are able to be generalized across new objects and scenes.

\setlength{\columnsep}{10pt}
\begin{table}
\caption{Instance segmentation results of all ablation experiments on Area 5 of S3DIS.}
\centering
\label{tab:ins_s3dis_ablation}
\tabcolsep= 0.8cm 
\begin{tabular}{ l|c c}
\hline
& mPrec & mRec \\
\hline
 (1) Remove Box Score Sub-branch &50.9&40.9\\
 (2) Euclidean Distance Only &53.8&\textbf{41.1} \\
 (3) Soft IoU Cost Only &55.2&40.6 \\
 (4) Cross-Entropy Score Only &51.8&37.8 \\
 (5) Do Not Supervise Box Prediction &37.3&28.5 \\
 (6) Remove Focal Loss &50.8&39.2 \\
 \textbf{(7) The Full Framework} & \textbf{57.5} &40.2 \\
\hline
\end{tabular}
\end{table}

\subsection{Ablation Study}
To evaluate the effectiveness of each component of our framework, we conduct $6$ groups of ablation experiments on the largest Area 5 of S3DIS dataset.

(1) \textbf{Remove Box Score Prediction Sub-branch.} Basically, the box score serves as an indicator and regularizer for valid bounding box prediction. After removing it, we train the network with:
\begin{equation*}
\setlength{\belowdisplayskip}{-4pt}
\ell_{ab1} = \ell_{sem}+ \ell_{bbox} +\ell_{pmask} 
\end{equation*}

Initially, the multi-criteria loss function is a simple unweighted combination of the Euclidean distance, the soft IoU cost, and the cross-entropy score. However, this may not be optimal, because the density of input point clouds is usually inconsistent and tends to prefer different criteria. We conduct $3$ groups of experiments on ablated bounding box loss functions.

(2) \textbf{Use Single Criterion: Euclidean Distance.} Only the Euclidean Distance criterion is used for the box association and loss $\ell_{bbox}$.
\begin{equation*}
\ell_{ab2} = \ell_{sem} + 
\frac{1}{T}\sum_{t=1}^{T} \boldsymbol{C}^{ed}_{t,t}
+\ell_{bbs} +\ell_{pmask}
\end{equation*}

(3) \textbf{Use Single Criterion: Soft IoU.} Only the Soft IoU criterion is used for the box association and loss $\ell_{bbox}$.
\begin{equation*}
\ell_{ab3} = \ell_{sem} + 
\frac{1}{T}\sum_{t=1}^{T} \boldsymbol{C}^{sIoU}_{t,t}
+\ell_{bbs} +\ell_{pmask}
\end{equation*}

(4) \textbf{Use Single Criterion: Cross-Entropy Score.} Only the Cross-Entropy Score criterion is used for the box association and loss $\ell_{bbox}$.
\begin{equation*}
\ell_{ab4} = \ell_{sem} + 
\frac{1}{T}\sum_{t=1}^{T} \boldsymbol{C}^{ces}_{t,t}
+\ell_{bbs} +\ell_{pmask}
\end{equation*}

(5) \textbf{Do Not Supervise Box Prediction.} The predicted boxes are still associated according to the three criteria, but we remove the box supervision signal. The framework is trained with:
\begin{equation*}
\ell_{ab5} = \ell_{sem} +\ell_{bbs} +\ell_{pmask}
\end{equation*}

(6) \textbf{Remove Focal Loss for Point Mask Prediction.} In the point mask prediction branch, the focal loss is replaced by the standard cross-entropy loss for comparison. 

\textbf{Analysis.} Table \ref{tab:ins_s3dis_ablation} shows the scores for ablation experiments. 
\begin{itemize}
    \item The box score sub-branch indeed benefits the overall instance segmentation performance, as it tends to penalize duplicated box predictions. 
    \item Compared with Euclidean distance and cross-entropy score, the sIoU cost tends to be better for box association and supervision, thanks to our differentiable Algorithm \ref{alg:ppp}. As the three individual criteria prefer different types of point structures, a simple combination of three criteria may not always be optimal on a specific dataset.
    \item Without the supervision for box prediction, the performance drops significantly, primarily because the network is unable to infer satisfactory instance 3D boundaries and the quality of predicted point masks deteriorates accordingly. 
    \item Compared with focal loss, the standard cross entropy loss is less effective for point mask prediction due to the imbalance of instance and background point numbers. 
\end{itemize}

\subsection{Computation Analysis}
The computation complexity of all existing approaches are analysed as follows:

(1) For point feature clustering based approaches including SGPN \cite{Wang2018d}, ASIS \cite{Wang2019}, JSIS3D \cite{Pham2019}, 3D-BEVIS \cite{Elich2019}, MASC \cite{Liu2019}, and \cite{Liang2019}, the computation complexity of the post clustering algorithm such as Mean Shift \cite{Comaniciu2002} tends towards $\mathcal{O}(TN^2)$, where $T$ is the number of instances and $N$ is the number of input points. 

(2) For dense proposal-based methods including GSPN \cite{Yi2019}, 3D-SIS \cite{Hou2019} and PanopticFusion \cite{Narita2019}, region proposal network and non-maximum suppression are usually required to generate and prune the dense proposals, which is computationally expensive \cite{Narita2019}. 

(3) Both PartNet baseline \cite{Mo2019} and our \nicknameBoNet{} have similar efficient computation complexity $\mathcal{O}(N)$. Empirically, our \nicknameBoNet{} takes around $20$ ms GPU time to process $4k$ points, while most approaches in (1)(2) need more than 200ms GPU/CPU time to process the same number of points.

Table \ref{tab:time_con} compares the time consumption of four existing approaches using their released codes on the validation split (312 scenes) of ScanNet(v2) dataset. Normally, each scene has dozens of objects with around 80k points spanning up to $10 \times 20 \times 5$ meters in 3D space. SGPN \cite{Wang2018d}, ASIS \cite{Wang2019}, GSPN \cite{Yi2019} and our 3D-BoNet are implemented by Tensorflow 1.4, 3D-SIS  \cite{Hou2019} by Pytorch 0.4. All approaches are running on a single Titan X GPU and the pre/post-processing steps on an i7 CPU core with a single thread. Note that 3D-SIS automatically uses CPU for computing when some large scenes are unable to be processed by the single GPU. 

It can be seen that our approach is much more computationally efficient than existing methods, and is up to 20$\times$ faster than ASIS \cite{Wang2019}. However, the majority of time spent by our 3D-BoNet is within the \textit{block merging} step, which is used to assemble the partitioned blocks of point clouds back to large-scale scenes. Fundamentally, this is because the backbone network used in our framework, \ie{} PointNet++, is unable to take large-scale point clouds as input due to its expensive sampling operations. In this regard, the efficiency of our framework can be further improved if a more advanced backbone network is available to process large-scale point clouds. We leave it as our future exploration.

\begin{table}[h]
\centering
\tiny{
\caption{\small{Time consumption of different approaches on the validation split (312 scenes) of  ScanNet(v2) (seconds).}}
\vspace{-0.16cm}
\centering
\label{tab:time_con}
\tabcolsep=0.0cm
\begin{tabular}{ l|c|c|c|c|c}
\hline
& SGPN \cite{Wang2018d} & ASIS \cite{Wang2019} & GSPN \cite{Yi2019} & 3D-SIS \cite{Hou2019} & \textbf{3D-BoNet(Ours)}  \\
\hline
&\begin{tabular}{@{}c@{}c@{}} network(GPU): 650 \\group merge(CPU): 46562 \\block merge(CPU): 2221 \end{tabular} 
&\begin{tabular}{@{}c@{}c@{}} network(GPU): 650 \\mean shift(CPU): 53886 \\block merge(CPU): 2221 \end{tabular}
&\begin{tabular}{@{}c@{}c@{}} network(GPU): 500 \\point sampling(GPU): 2995 \\neighbour search(CPU): 468 \end{tabular}
&\begin{tabular}{@{}c@{}c@{}} voxelization, projection,\\ network.. (GPU+CPU): \\ 38841 \end{tabular}
&\begin{tabular}{@{}c@{}c@{}} network(GPU): 650 \\ \textit{SCN (GPU parallel): 208} \\block merge(CPU): 2221 \end{tabular}\\
\hline
total &49433 &56757 &3963 &38841 &\textbf{2871} \\
\hline
\end{tabular}
}
\end{table}
\section{Conclusion}
In this chapter, we proposed a simple, effective and efficient framework for instance segmentation on 3D point clouds. The framework consists of a bounding box branch and a point mask prediction branch. The bounding box branch directly regresses a set of 3D bounding boxes to roughly detect all object instances, whereas the point mask branch focuses on each predicted bounding box to classify whether each 3D point belongs to the foreground instance or the background clutter. By training the branches in an end-to-end fashion, we obtain the instance segmentation results in a single forward pass. Compared with all existing works, our framework only consists of feed-forward MLPs without requiring any heavy post-processing steps such as non-maximum-suppression or feature clustering, therefore being lightweight and efficient. 

However, it also has some limitations which open up new directions to future work. (1) Instead of using unweighted combination of three criteria, it could be better to design a module to automatically learn the weights, in order to be able to adapt to different types of input point clouds. (2) Instead of training a separate branch for semantic prediction, more advanced feature fusion modules can be introduced to mutually improve both semantic and instance segmentation. (3) Our framework follows the MLP design and is therefore agnostic to the number and order of input points. It is desirable to directly train and test on large-scale input point clouds instead of the divided small blocks, by drawing on recent work \cite{Engelmann2017}\cite{Landrieu2018}.

\chapter{Conclusion and Future Work}
\label{chap:conclusion}
\section{Summary of Key Contributions}
The overarching purpose of this thesis has been to build intelligent systems that 
understand the geometric structure and semantics of the 3D real-world environments. To achieve this goal, we proposed solutions from the object level to the more complex scene level. 

In Chapter \ref{chap:rec_obj_sv}, we introduced an encoder-decoder architecture together with adversarial training to accurately estimate dense 3D shape of objects from a single depth view. To enable the encoder-decoder to learn more fine-grained geometric details, we integrated an adversarial component conditioned on the input single view. However, the adversarial component is hard to train due to the extremely high-dimensional 3D data. To stabilise the end-to-end training procedure, we proposed a simple mean feature neural layer to discriminate the real-world 3D shapes and the predicted ones, allowing the entire network to recover more plausible 3D structures. Since the existing datasets are based on 3D CAD models and synthesized images, we therefore collected a reasonable amount of real-world datasets using the Kinect device, demonstrating strong generalization of our approach from simulated data to real-world noisy environments. Compared with existing methods, our approach is distinguished in the following ways: 1) it estimates dense 3D shapes within high-resolution voxel grids (\ie{} $256^3$) which represent compelling geometric details. 2) Thanks to the adversarial training, the estimated 3D shapes tend to be more realistic because the priors of object shapes can be well learnt and propagated from our stable discriminator. 

In Chapter \ref{chap:rec_obj_mv}, we aimed to reconstruct a better 3D shape of an object from multiple views instead of merely a single view. Multiple views naturally provide much more valuable information to infer the 3D structure. However, to effectively integrate this valuable information is not easy. In this chapter, we introduced an attentive aggregation module to selectively fuse the deep visual features for precise 3D shape estimation from multiple views. By formulating the multi-view 3D reconstruction problem as an aggregation step for a set of deep features, our attention based method learns an attention score for each visual feature of all multiple input views. These scores are then used to weigh the corresponding features. After that, all the weighted features are summed and integrated across multiple views, generating the final 3D shapes. Fundamentally, the proposed module serves as a mask to filter out the less informative deep features, whereas preserving the relative important information according to the input multiple views.
However, we observed that a naive end-to-end training strategy would result in the whole network not being robust to an arbitrary number of input images. To overcome this problem, we theoretically investigated the principle underlying this issue and presented a two-stage training algorithm. Our algorithm separately optimizes the base encoder-decoder and the proposed attention module, achieving superior performance at reconstructing 3D shapes. Compared with existing RNN based methods, our approach is permutation invariant and estimates consist of 3D shapes regardless of the different orderings of input views. In contrast to the max/mean/sum poolings which usually ignore the majority of information from multiple images, our module learns to attentively preserve useful features and infers better 3D shapes. Unlike the existing attention based methods which only operate on a fixed number of input images, our module uses a novel training strategy, forcing the robustness of predicted 3D shapes given an arbitrary number of input images.

In Chapter \ref{chap:seg_obj_pc}, we extended object-level perception to scene-level understanding. In real-world scenarios, the 3D scenes are usually cluttered and include multiple objects. In this chapter, we proposed an efficient neural pipeline to identify and segment all individual 3D objects from real-world and large-scale point clouds. In contrast to segmenting 2D images, to identify objects in 3D space is extremely challenging, due to the irregular and incomplete statistics of 3D point clouds. We solved this difficult task by directly regressing a set of bounding boxes to roughly detect individual 3D objects and then segmenting each object within its box. By leveraging pioneering work on neural networks for 3D point clouds processing, we designed a bounding box prediction branch and a point mask prediction branch based on the the learnt per-point features and global features. The bounding box prediction branch directly regresses a set of 3D boxes to detect all objects, whereas the point mask prediction branch serves to classify whether each 3D point belongs to the foreground object instance or the background clutter. To supervise the bounding box prediction branch, we drew on the seminal work on data association and carefully designed loss functions to train the entire network in an end-to-end fashion. The proposed neural pipeline demonstrates accurate and efficient segmentation results on multiple large-scale real-world datasets. Compared with the existing proposal-free approaches, our framework does not require computationally expensive algorithms to cluster point features, and the explicitly predicted bounding boxes guarantee high objectness for the final segmented instances. In contrast to the existing proposal-based methods, our pipeline does not rely on spatial anchors to generate numerous candidate bounding boxes, and thus does not require computationally heavy post-processing steps such as non-maximum-suppression.

\section{Limitations and Future Work}

The work in this thesis has opened up a number of new directions for future work: 

\textbf{Using 2D Supervision.} The proposed architecture in this thesis mainly relies on ground truth 3D data for supervision. However, it is labour-intensive to acquire large-scale 3D labels. In addition, the learnt representations tend to be fitted to specific 3D datasets and are unlikely to generalize to completely unseen scenarios due to domain gaps. An alternative approach is to leverage widely available 2D images as supervision signal. Unlike 3D supervision, 2D images inherently have weaker prior knowledge. Nevertheless, the explicit geometric consistency would provide valuable information to supervise the networks.

\textbf{Better 3D Representation.} The widely used voxel grid requires extremely heavy computation and memory if being used to represent large-scale 3D scenes. Point clouds are efficient to represent complex 3D structures, but they are unable to present the object surface which is useful for high-level applications such as rendering and grasping. By contrast, 3D meshes have been recently integrated into deep neural networks and emerge as an efficient and promising approach to shape representation.

\textbf{Disentanglement of 3D Scenes.} The proposed approaches in this thesis usually learn a singular representation for an object or a scene, without decomposing the shape components explicitly. However, learning to factorize the objects or scenes into more detailed elements would be desirable, as it allows more robustness and generalization across different scenarios. 

In conclusion, this thesis has taken a step towards 3D scene understanding by learning to reconstruct and segment 3D objects. We hope it can inspire researchers to develop more advanced systems that would endow machines with the ability to perceive and interact with our environment, thus benefiting us humans in a variety of real-world applications.

\appendix
\include{appendix1}
\include{appendix2}

\addcontentsline{toc}{chapter}{Bibliography}
\bibliography{references}  

\begin{thebibliography}{100}

\bibitem{Agarwal2009}
Sameer Agarwal, Noah Snavely, Ian Simon, Steven~M. Seitz, and Richard Szeliski.
\newblock {Building Rome in a Day}.
\newblock {\em IEEE International Conference on Computer Vision}, pages
  105--112, 2009.

\bibitem{Ali2018}
Waleed Ali, Sherif Abdelkarim, Mohamed Zahran, Mahmoud Zidan, and Ahmad~El
  Sallab.
\newblock {YOLO3D : End-to-end real-time 3D Oriented Object Bounding Box
  Detection from LiDAR Point Cloud}.
\newblock {\em European Conference on Computer Vision Workshops}, 2018.

\bibitem{Arase2019}
Kosuke Arase, Yusuke Mukuta, and Tatsuya Harada.
\newblock {Rethinking Task and Metrics of Instance Segmentation on 3D Point
  Clouds}.
\newblock {\em IEEE International Conference on Computer Vision Workshops},
  2019.

\bibitem{Arjovsky2017c}
Martin Arjovsky and Leon Bottou.
\newblock {Towards Principled Methods for Training Generative Adversarial
  Networks}.
\newblock {\em International Conference on Learning Representations}, 2017.

\bibitem{Arjovsky2017a}
Martin Arjovsky, Soumith Chintala, and Léon Bottou.
\newblock {Wasserstein Generative Adversarial Networks}.
\newblock {\em International Conference on Machine Learning}, pages 214--223,
  2017.

\bibitem{Armeni2016}
Iro Armeni, Ozan Sener, AR~Zamir, and Helen Jiang.
\newblock {3D Semantic Parsing of Large-Scale Indoor Spaces}.
\newblock {\em IEEE Conference on Computer Vision and Pattern Recognition},
  pages 1534--1543, 2016.

\bibitem{Asvadi2017}
Alireza Asvadi, Luis Garrote, Cristiano Premebida, Paulo Peixoto, and Urbano~J.
  Nunes.
\newblock {DepthCN: Vehicle detection using 3D-LIDAR and ConvNet}.
\newblock {\em IEEE International Conference on Intelligent Transportation
  Systems}, pages 1--6, 2017.

\bibitem{Bahdanau2015}
Dzmitry Bahdanau, Kyunghyun Cho, and Yoshua Bengio.
\newblock {Neural Machine Translation by Jointly Learning to Align and
  Translate}.
\newblock {\em International Conference on Learning Representations}, 2015.

\bibitem{Bao2017b}
Jianmin Bao, Dong Chen, Fang Wen, Houqiang Li, and Gang Hua.
\newblock {CVAE-GAN: Fine-Grained Image Generation through Asymmetric
  Training}.
\newblock {\em IEEE International Conference on Computer Vision}, pages
  2745--2754, 2017.

\bibitem{Behley2019}
Jens Behley, Martin Garbade, Andres Milioto, Jan Quenzel, Sven Behnke, Cyrill
  Stachniss, and Juergen Gall.
\newblock {SemanticKITTI: A Dataset for Semantic Scene Understanding of LiDAR
  Sequences}.
\newblock {\em IEEE International Conference on Computer Vision}, pages
  9297--9307, 2019.

\bibitem{Beltran2018}
Jorge Beltran, Carlos Guindel, Francisco~Miguel Moreno, Daniel Cruzado,
  Fernando Garcia, and Arturo de~la Escalera.
\newblock {BirdNet: a 3D Object Detection Framework from LiDAR Information}.
\newblock {\em IEEE International Conference on Intelligent Transportation
  Systems}, pages 3517--3523, 2018.

\bibitem{Bengio1994}
Y.~Bengio, P.~Simard, and P.~Frasconi.
\newblock {Learning Long-term Dependencies with Gradient Descent is Difficult}.
\newblock {\em IEEE Transactions on Neural Networks}, 5(2):157--166, 1994.

\bibitem{Bengio2013}
Yoshua Bengio, Nicholas L{\'{e}}onard, and Aaron Courville.
\newblock {Estimating or Propagating Gradients Through Stochastic Neurons for
  Conditional Computation}.
\newblock {\em arXiv preprint arXiv:1308.3432}, 2013.

\bibitem{Blanz2003}
V.~Blanz and {T.Vetter}.
\newblock {Face Recognition based on Fitting a 3D Morphable Model}.
\newblock {\em IEEE Transactions on Pattern Analysis and Machine Intelligence},
  25(9):1063--1074, 2003.

\bibitem{Bloesch2018}
Michael Bloesch, Jan Czarnowski, Ronald Clark, Stefan Leutenegger, and
  Andrew~J. Davison.
\newblock {CodeSLAM - Learning a Compact, Optimisable Representation for
  DenseVisual SLAM}.
\newblock {\em IEEE Conference on Computer Vision and Pattern Recognition},
  pages 2560--2568, 2018.

\bibitem{Boulch2020}
Alexandre Boulch.
\newblock {ConvPoint: Continuous Convolutions for Point Cloud Processing}.
\newblock {\em Computers {\&} Graphics}, 2020.

\bibitem{Brock2016}
Andrew Brock, Theodore Lim, J.~M. Ritchie, and Nick Weston.
\newblock {Generative and Discriminative Voxel Modeling with Convolutional
  Neural Networks}.
\newblock {\em Conference on Neural Information Processing Systems Workshops},
  2016.

\bibitem{Cadena2016}
Cesar Cadena, Luca Carlone, Henry Carrillo, Yasir Latif, Davide Scaramuzza,
  Jose Neira, Ian~D. Reid, and John~J. Leonard.
\newblock {Past, Present, and Future of Simultaneous Localization and Mapping:
  Towards the Robust-Perception Age}.
\newblock {\em IEEE Transactions on Robotics}, 32(6):1309--1332, 2016.

\bibitem{Cao2018}
Yan-Pei Cao, Zheng-Ning Liu, Zheng-Fei Kuang, Leif Kobbelt, and Shi-Min Hu.
\newblock {Learning to Reconstruct High-quality 3D Shapes with Cascaded Fully
  Convolutional Networks}.
\newblock {\em European Conference on Computer Vision}, pages 616--633, 2018.

\bibitem{Cha2019}
Geonho Cha, Minsik Lee, and Songhwai Oh.
\newblock {Unsupervised 3D Reconstruction Networks}.
\newblock {\em International Conference on Computer Vision}, pages 3849--3858,
  2019.

\bibitem{Chang2015}
Angel~X. Chang, Thomas Funkhouser, Leonidas Guibas, Pat Hanrahan, Qixing Huang,
  Zimo Li, Silvio Savarese, Manolis Savva, Shuran Song, Hao Su, Jianxiong Xiao,
  Li~Yi, and Fisher Yu.
\newblock {ShapeNet: An Information-Rich 3D Model Repository}.
\newblock {\em arXiv preprint arXiv:1512.03012}, 2015.

\bibitem{Chen2019c}
Chao Chen, Guanbin Li, Ruijia Xu, Tianshui Chen, Meng Wang, and Liang Lin.
\newblock {ClusterNet: Deep Hierarchical Cluster Network with Rigorously
  Rotation-Invariant Representation for Point Cloud Analysis}.
\newblock {\em IEEE Conference on Computer Vision and Pattern Recognition},
  pages 4994--5002, 2019.

\bibitem{Chen2019h}
Qi~Chen, Lin Sun, Zhixin Wang, Kui Jia, and Alan Yuille.
\newblock {Object as Hotspots: An Anchor-Free 3D Object Detection Approach via
  Firing of Hotspots}.
\newblock {\em arXiv:1912.12791}, 2019.

\bibitem{Chen2019e}
Rui Chen, Songfang Han, Jing Xu, and Hao Su.
\newblock {Point-Based Multi-View Stereo Network}.
\newblock {\em IEEE International Conference on Computer Vision}, pages
  1538--1547, 2019.

\bibitem{Chen2019}
Siheng Chen, Sufeng Niu, Tian Lan, and Baoan Liu.
\newblock {PCT: Large-Scale 3D Point Cloud Representations via Graph Inception
  Networks with Applications to Autonomous Driving}.
\newblock {\em IEEE International Conference on Image Processing}, pages
  4395--4399, 2019.

\bibitem{Chen2016a}
Xi~Chen, Yan Duan, Rein Houthooft, John Schulman, Ilya Sutskever, and Pieter
  Abbeel.
\newblock {InfoGAN: Interpretable Representation Learning by Information
  Maximizing Generative Adversarial Nets}.
\newblock {\em Advances in Neural Information Processing Systems}, pages
  2172--2180, 2016.

\bibitem{Chen2017b}
Xiaozhi Chen, Huimin Ma, Ji~Wan, Bo~Li, and Tian Xia.
\newblock {Multi-View 3D Object Detection Network for Autonomous Driving}.
\newblock {\em IEEE Conference on Computer Vision and Pattern Recognition},
  pages 1907--1915, 2017.

\bibitem{Chen2019f}
Yilun Chen, Shu Liu, Xiaoyong Shen, and Jiaya Jia.
\newblock {Fast Point R-CNN}.
\newblock {\em IEEE International Conference on Computer Vision}, pages
  9775--9784, 2019.

\bibitem{Chen2020}
Zhiqin Chen, Andrea Tagliasacchi, and Hao Zhang.
\newblock {BSP-Net: Generating Compact Meshes via Binary Space Partitioning}.
\newblock {\em IEEE Conference on Computer Vision and Pattern Recognition},
  2020.

\bibitem{Chen2019g}
Zhiqin Chen and Hao Zhang.
\newblock {Learning Implicit Fields for Generative Shape Modeling}.
\newblock {\em IEEE Conference on Computer Vision and Pattern Recognition},
  pages 5939--5948, 2019.

\bibitem{Choy2019}
Christopher Choy, JunYoung Gwak, and Silvio Savarese.
\newblock {4D Spatio-Temporal ConvNets: Minkowski Convolutional Neural
  Networks}.
\newblock {\em IEEE Conference on Computer Vision and Pattern Recognition},
  pages 3075--3084, 2019.

\bibitem{Chan2016}
Christopher~B. Choy, Danfei Xu, JunYoung Gwak, Kevin Chen, and Silvio Savarese.
\newblock {3D-R2N2: A Unified Approach for Single and Multi-view 3D Object
  Reconstruction}.
\newblock {\em European Conference on Computer Vision}, pages 628--644, 2016.

\bibitem{Christian2017}
H~Christian, Shubham Tulsiani, and Jitendra Malik.
\newblock {Hierarchical Surface Prediction for 3D Object Reconstruction}.
\newblock {\em International Conference on 3D Vision}, pages 412--420, 2017.

\bibitem{Chua1997}
Chin~Seng Chua and Ray Jarvis.
\newblock {Point signatures: A new representation for 3d object recognition}.
\newblock {\em International Journal of Computer Vision}, 25(1):63--85, 1997.

\bibitem{Comaniciu2002}
D.~Comaniciu and P.~Meer.
\newblock {Mean Shift: A Robust Approach toward Feature Space Analysis}.
\newblock {\em IEEE Transactions on Pattern Analysis and Machine Intelligence},
  24(5):603--619, 2002.

\bibitem{Crandall2011}
David Crandall, Andrew Owens, Noah Snavely, and Dan Huttenlocher.
\newblock {Discrete-continuous Optimization for Large-scale Structure from
  Motion}.
\newblock {\em IEEE Conference on Computer Vision and Pattern Recognition},
  page 3001–3008, 2011.

\bibitem{Curless1996}
Brian Curless and Marc Levoy.
\newblock {A Volumetric Method for Building Complex Models from Range Images}.
\newblock {\em Conference on Computer Graphics and Interactive Techniques},
  pages 303--312, 1996.

\bibitem{Czarnowski2019}
Jan Czarnowski, Tristan Laidlow, Ronald Clark, and Andrew~J. Davison.
\newblock {DeepFactors: Real-Time Probabilistic DenseMonocular SLAM}.
\newblock {\em IEEE Robotics and Automation Letters}, 2019.

\bibitem{Dai2017}
Angela Dai, Angel~X. Chang, Manolis Savva, Maciej Halber, Thomas Funkhouser,
  and Matthias Nie{\ss}ner.
\newblock {ScanNet: Richly-annotated 3D Reconstructions of Indoor Scenes}.
\newblock {\em IEEE Conference on Computer Vision and Pattern Recognition},
  pages 5828--5839, 2017.

\bibitem{Dai2017a}
Angela Dai, Matthias Niessner, Michael Zollhofer, Shahram Izadi, and Christian
  Theobalt.
\newblock {BundleFusion: Real-time Globally Consistent 3D Reconstruction using
  On-the-fly Surface Re-integration}.
\newblock {\em ACM Transactions on Graphics}, 36(3), 2017.

\bibitem{Dai2017b}
Angela Dai, Charles~Ruizhongtai Qi, and Matthias Nie{\ss}ner.
\newblock {Shape Completion using 3D-Encoder-Predictor CNNs and Shape
  Synthesis}.
\newblock {\em IEEE Conference on Computer Vision and Pattern Recognition},
  pages 5868--5877, 2017.

\bibitem{Davison2007}
Andrew~J Davison, Ian~D Reid, Nicholas~D Molton, and Oliver Stasse.
\newblock {MonoSLAM: Real-time single camera SLAM}.
\newblock {\em IEEE Transactions on Pattern Analysis and Machine Intelligence},
  29(6):1052--1067, 2007.

\bibitem{Delaunoy2014}
Amael Delaunoy and Marc Pollefeys.
\newblock {Photometric Bundle Adjustment for Dense Multi-View 3D Modeling}.
\newblock {\em IEEE Conference on Computer Vision and Pattern Recognition},
  pages 1486--1493, 2014.

\bibitem{Deprelle2019}
Theo Deprelle, Thibault Groueix, Matthew Fisher, Vladimir~G. Kim, Bryan~C.
  Russell, and Mathieu Aubry.
\newblock {Learning Elementary Structures for 3D Shape Generation and
  Matching}.
\newblock {\em Advances in Neural Information Processing Systems}, pages
  7433--7443, 2019.

\bibitem{Donahue2018}
Chris Donahue, Julian McAuley, and Miller Puckette.
\newblock {Synthesizing Audio with GANs}.
\newblock {\em International Conference on Learning Representations Workshops},
  2018.

\bibitem{Dong2018}
Wei Dong, Qiuyuan Wang, Xin Wang, and Hongbin Zha.
\newblock {PSDF Fusion: Probabilistic Signed Distance Function for On-the-fly
  3D Data Fusion and Scene Reconstruction}.
\newblock {\em European Conference on Computer Vision}, pages 714--730, 2018.

\bibitem{Dou2017}
Pengfei Dou, Shishir~K Shah, and Ioannis~A Kakadiaris.
\newblock {End-to-end 3D Face Reconstruction with Deep Neural Networks}.
\newblock {\em IEEE Conference on Computer Vision and Pattern Recognition},
  pages 5908--5917, 2017.

\bibitem{Duan2019}
Yueqi Duan, Yu~Zheng, Jiwen Lu, Jie Zhou, and Qi~Tian.
\newblock {Structural Relational Reasoning of Point Clouds}.
\newblock {\em IEEE Conference on Computer Vision and Pattern Recognition},
  pages 949--958, 2019.

\bibitem{Elich2019}
Cathrin Elich, Francis Engelmann, Jonas Schult, Theodora Kontogianni, and
  Bastian Leibe.
\newblock {3D-BEVIS: Birds-Eye-View Instance Segmentation}.
\newblock {\em German Conference on Pattern Recognition}, pages 48--61, 2019.

\bibitem{Emami2018}
Patrick Emami and Sanjay Ranka.
\newblock {Learning Permutations with Sinkhorn Policy Gradient}.
\newblock {\em arXiv preprint arXiv:1805.07010}, 2018.

\bibitem{Engelcke2016}
Martin Engelcke, Dushyant Rao, Dominic~Zeng Wang, Chi~Hay Tong, and Ingmar
  Posner.
\newblock {Vote3Deep: Fast Object Detection in 3D Point Clouds Using Efficient
  Convolutional Neural Networks}.
\newblock {\em IEEE International Conference on Robotics and Automation}, pages
  1355--1361, 2017.

\bibitem{Engelmann2017}
Francis Engelmann, Theodora Kontogianni, Alexander Hermans, and Bastian Leibe.
\newblock {Exploring Spatial Context for 3D Semantic Segmentation of Point
  Clouds}.
\newblock {\em IEEE International Conference on Computer Vision Workshops},
  pages 716--724, 2017.

\bibitem{Eslami2018}
S.M.~Ali Eslami, Danilo~Jimenez Rezende, Frederic Besse, Fabio Viola, Ari~S.
  Morcos, Marta Garnelo, Avraham Ruderman, Andrei~A. Rusu, Ivo Danihelka, Karol
  Gregor, David~P. Reichert, Lars Buesing, Theophane Weber, Oriol Vinyals, Dan
  Rosenbaum, Neil Rabinowitz, Helen King, Chloe Hillier, Matt Botvinick, Daan
  Wierstra, Koray Kavukcuoglu, and Demis Hassabis.
\newblock {Neural Scene Representation and Rendering}.
\newblock {\em Science}, 360(6394):1204--1210, 2018.

\bibitem{Everingham2010}
Mark Everingham, Luc~Van Gool, Christopher K.~I. Williams, John Winn, and
  Andrew Zisserman.
\newblock {The PASCAL Visual Object Classes (VOC) Challenge}.
\newblock {\em International Journal of Computer Vision}, (88):303--338, 2010.

\bibitem{Fan2017}
Haoqiang Fan, Hao Su, and Leonidas Guibas.
\newblock {A Point Set Generation Network for 3D Object Reconstruction from a
  Single Image}.
\newblock {\em IEEE Conference on Computer Vision and Pattern Recognition},
  pages 605--613, 2017.

\bibitem{Firman2016}
Michael Firman, Oisin~Mac Aodha, Simon Julier, and Gabriel~J. Brostow.
\newblock {Structured Prediction of Unobserved Voxels From a Single Depth
  Image}.
\newblock {\em IEEE Conference on Computer Vision and Pattern Recognition},
  pages 5431--5440, 2016.

\bibitem{Frahm2010}
Jan-Michael Frahm, Pierre Fite-Georgel, David Gallup, Tim Johnson, Rahul
  Raguram, Changchang Wu, Yi-Hung Jen, Enrique Dunn, Brian Clipp, Svetlana
  Lazebnik, and Marc Pollefeys.
\newblock {Building Rome on a Cloudless Day}.
\newblock {\em European Conference on Computer Vision}, pages 368--381, 2010.

\bibitem{Gadelha2016}
Matheus Gadelha, Subhransu Maji, and Rui Wang.
\newblock {3D Shape Induction from 2D Views of Multiple Objects}.
\newblock {\em International Conference on 3D Vision}, pages 402--411, 2017.

\bibitem{Gardner2017a}
Andrew Gardner, Jinko Kanno, Christian~A. Duncan, and Rastko~R. Selmic.
\newblock {Classifying Unordered Feature Sets with Convolutional Deep Averaging
  Networks}.
\newblock {\em IEEE International Conference on Systems, Man and Cybernetics},
  pages 3447--3453, 2019.

\bibitem{Gherardi2010}
Riccardo Gherardi, Michela Farenzena, and Andrea Fusiello.
\newblock {Improving the Efficiency of Hierarchical Structure-and-motion}.
\newblock {\em IEEE Conference on Computer Vision and Pattern Recognition},
  pages 1594--1600, 2010.

\bibitem{Ghosh2020}
Pallabi Ghosh, Larry~S Davis, and Neel Joshi.
\newblock {Deep Depth Prior for Multi-View Stereo}.
\newblock {\em arXiv:2001.07791}, 2020.

\bibitem{Girdhar}
Rohit Girdhar, David~F Fouhey, Mikel Rodriguez, and Abhinav Gupta.
\newblock {Learning a Predictable and Generative Vector Representation for
  Objects}.
\newblock {\em European Conference on Computer Vision}, pages 484--499, 2016.

\bibitem{Girdhar2017a}
Rohit Girdhar and Deva Ramanan.
\newblock {Attentional Pooling for Action Recognition}.
\newblock {\em Advances in Neural Information Processing Systems}, pages
  34--45, 2017.

\bibitem{Goldlucke2014}
Bastian Goldl{\"{u}}cke, Mathieu Aubry, Kalin Kolev, and Daniel Cremers.
\newblock {A Super-Resolution Framework for High-Accuracy Multiview
  Reconstruction}.
\newblock {\em International Journal of Computer Vision}, 106:172–191, 2014.

\bibitem{Goodfellow2014}
Ian~J. Goodfellow, Jean Pouget-Abadie, Mehdi Mirza, Bing Xu, David
  Warde-Farley, Sherjil Ozair, Aaron Courville, and Yoshua Bengio.
\newblock {Generative Adversarial Nets}.
\newblock {\em Advances in Neural Information Processing Systems}, pages
  2672--2680, 2014.

\bibitem{Graham2018}
Benjamin Graham, Martin Engelcke, and Laurens van~der Maaten.
\newblock {3D Semantic Segmentation with Submanifold Sparse Convolutional
  Networks}.
\newblock {\em IEEE Conference on Computer Vision and Pattern Recognition},
  pages 9224--9232, 2018.

\bibitem{B2016f}
Edward Grant, Pushmeet Kohli, and Marcel~Van Gerven.
\newblock {Deep Disentangled Representations for Volumetric Reconstruction}.
\newblock {\em European Conference on Computer Vision Workshops}, pages
  266--279, 2016.

\bibitem{Groueix2018}
Thibault Groueix, Matthew Fisher, Vladimir~G Kim, Bryan~C Russell, and Mathieu
  Aubry.
\newblock {A Papier-Mache Approach to Learning 3D Surface Generation}.
\newblock {\em IEEE Conference on Computer Vision and Pattern Recognition},
  pages 216--224, 2018.

\bibitem{Grover2019}
Aditya Grover, Eric Wang, Aaron Zweig, and Stefano Ermon.
\newblock {Stochastic Optimization of Sorting Networks via Continuous
  Relaxations}.
\newblock {\em International Conference on Learning Representations}, 2019.

\bibitem{Guerrero2018}
Paul Guerrero, Yanir Kleiman, Maks Ovsjanikov, and Niloy~J. Mitra.
\newblock {PCPNET: Learning Local Shape Properties from Raw Point Clouds}.
\newblock {\em Computer Graphics Forum}, 37(2):75--85, 2018.

\bibitem{Gulrajani2017}
Ishaan Gulrajani, Faruk Ahmed, Martin Arjovsky, Vincent Dumoulin, and Aaron
  Courville.
\newblock {Improved Training of Wasserstein GANs}.
\newblock {\em Advances in Neural Information Processing Systems}, pages
  5767--5777, 2017.

\bibitem{Gwak2017}
JunYoung Gwak, Christopher~B. Choy, Manmohan Chandraker, Animesh Garg, and
  Silvio Savarese.
\newblock {Weakly supervised 3D Reconstruction with Adversarial Constraint}.
\newblock {\em International Conference on 3D Vision}, pages 263--272, 2017.

\bibitem{Hackel2017}
Timo Hackel, N.~Savinov, L.~Ladicky, Jan~D. Wegner, K.~Schindler, and
  M.~Pollefeys.
\newblock {SEMANTIC3D.NET: A New Large-scale Point Cloud Classification
  Benchmark}.
\newblock {\em ISPRS Annals of the Photogrammetry, Remote Sensing and Spatial
  Information Sciences}, IV-1-W1:91--98, 2017.

\bibitem{Han2017}
Xiaoguang Han, Zhen Li, Haibin Huang, Evangelos Kalogerakis, and Yizhou Yu.
\newblock {High-Resolution Shape Completion Using Deep Neural Networks for
  Global Structure and Local Geometry Inference}.
\newblock {\em IEEE International Conference on Computer Vision}, pages 85--93,
  2017.

\bibitem{Han2019b}
Zhizhong Han, Honglei Lu, Zhenbao Liu, Chi~Man Vong, Yu~Shen Liu, Matthias
  Zwicker, Junwei Han, and C.~L. Philip~Chen.
\newblock {3D2SeqViews: Aggregating Sequential Views for 3D Global Feature
  Learning by CNN with Hierarchical Attention Aggregation}.
\newblock {\em IEEE Transactions on Image Processing}, 28(8):3986--3999, 2019.

\bibitem{Hartley2004}
Richard Hartley and Andrew Zisserman.
\newblock {\em {Multiple View Geometry in Computer Vision}}.
\newblock Cambridge University Press, 2004.

\bibitem{He2017a}
Kaiming He, Georgia Gkioxari, Piotr Dollar, and Ross Girshick.
\newblock {Mask R-CNN}.
\newblock {\em IEEE International Conference on Computer Vision}, pages
  2961--2969, 2017.

\bibitem{He2016b}
Kaiming He, Xiangyu Zhang, Shaoqing Ren, and Jian Sun.
\newblock {Deep Residual Learning for Image Recognition}.
\newblock {\em IEEE Conference on Computer Vision and Pattern Recognition},
  pages 770--778, 2016.

\bibitem{He2020a}
Tong He, Dong Gong, Zhi Tian, and Chunhua Shen.
\newblock {Learning and Memorizing Representative Prototypes for 3D Point Cloud
  Semantic and Instance Segmentation}.
\newblock {\em arXiv:2001.01349}, 2020.

\bibitem{He2020}
Xingzhe He, Helen~Lu Cao, and Bo~Zhu.
\newblock {AdvectiveNet: An Eulerian-Lagrangian Fluidic reservoir for Point
  Cloud Processing}.
\newblock {\em International Conference on Learning Representations}, 2020.

\bibitem{Henry2012}
Peter Henry, Michael Krainin, Evan Herbst, Xiaofeng Ren, and Dieter Fox.
\newblock {RGB-D Mapping: Using Kinect-style Depth Cameras for Dense 3D
  Modeling of Indoor Environments}.
\newblock {\em International Journal of Robotics Research}, 31(5):647--663,
  2012.

\bibitem{Hermosilla2018}
Pedro Hermosilla, Tobias Ritschel, Pere-Pau Vazquez, Alvar Vinacua, and Timo
  Ropinski.
\newblock {Monte Carlo Convolution for Learning on Non-Uniformly Sampled Point
  Clouds}.
\newblock {\em ACM Transactions on Graphics}, 37(6):1--12, 2018.

\bibitem{Hou2019}
Ji~Hou, Angela Dai, and Matthias Nie{\ss}ner.
\newblock {3D-SIS: 3D Semantic Instance Segmentation of RGB-D Scans}.
\newblock {\em IEEE Conference on Computer Vision and Pattern Recognition},
  pages 4421--4430, 2019.

\bibitem{Hu2020a}
Peiyun Hu, Jason Ziglar, David Held, and Deva Ramanan.
\newblock {What You See is What You Get: Exploiting Visibility for 3D Object
  Detection}.
\newblock {\em IEEE Conference on Computer Vision and Pattern Recognition},
  2020.

\bibitem{Hu2020}
Qingyong Hu, Bo~Yang, Linhai Xie, Stefano Rosa, Yulan Guo, Zhihua Wang, Niki
  Trigoni, and Andrew Markham.
\newblock {RandLA-Net: Efficient Semantic Segmentation of Large-Scale Point
  Clouds}.
\newblock {\em IEEE Conference on Computer Vision and Pattern Recognition},
  pages 11108--11117, 2020.

\bibitem{Hu2017d}
Zhiting Hu, Zichao Yang, Xiaodan Liang, Ruslan Salakhutdinov, and Eric~P. Xing.
\newblock {Toward Controlled Generation of Text}.
\newblock {\em International Conference on Machine Learning}, pages 1587--1596,
  2017.

\bibitem{Hua2016}
Binh-Son Hua, Quang-Hieu Pham, Duc~Thanh Nguyen, Minh-Khoi Tran, Lap-Fai Yu,
  and Sai-Kit Yeung.
\newblock {SceneNN: A Scene Meshes Dataset with Annotations}.
\newblock {\em International Conference on 3D Vision}, 2016.

\bibitem{Hua2018}
Binh-Son Hua, Minh-Khoi Tran, and Sai-Kit Yeung.
\newblock {Pointwise Convolutional Neural Networks}.
\newblock {\em IEEE Conference on Computer Vision and Pattern Recognition},
  pages 984--993, 2018.

\bibitem{Huang2015a}
Haibin Huang, Evangelos Kalogerakis, and Benjamin Marlin.
\newblock {Analysis and Synthesis of 3D Shape Families via Deep-learned
  Generative Models of Surfaces}.
\newblock {\em Computer Graphics Forum}, 34(5):25--38, 2015.

\bibitem{Huang2016a}
Jing Huang and Suya You.
\newblock {Point Cloud Labeling using 3D Convolutional Neural Network}.
\newblock {\em International Conference on Pattern Recognition}, pages
  2670--2675, 2016.

\bibitem{Huang2018}
Po-Han Huang, Kevin Matzen, Johannes Kopf, Narendra Ahuja, and Jia-Bin Huang.
\newblock {DeepMVS: Learning Multi-view Stereopsis}.
\newblock {\em IEEE Conference on Computer Vision and Pattern Recognition},
  pages 2821--2830, 2018.

\bibitem{Huang2018a}
Qiangui Huang, Weiyue Wang, and Ulrich Neumann.
\newblock {Recurrent Slice Networks for 3D Segmentation of Point Clouds}.
\newblock {\em IEEE Conference on Computer Vision and Pattern Recognition},
  pages 2626--2635, 2018.

\bibitem{Ilse2018}
Maximilian Ilse, Jakub~M. Tomczak, and Max Welling.
\newblock {Attention-based Deep Multiple Instance Learning}.
\newblock {\em International Conference on Machine Learning}, pages 2127--2136,
  2018.

\bibitem{Insafutdinov2018}
Eldar Insafutdinov and Alexey Dosovitskiy.
\newblock {Unsupervised Learning of Shape and Pose with Differentiable Point
  Clouds}.
\newblock {\em Advances in Neural Information Processing Systems}, pages
  2802--2812, 2018.

\bibitem{Ionescu2015}
Catalin Ionescu, Orestis Vantzos, and Cristian Sminchisescu.
\newblock {Matrix backpropagation for deep networks with structured layers}.
\newblock {\em IEEE International Conference on Computer Vision}, pages
  2965--2973, 2015.

\bibitem{Ji2017b}
Mengqi Ji, Juergen Gall, Haitian Zheng, Yebin Liu, and Lu~Fang.
\newblock {SurfaceNet: An End-to-end 3D Neural Network for Multiview
  Stereopsis}.
\newblock {\em IEEE International Conference on Computer Vision}, pages
  2326--2334, 2017.

\bibitem{Jiang2018}
Li~Jiang, Shaoshuai Shi, Xiaojuan Qi, and Jiaya Jia.
\newblock {GAL: Geometric adversarial loss for single-view 3D-object
  reconstruction}.
\newblock {\em European Conference on Computer Vision}, pages 820--834, 2018.

\bibitem{Jiang2019a}
Li~Jiang, Hengshuang Zhao, Shu Liu, Xiaoyong Shen, Chi-Wing Fu, and Jiaya Jia.
\newblock {Hierarchical Point-Edge Interaction Network for Point Cloud Semantic
  Segmentation}.
\newblock {\em IEEE International Conference on Computer Vision}, pages
  10433--10441, 2019.

\bibitem{JieHu2018}
{Jie Hu}, {Li Shen}, Samuel Albanie, Gang Sun, and Enhua Wu.
\newblock {Squeeze-and-Excitation Networks}.
\newblock {\em IEEE Conference on Computer Vision and Pattern Recognition},
  pages 7132--7141, 2018.

\bibitem{Johnston2017}
Adrian Johnston, Ravi Garg, Gustavo Carneiro, Ian Reid, and Anton van~den
  Hengel.
\newblock {Scaling CNNs for High Resolution Volumetric Reconstruction from a
  Single Image}.
\newblock {\em IEEE International Conference on Computer Vision Workshops},
  2017.

\bibitem{Kanade1981}
Takeo Kanade.
\newblock {Recovery of the Three-Dimensional Shape of an Object from a Single
  View}.
\newblock {\em Artificial Intelligence}, 17(1-3):409--460, 1981.

\bibitem{Kar2017}
Abhishek Kar, Christian H{\"{a}}ne, and Jitendra Malik.
\newblock {Learning a Multi-View Stereo Machine}.
\newblock {\em International Conference on Neural Information Processing
  Systems}, pages 364--375, 2017.

\bibitem{Kar2015a}
Abhishek Kar, Shubham Tulsiani, Joao Carreira, and Jitendra Malik.
\newblock {Category-specific Object Reconstruction from a Single Image}.
\newblock {\em IEEE Conference on Computer Vision and Pattern Recognition},
  pages 1966--1974, 2015.

\bibitem{Karras2017}
Tero Karras, Timo Aila, Samuli Laine, and Jaakko Lehtinen.
\newblock {Progressive Growing of GANs for Improved Quality, Stability, and
  Variation}.
\newblock {\em International Conference on Learning Representations}, 2018.

\bibitem{Kato2017}
Hiroharu Kato, Yoshitaka Ushiku, and Tatsuya Harada.
\newblock {Neural 3D Mesh Renderer}.
\newblock {\em IEEE Conference on Computer Vision and Pattern Recognition},
  pages 3907--3916, 2018.

\bibitem{Kazhdan2006}
Michael Kazhdan, Matthew Bolitho, and Hugues Hoppe.
\newblock {Poisson Surface Reconstruction}.
\newblock {\em Symposium on Geometry Processing}, 2006.

\bibitem{Kazhdan2013}
Michael Kazhdan and Hugues Hoppe.
\newblock {Screened Poisson Surface Reconstruction}.
\newblock {\em ACM Transactions on Graphics}, 32(3):1--13, 2013.

\bibitem{Khoshelham2012}
Kourosh Khoshelham and Sander~Oude Elberink.
\newblock {Accuracy and Resolution of Kinect Depth Data for Indoor Mapping
  Applications}.
\newblock {\em Sensors}, 12:1437--1454, 2012.

\bibitem{Kim2012}
Young~Min Kim, Niloy~J. Mitra, Dong-Ming Yan, and Leonidas Guibas.
\newblock {Acquiring 3D Indoor Environments with Variability and Repetition}.
\newblock {\em ACM Transactions on Graphics}, 31(6), 2012.

\bibitem{Kingma2015a}
Diederik~P. Kingma and Jimmy Ba.
\newblock {Adam: A Method for Stochastic Optimization}.
\newblock {\em International Conference on Learning Representations}, 2015.

\bibitem{Kingma2014}
Diederik~P Kingma and Max Welling.
\newblock {Auto-Encoding Variational Bayes}.
\newblock {\em International Conference on Learning Representations}, 2014.

\bibitem{Klokov2017}
Roman Klokov and Victor Lempitsky.
\newblock {Escape from Cells: Deep Kd-Networks for The Recognition of 3D Point
  Cloud Models}.
\newblock {\em IEEE International Conference on Computer Vision}, pages
  863--872, 2017.

\bibitem{Koch2018}
Sebastian Koch, Albert Matveev, Zhongshi Jiang, Francis Williams, Alexey
  Artemov, Evgeny Burnaev, Marc Alexa, Denis Zorin, and Daniele Panozzo.
\newblock {ABC: A Big CAD Model Dataset For Geometric Deep Learning}.
\newblock {\em IEEE Conference on Computer Vision and Pattern Recognition},
  pages 9601--9611, 2019.

\bibitem{Kolen2001}
John~F. Kolen and Stefan~C. Kremer.
\newblock {Gradient Flow in Recurrent Nets: The Difficulty of Learning Long
  Term Dependencies}.
\newblock {\em A Field Guide to Dynamical Recurrent Networks}, 2001.

\bibitem{Komarichev2019}
Artem Komarichev, Zichun Zhong, and Jing Hua.
\newblock {A-CNN: Annularly Convolutional Neural Networks on Point Clouds}.
\newblock {\em IEEE Conference on Computer Vision and Pattern Recognition},
  pages 7421--7430, 2019.

\bibitem{Kong2017}
Chen Kong, Chen-Hsuan Lin, and Simon Lucey.
\newblock {Using Locally Corresponding CAD Models for Dense 3D Reconstructions
  from a Single Image}.
\newblock {\em IEEE Conference on Computer Vision and Pattern Recognition},
  pages 4857--4865, 2017.

\bibitem{Krizhevsky2012b}
Alex Krizhevsky, Ilya Sutskever, and Geoffrey~E. Hinton.
\newblock {Imagenet Classification with Deep Convolutional Neural Networks}.
\newblock {\em Advances in Neural Information Processing Systems}, pages
  1097--1105, 2012.

\bibitem{Ku2018}
Jason Ku, Melissa Mozifian, Jungwook Lee, Ali Harakeh, and Steven~L. Waslander.
\newblock {Joint 3D Proposal Generation and Object Detection from View
  Aggregation}.
\newblock {\em IEEE/RSJ International Conference on Intelligent Robots and
  Systems}, pages 1--8, 2018.

\bibitem{Kuhn1955}
Harold~W. Kuhn.
\newblock {The Hungarian Method for the assignment problem}.
\newblock {\em Naval Research Logistics Quarterly}, 2(1-2):83--97, 1955.

\bibitem{Kuhn1956}
Harold~W. Kuhn.
\newblock {Variants of the hungarian method for assignment problems}.
\newblock {\em Naval Research Logistics Quarterly}, 3(4):253--258, 1956.

\bibitem{Kulkarni2015}
Tejas~D Kulkarni, William~F Whitney, Pushmeet Kohli, and Joshua~B Tenenbaum.
\newblock {Deep Convolutional Inverse Graphics Network}.
\newblock {\em Advances in Neural Information Processing Systems}, pages
  2539--2547, 2015.

\bibitem{Kumar2017}
Suryansh Kumar, Yuchao Dai, and Hongdong Li.
\newblock {Monocular Dense 3D Reconstruction of a Complex Dynamic Scene from
  Two Perspective Frames}.
\newblock {\em IEEE International Conference on Computer Vision}, pages
  4649--4657, 2017.

\bibitem{Kurenkov2017}
Andrey Kurenkov, Jingwei Ji, Animesh Garg, Viraj Mehta, JunYoung Gwak,
  Christopher Choy, and Silvio Savarese.
\newblock {DeformNet: Free-Form Deformation Network for 3D Shape Reconstruction
  from a Single Image}.
\newblock {\em IEEE Winter Conference on Applications of Computer Vision},
  pages 858--866, 2017.

\bibitem{L2019}
Navaneet~K L, Priyanka Mandikal, Mayank Agarwal, and R.~Venkatesh Babu.
\newblock {CAPNet: Continuous Approximation Projection For 3D Point Cloud
  Reconstruction Using 2D Supervision}.
\newblock {\em AAAI Conference on Artificial Intelligence}, 2019.

\bibitem{Lahoud2019}
Jean Lahoud, Bernard Ghanem, Marc Pollefeys, and Martin~R. Oswald.
\newblock {3D Instance Segmentation via Multi-Task Metric Learning}.
\newblock {\em IEEE International Conference on Computer Vision}, pages
  9256--9266, 2019.

\bibitem{Lan2019}
Shiyi Lan, Ruichi Yu, Gang Yu, and Larry~S. Davis.
\newblock {Modeling Local Geometric Structure of 3D Point Clouds using
  Geo-CNN}.
\newblock {\em IEEE Conference on Computer Vision and Pattern Recognition},
  pages 998--1008, 2019.

\bibitem{Landrieu2018}
Loic Landrieu and Martin Simonovsky.
\newblock {Large-scale Point Cloud Semantic Segmentation with Superpoint
  Graphs}.
\newblock {\em IEEE Conference on Computer Vision and Pattern Recognition},
  pages 4558--4567, 2018.

\bibitem{Lang2019}
Alex~H. Lang, Sourabh Vora, Holger Caesar, Lubing Zhou, Jiong Yang, and Oscar
  Beijbom.
\newblock {PointPillars: Fast Encoders for Object Detection from Point Clouds}.
\newblock {\em IEEE Conference on Computer Vision and Pattern Recognition},
  pages 12697--12705, 2019.

\bibitem{Lea}
Truc Le and Ye~Duan.
\newblock {PointGrid: A Deep Network for 3D Shape Understanding}.
\newblock {\em IEEE Conference on Computer Vision and Pattern Recognition},
  pages 9204--9214, 2018.

\bibitem{LeCun2015a}
Yann LeCun, Yoshua Bengio, and Geoffrey Hinton.
\newblock {Deep Learning}.
\newblock {\em Nature}, 521:436--444, 2015.

\bibitem{Ledig2016}
Christian Ledig, Lucas Theis, Ferenc Huszar, Jose Caballero, Andrew Cunningham,
  Alejandro Acosta, Andrew Aitken, Alykhan Tejani, Johannes Totz, Zehan Wang,
  and Wenzhe Shi.
\newblock {Photo-Realistic Single Image Super-Resolution Using a Generative
  Adversarial Network}.
\newblock {\em IEEE Conference on Computer Vision and Pattern Recognition},
  pages 4681--4690, 2017.

\bibitem{Lee2019a}
Juho Lee, Yoonho Lee, Jungtaek Kim, Adam~R. Kosiorek, Seungjin Choi, and
  Yee~Whye Teh.
\newblock {Set Transformer: A Framework for Attention-based
  Permutation-Invariant Neural Networks}.
\newblock {\em International Conference on Machine Learning}, pages 3744--3753,
  2019.

\bibitem{Lei2019}
Huan Lei, Naveed Akhtar, and Ajmal Mian.
\newblock {Octree guided CNN with Spherical Kernels for 3D Point Clouds}.
\newblock {\em IEEE Conference on Computer Vision and Pattern Recognition},
  pages 9631--9640, 2019.

\bibitem{Li2017j}
Bo~Li.
\newblock {3D Fully Convolutional Network for Vehicle Detection in Point
  Cloud}.
\newblock {\em IEEE/RSJ International Conference on Intelligent Robots and
  Systems}, pages 1513--1518, 2017.

\bibitem{Li2016f}
Bo~Li, Tianlei Zhang, and Tian Xia.
\newblock {Vehicle Detection from 3D Lidar Using Fully Convolutional Network}.
\newblock {\em Robotics: Science and Systems}, 2016.

\bibitem{Li2018a}
Hanchao Li, Pengfei Xiong, Jie An, and Lingxue Wang.
\newblock {Pyramid Attention Network for Semantic Segmentation}.
\newblock {\em arXiv}, 1805.10180, 2018.

\bibitem{Li2018}
Jiaxin Li, Ben~M. Chen, and Gim~Hee Lee.
\newblock {SO-Net: Self-Organizing Network for Point Cloud Analysis}.
\newblock {\em IEEE Conference on Computer Vision and Pattern Recognition},
  pages 9397--9406, 2018.

\bibitem{Li2019b}
Kejie Li, Ravi Garg, Ming Cai, and Ian Reid.
\newblock {Single-view Object Shape Reconstruction Using Deep Shape Prior and
  Silhouette}.
\newblock {\em British Machine Vision Conference}, 2019.

\bibitem{Li2018f}
Yangyan Li, Rui Bu, Mingchao Sun, Wei Wu, Xinhan Di, and Baoquan Chen.
\newblock {PointCNN : Convolution On X -Transformed Points}.
\newblock {\em Advances in Neural Information Processing Systems}, pages
  820--830, 2018.

\bibitem{Li2015a}
Yangyan Li, Angela Dai, Leonidas Guibas, and Matthias Nie{\ss}ner.
\newblock {Database-Assisted Object Retrieval for Real-Time 3D Reconstruction}.
\newblock {\em Computer Graphics Forum}, 34(2):435--446, 2015.

\bibitem{Liang2019}
Zhidong Liang, Ming Yang, and Chunxiang Wang.
\newblock {3D Graph Embedding Learning with a Structure-aware Loss Function for
  Point Cloud Semantic Instance Segmentation}.
\newblock {\em arXiv preprint arXiv:1902.05247}, 2019.

\bibitem{Lin2017a}
Chen-Hsuan Lin, Chen Kong, and Simon Lucey.
\newblock {Learning Efficient Point Cloud Generation for Dense 3D Object
  Reconstruction}.
\newblock {\em AAAI Conference on Artificial Intelligence}, 2018.

\bibitem{Lin2019}
Chen-Hsuan Lin, Oliver Wang, Bryan~C. Russell, Eli Shechtman, Vladimir~G. Kim,
  Matthew Fisher, and Simon Lucey.
\newblock {Photometric Mesh Optimization for Video-Aligned 3D Object
  Reconstruction}.
\newblock {\em IEEE Conference on Computer Vision and Pattern Recognition},
  pages 969--978, 2019.

\bibitem{Lin2018c}
Shuyu Lin, Bo~Yang, Robert Birke, and Ronald Clark.
\newblock {Learning Semantically Meaningful Embeddings Using Linear
  Constraints}.
\newblock {\em IEEE Conference on Computer Vision and Pattern Recognition
  Workshops (CVPRW)}, pages 53--56, 2018.

\bibitem{Lin2017c}
Tsung-Yi Lin, Priya Goyal, Ross Girshick, Kaiming He, and Piotr Dollar.
\newblock {Focal Loss for Dense Object Detection}.
\newblock {\em ICCV}, 2017.

\bibitem{Lin2014}
Tsung-Yi Lin, Michael Maire, Serge Belongie, James Hays, Pietro Perona, Deva
  Ramanan, Piotr Doll{\'{a}}r, and C.~Lawrence Zitnick.
\newblock {Microsoft COCO: Common Objects in Context}.
\newblock {\em European Conference on Computer Vision}, pages 740--755, 2014.

\bibitem{Lin2017b}
Tsung-Yu Lin and Subhransu Maji.
\newblock {Improved Bilinear Pooling with CNNs}.
\newblock {\em British Machine Vision Conference}, 2017.

\bibitem{Lin2018}
Tsung-Yu Lin, Subhransu Maji, and Piotr Koniusz.
\newblock {Second-order Democratic Aggregation}.
\newblock {\em European Conference on Computer Vision}, pages 620--636, 2018.

\bibitem{Lin2015}
Tsung~Yu Lin, Aruni Roychowdhury, and Subhransu Maji.
\newblock {Bilinear CNN Models for Fine-grained Visual Recognition}.
\newblock {\em IEEE International Conference on Computer Vision}, pages
  1449--1457, 2015.

\bibitem{Lin2020}
Yiqun Lin, Zizheng Yan, Haibin Huang, Dong Du, Ligang Liu, Shuguang Cui, and
  Xiaoguang Han.
\newblock {FPConv: Learning Local Flattening for Point Convolution}.
\newblock {\em IEEE Conference on Computer Vision and Pattern Recognition},
  2020.

\bibitem{Liu2019}
Chen Liu and Yasutaka Furukawa.
\newblock {MASC: Multi-scale Affinity with Sparse Convolution for 3D Instance
  Segmentation}.
\newblock {\em arXiv preprint arXiv:1902.04478}, 2019.

\bibitem{Liu2017d}
Fangyu Liu, Shuaipeng Li, Liqiang Zhang, Chenghu Zhou, Rongtian Ye, Yuebin
  Wang, and Jiwen Lu.
\newblock {3DCNN-DQN-RNN: A Deep Reinforcement Learning Framework for Semantic
  Parsing of Large-scale 3D Point Clouds}.
\newblock {\em IEEE International Conference on Computer Vision}, pages
  5678--5687, 2017.

\bibitem{Liu2019f}
Jinxian Liu, Bingbing Ni, Caiyuan Li, Jiancheng Yang, and Qi~Tian.
\newblock {Dynamic Points Agglomeration for Hierarchical Point Sets Learning}.
\newblock {\em IEEE International Conference on Computer Vision}, pages
  7546--7555, 2019.

\bibitem{Liu2020}
Minghua Liu, Lu~Sheng, Sheng Yang, Jing Shao, and Shi-Min Hu.
\newblock {Morphing and Sampling Network for Dense Point Cloud Completion}.
\newblock {\em AAAI Conference on Artificial Intelligence}, 2020.

\bibitem{Liu2018a}
Sainan Liu, Saining Xie, Zeyu Chen, and Zhuowen Tu.
\newblock {Attentional ShapeContextNet for Point Cloud Recognition}.
\newblock {\em IEEE Conference on Computer Vision and Pattern Recognition},
  pages 4606--4615, 2018.

\bibitem{Liu2019e}
Shichen Liu, Tianye Li, Weikai Chen, and Hao Li.
\newblock {Soft Rasterizer: A Differentiable Renderer for Image-based 3D
  Reasoning}.
\newblock {\em IEEE International Conference on Computer Vision}, pages
  7708--7717, 2019.

\bibitem{Liu2019b}
Shichen Liu, Shunsuke Saito, Weikai Chen, and Hao Li.
\newblock {Learning to Infer Implicit Surfaces without 3D Supervision}.
\newblock {\em Advances in Neural Information Processing Systems}, pages
  8293--8304, 2019.

\bibitem{Liu2019g}
Shikun Liu, Edward Johns, and Andrew~J. Davison.
\newblock {End-to-End Multi-Task Learning with Attention}.
\newblock {\em IEEE Conference on Computer Vision and Pattern Recognition},
  pages 1871--1880, 2019.

\bibitem{Liu2018e}
Xiaofeng Liu, B.V.K~Vijaya Kumar, Chao Yang, Qingming Tang, and Jane You.
\newblock {Dependency-aware Attention Control for Unconstrained Face
  Recognition with Image Sets}.
\newblock {\em European Conference on Computer Vision}, pages 548--565, 2018.

\bibitem{Liu2018g}
Xinhai Liu, Zhizhong Han, Yu-Shen Liu, and Matthias Zwicker.
\newblock {Point2Sequence: Learning the Shape Representation of 3D Point Clouds
  with an Attention-based Sequence to Sequence Network}.
\newblock {\em AAAI Conference on Artificial Intelligence}, pages 8778--8785,
  2019.

\bibitem{Liu2019c}
Yongcheng Liu, Bin Fan, Shiming Xiang, and Chunhong Pan.
\newblock {Relation-Shape Convolutional Neural Network for Point Cloud
  Analysis}.
\newblock {\em IEEE Conference on Computer Vision and Pattern Recognition},
  pages 8895--8904, 2019.

\bibitem{Lowe2004}
David~G. Lowe.
\newblock {Distinctive Image Features from Scale-Invariant Keypoints}.
\newblock {\em International Journal of Computer Vision}, 60(2):91--110, 2004.

\bibitem{Lu2019}
Qiang Lu, Mingjie Xiao, Yiyang Lu, Xiaohui Yuan, and Ye~Yu.
\newblock {Attention-Based Dense Point Cloud Reconstruction from a Single
  Image}.
\newblock {\em IEEE Access}, 7:137420--137431, 2019.

\bibitem{Maaten2008}
Laurens van~der Maaten and Geoffrey Hinton.
\newblock {Visualizing Data using t-SNE}.
\newblock {\em Journal of Machine Learning Research}, 9(Nov):2579--2605, 2008.

\bibitem{Mandikal2019}
Priyanka Mandikal and R.~Venkatesh Babu.
\newblock {Dense 3D point cloud reconstruction using a deep pyramid network}.
\newblock {\em IEEE Winter Conference on Applications of Computer Vision},
  pages 1052--1060, 2019.

\bibitem{Mao2019}
Jiageng Mao, Xiaogang Wang, and Hongsheng Li.
\newblock {Interpolated Convolutional Networks for 3D Point Cloud
  Understanding}.
\newblock {\em IEEE International Conference on Computer Vision}, pages
  1578--1587, 2019.

\bibitem{Martin2018}
Eric Martin and Chris Cundy.
\newblock {Parallelizing Linear Recurrent Neural Nets over Sequence Length}.
\newblock {\em International Conference on Learning Representations}, 2018.

\bibitem{McCormac2018}
John McCormac, Ronald Clark, Michael Bloesch, Andrew~J. Davison, and Stefan
  Leutenegger.
\newblock {Fusion++: Volumetric Object-Level SLAM}.
\newblock {\em International Conference on 3D Vision}, pages 32--41, 2018.

\bibitem{McCormac2017}
John McCormac, Ankur Handa, Stefan Leutenegger, and Andrew~J. Davison.
\newblock {SceneNet RGB-D: 5M Photorealistic Images of Synthetic Indoor
  Trajectories with Ground Truth}.
\newblock {\em IEEE International Conference on Computer Vision}, pages
  2697--2706, 2017.

\bibitem{Meng2019}
Hsien-Yu Meng, Gao Lin, Yu-Kun Lai, and Dinesh Manocha.
\newblock {VV-Net: Voxel VAE Net with Group Convolutions for Point Cloud
  Segmentation}.
\newblock {\em IEEE International Conference on Computer Vision}, pages
  8500--8508, 2019.

\bibitem{Mescheder2019}
Lars Mescheder, Michael Oechsle, Michael Niemeyer, Sebastian Nowozin, and
  Andreas Geiger.
\newblock {Occupancy Networks: Learning 3D Reconstruction in Function Space}.
\newblock {\em IEEE Conference on Computer Vision and Pattern Recognition},
  pages 4455--4465, 2019.

\bibitem{Meyer2019}
Gregory~P. Meyer, Jake Charland, Darshan Hegde, Ankit Laddha, and Carlos
  Vallespi-Gonzalez.
\newblock {Sensor Fusion for Joint 3D Object Detection and Semantic
  Segmentation}.
\newblock {\em IEEE Conference on Computer Vision and Pattern Recognition
  Workshops}, 2019.

\bibitem{Minemura2018}
Kazuki Minemura, Hengfui Liau, Abraham Monrroy, and Shinpei Kato.
\newblock {LMNet: Real-time Multiclass Object Detection on CPU using 3D LiDAR}.
\newblock {\em Asia-Pacific Conference on Intelligent Robot Systems}, pages
  28--34, 2018.

\bibitem{Mirza2014}
Mehdi Mirza and Simon Osindero.
\newblock {Conditional Generative Adversarial Nets}.
\newblock {\em arXiv preprint arXiv:1411.1784}, 2014.

\bibitem{Mitra2006}
Niloy~J Mitra, Leonidas~J Guibas, and Mark Pauly.
\newblock {Partial and Approximate Symmetry Detection for 3D Geometry}.
\newblock {\em SIGGRAPH}, 25(3):560--568, 2006.

\bibitem{Mo2019}
Kaichun Mo, Shilin Zhu, Angel~X. Chang, Li~Yi, Subarna Tripathi, Leonidas~J.
  Guibas, and Hao Su.
\newblock {PartNet: A Large-scale Benchmark for Fine-grained and Hierarchical
  Part-level 3D Object Understanding}.
\newblock {\em IEEE Conference on Computer Vision and Pattern Recognition},
  pages 909--918, 2019.

\bibitem{Monszpart2015a}
Aron Monszpart, Nicolas Mellado, Gabriel~J. Brostow, and Niloy~J. Mitra.
\newblock {RAPter: Rebuilding Man-made Scenes with Regular Arrangements of
  Planes}.
\newblock {\em ACM Transactions on Graphics}, 34(4):1--12, 2015.

\bibitem{Mroueh2017b}
Youssef Mroueh, Tom Sercu, and Vaibhava Goel.
\newblock {McGAN: Mean and Covariance Feature Matching GAN}.
\newblock {\em International Conference on Machine Learning}, pages 2527--2535,
  2017.

\bibitem{Mukherjee1995}
Dipti~Prasad Mukherjee, Andrew Zisserman, and Michael Brady.
\newblock {Shape from Symmetry: Detecting and Exploiting Symmetry in Affine
  Images}.
\newblock {\em Philosophical Transactions: Physical Sciences and Engineering},
  351(1695):77--106, 1995.

\bibitem{Mur-Artal2015}
Raul Mur-Artal, JMM M~M Montiel, and Juan~D Tardos.
\newblock {ORB-SLAM: a versatile and accurate monocular SLAM system}.
\newblock {\em IEEE Trans on Robotics}, 31(5):1147--1163, 2015.

\bibitem{Mur-Artal2016}
Raul Mur-Artal and Juan~D. Tardos.
\newblock {ORB-SLAM2: an Open-Source SLAM System for Monocular, Stereo and
  RGB-D Cameras}.
\newblock {\em IEEE Transactions on Robotics}, 33(5):1255 -- 1262, 2017.

\bibitem{Murthy2016}
J~Krishna Murthy, G~V~Sai Krishna, Falak Chhaya, and K~Madhava Krishna.
\newblock {Reconstructing Vechicles from a Single Image: Shape Priors for Road
  Scene Understanding}.
\newblock {\em IEEE International Conference on Robotics and Automation}, pages
  724--731, 2017.

\bibitem{Nakka2018}
Krishna~Kanth Nakka and Mathieu Salzmann.
\newblock {Deep Attentional Structured Representation Learning for Visual
  Recognition}.
\newblock {\em British Machine Vision Conference}, 2018.

\bibitem{Nan}
Liangliang Nan, Ke~Xie, and Andrei Sharf.
\newblock {A Search-Classify Approach for Cluttered Indoor Scene
  Understanding}.
\newblock {\em ACM Transactions on Graphics}, 31(6):1--10, 2012.

\bibitem{Narita2019}
Gaku Narita, Takashi Seno, Tomoya Ishikawa, and Yohsuke Kaji.
\newblock {PanopticFusion: Online Volumetric Semantic Mapping at the Level of
  Stuff and Things}.
\newblock {\em IEEE/RSJ International Conference on Intelligent Robots and
  Systems}, 2019.

\bibitem{Nash2020}
Charlie Nash, Yaroslav Ganin, S.~M.~Ali Eslami, and Peter~W. Battaglia.
\newblock {PolyGen: An Autoregressive Generative Model of 3D Meshes}.
\newblock {\em International Conference on Machine Learning}, 2020.

\bibitem{Nealen2006}
Andrew Nealen, Takeo Igarashi, Olga Sorkine, and Marc Alexa.
\newblock {Laplacian Mesh Optimization}.
\newblock {\em International Conference on Computer Graphics and Interactive
  Techniques in Australasia and Southeast Asia}, pages 381--389, 2006.

\bibitem{Newcombe2011a}
Richard~A. Newcombe, Shahram Izadi, Otmar Hilliges, David Molyneaux, David Kim,
  Andrew~J. Davison, Pushmeet Kohli, Jamie Shotton, Steve Hodges, and Andrew
  Fitzgibbon.
\newblock {KinectFusion: Real-time Dense Surface Mapping and Tracking}.
\newblock {\em International Symposium on Mixed and Augmented Reality}, pages
  127--136, 2011.

\bibitem{Newcombe2011b}
Richard~A. Newcombe, Steven~J. Lovegrove, and Andrew~J. Davision.
\newblock {DTAM: Dense Tracking and Mapping in Real-time}.
\newblock {\em IEEE International Conference on Computer Vision}, pages
  2320--2327, 2011.

\bibitem{Nicastro2019}
Andrea Nicastro, Ronald Clark, and Stefan Leutenegger.
\newblock {X-Section: Cross-Section Prediction for Enhanced RGBD Fusion}.
\newblock {\em IEEE International Conference on Computer Vision}, pages
  1517--1526, 2019.

\bibitem{Niemeyer2019}
Michael Niemeyer, Lars Mescheder, Michael Oechsle, and Andreas Geiger.
\newblock {Differentiable Volumetric Rendering: Learning Implicit 3D
  Representations without 3D Supervision}.
\newblock {\em IEEE Conference on Computer Vision and Pattern Recognition},
  pages 3504--3515, 2020.

\bibitem{Niener2013}
Matthias Nie{\ss}ner, Michael Zollh{\"{o}}fer, Shahram Izadi, and Marc
  Stamminger.
\newblock {Real-time 3D Reconstruction at Scale using Voxel Hashing}.
\newblock {\em ACM Transactions on Graphics}, 32(6):1--11, 2013.

\bibitem{Odena2017}
Augustus Odena, Christopher Olah, and Jonathon Shlens.
\newblock {Conditional Image Synthesis with Auxiliary Classifier GANs}.
\newblock {\em International Conference on Machine Learning}, page 2642–2651,
  2017.

\bibitem{Ozyesil2017}
Onur Ozyesil, Vladislav Voroninski, Ronen Basri, and Amit Singer.
\newblock {A Survey of Structure from Motion}.
\newblock {\em Acta Numerica}, 26:305--364, 2017.

\bibitem{Paigwar2019}
Anshul Paigwar, Ozgur Erkent, Christian Wolf, and Christian Laugier.
\newblock {Attentional PointNet for 3D-Object Detection in Point Clouds}.
\newblock {\em IEEE Conference on Computer Vision and Pattern Recognition
  Workshops}, 2019.

\bibitem{Pan2019}
Junyi Pan, Xiaoguang Han, Weikai Chen, Jiapeng Tang, and Kui Jia.
\newblock {Deep Mesh Reconstruction from Single RGB Images via Topology
  Modification Networks}.
\newblock {\em IEEE International Conference on Computer Vision}, pages
  9964--9973, 2019.

\bibitem{Pan2020}
Yancheng Pan, Biao Gao, Jilin Mei, Sibo Geng, Chengkun Liy, and Huijing Zhao.
\newblock {SemanticPOSS: A Point Cloud Dataset with Large Quantity of Dynamic
  Instances}.
\newblock {\em arXiv:2002.09147}, 2020.

\bibitem{Park2019}
Jeong~Joon Park, Peter Florence, Julian Straub, Richard Newcombe, and Steven
  Lovegrove.
\newblock {DeepSDF: Learning Continuous Signed Distance Functions for Shape
  Representation}.
\newblock {\em IEEE Conference on Computer Vision and Pattern Recognition},
  pages 165--174, 2019.

\bibitem{Paschalidou2018}
Despoina Paschalidou, Ali~Osman Ulusoy, Carolin Schmitt, Luc Van~Gool, and
  Andreas Geiger.
\newblock {RayNet: Learning Volumetric 3D Reconstruction with Ray Potentials}.
\newblock {\em IEEE Conference on Computer Vision and Pattern Recognition},
  pages 3897--3906, 2018.

\bibitem{Pauly2008}
Mark Pauly, Niloy~J. Mitra, Johannes Wallner, Helmut Pottmann, and Leonidas~J.
  Guibas.
\newblock {Discovering Structural Regularity in 3D Geometry}.
\newblock {\em ACM Transactions on Graphics}, 27(3):1, 2008.

\bibitem{Peng2020}
Yanjun Peng, Ming Chang, Qiong Wang, Yinling Qian, Yingkui Zhang, Mingqiang
  Wei, and Xiangyun Liao.
\newblock {Sparse-to-Dense Multi-Encoder Shape Completion of Unstructured Point
  Cloud}.
\newblock {\em IEEE Access}, 2020.

\bibitem{Pham2019}
Quang-Hieu Pham, Duc~Thanh Nguyen, Binh-Son Hua, Gemma Roig, and Sai-Kit Yeung.
\newblock {JSIS3D: Joint Semantic-Instance Segmentation of 3D Point Clouds with
  Multi-Task Pointwise Networks and Multi-Value Conditional Random Fields}.
\newblock {\em IEEE Conference on Computer Vision and Pattern Recognition},
  pages 8827--8836, 2019.

\bibitem{Qi2020}
Charles~R Qi, Xinlei Chen, Or~Litany, and Leonidas~J. Guibas.
\newblock {ImVoteNet : Boosting 3D Object Detection in Point Clouds with Image
  Votes}.
\newblock {\em IEEE Conference on Computer Vision and Pattern Recognition},
  2020.

\bibitem{Qi2019}
Charles~R. Qi, Or~Litany, Kaiming He, and Leonidas~J. Guibas.
\newblock {Deep Hough Voting for 3D Object Detection in Point Clouds}.
\newblock {\em IEEE International Conference on Computer Vision}, pages
  9277--9286, 2019.

\bibitem{Qi2018}
Charles~R. Qi, Wei Liu, Chenxia Wu, Hao Su, and Leonidas~J. Guibas.
\newblock {Frustum PointNets for 3D Object Detection from RGB-D Data}.
\newblock {\em IEEE Conference on Computer Vision and Pattern Recognition},
  pages 918--927, 2018.

\bibitem{Qi2016}
Charles~R. Qi, Hao Su, Kaichun Mo, and Leonidas~J. Guibas.
\newblock {PointNet: Deep Learning on Point Sets for 3D Classification and
  Segmentation}.
\newblock {\em IEEE Conference on Computer Vision and Pattern Recognition},
  pages 652--660, 2017.

\bibitem{Qi2016a}
Charles~R Qi, Hao Su, Matthias Nie{\ss}ner, Angela Dai, Mengyuan Yan, and
  Leonidas~J Guibas.
\newblock {Volumetric and Multi-View CNNs for Object Classification on 3D
  Data}.
\newblock {\em IEEE Conference on Computer Vision and Pattern Recognition},
  pages 5648--5656, 2016.

\bibitem{Qi2017}
Charles~R. Qi, Li~Yi, Hao Su, and Leonidas~J. Guibas.
\newblock {PointNet++: Deep Hierarchical Feature Learning on Point Sets in a
  Metric Space}.
\newblock {\em Advances in Neural Information Processing Systems}, pages
  5099--5108, 2017.

\bibitem{Radford2016}
Alec Radford, Luke Metz, and Soumith Chintala.
\newblock {Unsupervised Representation Learning with Deep Convolutional
  Generative Adversarial Networks}.
\newblock {\em International Conference on Learning Representations}, 2016.

\bibitem{Raffel2016}
Colin Raffel and Daniel P.~W. Ellis.
\newblock {Feed-Forward Networks with Attention Can Solve Some Long-Term Memory
  Problems}.
\newblock {\em International Conference on Learning Representations Workshops},
  2016.

\bibitem{Rao2019}
Yongming Rao, Jiwen Lu, and Jie Zhou.
\newblock {Spherical Fractal Convolutional Neural Networks for Point Cloud
  Recognition}.
\newblock {\em IEEE Conference on Computer Vision and Pattern Recognition},
  pages 452--460, 2019.

\bibitem{Rashed2019}
Hazem Rashed, Mohamed Ramzy, Victor Vaquero, Ahmad~El Sallab, Ganesh Sistu, and
  Senthil Yogamani.
\newblock {FuseMODNet: Real-Time Camera and LiDAR based Moving Object Detection
  for robust low-light Autonomous Driving}.
\newblock {\em IEEE International Conference on Computer Vision Workshops},
  2019.

\bibitem{Reed2016}
Scott Reed, Zeynep Akata, Xinchen Yan, Lajanugen Logeswaran, Bernt Schiele, and
  Honglak Lee.
\newblock {Generative Adversarial Text to Image Synthesis}.
\newblock {\em International Conference on Machine Learning}, pages 1060--1069,
  2016.

\bibitem{Reed2016a}
Scott~E. Reed, Zeynep Akata, Santosh Mohan, Samuel Tenka, Bernt Schiele, and
  Honglak Lee.
\newblock {Learning What and Where to Draw}.
\newblock {\em Advances in Neural Information Processing Systems}, pages
  217--225, 2016.

\bibitem{Ren2015a}
Shaoqing Ren, Kaiming He, Ross Girshick, and Jian Sun.
\newblock {Faster R-CNN: Towards Real-time Object Detection with Region
  Proposal Networks}.
\newblock {\em Advances in Neural Information Processing Systems}, pages
  91--99, 2015.

\bibitem{Rethage2018}
Dario Rethage, Johanna Wald, Jürgen Sturm, Nassir Navab, and Federico Tombari.
\newblock {Fully-Convolutional Point Networks for Large-Scale Point Clouds}.
\newblock {\em European Conference on Computer Vision}, pages 596--611, 2018.

\bibitem{Riegler2017}
Gernot Riegler, Ali~Osman Ulusoy, Horst Bischof, and Andreas Geiger.
\newblock {OctNetFusion: Learning Depth Fusion from Data}.
\newblock {\em International Conference on 3D Vision}, pages 57--66, 2017.

\bibitem{Riegler2017a}
Gernot Riegler, Ali~Osman Ulusoy, and Andreas Geiger.
\newblock {OctNet: Learning Deep 3D Representations at High Resolutions}.
\newblock {\em IEEE Conference on Computer Vision and Pattern Recognition},
  pages 3577--3586, 2017.

\bibitem{Rock2015}
Jason Rock, Tanmay Gupta, Justin Thorsen, JunYoung Gwak, Daeyun Shin, and Derek
  Hoiem.
\newblock {Completing 3D Object Shape from One Depth Image}.
\newblock {\em IEEE Conference on Computer Vision and Pattern Recognition},
  pages 2484--2493, 2015.

\bibitem{Rodriguez2018}
Pau Rodr{\'{i}}guez, Josep~M. Gonfaus, Guillem Cucurull, F.~Xavier Roca, and
  Jordi Gonz{\`{a}}lez.
\newblock {Attend and Rectify: a Gated Attention Mechanism for Fine-Grained
  Recovery}.
\newblock {\em European Conference on Computer Vision}, pages 349--364, 2018.

\bibitem{Ronneberger2015}
Olaf Ronneberger, Philipp Fischer, and Thomas Brox.
\newblock {U-Net : Convolutional Networks for Biomedical Image Segmentation}.
\newblock {\em International Conference on Medical Image Computing and
  Computer-Assisted Intervention}, pages 234--241, 2015.

\bibitem{Russakovsky2015}
Olga Russakovsky, Jia Deng, Hao Su, Jonathan Krause, Sanjeev Satheesh, Sean Ma,
  Zhiheng Huang, Andrej Karpathy, Aditya Khosla, Michael Bernstein,
  Alexander~C. Berg, and Li~Fei-Fei.
\newblock {ImageNet Large Scale Visual Recognition Challenge}.
\newblock {\em International Journal of Computer Vision}, (115):211--252, 2015.

\bibitem{Rusu2009a}
Radu~Bogdan Rusu, Nico Blodow, and Michael Beetz.
\newblock {Fast Point Feature Histograms (fpfh) for 3D Registration}.
\newblock {\em IEEE International Conference on Robotics and Automation}, pages
  3212--3217, 2009.

\bibitem{Salas-Moreno2013}
Renato~F. Salas-Moreno, Richard~A. Newcombe, Hauke Strasdat, Paul H.~J. Kelly,
  and Andrew~J. Davison.
\newblock {SLAM++: Simultaneous Localisation and Mapping at the Level of
  Objects}.
\newblock {\em IEEE Conference on Computer Vision and Pattern Recognition},
  pages 1352--1359, 2013.

\bibitem{Sarafianos2018}
Nikolaos Sarafianos, Xiang Xu, and Ioannis~A. Kakadiaris.
\newblock {Deep Imbalanced Attribute Classification using Visual Attention
  Aggregation}.
\newblock {\em European Conference on Computer Vision}, pages 680--697, 2018.

\bibitem{Schonberger2016}
Johannes~L. Schonberger and Jan-Michael Frahm.
\newblock {Structure-from-Motion Revisited}.
\newblock {\em IEEE Conference on Computer Vision and Pattern Recognition},
  pages 4104--4113, 2016.

\bibitem{Schops2019}
Thomas Schops, Torsten Sattler, and Marc Pollefeys.
\newblock {BAD SLAM: Bundle Adjusted Direct RGB-D SLAM}.
\newblock {\em IEEE Conference on Computer Vision and Pattern Recognition},
  pages 134--144, 2019.

\bibitem{Shao2012}
Tianjia Shao, Weiwei Xu, Kun Zhou, Jingdong Wang, Dongping Li, and Baining Guo.
\newblock {An Interactive Approach to Semantic Modeling of Indoor Scenes with
  an RGBD Camera}.
\newblock {\em ACM Transactions on Graphics}, 31(6):1--11, 2012.

\bibitem{B2016e}
Abhishek Sharma, Oliver Grau, and Mario Fritz.
\newblock {VConv-DAE : Deep Volumetric Shape Learning Without Object Labels}.
\newblock {\em European Conference on Computer Vision}, pages 236--250, 2016.

\bibitem{Shen2017a}
Yiru Shen, Chen Feng, Yaoqing Yang, and Dong Tian.
\newblock {Mining Point Cloud Local Structures by Kernel Correlation and Graph
  Pooling}.
\newblock {\em IEEE Conference on Computer Vision and Pattern Recognition},
  pages 4548--4557, 2018.

\bibitem{Shi2019}
Shaoshuai Shi, Xiaogang Wang, and Hongsheng Li.
\newblock {PointRCNN: 3D Object Proposal Generation and Detection from Point
  Cloud}.
\newblock {\em IEEE Conference on Computer Vision and Pattern Recognition},
  pages 770--779, 2019.

\bibitem{Shi2016}
Yifei Shi, Pinxin Long, Kai Xu, Hui Huang, and Yueshan Xiong.
\newblock {Data-driven Contextual Modeling for 3D Scene Understanding}.
\newblock {\em Computers {\&} Graphics}, 55:55--67, 2016.

\bibitem{Shin2019}
Daeyun Shin, Zhile Ren, Erik~B. Sudderth, and Charless~C. Fowlkes.
\newblock {3D Scene Reconstruction with Multi-layer Depth and Epipolar
  Transformers}.
\newblock {\em IEEE International Conference on Computer Vision}, pages
  2172--2182, 2019.

\bibitem{Simon2018}
Martin Simon, Stefan Milz, Karl Amende, and Horst-michael Gross.
\newblock {Complex-YOLO: An Euler-Region-Proposal for Real-time 3D Object
  Detection on Point Clouds}.
\newblock {\em European Conference on Computer Vision Workshops}, 2018.

\bibitem{Simonyan2015}
Karen Simonyan and Andrew Zisserman.
\newblock {Very Deep Convolutional Networks for Large-Scale Image Recognition}.
\newblock {\em International Conference on Learning Representations}, 2015.

\bibitem{Sipiran2014}
Ivan Sipiran, Robert Gregor, and Tobias Schreck.
\newblock {Approximate Symmetry Detection in Partial 3D Meshes}.
\newblock {\em Computer Graphics Forum}, 33(7):131--140, 2014.

\bibitem{Sitzmann2019}
Vincent Sitzmann, Michael Zollh{\"{o}}fer, and Gordon Wetzstein.
\newblock {Scene Representation Networks: Continuous 3D-Structure-Aware Neural
  Scene Representations}.
\newblock {\em Advances in Neural Information Processing Systems}, pages
  1119--1130, 2019.

\bibitem{Slavcheva2016}
Miroslava Slavcheva, Wadim Kehl, Nassir Navab, and Slobodan Ilic.
\newblock {SDF-2-SDF: Highly Accurate 3D Object Reconstruction}.
\newblock {\em European Conference on Computer Vision}, pages 680--696, 2016.

\bibitem{Smith2017}
Edward Smith and David Meger.
\newblock {Improved Adversarial Systems for 3D Object Generation and
  Reconstruction}.
\newblock {\em Conference on Robot Learning}, pages 87--96, 2017.

\bibitem{Soltani2017}
Amir~Arsalan Soltani, Haibin Huang, Jiajun Wu, Tejas~D. Kulkarni, and Joshua~B.
  Tenenbaum.
\newblock {Synthesizing 3D Shapes via Modeling Multi-View Depth Maps and
  Silhouettes with Deep Generative Networks}.
\newblock {\em IEEE Conference on Computer Vision and Pattern Recognition},
  pages 1511--1519, 2017.

\bibitem{Song2017}
Shuran Song, Fisher Yu, Andy Zeng, Angel~X. Chang, Manolis Savva, and Thomas
  Funkhouser.
\newblock {Semantic Scene Completion from a Single Depth Image}.
\newblock {\em IEEE Conference on Computer Vision and Pattern Recognition},
  pages 1746--1754, 2017.

\bibitem{Speciale2016}
Pablo Speciale, Martin~R. Oswald, Andrea Cohen, and Marc Pollefeys.
\newblock {A Symmetry Prior for Convex Variational 3D Reconstruction}.
\newblock {\em European Conference on Computer Vision}, pages 313--328, 2016.

\bibitem{Sridhar2019}
Srinath Sridhar, Davis Rempe, Julien Valentin, Sofien Bouaziz, and Leonidas~J.
  Guibas.
\newblock {Multiview Aggregation for Learning Category-Specific Shape
  Reconstruction}.
\newblock {\em Advances in Neural Information Processing Systems}, pages
  2348--2359, 2019.

\bibitem{Steinbr2013}
Frank Steinbrucker, Christian Kerl, Jurgen Sturm, and Daniel Cremers.
\newblock {Large-Scale Multi-Resolution Surface Reconstruction from RGB-D
  Sequences}.
\newblock {\em IEEE International Conference on Computer Vision}, pages
  3264--3271, 2013.

\bibitem{Straub2019}
Julian Straub, Thomas Whelan, Lingni Ma, Yufan Chen, Erik Wijmans, Simon Green,
  Jakob~J Engel, Raul Mur-artal, Carl Ren, Shobhit Verma, Anton Clarkson,
  Mingfei Yan, Brian Budge, Yajie Yan, Xiaqing Pan, June Yon, Yuyang Zou,
  Kimberly Leon, Nigel Carter, Jesus Briales, Tyler Gillingham, Elias Mueggler,
  Luis Pesqueira, Manolis Savva, Dhruv Batra, and Hauke~M Strasdat.
\newblock {The Replica Dataset : A Digital Replica of Indoor Spaces}.
\newblock {\em arXiv:1906.05797}, 2019.

\bibitem{Su2018}
Hang Su, Varun Jampani, Deqing Sun, Subhransu Maji, Evangelos Kalogerakis,
  Ming-Hsuan Yang, and Jan Kautz.
\newblock {SPLATNet: Sparse Lattice Networks for Point Cloud Processing}.
\newblock {\em IEEE Conference on Computer Vision and Pattern Recognition},
  pages 2530--2539, 2018.

\bibitem{Su2015}
Hang Su, Subhransu Maji, Evangelos Kalogerakis, and Erik Learned-Miller.
\newblock {Multi-view Convolutional Neural Networks for 3D Shape Recognition}.
\newblock {\em IEEE International Conference on Computer Vision}, pages
  945--953, 2015.

\bibitem{Sweeney2015}
Chris Sweeney, Torsten Sattler, Tobias H{\"{o}}llerer, Matthew Turk, and Marc
  Pollefeys.
\newblock {Optimizing the Viewing Graph for Structure-from-Motion}.
\newblock {\em IEEE International Conference on Computer Vision}, pages
  801--809, 2015.

\bibitem{Tang2019}
Jiapeng Tang, Xiaoguang Han, Junyi Pan, Kui Jia, and Xin Tong.
\newblock {A Skeleton-bridged Deep Learning Approach for Generating Meshes of
  Complex Topologies from Single RGB Images}.
\newblock {\em IEEE Conference on Computer Vision and Pattern Recognition},
  pages 4541--4550, 2019.

\bibitem{Tatarchenko2017}
Maxim Tatarchenko, Alexey Dosovitskiy, and Thomas Brox.
\newblock {Octree Generating Networks: Efficient Convolutional Architectures
  for High-resolution 3D Outputs}.
\newblock {\em IEEE International Conference on Computer Vision}, pages
  2088--2096, 2017.

\bibitem{Tchapmi2017}
Lyne~P. Tchapmi, Christopher~B Choy, Iro Armeni, JunYoung Gwak, and Silvio
  Savarese.
\newblock {SEGCloud: Semantic Segmentation of 3D Point Clouds}.
\newblock {\em International Conference on 3D Vision}, pages 537--547, 2017.

\bibitem{Te2018}
Gusi Te, Wei Hu, Zongming Guo, and Amin Zheng.
\newblock {RGCNN : Regularized Graph CNN for Point Cloud Segmentation}.
\newblock {\em ACM Multimedia Conference on Multimedia Conference}, pages
  746--754, 2018.

\bibitem{Terzopoulos1988}
Demetri Terzopoulos, Andrew Witkin, and Michael Kass.
\newblock {Symmetry-seeking Models and 3D Object Reconstruction}.
\newblock {\em International Journal of Computer Vision}, 1:211--221, 1988.

\bibitem{Thomas2018a}
Hugues Thomas, Jean-emmanuel Deschaud, Beatriz Marcotegui, Francois Goulette,
  and Yann Legall.
\newblock {Semantic Classification of 3D Point Clouds with Multiscale Geometric
  Neighborhoods}.
\newblock {\em International Conference on 3D Vision}, pages 390--398, 2018.

\bibitem{Thomas2019}
Hugues Thomas, Charles~R. Qi, Jean-Emmanuel Deschaud, Beatriz Marcotegui,
  François Goulette, and Leonidas~J. Guibas.
\newblock {KPConv: Flexible and Deformable Convolution for Point Clouds}.
\newblock {\em IEEE International Conference on Computer Vision}, pages
  6411--6420, 2019.

\bibitem{Thrun2005}
Sebastian Thrun and Ben Wegbreit.
\newblock {Shape from Symmetry}.
\newblock {\em IEEE International Conference on Computer Vision}, pages
  1824--1831, 2005.

\bibitem{Triggs2000}
Bill Triggs, Philip~F. McLauchlan, Richard~I. Hartley, and Andrew~W.
  Fitzgibbon.
\newblock {Bundle Adjustment - A Modern Synthesis}.
\newblock {\em International Workshop on Vision Algorithms}, 1999.

\bibitem{Tulsiani2017}
Shubham Tulsiani, Tinghui Zhou, Alexei~A. Efros, and Jitendra Malik.
\newblock {Multi-view Supervision for Single-view Reconstruction via
  Differentiable Ray Consistency}.
\newblock {\em IEEE Conference on Computer Vision and Pattern Recognition},
  pages 2626--2634, 2017.

\bibitem{Vaquero2017}
Víctor Vaquero, Ivan Del~Pino, Francesc Moreno-Noguer, Joan So{\`{i}}, Alberto
  Sanfeliu, and Juan Andrade-Cetto.
\newblock {Deconvolutional Networks for Point-Cloud Vehicle Detection and
  Tracking in Driving Scenarios}.
\newblock {\em European Conference on Mobile Robots}, pages 1--7, 2017.

\bibitem{Varley2017}
Jacob Varley, Chad Dechant, Adam Richardson, Joaquín Ruales, and Peter Allen.
\newblock {Shape Completion Enabled Robotic Grasping}.
\newblock {\em International Conference on Intelligent Robots and Systems},
  pages 2442--2447, 2017.

\bibitem{Vaswani2017}
Ashish Vaswani, Noam Shazeer, Niki Parmar, Jakob Uszkoreit, Llion Jones,
  Aidan~N. Gomez, Lukasz Kaiser, and Illia Polosukhin.
\newblock {Attention Is All You Need}.
\newblock {\em Advances in Neural Information Processing Systems}, pages
  5998--6008, 2017.

\bibitem{Vinyals2016a}
Oriol Vinyals, Samy Bengio, and Manjunath Kudlur.
\newblock {Order Matters: Sequence to Sequence for Sets}.
\newblock {\em International Conference on Learning Representations}, 2015.

\bibitem{Wagstaff2019}
Edward Wagstaff, Fabian Fuchs, Martin Engelcke, Ingmar Posner, and Michael~A.
  Osborne.
\newblock {On the Limitations of Representing Functions on Sets}.
\newblock {\em International Conference on Machine Learning}, pages 6487--6494,
  2019.

\bibitem{Wang2018e}
Chu Wang, Babak Samari, and Kaleem Siddiqi.
\newblock {Local Spectral Graph Convolution for Point Set Feature Learning}.
\newblock {\em European Conference on Computer Vision}, pages 52--66, 2018.

\bibitem{Wang2019b}
Lei Wang, Yuchun Huang, Yaolin Hou, Shenman Zhang, and Jie Shan.
\newblock {Graph Attention Convolution for Point Cloud Semantic Segmentation}.
\newblock {\em IEEE Conference on Computer Vision and Pattern Recognition},
  pages 10296--10305, 2019.

\bibitem{Wang2017i}
Meng Wang, Lingjing Wang, and Yi~Fang.
\newblock {3DensiNet: A Robust Neural Network Architecture towards 3D
  Volumetric Object Prediction from 2D Image}.
\newblock {\em ACM international conference on Multimedia}, page 961–969,
  2017.

\bibitem{Wang2018f}
Nanyang Wang, Yinda Zhang, Zhuwen Li, Yanwei Fu, Wei Liu, and Yu-Gang Jiang.
\newblock {Pixel2Mesh: Generating 3D Mesh Models from Single RGB Images}.
\newblock {\em European Conference on Computer Vision}, pages 52--67, 2018.

\bibitem{Wang2019f}
Wei Wang, Muhamad Risqi~U Saputra, Peijun Zhao, Pedro Gusmao, Bo~Yang, Changhao
  Chen, Andrew Markham, and Niki Trigoni.
\newblock {DeepPCO: End-to-End Point Cloud Odometry through Deep Parallel
  Neural Network}.
\newblock {\em International Conference on Intelligent Robots and Systems
  (IROS)}, 2019.

\bibitem{Wang2017b}
Weiyue Wang, Qiangui Huang, Suya You, Chao Yang, and Ulrich Neumann.
\newblock {Shape Inpainting using 3D Generative Adversarial Network and
  Recurrent Convolutional Networks}.
\newblock {\em IEEE International Conference on Computer Vision}, pages
  2298--2306, 2017.

\bibitem{Wang2018d}
Weiyue Wang, Ronald Yu, Qiangui Huang, and Ulrich Neumann.
\newblock {SGPN: Similarity Group Proposal Network for 3D Point Cloud Instance
  Segmentation}.
\newblock {\em IEEE Conference on Computer Vision and Pattern Recognition},
  pages 2569--2578, 2018.

\bibitem{Wang2019}
Xinlong Wang, Shu Liu, Xiaoyong Shen, Chunhua Shen, and Jiaya Jia.
\newblock {Associatively Segmenting Instances and Semantics in Point Clouds}.
\newblock {\em IEEE Conference on Computer Vision and Pattern Recognition},
  pages 4096--4105, 2019.

\bibitem{Wang2019e}
Xu~Wang, Jingming He, and Lin Ma.
\newblock {Exploiting Local and Global Structure for Point Cloud Semantic
  Segmentation with Contextual Point Representations}.
\newblock {\em Advances in Neural Information Processing Systems}, pages
  4573--4583, 2019.

\bibitem{Wang2018c}
Yue Wang, Yongbin Sun, Ziwei Liu, Sanjay~E. Sarma, Michael~M. Bronstein, and
  Justin~M. Solomon.
\newblock {Dynamic Graph CNN for Learning on Point Clouds}.
\newblock {\em ACM Transactions on Graphics}, 38(5), 2019.

\bibitem{Wang2018a}
Zhihua Wang, Stefano Rosa, Linhai Xie, Bo~Yang, Sen Wang, Niki Trigoni, and
  Andrew Markham.
\newblock {Defo-Net: Learning Body Deformation Using Generative Adversarial
  Networks}.
\newblock {\em IEEE International Conference on Robotics and Automation}, pages
  2440--2447, 2018.

\bibitem{Wang2018l}
Zhihua Wang, Stefano Rosa, Bo~Yang, Sen Wang, Niki Trigoni, and Andrew Markham.
\newblock {3D-PhysNet: Learning the Intuitive Physics of Non-Rigid Object
  Deformations}.
\newblock {\em International Joint Conference on Artificial Intelligence}, page
  4958–4964, 2018.

\bibitem{Wang2018b}
Zining Wang, Wei Zhan, and Masayoshi Tomizuka.
\newblock {Fusing Bird View LIDAR Point Cloud and Front View Camera Image for
  Deep Object Detection}.
\newblock {\em arXiv:1711.06703}, 2017.

\bibitem{Wen2019}
Chao Wen, Yinda Zhang, Zhuwen Li, and Yanwei Fu.
\newblock {Pixel2Mesh++: Multi-View 3D Mesh Generation via Deformation}.
\newblock {\em IEEE International Conference on Computer Vision}, pages
  1042--1051, 2019.

\bibitem{Whelan2015}
Thomas Whelan, Stefan Leutenegger, Renato~F Salas-moreno, Ben Glocker, and
  Andrew~J Davison.
\newblock {ElasticFusion : Dense SLAM Without A Pose Graph}.
\newblock {\em Robotics: Science and Systems}, 2015.

\bibitem{Wiles2017}
Olivia Wiles and Andrew Zisserman.
\newblock {SilNet : Single- and Multi-View Reconstruction by Learning from
  Silhouettes}.
\newblock {\em British Machine Vision Conference}, 2017.

\bibitem{Wiles2018a}
Olivia Wiles and Andrew Zisserman.
\newblock {Learning to Predict 3D Surfaces of Sculptures from Single and
  Multiple Views}.
\newblock {\em International Journal of Computer Vision}, pages 1--21, 2018.

\bibitem{Wu2017e}
Bichen Wu, Alvin Wan, Xiangyu Yue, and Kurt Keutzer.
\newblock {SqueezeSeg: Convolutional Neural Nets with Recurrent CRF for
  Real-Time Road-Object Segmentation from 3D LiDAR Point Cloud}.
\newblock {\em IEEE International Conference on Robotics and Automation}, pages
  1887--1893, 2018.

\bibitem{Wu2013}
Changchang Wu.
\newblock {Towards Linear-Time Incremental Structure from Motion}.
\newblock {\em International Conference on 3D Vision}, 2013.

\bibitem{Wu2017d}
Jiajun Wu, Yifan Wang, Tianfan Xue, Xingyuan Sun, William~T Freeman, and
  Joshua~B Tenenbaum.
\newblock {MarrNet: 3D Shape Reconstruction via 2.5D Sketches}.
\newblock {\em Advances in Neural Information Processing Systems}, pages
  540--550, 2017.

\bibitem{Wu2016a}
Jiajun Wu, Chengkai Zhang, Tianfan Xue, William~T Freeman, and Joshua
  B.~Tenenbaum.
\newblock {Learning a Probabilistic Latent Space of Object Shapes via 3D
  Generative-Adversarial Modeling}.
\newblock {\em Advances in Neural Information Processing Systems}, pages
  82--90, 2016.

\bibitem{Wu2019a}
Pengxiang Wu, Chao Chen, Jingru Yi, and Dimitris Metaxas.
\newblock {Point Cloud Processing via Recurrent Set Encoding}.
\newblock {\em AAAI Conference on Artificial Intelligence}, pages 5441--5449,
  2019.

\bibitem{Wu2019}
Wenxuan Wu, Zhongang Qi, and Li~Fuxin.
\newblock {PointConv: Deep Convolutional Networks on 3D Point Clouds}.
\newblock {\em IEEE Conference on Computer Vision and Pattern Recognition},
  pages 9621--9630, 2019.

\bibitem{Wu2015}
Zhirong Wu, Shuran Song, Aditya Khosla, Fisher Yu, Linguang Zhang, Xiaoou Tang,
  and Jianxiong Xiao.
\newblock {3D ShapeNets: A Deep Representation for Volumetric Shapes}.
\newblock {\em IEEE Conference on Computer Vision and Pattern Recognition},
  pages 1912--1920, 2015.

\bibitem{Xie2019}
Haozhe Xie, Hongxun Yao, Xiaoshuai Sun, Shangchen Zhou, Shengping Zhang, and
  Xiaojun Tong.
\newblock {Pix2Vox: Context-aware 3D Reconstruction from Single and Multi-view
  Images}.
\newblock {\em IEEE International Conference on Computer Vision}, pages
  2690--2698, 2019.

\bibitem{Xie2019a}
Haozhe Xie, Hongxun Yao, Shangchen Zhou, Shengping Zhang, Xiaoshuai Sun, and
  Wenxiu Sun.
\newblock {Toward 3D Object Reconstruction from Stereo Images}.
\newblock {\em arXiv:1910.08223}, 2019.

\bibitem{Xu2019c}
Binbin Xu, Wenbin Li, Dimos Tzoumanikas, Michael Bloesch, Andrew Davison, and
  Stefan Leutenegger.
\newblock {MID-Fusion: Octree-based Object-Level Multi-Instance Dynamic SLAM}.
\newblock {\em IEEE International Conference on Robotics and Automation}, pages
  5231--5237, 2019.

\bibitem{Xu2018}
Danfei Xu, Dragomir Anguelov, and Ashesh Jain.
\newblock {PointFusion: Deep Sensor Fusion for 3D Bounding Box Estimation}.
\newblock {\em IEEE Conference on Computer Vision and Pattern Recognition},
  pages 244--253, 2018.

\bibitem{Xu2015b}
Kelvin Xu, Jimmy~Lei Ba, Ryan Kiros, Kyunghyun Cho, Aaron Courville, Ruslan
  Salakhutdinov, Richard~S. Zemel, and Yoshua Bengio.
\newblock {Show, Attend and Tell: Neural Image Caption Generation with Visual
  Attention}.
\newblock {\em International Conference on Machine Learning}, pages 2048--2057,
  2015.

\bibitem{Xu2020}
Qiangeng Xu, Xudong Sun, Cho-Ying Wu, Panqu Wang, and Ulrich Neumann.
\newblock {Grid-GCN for Fast and Scalable Point Cloud Learning}.
\newblock {\em IEEE Conference on Computer Vision and Pattern Recognition},
  2020.

\bibitem{Xu2018a}
Yifan Xu, Tianqi Fan, Mingye Xu, Long Zeng, and Yu~Qiao.
\newblock {SpiderCNN: Deep Learning on Point Sets with Parameterized
  Convolutional Filters}.
\newblock {\em European Conference on Computer Vision}, pages 87--102, 2018.

\bibitem{Yan2016}
Xinchen Yan, Jimei Yang, Ersin Yumer, Yijie Guo, and Honglak Lee.
\newblock {Perspective Transformer Nets: Learning Single-View 3D Object
  Reconstruction without 3D Supervision}.
\newblock {\em Advances in Neural Information Processing Systems}, pages
  1696--1704, 2016.

\bibitem{Yang2018d}
Bin Yang, Wenjie Luo, and Raquel Urtasun.
\newblock {PIXOR : Real-time 3D Object Detection from Point Clouds}.
\newblock {\em IEEE Conference on Computer Vision and Pattern Recognition},
  pages 7652--7660, 2018.

\bibitem{Yang2018b}
Bo~Yang, Zihang Lai, Xiaoxuan Lu, Shuyu Lin, Hongkai Wen, Andrew Markham, and
  Niki Trigoni.
\newblock {Learning 3D Scene Semantics and Structure from a Single Depth
  Image}.
\newblock {\em IEEE Conference on Computer Vision and Pattern Recognition
  Workshops}, pages 309--312, 2018.

\bibitem{Yang2018}
Bo~Yang, Stefano Rosa, Andrew Markham, Niki Trigoni, and Hongkai Wen.
\newblock {Dense 3D Object Reconstruction from a Single Depth View}.
\newblock {\em IEEE Transactions on Pattern Analysis and Machine Intelligence},
  41(12):2820 -- 2834, 2019.

\bibitem{Yang2019d}
Bo~Yang, Jianan Wang, Ronald Clark, Qingyong Hu, Sen Wang, Andrew Markham, and
  Niki Trigoni.
\newblock {Learning Object Bounding Boxes for 3D Instance Segmentation on Point
  Clouds}.
\newblock {\em Advances in Neural Information Processing Systems}, pages
  6737--6746, 2019.

\bibitem{Yang2020}
Bo~Yang, Sen Wang, Andrew Markham, and Niki Trigoni.
\newblock {Robust Attentional Aggregation of Deep Feature Sets for Multi-view
  3D Reconstruction}.
\newblock {\em International Journal of Computer Vision}, 128:53--73, 2020.

\bibitem{Yang2017b}
Bo~Yang, Hongkai Wen, Sen Wang, Ronald Clark, Andrew Markham, and Niki Trigoni.
\newblock {3D Object Reconstruction from a Single Depth View with Adversarial
  Learning}.
\newblock {\em IEEE International Conference on Computer Vision Workshops},
  pages 679--688, 2017.

\bibitem{Yang2019b}
Jiancheng Yang, Qiang Zhang, Bingbing Ni, Linguo Li, Jinxian Liu, Mengdie Zhou,
  and Qi~Tian.
\newblock {Modeling Point Clouds with Self-Attention and Gumbel Subset
  Sampling}.
\newblock {\em IEEE Conference on Computer Vision and Pattern Recognition},
  pages 3323--3332, 2019.

\bibitem{Yang2018c}
Xin Yang, Yuanbo Wang, Yaru Wang, Baocai Yin, Qiang Zhang, Xiaopeng Wei, and
  Hongbo Fu.
\newblock {Active Object Reconstruction Using a Guided View Planner}.
\newblock {\em International Joint Conference on Artificial Intelligence},
  pages 4965--4971, 2018.

\bibitem{Yang2019c}
Zetong Yang, Yanan Sun, Shu Liu, Xiaoyong Shen, and Jiaya Jia.
\newblock {STD: Sparse-to-Dense 3D Object Detector for Point Cloud}.
\newblock {\em IEEE International Conference on Computer Vision}, pages
  1951--1960, 2019.

\bibitem{Yang2016}
Zichao Yang, Xiaodong He, Jianfeng Gao, Li~Deng, and Alex Smola.
\newblock {Stacked Attention Networks for Image Question Answering}.
\newblock {\em IEEE Conference on Computer Vision and Pattern Recognition},
  pages 21--29, 2016.

\bibitem{Yao2018}
Yao Yao, Zixin Luo, Shiwei Li, Tian Fang, and Long Quan.
\newblock {MVSNet: Depth Inference for Unstructured Multi-view Stereo}.
\newblock {\em European Conference on Computer Vision}, pages 767--783, 2018.

\bibitem{Ye2018}
Xiaoqing Ye, Jiamao Li, Hexiao Huang, Liang Du, and Xiaolin Zhang.
\newblock {3D Recurrent Neural Networks with Context Fusion for Point Cloud
  Semantic Segmentation}.
\newblock {\em European Conference on Computer Vision}, pages 403--417, 2018.

\bibitem{Yi2019}
Li~Yi, Wang Zhao, He~Wang, Minhyuk Sung, and Leonidas Guibas.
\newblock {GSPN: Generative Shape Proposal Network for 3D Instance Segmentation
  in Point Cloud}.
\newblock {\em IEEE Conference on Computer Vision and Pattern Recognition},
  pages 3947--3956, 2019.

\bibitem{Yin2019}
Penghang Yin, Jiancheng Lyu, Shuai Zhang, Stanley Osher, Yingyong Qi, and Jack
  Xin.
\newblock {Understanding Straight-Through Estimator in Training Activation
  Quantized Neural Nets}.
\newblock {\em International Conference on Learning Representations}, 2019.

\bibitem{Yu2018b}
Kaicheng Yu and Mathieu Salzmann.
\newblock {Statistically Motivated Second Order Pooling}.
\newblock {\em European Conference on Computer Vision}, pages 600--616, 2018.

\bibitem{Yu2017c}
Lantao Yu, Weinan Zhang, Jun Wang, and Yong Yu.
\newblock {SeqGAN: Sequence Generative Adversarial Nets with Policy Gradient}.
\newblock {\em AAAI Conference on Artificial Intelligence}, 2017.

\bibitem{Yu2018a}
Tan Yu, Jingjing Meng, and Junsong Yuan.
\newblock {Multi-view Harmonized Bilinear Network for 3D Object Recognition}.
\newblock {\em IEEE Conference on Computer Vision and Pattern Recognition},
  pages 186--194, 2018.

\bibitem{Zaheer2017}
Manzil Zaheer, Satwik Kottur, Siamak Ravanbakhsh, Barnabas Poczos, Ruslan
  Salakhutdinov, and Alexander Smola.
\newblock {Deep Sets}.
\newblock {\em International Conference on Neural Information Processing
  Systems}, pages 3391--3401, 2017.

\bibitem{Zeng2018}
Yiming Zeng, Yu~Hu, Shice Liu, Jing Ye, Yinhe Han, Xiaowei Li, and Ninghui Sun.
\newblock {RT3D: Real-Time 3D Vehicle Detection in LiDAR Point Cloud for
  Autonomous Driving}.
\newblock {\em IEEE Robotics and Automation Letters}, 3(4):3434--3440, 2018.

\bibitem{Zhang2018b}
Han Zhang, Ian Goodfellow, Dimitris Metaxas, and Augustus Odena.
\newblock {Self-Attention Generative Adversarial Networks}.
\newblock {\em International Conference on Machine Learning}, pages 7354--7363,
  2019.

\bibitem{Zhang2017e}
Han Zhang, Tao Xu, Hongsheng Li, Shaoting Zhang, Xiaogang Wang, Xiaolei Huang,
  and Dimitris~N. Metaxas.
\newblock {StackGAN: Text to Photo-Realistic Image Synthesis With Stacked
  Generative Adversarial Networks}.
\newblock {\em IEEE International Conference on Computer Vision}, pages
  5907--5915, 2017.

\bibitem{Zhang2019}
Wenxiao Zhang and Chunxia Xiao.
\newblock {PCAN: 3D Attention Map Learning Using Contextual Information for
  Point Cloud Based Retrieval}.
\newblock {\em IEEE Conference on Computer Vision and Pattern Recognition},
  pages 12436--12445, 2019.

\bibitem{Zhang2019c}
Zhiyuan Zhang, Binh-Son Hua, and Sai-Kit Yeung.
\newblock {ShellNet: Efficient Point Cloud Convolutional Neural Networks using
  Concentric Shells Statistics}.
\newblock {\em IEEE International Conference on Computer Vision}, pages
  1607--1616, 2019.

\bibitem{Zhao2019a}
Hengshuang Zhao, Li~Jiang, Chi-wing~Fu Jiaya, and Jiaya Jia.
\newblock {PointWeb : Enhancing Local Neighborhood Features for Point Cloud
  Processing}.
\newblock {\em IEEE Conference on Computer Vision and Pattern Recognition},
  pages 5565--5573, 2019.

\bibitem{Zhao2020}
Lin Zhao and Wenbing Tao.
\newblock {JSNet: Joint Instance and Semantic Segmentation of 3D Point Clouds}.
\newblock {\em AAAI Conference on Artificial Intelligence}, 2020.

\bibitem{Zhao2007}
Wei Zhao, Shuming Gao, and Hongwei Lin.
\newblock {A Robust Hole-filling Algorithm for Triangular Mesh}.
\newblock {\em The Visual Computer}, 23(12):987--997, 2007.

\bibitem{Zhi2019}
Shuaifeng Zhi, Michael Bloesch, Stefan Leutenegger, and Andrew~J. Davison.
\newblock {SceneCode: Monocular Dense Semantic Reconstruction using Learned
  EncodedScene Representations}.
\newblock {\em IEEE Conference on Computer Vision and Pattern Recognition},
  pages 11776--11785, 2019.

\bibitem{Zhou2018a}
Yin Zhou and Oncel Tuzel.
\newblock {VoxelNet: End-to-End Learning for Point Cloud Based 3D Object
  Detection}.
\newblock {\em IEEE Conference on Computer Vision and Pattern Recognition},
  pages 4490--4499, 2018.

\bibitem{Zhu2017b}
Jun-Yan Zhu, Taesung Park, Phillip Isola, and Alexei~A. Efros.
\newblock {Unpaired Image-To-Image Translation Using Cycle-Consistent
  Adversarial Networks}.
\newblock {\em IEEE International Conference on Computer Vision}, pages
  2223--2232, 2017.

\bibitem{Zhu2018a}
Yingying Zhu, Jiong Wang, Lingxi Xie, and Liang Zheng.
\newblock {Attention-based Pyramid Aggregation Network for Visual Place
  Recognition}.
\newblock {\em ACM International Conference on Multimedia}, pages 99--107,
  2018.

\bibitem{Zou2020}
Chuhang Zou and Derek Hoiem.
\newblock {Silhouette Guided Point Cloud Reconstruction beyond Occlusion}.
\newblock {\em Winter Conference on Applications of Computer Vision}, 2020.

\bibitem{Zou2017}
Chuhang Zou, Ersin Yumer, Jimei Yang, Duygu Ceylan, and Derek Hoiem.
\newblock {3D-PRNN: Generating Shape Primitives with Recurrent Neural
  Networks}.
\newblock {\em IEEE International Conference on Computer Vision}, pages
  900--909, 2017.

\end{thebibliography}
\bibliographystyle{plain}  

\end{document}